\RequirePackage{silence}
\newcommand{\thesisTheme}{ucph} 

\documentclass[%
	table,
	paper=A4,					
	12pt,						
	twoside=false,				
	openright,					
	parskip=full,				
	chapterprefix=true,			
	headings=normal,			
	bibliography=totoc,			
	listof=totoc,				
	titlepage=on,				
	captions=tableabove,		
	draft=false,				
	abstract=on,                
]{scrreprt}

\usepackage [table]{xcolor}
\usepackage[utf8]{inputenc}
\usepackage[english]{babel}     
\DeclareOldFontCommand{\bf}{\normalfont\bfseries}{\mathbf}
\DeclareUnicodeCharacter{0301}{\'{e}}
\usepackage{rotating}

\usepackage{amsmath}
\usepackage{euscript}
\usepackage{latexsym}
\usepackage{tabularx}
\usepackage{booktabs}       
\usepackage{graphicx}       
\usepackage{amsfonts}       
\usepackage{natbib}
\usepackage{microtype}
\newcolumntype{P}[1]{>{\centering\arraybackslash}p{#1}}
\usepackage{lscape}
\usepackage{arydshln}

\usepackage{multirow, makecell}
\usepackage{tikz}
\usepackage{color, colortbl}
\usepackage{comment}
\usepackage{soul}
\usetikzlibrary{calc}
\usetikzlibrary{decorations.pathmorphing}
\usepackage{floatrow}

\newcommand*{\ie}{i.e.\@\xspace}

\usepackage{colortbl}
\definecolor{linksblue}{RGB}{2, 82, 189}

\definecolor{mygreen}{RGB}{179, 253, 179}

\definecolor{myred}{RGB}{255, 205, 205}

\definecolor{myblue}{RGB}{204, 229, 255}

\definecolor{myyellow}{RGB}{253, 253, 153}

\definecolor{mylgreen}{RGB}{181, 234, 245}

\definecolor{mydblue}{RGB}{20, 136, 199}

\definecolor{bad_res}{HTML}{800000}

\DeclareMathOperator*{\argmax}{arg\,max}

\DeclareMathOperator*{\argsort}{arg\,sort}

\definecolor{orange}{HTML}{D55E00}
\definecolor{blue_editing}{HTML}{56B4E9}
\definecolor{green_editing}{HTML}{009E73}
\definecolor{purple_editing}{HTML}{882255}

\usepackage{multirow}
\usepackage{stmaryrd}
\usepackage{paralist}
\usepackage{amsthm}

\theoremstyle{definition}
\newtheorem{example}{Example}[section]

\newcommand{\cqda}{
\texorpdfstring{CQD$^{\mathcal{A}}$}{CQD\^A}\xspace}
\newcommand{\flare}{\texorpdfstring{FLARE}\xspace}

\usepackage{url}
\usepackage{amsthm}
\usepackage{algorithm}
\usepackage{algorithmic}
\usepackage{multicol}
\usepackage{paralist} 
\usepackage{amssymb} 
\usepackage{breqn}
\usepackage{times}
\usepackage{latexsym}

\usepackage[T1]{fontenc}

\usepackage[utf8]{inputenc}

\usepackage{microtype}

\usepackage{inconsolata}
\usepackage{bbm}
\usepackage{times}
\usepackage{amsmath}
\usepackage{amsthm}
\usepackage{booktabs}
\usepackage{multirow}
\usepackage{latexsym}
\usepackage{subcaption}
\usepackage{graphicx}
\usepackage{hyperref}
\usepackage[T1]{fontenc}
\usepackage{url}
\usepackage[utf8]{inputenc}
\usepackage{tikz}
\usetikzlibrary{shapes.geometric, arrows}
\definecolor{mygray}{rgb}{0.86,0.86,0.86}
\usepackage{inconsolata}
\usepackage{enumitem}
\usepackage{appendix}
\usepackage{booktabs} 
\usepackage[algo2e]{algorithm2e}
\usepackage{Carrickc,lettrine}
\usepackage{GoudyIn}

\usepackage{eso-pic} 

\setulcolor{black}
\usepackage[super]{nth}

\definecolor{Blue}{rgb}{0, 0, 255}

\newcolumntype{+}{
  >{\@rowstyle}%
}

\newcommand*{\rowstyle}[1]{
  \gdef\@rowstyle{#1}%
  \@rowstyle\ignorespaces%
}

\newcolumntype{=}{
  >{\gdef\@rowstyle{}}%
}

\newcommand{\thesisTitle}{Reasoning Inconsistencies and How to Mitigate Them in Deep Learning}
\newcommand{\thesisSubtitle}{}
\newcommand{\thesisName}{Erik Arakelyan}
\newcommand{\thesisDate}{Submission date: 31$^{st}$ December 2024}

\newcommand{\thesisInternalSupervisor}{Isabelle Augenstein and Pasquale Minervini}
\newcommand{\thesisSubject}{\footnotesize This thesis has been accepted by the PhD School of The Faculty of Science, University of Copenhagen}

\usepackage[					
    figuresep=colon,        
    sansserif=false,        
    hangfigurecaption=false,
    hangsection=true,       
    hangsubsection=true,    
    colorize=full,          
    colortheme={\thesisTheme},  
    bibsys=biber,
    bibfile=references,       
    bibstyle=authoryear,        
]{ucphthesis}
\hypersetup{					
	pdftitle={\thesisTitle},	
	pdfsubject={\thesisSubject},
	pdfauthor={\thesisName},	
	plainpages=false,			
	colorlinks=true,
	citecolor=linksblue,
	allcolors=linksblue,
	linkcolor=linksblue, 
	pdfborder={0 0 0},			
	breaklinks=true,			
	bookmarksnumbered=true,		%
	bookmarksopen=true			%
    }

\subject{\vspace{4.5cm} \thesisSubject}
\title{\thesisTitle}
\subtitle{\thesisSubtitle}
\author{\thesisName \\
        \small{Supervised by {\thesisInternalSupervisor}}
    }
\date{\thesisDate}

\usepackage{cleveref}
\usepackage{adjustbox}
\usepackage{wrapfig}
\usepackage{float}
\usepackage{dsfont}
\usepackage{wrapfig}
\usepackage{longtable}
\usepackage{tabularx}
\usepackage{listings}
\usepackage{caption}
\usepackage{amsmath}
\usepackage{tabularx}
\usepackage{floatrow}
\newfloatcommand{capbtabbox}{table}[][\FBwidth]

\newcommand{\PreCoverBackground}{%
  \AddToShipoutPicture*{%
    \put(0,0){%
      \parbox[b][\paperheight]{\paperwidth}{%
        \vfill
        \centering
        \includegraphics[width=\paperwidth,height=\paperheight]{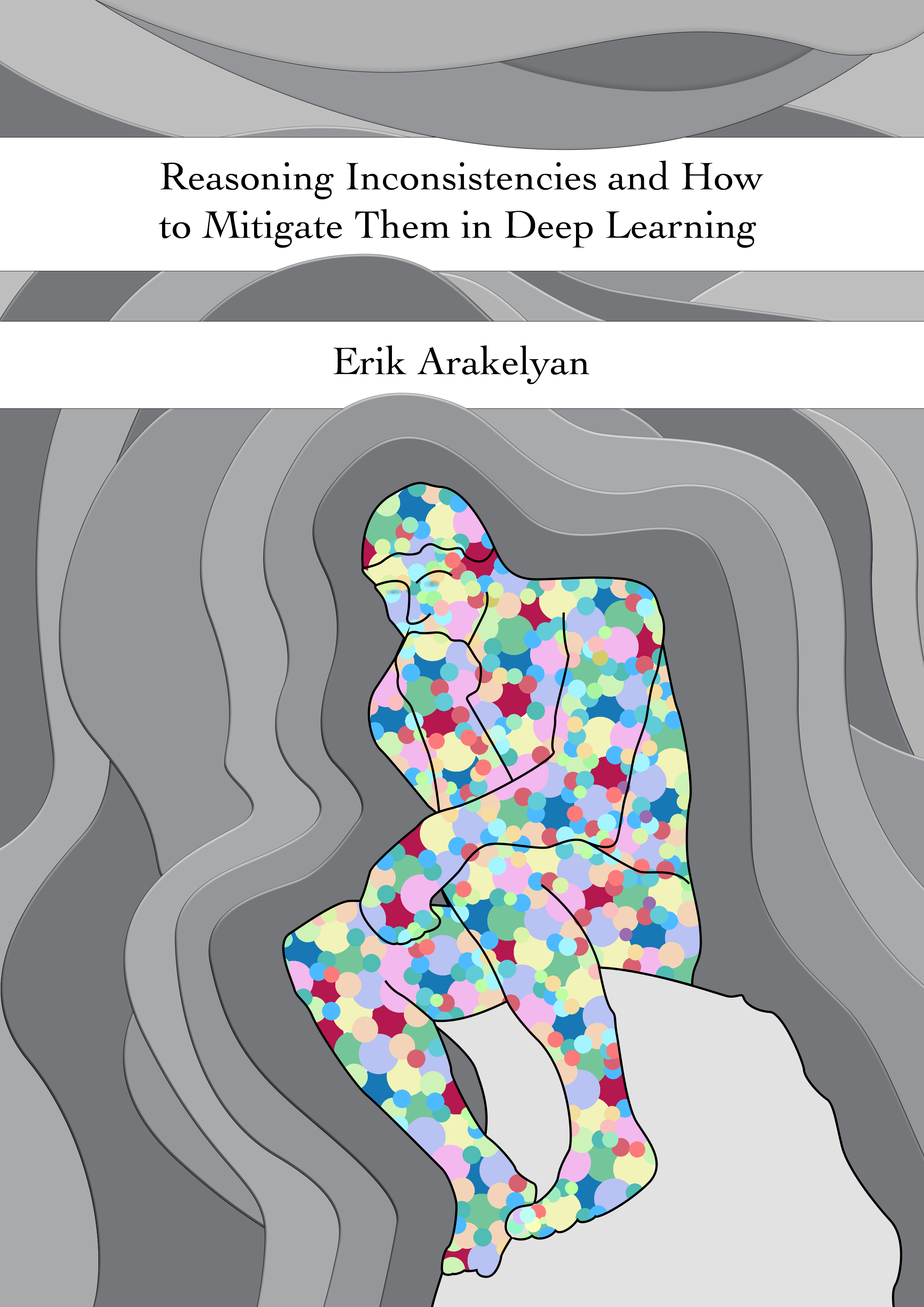}%
        \vfill
      }%
    }%
  }%
}

\begin{document}
\parskip=0pt
\pagenumbering{roman}
\pagestyle{empty}
\begin{titlepage}
  \PreCoverBackground
  \mbox{} 
\end{titlepage}

\AddToShipoutPicture*{\TitleBackground}     
\maketitle                                  
\pagestyle{plain}
\begin{center}
    \textit{"This thesis is dedicated to the cherished memory of Professor Norair Arakelian - a brilliant mathematician and an even more extraordinary grandfather whose legacy echoes lovingly through the corridors of time."}
\end{center}

\begin{figure}
    \centering
    \includegraphics[width=0.5\linewidth]{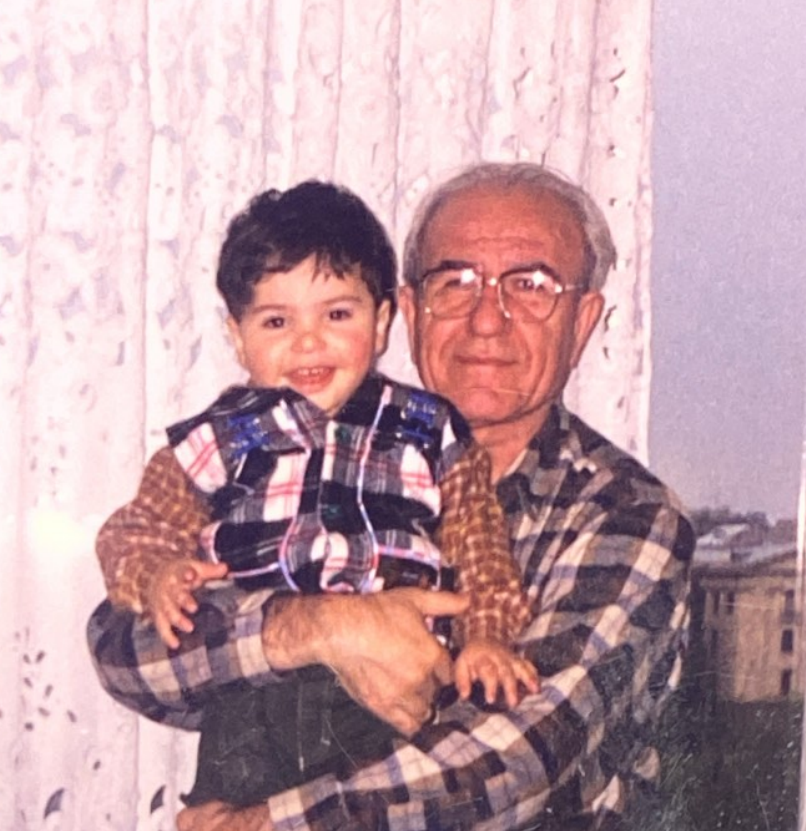}
    \label{fig:grandpa_ea}
\end{figure}

\pdfbookmark[0]{Acknowledgements}{Acknowledgements}
\chapter*{Acknowledgements}
\label{chap:acknowl}

As I am writing this segment, the pouring words inadvertently remind me of the past three years full of seemingly insurmountable perils and great triumphs. This journey of self-discovery would not have been possible without the support of many people who will always have a warm spot in my heart. These are the people who have shown me guidance and tough and tender love through their critiques and reassurances. 

First of all, I would like to thank both of my excellent supervisors, who have been there with me and for me each step of the way. I want to express my gratitude to Isabelle for believing in me, providing me with this incredible journey and this life-changing experience, and mentoring me on the true way of (ninja) being a scientist. Her pursuit of excellence and outstanding scientific narratives is something I took to heart. I want to dearly thank Pasquale for all the time and effort he spent helping me in all things professional and personal; without those late-night guidance sessions and talks, PhD would have felt tougher than bringing the one ring to Mordor. I am deeply indebted that he introduced me to the world of academic research during my MSc days leading me to where I am today.

This journey would not have been possible without my amazing collaborations - Daniel Daza and Michael Cochez, who have been a blessing to work with, constantly inspiring me for greater scientific discoveries and curious undertakings. 

I also want to thank Zhaoqi Liu and Gayane Ghazaryan, whom I was lucky to mentor during their time at the university. It was very inspiring to see such bright minds at the start of their careers but showing maturity in their research.

I definitely want to express my gratitude to all of my friends I met at the CopeNLU lab and for our fun times together - Marta, Sara, Haeun, Nadav, Jingyi, Siddhesh, Dustin and Greta. A profound thanks goes to Arnav and Pepa, with whom we had a fantastic time talking and laughing about all sorts of things meaningful, funny or mundane in the office. Thanks for all the support.

A very special corner of this acknowledgement goes to my dear friends Karolina and Karen, with whom we share countless hours of deep scientific discussions and heartwarming, friendly banter. I will always cherish these times with a nostalgic smile on my face.

This journey would not have been possible without my amazing family's complete support. Words cannot express my deep gratitude for the unconditional love, understanding, support and encouragement that I have received from my wife and soulmate, Tamara and my mother, Araksya. That gratitude also goes to my grandmother Marietta, grandfather Norair, my aunt Anahit, her husband Armen, and my brothers Vahan and David, who have been the inspiration for me to pursue this endeavour. 

\pdfbookmark[0]{Abstract}{Abstract}
\chapter*{Abstract}
\label{chap:abstract}

\lettrine[lines=3]{T}{he} recent advancements in Deep Learning (DL) models and techniques have led to significant strides in performance across diverse tasks and modalities. However, while the overall capabilities of models show promising growth, our understanding of their internal reasoning processes remains limited, particularly concerning systematic inconsistencies or errors—patterns of logical or inferential flaws. These inconsistencies may manifest as contradictory outputs, failure to generalize across similar tasks, or erroneous conclusions in specific contexts. Even detecting and measuring such reasoning discrepancies is challenging, as they may arise from opaque internal procedures, biases and imbalances in training data, or the inherent complexity of the task. Without effective methods to detect, measure, and mitigate these errors, there is a significant risk of deploying models that are biased, exploitable, or logically unreliable.

This thesis aims to address these issues by producing novel methods for deep learning models that reason over knowledge graphs, natural language, and images.
Firstly, the thesis contributes two techniques for detecting and quantifying predictive inconsistencies originating from opaque internal procedures in natural language and image processing models. We systematically evaluate a wide range of model families within novel adversarial setups that explicitly expose those internal procedures, allowing us to quantify significant reasoning discrepancies within these models.
To mitigate inconsistencies from biases in training data, this thesis presents a data-efficient sampling method to improve fairness and performance and a synthetic dataset generation approach to rigorously evaluate and enhance reasoning in low-resource scenarios.
Finally, the thesis offers two novel techniques to explicitly optimize the models for complex reasoning tasks in natural language and knowledge graphs. These methods directly enhance model performance while allowing for more faithful and interpretable exploration and exploitation during inference.
Critically, by addressing reasoning inconsistencies through quantifying and mitigating them with deep learning models, this thesis provides a comprehensive framework to improve the robustness, fairness, and interpretability of deep learning models across diverse tasks and modalities.      
\pdfbookmark[0]{Resume}{Resume}
\chapter*{Resum{\'e}}
\label{chap:abstract_dk}

De seneste fremskridt inden for dyb læring (DL) modeller og teknikker har ført til en betydelig forbedring af ydeevnen på tværs af forskellige opgaver og modaliteter. Imidlertid, mens modellernes overordnede kapacitet viser lovende vækst, vores forståelse af deres interne tankevirksomheder forbliver begrænset, især med hensyn til systematiske uoverensstemmelser eller fejl — mønstre af logiske eller inferentielle mangler. Disse uoverensstemmelser kan manifestere sig som modstridende udgange, manglende generalisering på tværs af lignende opgaver eller fejlagtige konklusioner i specifikke sammenhænge. Selv det er en udfordring at opdage og måle sådanne tankeforskelle, da de kan opstå som følge af uigennemsigtige interne procedurer, skævheder og ubalancer i træningsdata eller fordi denne opgave er meget kompleks. Uden effektive metoder til at opdage, måle og afbøde disse fejl er der en betydelig risiko for at implementere modeller, der er partiske, udnyttelige eller logisk upålidelige.

Denne afhandling har til formål at løse disse problemer ved at producere nye metoder til dybe læringsmodeller, der ræsonnerer over videngrafer, naturligt sprog og billeder. Den første del af afhandlingen bidrager med to teknikker til at detektere og eksplicit kvantificere forudsigelige uoverensstemmelser, der stammer fra uigennemsigtige interne procedurer i naturlige sprog- og billedbehandlingsmodeller. Vi evaluerer systematisk en bred vifte af modelfamilier inden for nye kontradiktoriske opsætninger, der eksplicit udsætter disse interne procedurer, giver os mulighed for at kvantificere betydelige tankeforskelle inden for disse modeller. For at afbøde uoverensstemmelser fra fordomme i træningsdata, denne afhandling præsenterer en dataeffektiv prøveudtagningsmetode til forbedring af retfærdighed og ydeevne og en syntetisk datasætgenereringsmetode til nøje at evaluere og forbedre ræsonnement i scenarier med lav ressource. Endelig tilbyder afhandlingen to nye teknikker til eksplicit at optimere modellerne til komplekse tankeopgaver i naturlige sprog- og vidensgrafer. Disse metoder forbedrer direkte modelens ydeevne, samtidig med at de giver mulighed for mere trofast og fortolkbar udforskning og udnyttelse under inferens. Kritisk, ved at adressere tankeforskelle gennem kvantificering og afbødning af dem med dybe læringsmodeller, denne afhandling giver en omfattende ramme for at forbedre robustheden, retfærdighed, og fortolkbarhed af dybe læringsmodeller på tværs af forskellige opgaver og modaliteter.
\pdfbookmark[0]{Publications}{Publications}
\chapter*{Publications}
\label{chap:publications}
This thesis includes the following papers as chapters, listed in the order of their appearance ($^*$ denotes equal contribution):

\begin{enumerate}
    \item \cite{DBLP:conf/eacl/ArakelyanLA24} Erik Arakelyan$^*$, Zhaoqi Liu$^*$ and Isabelle Augenstein, \textit{Semantic Sensitivities and Inconsistent Predictions: Measuring the Fragility of NLI Models}, 2024, European Chapter of the Association for Computational Linguistics (EACL), \textit{outstanding paper award}, pages 432-444
    \item \cite{arakelyan2024greatbackbones} Erik Arakelyan, Karen Hambardzumyan, Davit Papikyan, Pasquale Minervini, Aram H. Markosyan, Albert Gordo and Isabelle Augenstein, \textit{With Great Backbones Comes Great Adversarial Transferability}, 2024, CoRR (Under Review for ICML 2025)
    \item \cite{DBLP:conf/acl/ArakelyanAA23} Erik Arakelyan, Arnav Arora and Isabelle Augenstein, 2023, \textit{Topic-Guided Sampling For Data-Efficient Multi-Domain Stance Detection}, Annual Meeting of the Association for Computational Linguistics (ACL), pages 13448-13464
    \item \cite{DBLP:journals/coling/abs-2406-14425} Gayane Ghazaryan$^*$, Erik Arakelyan$^*$, Pasquale Minervini and Isabelle Augenstein, \textit{SynDARin: Synthesising Datasets for Automated Reasoning in Low-Resource Languages}, 2024, Proceedings of the 31st International Conference on Computational Linguistics (COLING)
    \item \cite{DBLP:conf/nips/ArakelyanMDCA23} Erik Arakelyan$^*$, Pasquale Minervini$^*$, Daniel Daza, Michael Cochez and Isabelle Augenstein, \textit{Adapting Neural Link Predictors for Data-Efficient Complex Query Answering}, 2023, Advances in Neural Information Processing Systems 36 (NeurIPS)
    \item \cite{DBLP:journals/corr/arakelyan-flare} Erik Arakelyan, Pasquale Minervini, Pat Verga, Patrick S. H. Lewis and Isabelle Augenstein, \textit{FLARE: Faithful Logic-Aided Reasoning and Exploration}, 2024, CoRR (Under Review for ICLR 2025)
    \item \cite{DBLP:series/faia/CochezAABD0MNR23} Michael Cochez, Dimitrios Alivanistos, Erik Arakelyan, Max Berrendorf, Daniel Daza, Mikhail Galkin, Pasquale Minervini, Mathias Niepert and Hongyu Ren, \textit{Approximate Answering of Graph Queries}, 2023, Compendium of Neurosymbolic Artificial Intelligence, pages 373-386
\end{enumerate}

The survey paper $7$ is presented through definitions and problem formulations across different parts of the thesis, but not as a separate chapter.



\clearpage

\setcounter{secnumdepth}{3}
\setcounter{tocdepth}{2}		
{
\hypersetup{linkcolor=black}

\tableofcontents				
    \clearpage
\listoffigures
     \clearpage
\listoftables
     \clearpage
}
\pagenumbering{arabic}			
\setcounter{page}{1}			
\pagestyle{maincontentstyle} 	
\part{Executive Summary}
\chapter{Executive Summary}
\label{chap:intro}

\section{Introduction}

The emergence of data-driven learning and predictive approaches \citep{carbonell1983overview} has unequivocally led to the question \textit{"How do machines reason?"}. In the spirit of this, earlier machine learning methods \citep{anderson1983machine} were constructed with ingrained mechanisms that directly explain the complete sequence for arriving at the solution \citep{bratko1997machine}. The transparency of the reasoning procedure allowed for steering away from potential biases \citep{he2009learning} introduced through data imbalances and the evaluation of the complexity of the designated task in terms of the expressivity \citep{vapnik1999overview} of the model, as well as the sufficiency \citep{balasubramanian2014conformal,vapnik1999overview} of the learning methodology. 
With the advancement of computing resources \citep{lecun20191,sze2017hardware}, deep learning models have made considerable strides in pushing state-of-the-art performance in reasoning over natural language \citep{DBLP:conf/nips/WangPNSMHLB19, DBLP:journals/tmlr/SrivastavaRRSAF23}, images \citep{DBLP:journals/ijcv/RussakovskyDSKS15, DBLP:conf/eccv/LinMBHPRDZ14} and knowledge graphs \citep{DBLP:journals/eswa/ChenJX20}. The most recent advancements have been propelled by developing large models pre-trained with vast amounts of data \citep{DBLP:conf/nips/VaswaniSPUJGKP17, DBLP:conf/iclr/DosovitskiyB0WZ21}. However, not all that glitters is gold.
The added structural and algorithmic learning complexity within these models \citep{DBLP:conf/naacl/DevlinCLT19, radford2018improving} has significantly limited the potential to tractably interpret \citep{DBLP:conf/fat/BenderGMS21} or follow the reasoning processes within them \citep{DBLP:conf/emnlp/AtanasovaSLA20}. Consequently, a large portion of modern explainability methods attempt to create explanations from the final model predictions, i.e. \emph{post-hoc} \citep{madsen2022post}, through either attribution methods producing input saliency maps \citep{DBLP:journals/inffus/ArrietaRSBTBGGM20} or a lens for analysing a specific part of the model architecture \citep{vashishth2019attention}. Further, methods attempting a complete mechanistic interpretation have not been shown to be scalable as models get larger \citep{DBLP:journals/corr/abs-2404-14082}. Additionally, explainability methods have been shown to struggle with faithfulness towards the inner workings of the model, rationale and dataset consistency \citep{DBLP:conf/emnlp/AtanasovaSLA20}. 

With these issues at heart, my dear reader, this thesis aims to create a set of tools for directly detecting, measuring, and mitigating systematic errors in the decision-making of deep learning models. In particular, I am interested in reasoning errors originating from opaque processes that mask erroneous behaviour in these models, data imbalances, and complex reasoning tasks. Towards this end, the research output of this thesis offers two methods for formulating adversarial setups in which the erroneous model behaviour is detected and explicitly quantified. Following this, the consequent research contributions focus on assessing the impact of imbalances in the training data on the reasoning behaviour of the model and further suggesting methods that mitigate these biases. Finally, we present two techniques that directly optimize complex reasoning tasks in knowledge graphs and natural language.

The following introductory \crefrange{subsec:reasoning_what}{subsec:complex_query_answering} detail the background concepts and tasks relevant to the presented research contributions. This is followed by \cref{sec:scientific_contrib} with a detailed overview of the individual contributions present within this thesis. \cref{sec:discuss} provides a discussion of the mentioned research with suggested potential future exploration directions.

\subsection{What is Reasoning in Deep Learning?}
\label{subsec:reasoning_what}

The concept of \emph{"reasoning"} is one that has been studied and discussed vividly across the formulation of modern philosophy \citep{scriven1976reasoning}. While the philosophical composition of this idea is not explicitly linked to this thesis, let us go on a slight tangent, which will clarify some motivations and ideas within the presented research. The \emph{Dictionary of Philosophy} defines "reasoning" as the \textit{"The process of inferring conclusions from the statement"} \citep{angeles1981dictionary}. This definition garnered a famous critique \citep{walton1990reasoning} w.r.t. the use of the word \emph{"inferring"}, as it is ill-defined. The alternative formulation suggests defining an inference as the use of a rule for creating a connection between a set of propositions (statements). The initial set of propositions is the \emph{premise} from which the inference starts and moves towards the \emph{conclusions}. This allows to formalize reasoning as a directional process that links the premises to the conclusions through a rule. This, unsurprisingly, is rather similar to how models operate in Deep Learning \citep{DBLP:journals/nature/LeCunBH15}, with the aim of connecting the inputs to the outputs through a series of learned transformations/rules.

The advancements in deep learning have allowed the creation of architectures \citep{gu2018recent,DBLP:journals/neco/YuSHZ19} that simultaneously learn both local and global features \citep{DBLP:conf/nips/KavukcuogluSBGML10} through a series of intermediate transformations. Particularly with the emergence of transformers \citep{DBLP:conf/nips/VaswaniSPUJGKP17}, that utilise attention mechanisms \citep{DBLP:journals/tkde/BrauwersF23} w.r.t. the input and intermediate representations, the ability of deep learning models to process complex tasks has significantly improved. These diverse mechanisms allow the deep learning models to perform a series of intermediate transformations that lead to the final output. This directional process is the definition of reasoning in these models. 

As the deep models grow larger, we see their increased performance across a variety of tasks from natural language understanding and generation \citep{DBLP:conf/nips/WangPNSMHLB19, DBLP:journals/tmlr/SrivastavaRRSAF23, DBLP:conf/iclr/HendrycksBBZMSS21}, image processing \citep{DBLP:journals/ijcv/RussakovskyDSKS15, DBLP:conf/eccv/LinMBHPRDZ14} and complex query answering over knowledge graphs \citep{DBLP:journals/eswa/ChenJX20}. Although the performance increase is substantial, it must be noted that the tasks do not directly test the intermediate reasoning of the model and are evaluated strictly based on the final output. Some of the tasks used in the thesis are introduced in the following subsections and \cref{subsec:complex_query_answering}.

\subsubsection{Stance Detection}

\begin{figure}[t]
    \centering
    \includegraphics[width=\textwidth]{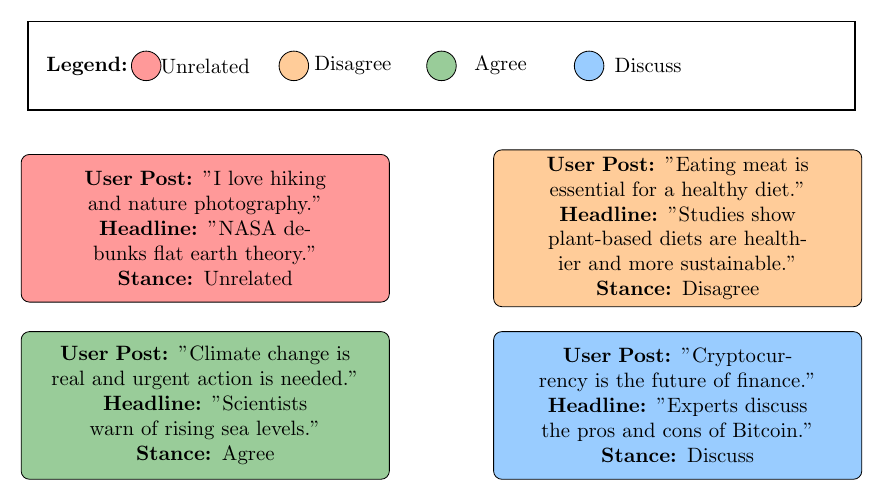}
    \caption{Examples of stance detection with labels: Unrelated, Disagree, Agree, and Discuss.}
    \label{fig:stance-detection}
\end{figure}

Stance Detection is a task in natural language processing where, given a piece of text, the model must identify the stance or attitude expressed towards the designated target \citep{DBLP:journals/csur/KucukC20}. This target can be anything from a political issue, a social event, or an entity. Although the label vocabulary might vary for diverse formulations of stance detection \citep{DBLP:conf/emnlp/HardalovANA21}, an example of a task can be seen in \cref{fig:stance-detection}. The task is important in several domains, such as social media monitoring \citep{DBLP:journals/ipm/AlDayelM21} or political analysis \citep{DBLP:journals/csl/LaiCFBPR20}, where understanding public sentiment can help predict trends, break down and study public opinion regarding existing discourse, or detect harmful content. It must be noted that accomplishing this requires the model to perform contextual and logical inference from the text and the fixed target to the linked attitude between them. This is where reasoning errors can manifest, as models may misinterpret nuanced expressions, irony, or indirect statements, especially when trained on imbalanced datasets.

The research output discussed in the thesis \citep[\textit{Paper 3}]{DBLP:conf/acl/ArakelyanAA23} aims to address these challenges by proposing novel methodologies to improve stance detection across diverse topics. Specifically, the contributions include introducing a topic-guided diversity sampling strategy and a contrastive learning objective, both designed to enhance the model's ability to generalize effectively while mitigating class imbalance issues. The topic-guided diversity sampling technique ensures that the training data is balanced not only across classes but also among topics. This is achieved by prioritizing the selection of examples that maximize topic diversity while maintaining a representative sample of stance labels. The method counters the skewed distributions commonly found in stance detection datasets, allowing models to learn more robust and generalized representations.

\subsubsection{Natural Language Inference}

\begin{figure}[t!]
    \centering
    \begin{subfigure}{0.32\textwidth}
        \includegraphics[width=\textwidth]{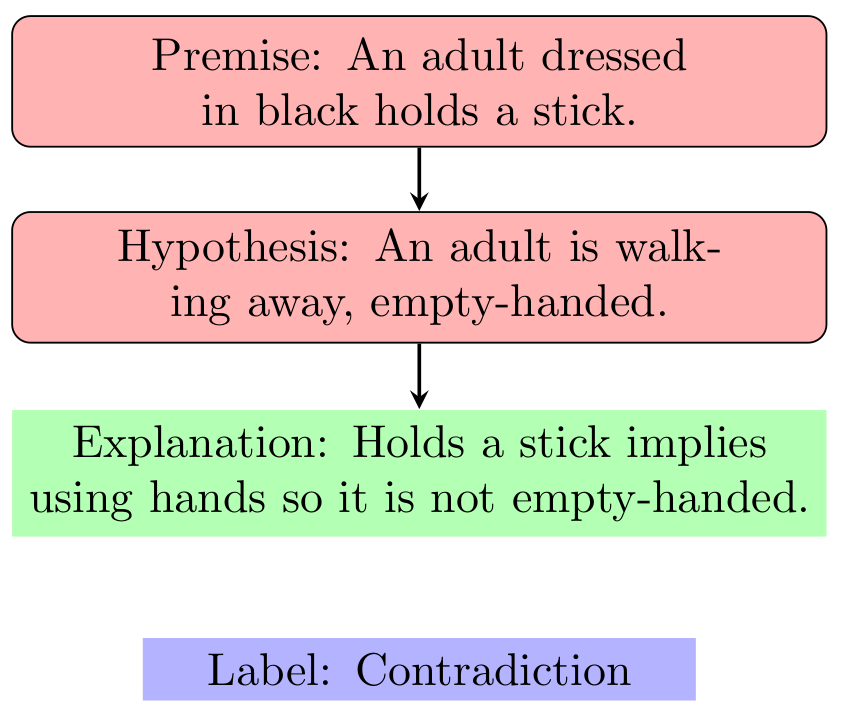}
        \caption{Contradiction}
        \label{fig:contr}
    \end{subfigure}
    \hfill
    \begin{subfigure}{0.32\textwidth}
        \includegraphics[width=\textwidth]{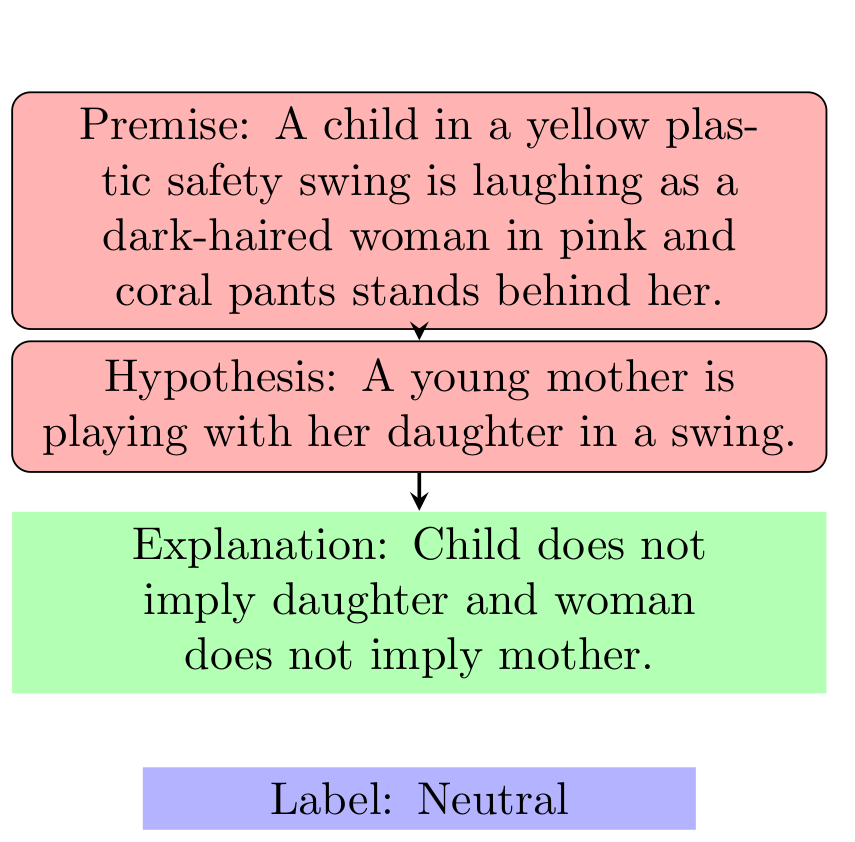}
        \caption{Entailment}
        \label{fig:ent}
    \end{subfigure}
    \hfill
    \begin{subfigure}{0.32\textwidth}
        \includegraphics[width=\textwidth]{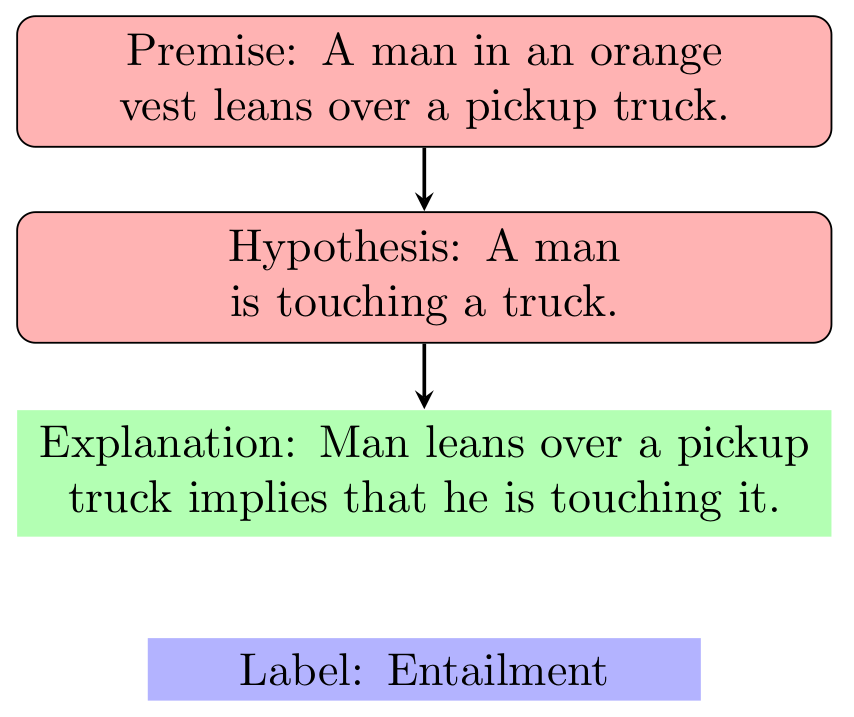}
        \caption{Neutral}
        \label{fig:neut}
    \end{subfigure}
    \caption{Examples of natural language inference (NLI) reasoning with explanations and labels.}
    \label{fig:nli_examples}
\end{figure}

The task of textual entailment \citep{dagan2005pascal}, otherwise referred to as Natural Language Inference \citep[NLI]{bowman2015large}, has been widely used to probe how well the models understand language \citep{condoravdi2003entailment, williams2017broad, nie2019adversarial}. This is a pairwise input task, where given a premise and a hypothesis, the objective is to predict if the premise \emph{entails, contradicts} or is \emph{neutral} towards the hypothesis. An example of this task can be seen in \cref{fig:nli_examples}. 

This task is rather suited for examining model reasoning patterns, as it demands a logical and contextual understanding to predict the relationships between the premise and the hypothesis. The main challenge arises from the complexity of language nuances, such as ambiguity in wording and latently implied meanings and ideas, which demand deeper semantic comprehension. To solve this task, the model must be capable of mapping a set of transformations from premises to hypotheses, effectively simulating a form of reasoning. Consequently, NLI has become a cornerstone task for evaluating reasoning in deep learning models due to its structured nature and availability of large-scale datasets like MNLI \citep{williams2017broad}, SNLI \citep{bowman2015large} and ANLI \citep{DBLP:conf/acl/NieWDBWK20}.

However, one must consider the limitations inherent in current benchmarks. For example, studies have shown that models often rely on shallow spurious correlations \citep{DBLP:conf/acl/McCoyPL19, DBLP:conf/emnlp/SchusterSYFSB19}, such as lexical overlap \citep{DBLP:conf/emnlp/RajaeeYP22}, lack generalisation out of distribution \citep{DBLP:journals/corr/abs-2110-01518} or fail to acquire capabilities for abstract or logical reasoning \citep{DBLP:journals/tacl/TalmorEGB20}. Models have even been shown to achieve high performance without the presence of hypothesis \citep{DBLP:conf/naacl/GururanganSLSBS18}. This misalignment underscores the importance of developing evaluation methods that test reasoning directly, as opposed to proxy metrics.

In this thesis, we extend the exploration of NLI beyond conventional datasets by introducing adversarially constructed examples aimed at exposing reasoning flaws \cite[\textit{Paper 1}]{DBLP:conf/eacl/ArakelyanLA24}. We demonstrate that state-of-the-art Natural Language Inference models are sensitive towards minor surface-form variations that preserve semantics, which can cause significant inconsistencies in their inference decisions. Critically, this behaviour contrasts with a genuine, nuanced understanding of compositional semantics. However, it remains undetected when assessing model accuracy on traditional benchmarks or when probing for syntactic, monotonic, and logical reasoning capabilities. To analyze this phenomenon, we test NLI models on adversarially crafted examples featuring semantics-preserving surface-level noise. These examples are generated using conditional text generation, with a specific requirement that the NLI model identifies the relationship between the original and adversarial inputs as symmetric equivalence entailment.

\begin{figure}[t!]
\centering
\includegraphics[width=\textwidth]{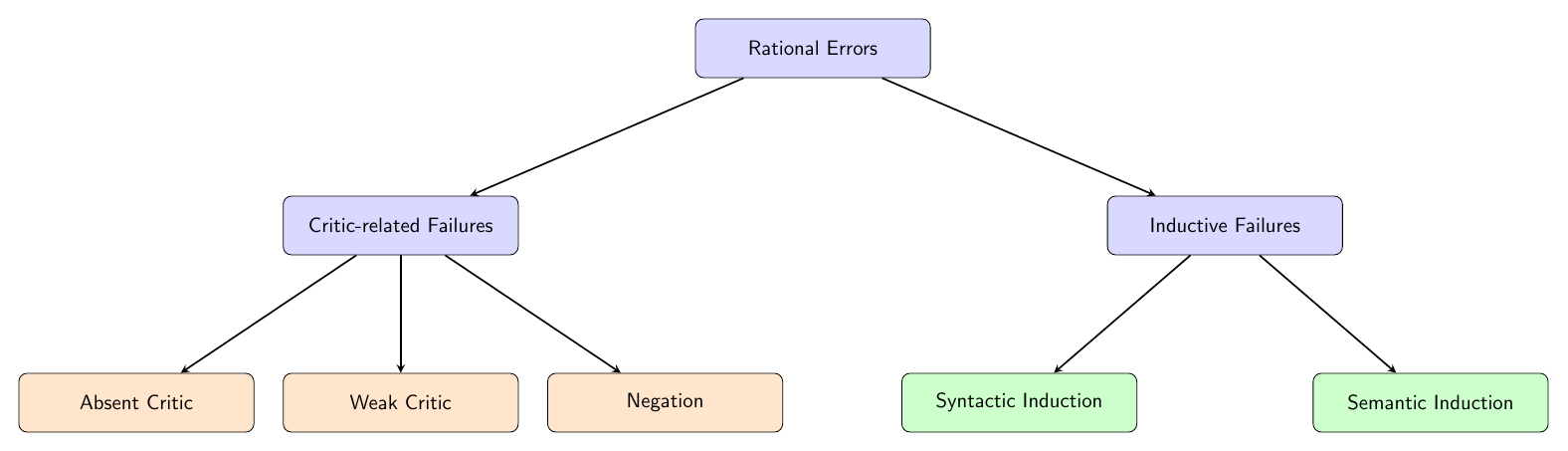}
\caption{Taxonomy of Rationel Errors in human cognition \citep{ben1998rational}: Critic-related and Inductive Failures.}
\label{fig:error-taxon}
\end{figure}

\subsection{Why does reasoning go wrong?}
\label{subsec:reasoning_wrong}

After we defined the notion of reasoning in the previous section, the next important task would be trying to formalise what a reasoning error is and gauge the potential origins of those errors in deep learning. Prior research of errors in human rationale \citep{ben2012erroneous} has suggested that systematic reasoning flaws observed in human cognition,  tend to emerge not from randomness but from deliberate, rule-based processes \citep{ben1998rational}. As seen in \cref{fig:error-taxon}, the errors arise from misapplications of principles during learning or problem-solving, which are categorised into \emph{critic-related} failures and \emph{inductive misgeneralizations}. Critic-related failures occur when mechanisms to detect inconsistencies are missing, weak, or suppressed. For example, a model may fail to identify contradictions in the text due to inadequate intermediate validation. Inductive misgeneralizations arise from overgeneralizing or overspecializing rules based on patterns priorly internalised patterns. Semantic induction involves errors due to flawed analogies or understanding of ambiguous concepts. 

This is directly related to reasoning inconsistencies we find in deep learning models. Models often exploit spurious correlations \citep{DBLP:conf/nips/IzmailovKGW22} internalized from the training data, such as associating specific keywords with outcomes, which breaks down in nuanced contexts \citep{wang-etal-2022-identifying}. A model might also be unable to create a comprehensive internal representation for further reasoning because of the complexity of the task \citep{vu-etal-2020-exploring}. For instance, a model might misinterpret sarcasm by focusing on isolated words rather than broader context \citep{verma2021techniques}, or fail to produce a cohesive set of inferences when solving mathematical tasks \citep{patel-etal-2021-nlp}.

\begin{figure}[!t]
    \centering
    \includegraphics[width=\textwidth]{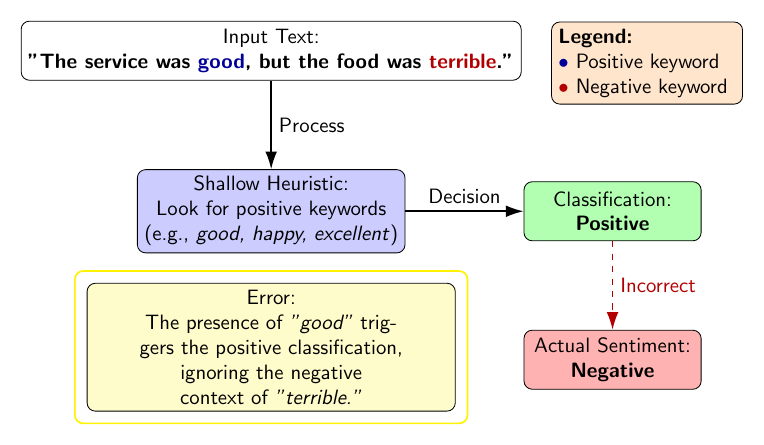}
    \caption{Example of a shallow heuristic used in a simple sentiment analysis reasoning process.}
    \label{fig:shallow-heruistic}
\end{figure}

\begin{figure}[!t]
    \centering
    \includegraphics[width=\textwidth]{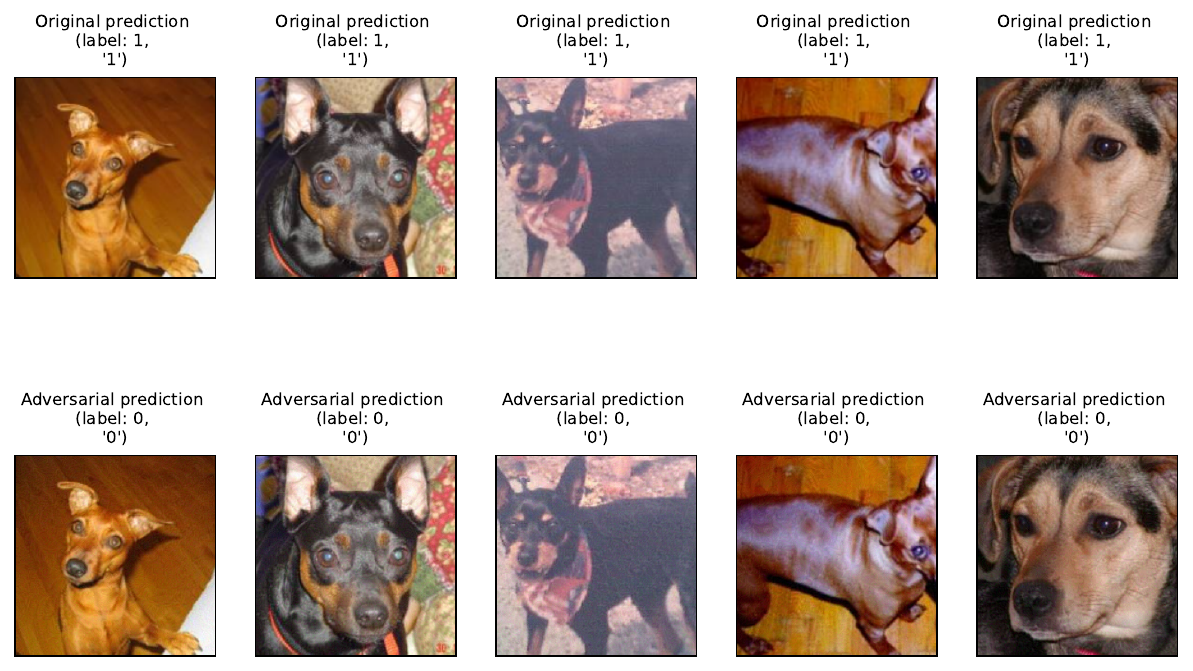}
    \caption{Example of an imperceptible adversarial noise added (bottom) to the original image (top) that changes the final prediction of the model.}
    \label{fig:adversarial-example}
\end{figure}
\subsubsection{Internal Procedures}

Errors arising from the internal procedures of a model often stem from a misalignment between the learned representations and their transitions into the final output and the ground truth. Deep learning models perform a sequence of transformations on input data, mapping it to intermediate representations and eventually to outputs. However, these transformations may inadvertently optimize for surface-level patterns rather than deeper semantic or logical relationships \citep{DBLP:conf/acl/McCoyPL19}. For instance, models frequently rely on shallow heuristics, as shown in \cref{fig:shallow-heruistic}, where specific tokens or phrases disproportionately influence predictions. These heuristics often lead to spurious correlations that degrade performance in nuanced contexts or under distributional shifts \citep{DBLP:conf/emnlp/SchusterSYFSB19}. Moreover, models fail to perform necessary internal validation, leaving inconsistencies undetected and unresolved \citep{DBLP:conf/icml/Minervini0SGR20}. 

Another reasoning inconsistency in deep learning models is the adversarial exploitability of their internal representations \citep{DBLP:journals/access/AkhtarMKS21}. These representations encode intermediate abstractions of input data through high-dimensional embeddings, ideally capturing semantic and logical relationships and reasoning required for the task. However, adversarial attacks exploit discrepancies within these representations, by introducing imperceptible noise to the input as seen in \cref{fig:adversarial-example} \citep{bhambri2019survey}. This noise can significantly shift the internal representations, causing erroneous predictions, thus highlighting the susceptibility of the models towards non-semantic perturbations. This means maintaining consistency in inference is a non-trivial task in deep learning.

In this thesis, I systematically explore the reasoning inconsistencies stemming from both inductive and critic-related failures. Our findings \citep[\textit{Paper 2}]{arakelyan2024greatbackbones} show that models trained from pre-trained backbones, like ResNet and ViT, are highly vulnerable to adversarial attacks, even when attackers possess only partial knowledge of the target model's tuning details. I introduce \emph{backbone attacks}, which solely rely on the available feature extractors, and show that even such knowledge can induce significant disruptions, often matching the effectiveness of \emph{white-box} attack strategies. 

\subsubsection{Data Imbalances}
\label{subsubsec:data_imbalances}

Imbalances in the data are a common origin for reasoning inconsistencies and discrepancies in deep learning models \citep{DBLP:journals/csur/kaurPM19}. They occur when certain classes, features, or relationships are over or underrepresented in the training data, leading to biases emerging in the predictions of the model. Imbalances can manifest in various forms, such as class imbalance \citep{DBLP:journals/jbd/JohnsonK19}, topic imbalance, or semantic skew \citep{garrido2021survey}, limiting the generalization capabilities of the model.

The presence of class imbalances means that a particular label has a dominant presence in the training data, which can cause the model to overfit the majority class while maintaining suboptimal reasoning patterns for minority classes, thus biasing predictions because of over-reliance on spurious correlations \citep{wang-culotta-2020-identifying}. The topic or semantic imbalances limit the diversity of relationships the model internalizes, not allowing the model to adapt its reasoning to diverse contexts \citep{johnson2019survey}. Although strategies for mitigating class imbalances exist \citep{DBLP:journals/jbd/HasaninKLB19,rendon2020data}, the pursuit of the same success for imbalances of semantic representation has been studied to a lesser degree.

In this thesis, I expand the current research on data-efficient sampling, introducing a topic-guided diversity sampling method that ensures that the training data is balanced not only across classes but also across topics and semantic nuances \cite[\textit{Paper 3}]{DBLP:conf/acl/ArakelyanAA23}. By integrating this technique into model training, we show significant improvements in model accuracy, robustness on out-of-domain evaluation and reasoning consistency across domains.

\subsubsection{Task Complexity}

The last potential cause of reasoning inconsistencies discussed in this thesis is connected to task complexity. Since the emergence of probabilistic predictive methods, various frameworks have been proposed to measure and analyse the complexity of tasks w.r.t. the capacity of the predictive model \citep{DBLP:journals/jacm/BlumerEHW89,DBLP:conf/aaai/Haussler90} and the capability of the learning algorithms \citep{kearns1994introduction} to find the optimal model for the designated task, forming what's known as its \emph{effective capacity}. 

Assessing the capacity of a deep learning model is challenging as the effective capacity depends on the chosen optimization algorithm in a non-convex setting, offering little theoretical insight \citep{DBLP:journals/Kais/HuCPLB21}. Consequently, quantifying the difficulty of the task given the designated model becomes an insurmountable challenge. The task complexity can be compounded by a plethora of factors, such as the number of reasoning steps required, the presence of hierarchical or nested dependencies, ambiguity in data representation, or the inherent difficulty of capturing abstract relationships. The inability to explicitly measure the complexity of the task, along with these challenges, limits the means to optimize a model for generalizing towards a comprehensive logical representation for that task, resulting in reasoning inconsistencies. For instance, in compositional generalization tasks \citep{DBLP:conf/iclr/KeysersSSBFKMSS20}, where the goal is to generalize learned components to novel combinations, models often struggle to extrapolate rules to unseen contexts. Similarly, in mathematical reasoning \citep{DBLP:conf/iclr/SaxtonGHK19}, solving problems with nested or multi-step operations requires structured reasoning pathways, retaining and recurrently reusing prior inductions and deductions across different reasoning stages. However, models often fail to maintain consistency in intermediate representations, leading to errors that accumulate over inference steps.

To address these challenges, the research output in the thesis presents a novel reasoning method over knowledge graphs \citep[\textit{Paper 5}]{DBLP:conf/nips/ArakelyanMDCA23} that includes learnable adaptation layers that directly optimise the intermediate answers and representations during the inference. This boosts the generalisation towards unseen types of queries and increases the \emph{effective capacity} of the model, along with the added benefit that the method remains data-efficient.

Another discrepancy present in modern Large Language Models (LLM) is that while they exhibit strong performance on numerous language reasoning tasks, they often lack a structured and faithful inference mechanism when solving complex queries. This means that while the model might output tokens of intermediate reasoning, their exact impact on the final answer is not explicitly known. Moreover, the reasoning written in natural language lacks explicit verifiability because it is inherently freeform. To overcome this, we introduce \textbf{F}aithful \textbf{L}ogic-\textbf{A}ided \textbf{R}easoning and \textbf{E}xploration (\textbf{\flare}) \citep[\textit{Paper 6}]{DBLP:journals/corr/arakelyan-flare}, a novel interpretable approach for traversing the problem space using task decompositions. The method enhances reasoning interpretability and faithfulness by combining task decomposition, Prolog-like logical formalization, and LLM simulated search. Critically, {\flare} addresses task complexity by enhancing the reasoning capacity of LLMs without solely relying on deterministic algorithms. It supports multi-hop reasoning, task decomposition, and logical consistency verification. The results highlight FLARE's state-of-the-art performance on several datasets, achieving significant improvements in reasoning faithfulness and task accuracy. To overcome this, we introduce \textbf{F}aithful \textbf{L}ogic-\textbf{A}ided \textbf{R}easoning and \textbf{E}xploration (\textbf{\flare}), a novel interpretable approach designed to navigate the problem space through task decompositions.




\subsection{Reasoning with Complex Questions}

\label{subsec:complex_query_answering}

The emergence of strong deep learning models in natural language \citep{vaswani2017attention,DBLP:journals/corr/abs-2307-09288}, image processing \citep{DBLP:conf/iclr/DosovitskiyB0WZ21,DBLP:journals/corr/abs-2405-09818} and query answering over knowledge graphs \citep{DBLP:conf/iclr/0001YM0Z24}, created the necessity for more elaborate evaluation benchmarks. The main added components for reasoning over these new datasets \citep{DBLP:conf/emnlp/Yang0ZBCSM18,DBLP:journals/tacl/KwiatkowskiPRCP19,DBLP:journals/corr/abs-2110-14168}, was that the models needed to adapt to the presence of semantic ambiguity \citep{DBLP:journals/tacl/GevaKSKRB21} within the questions, the necessity for multi-hop reasoning \citep{DBLP:conf/emnlp/Yang0ZBCSM18}, the need for adaptability to diverse logical paradigms \citep{DBLP:conf/iclr/Saparov023,DBLP:journals/corr/abs-2104-06598,DBLP:journals/tmlr/SrivastavaRRSAF23} and the ability for more rigorous task formalization, decomposition and exploration \citep{DBLP:conf/nips/HendrycksBKABTS21,glazer2024frontiermath}. In this thesis, we detect and mitigate reasoning inconsistencies in deep learning models that operate over natural language and knowledge graphs.

\subsubsection{Complex Multi-hop Question Answering}

Complex multi-hop question-answering tasks require models to reason over ambiguously worded information distributed across various contexts, sources, and commonsense or logical implications. Unlike single-hop tasks, where a direct relationship exists between the query and the answer, multi-hop reasoning involves intermediate steps, where the output of one step serves as input for the next. Examples include connecting facts across sentences, documents, or knowledge graph entities to arrive at a final answer. For instance, consider a question like, "Which author wrote the book that inspired the movie 'Blade Runner'?" To answer this, a model must connect multiple pieces of information: identifying that Blade Runner was inspired by the book \emph{"Do Androids Dream of Electric Sheep?"} and then recognizing that the book's author is Philip K. Dick. Such tasks demand robust semantic understanding, logical consistency, and precise chaining of inferences. urrent benchmarks, such as HotpotQA \citep{DBLP:conf/emnlp/Yang0ZBCSM18} and ComplexWebQuestions \citep{DBLP:conf/naacl/TalmorB18}, aim to evaluate these multi-step reasoning abilities but are insufficient to assess reasoning faithfulness. Models might produce correct answers without following valid reasoning paths even if they produce tokens that seem like correct intermediate justificaitions.

\subsubsection{Complex Logical Query Answering over Knowledge Graphs}

A Knowledge Graph (KG) is a knowledge base representing the relationships between entities in a relational graph structure.
The flexibility of this knowledge representation formalism allows KGs to be widely used in various domains.
A Knowledge Graph $\mathcal{G} \subseteq \mathcal{E} \times \mathcal{R} \times \mathcal{E}$ can be defined as a set of subject-predicate-object $\langle s, p, o \rangle$ triples, where each triple encodes a relationship of type $p \in \mathcal{R}$ between the subject $s \in \mathcal{E}$ and the object $o \in \mathcal{E}$ of the triple, where $\mathcal{E}$ and $\mathcal{R}$ denote the set of all entities and relation types, respectively.
A Knowledge Graph can be represented as a First-Order Logic Knowledge Base, where each triple $\langle s, p, o \rangle$ denotes an atomic formula $p(s, o)$, with $p \in \mathcal{R}$ a binary predicate and $s, o \in \mathcal{E}$ its arguments.
We are concerned with answering logical queries over incomplete knowledge graphs.
We consider queries that use existential quantification ($\exists$) and conjunction ($\land$) operations.
Furthermore, we include disjunctions ($\lor$) and atomic negations ($\neg$).

Consider the question ``\emph{Which people are German and produced the music for the film Constantine?}''.
It can be formalised as a complex query $\mathcal{Q} \equiv\ ?T: \text{country}(\text{Germany}, T) \land \text{producerOf}(\text{Constantine}, T)$, where \emph{Germany} and \emph{Constantine} are anchor nodes, and $T$ is the target of the query.
The answer $[ \mathcal{Q} ]$ corresponds to all the entities in the knowledge graph that are German composers for the film Constantine. We propose a novel method for reasoning over knowledge graphs introduced in \citep{DBLP:conf/nips/ArakelyanMDCA23}.

\section{Scientific Contributions}
\label{sec:scientific_contrib}
\subsection{Reasoning Inconsistencies from Internal Processes}

\subsubsection{Paper 1: Semantic Sensitivities and Inconsistent Predictions: Measuring the Fragility of NLI Models}

\begin{figure}[t!]
    \centering
    \includegraphics[width=\textwidth]{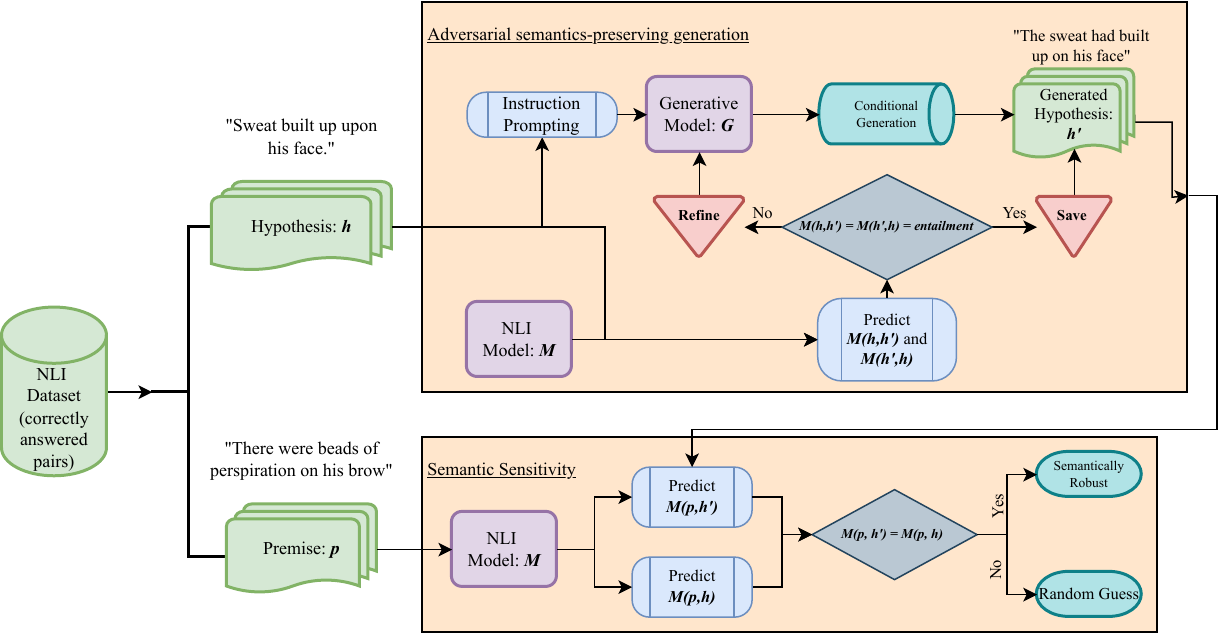}
    \caption{The proposed framework is comprised of two components. (i) a module for generating semantics-preserving surface-form hypothesis variations and (ii) using the generated surface for measuring semantic sensitivity and predictive inconsistency.}
    \label{fig:framework_sensit}
\end{figure}

Recent studies of the emergent capabilities of transformer-based Natural Language Understanding (NLU) models have indicated that they have an understanding of lexical and compositional semantics.
I provide evidence that suggests these claims should be taken with a grain of salt: finding that state-of-the-art Natural Language Inference (NLI) models are sensitive towards minor semantics preserving surface-form variations, which lead to sizable inconsistent model decisions during inference. 
This behavior diverges from a genuine and robust comprehension of compositional semantics. Notably, it does not explicitly emerge when evaluating model accuracy on standard benchmarks or during probing for syntactic, monotonic, and logically robust reasoning.
To address this, I propose a novel framework, illustrated in \cref{fig:framework_sensit}, for quantifying semantic sensitivity. This framework evaluates NLI models on adversarially generated examples containing minor semantics-preserving surface-form variations. These adversarial examples are created using conditional text generation, with the explicit condition that the NLI model should predict the relationship between the original and adversarial inputs as a symmetric equivalence entailment.

I systematically examine the effects of this phenomenon across NLI models in both \emph{in-domain} and \emph{out-of-domain} settings. Experimental results reveal that semantic sensitivity leads to performance degradations of $12.92\%$ and $23.71\%$ on average for \emph{in-domain} and \emph{out-of-domain} settings, respectively. Furthermore, through ablation studies, I analyze this phenomenon across various models, datasets, and inference variations, demonstrating that semantic sensitivity can cause significant inconsistencies in model predictions.

\subsubsection{Paper 2: With Great Backbones Comes Great Adversarial Transferability}

\begin{figure*}[t!]
    \centering
    \includegraphics[clip=true,width=\textwidth]{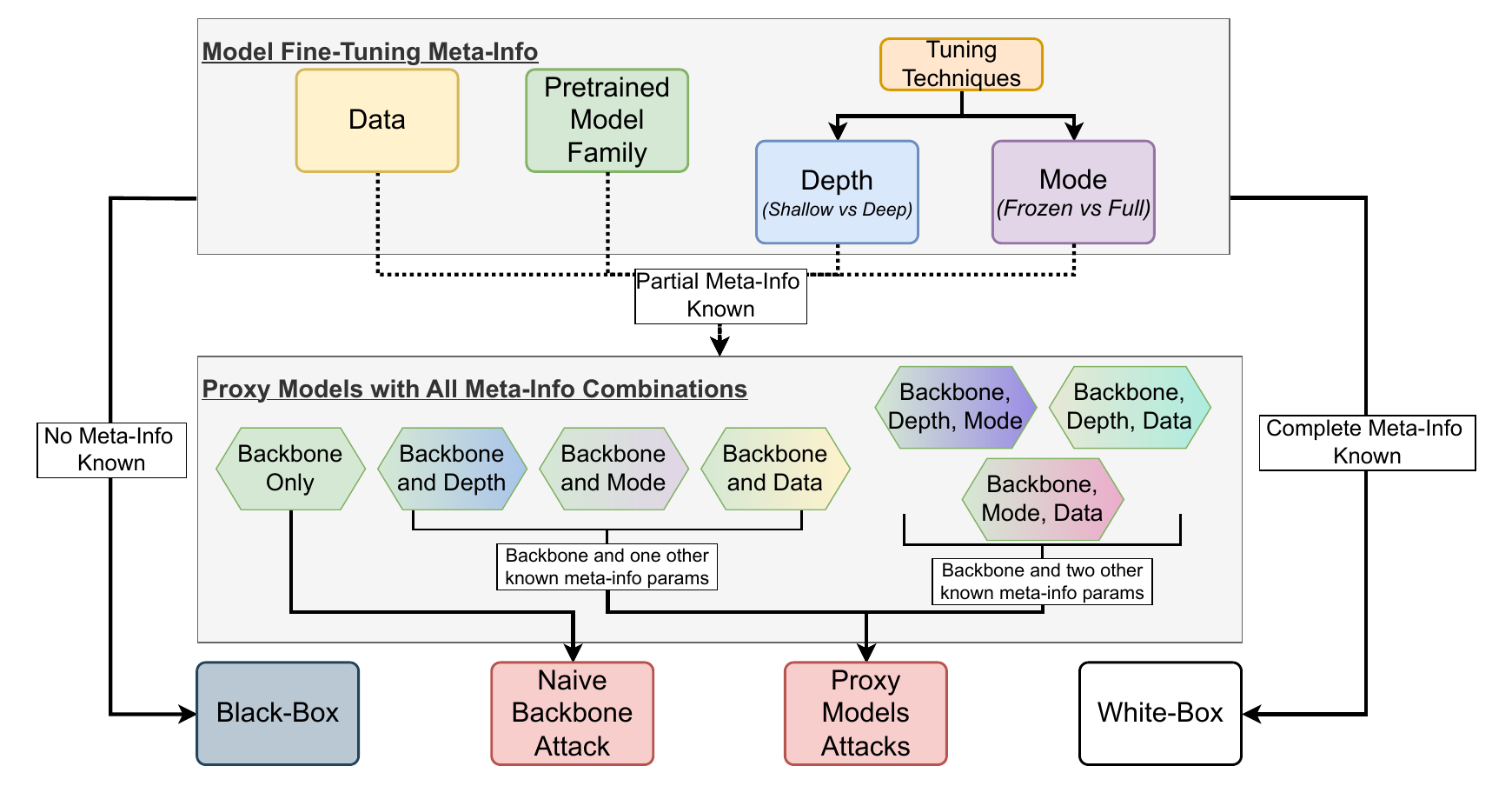}
    \caption{The figure depicts all of the settings used to evaluate adversarial vulnerabilities given different information of the target model construction. From left to right, I simulate exhaustive varying combinations of meta-information available about the target model during adversarial attack construction. All of the created proxy models are used separately to assess adversarial transferability.}
    \label{fig:framework_advers}
\end{figure*}

Advancements in self-supervised learning (SSL) for machine vision have enhanced representation robustness and model performance, leading to the emergence of publicly shared pre-trained backbones, such as \emph{ResNet} and \emph{ViT} models tuned with SSL methods like \emph{SimCLR}. Due to the computational and data demands of pre-training, the utilization of such backbones becomes a strenuous necessity. However, employing such backbones may imply adhering to the existing vulnerabilities towards adversarial attacks.
Prior research on adversarial robustness typically examines attacks with either full (\emph{white-box}) or no access (\emph{black-box}) to the target model, but the adversarial robustness of models tuned on known pre-trained backbones remains largely unexplored. Furthermore, it is unclear which tuning meta-information is critical for mitigating exploitation risks. In this work, I systematically study the adversarial robustness of models that use such backbones, evaluating $20000$ combinations of tuning meta-information, including fine-tuning techniques, backbone families, datasets, and attack types, as seen in \cref{fig:framework_advers}. 
 To uncover and exploit potential vulnerabilities, I propose using proxy (surrogate) models to transfer adversarial attacks, fine-tuning these proxies with various tuning variations to simulate different levels of knowledge about the target. Our findings show that proxy-based attacks can reach close performance to strong \emph{black-box} methods with sizable budgets and closing to \emph{white-box} methods, exposing vulnerabilities even with minimal tuning knowledge. Additionally, we introduce a naive "backbone attack", leveraging only the shared backbone to create adversarial samples, demonstrating an efficacy surpassing \emph{black-box} and close to \emph{white-box} attacks and exposing critical risks in model-sharing practices. Finally, our ablations reveal how increasing tuning meta-information impacts attack transferability, measuring each meta-information combination.

\subsection{Reasoning Inconsistencies from Data}

\subsubsection{Paper 3: Topic-Guided Sampling For Data-Efficient Multi-Domain Stance Detection}

\begin{figure}[!t]
\centering
\includegraphics[width=\columnwidth]{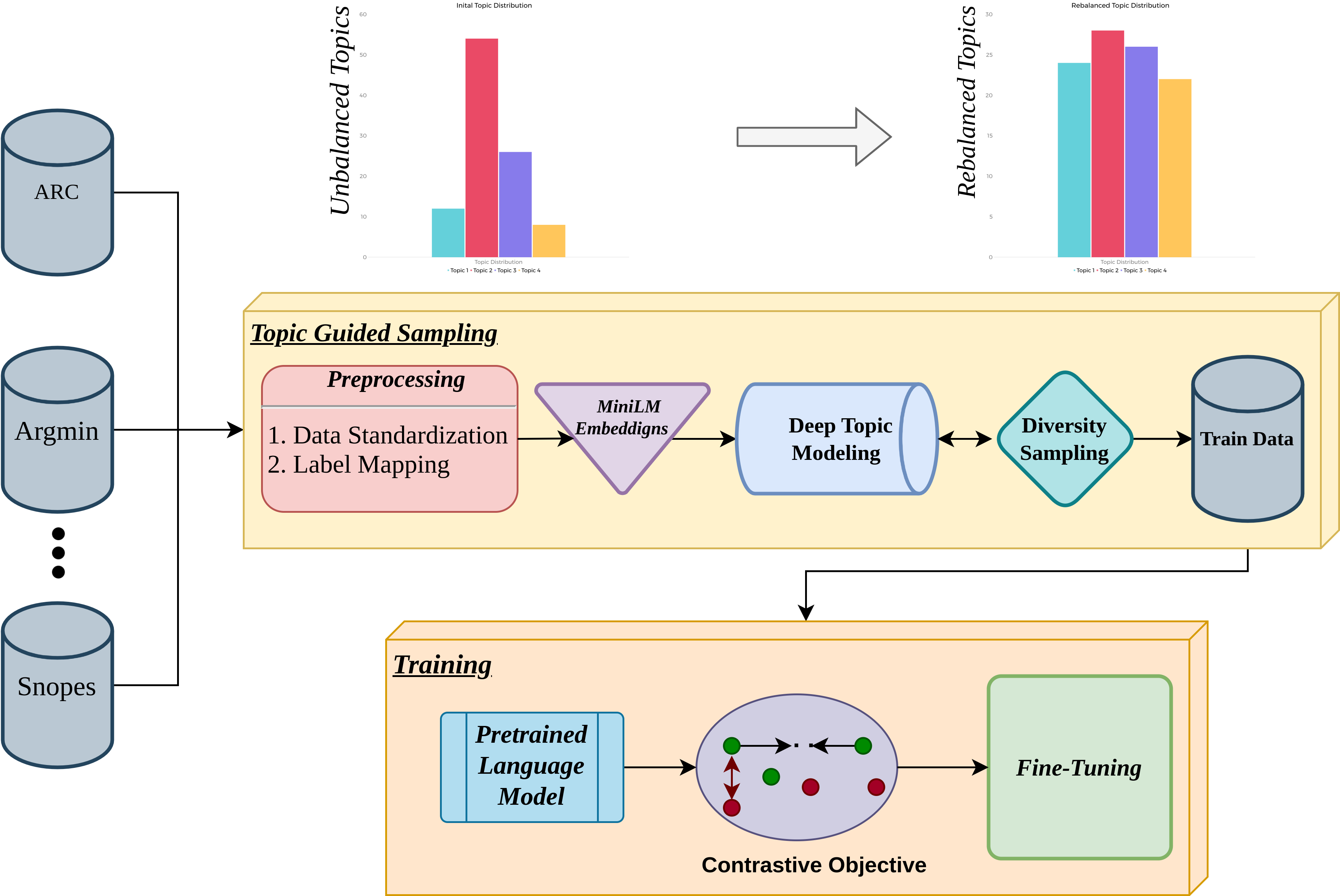}
\caption{The two components of TESTED: Topic Guided Sampling (top) and training with contrastive objective (bottom).}
\label{fig:framework_TESTED}
\end{figure}

Stance Detection is concerned with identifying the attitudes expressed by an author towards a target of interest. This task spans a variety of domains ranging from social media opinion identification to detecting the stance for a legal claim. However, the framing of the task varies within these domains, in terms of the data collection protocol, the label dictionary and the number of available annotations. Furthermore, these stance annotations are significantly imbalanced on a per-topic and inter-topic basis. These make multi-domain stance detection a challenging task, requiring standardization and domain adaptation. To overcome this challenge, I propose \textbf{T}opic \textbf{E}fficient \textbf{St}anc\textbf{E} \textbf{D}etection	(TESTED), seen in \cref{fig:framework_TESTED}, consisting of a topic-guided diversity sampling technique and a contrastive objective that is used for fine-tuning a stance classifier. I evaluate the method on an existing benchmark of $16$ datasets with in-domain, i.e. all topics seen and out-of-domain, i.e. unseen topics, experiments. The results show that our method outperforms the state-of-the-art with an average of $3.5$ F1 points increase in-domain, and is more generalizable with an averaged increase of $10.2$ F1 on out-of-domain evaluation while using $\leq10\%$ of the training data. I show that our sampling technique mitigates both inter- and per-topic class imbalances. Finally, our analysis demonstrates that the contrastive learning objective allows the model a more pronounced segmentation of samples with varying labels.

\subsubsection{Paper 4: SynDARin: Synthesising Datasets for Automated Reasoning in Low-Resource Languages}

\begin{figure}[t!]
    \centering
    \includegraphics[trim=0.0cm 0.0cm 0.0cm 0.0cm,clip=true,width=\textwidth]{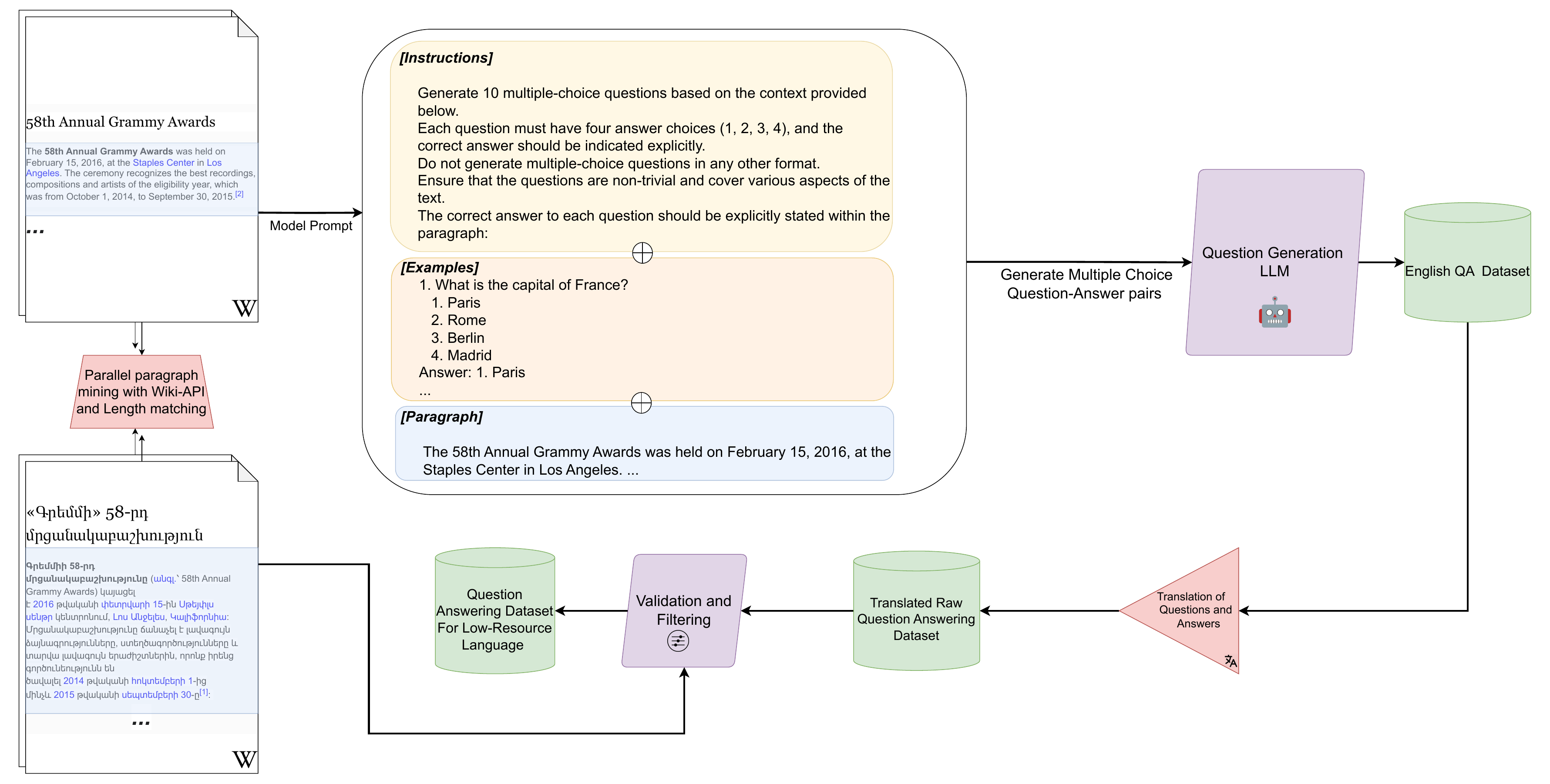}
    \caption{The proposed framework is comprised of three components: (i) a module for mining parallel paragraphs using wiki-API and length matching; (ii) generating a synthetic question-answering dataset with an LLM using the mined English paragraphs; (iii) translating the question-answer pairs and Filtering/Validating them for obtaining a high-quality synthetic QA dataset in the low-resource language.}
    \label{fig:framework_syndarin}
\end{figure}

Question Answering (QA) datasets have been instrumental in developing and evaluating Large Language Model (LLM) capabilities.
However, such datasets are scarce for languages other than English due to the cost and difficulties of collection and manual annotation.
This means that producing novel models and measuring the performance of multilingual LLMs in low-resource languages is challenging. 
To mitigate this, I propose \textbf{S}yn\textbf{DAR}in, a method for generating and validating QA datasets for low-resource languages, seen in \cref{fig:framework_syndarin}.
I utilize parallel content mining to obtain \emph{human-curated} paragraphs between English and the target language.
I use the English data as context to \emph{generate} synthetic multiple-choice (MC) question-answer pairs, which are automatically translated and further validated for quality.
Combining these with their designated non-English \emph{human-curated} paragraphs from the final QA dataset. 
The method allows to maintain content quality, reduces the likelihood of factual errors, and circumvents the need for costly annotation.
To test the method, I created a QA dataset with $1.2$K samples for the Armenian language.
The human evaluation shows that $98\%$ of the generated English data maintains quality and diversity in the question types and topics, while the translation validation pipeline can filter out $\sim70\%$ of data of poor quality.
I use the dataset to benchmark state-of-the-art LLMs, showing their inability to achieve human accuracy with some model performances closer to random chance.
This shows that the generated dataset is non-trivial and can be used to evaluate reasoning capabilities in low-resource language.

\subsection{Reasoning Inconsistencies from Task Complexity}
\subsubsection{Paper 5: Adapting Neural Link Predictors for Data-Efficient Complex Query Answering}

\begin{figure}[!t]
    \centering
    \includegraphics[width=\textwidth]{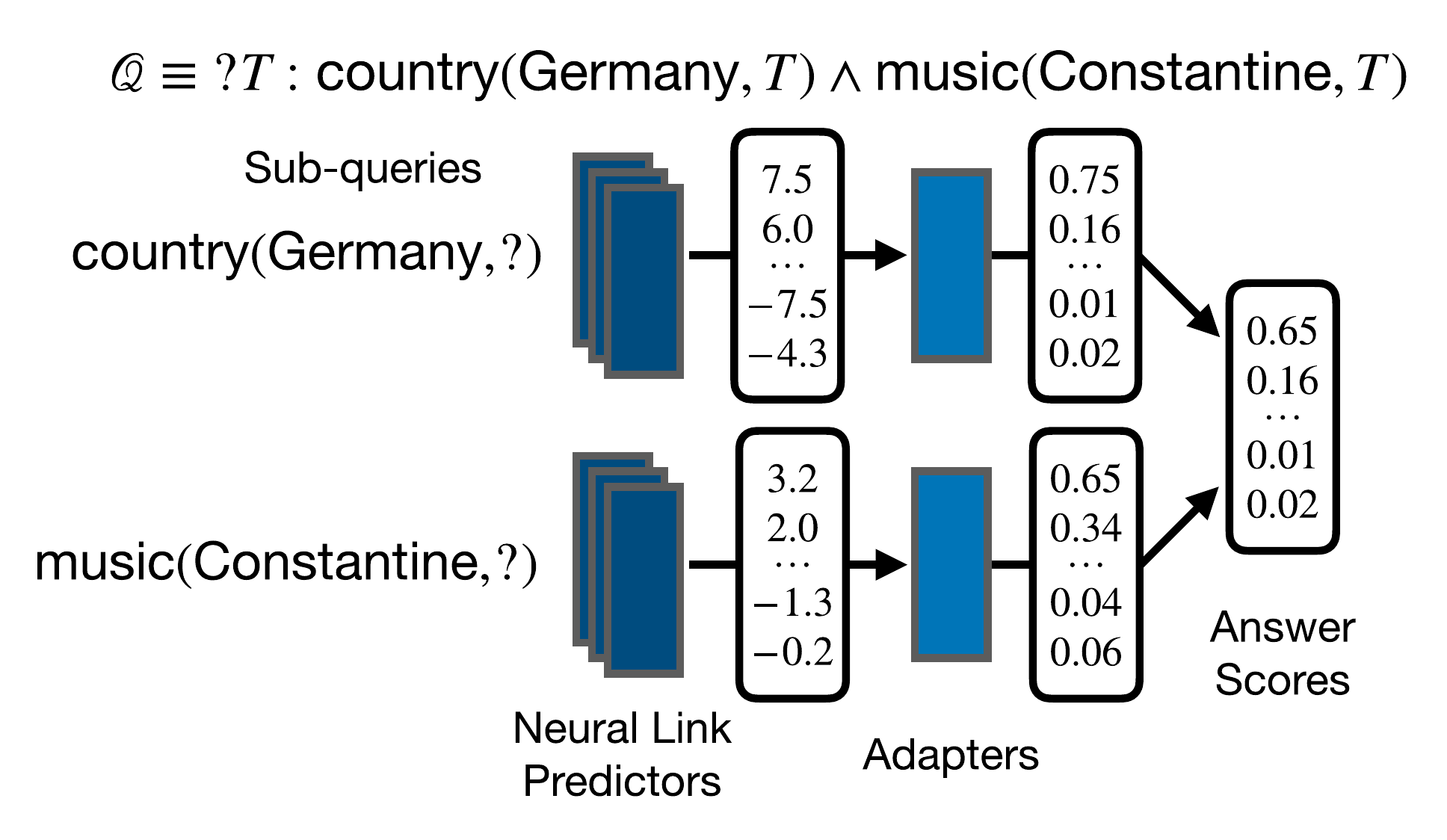}
\caption{Given a complex query $\mathcal{Q}$, \cqda adapts the neural link prediction scores for the sub-queries to improve the interactions between them.
} 
\label{fig:frmaework_cqda}
\end{figure}

Answering complex queries on incomplete knowledge graphs is a challenging task where a model needs to answer complex logical queries in the presence of missing knowledge.
Prior work in the literature has proposed to address this problem by designing architectures trained end-to-end for the complex query answering task with a reasoning process that is hard to interpret while requiring data and resource-intensive training. 
Other lines of research have proposed re-using simple neural link predictors to answer complex queries, reducing the amount of training data by orders of magnitude while providing interpretable answers.
The neural link predictor used in such approaches is not explicitly optimised for the complex query answering task, implying that its scores are not calibrated to interact together.
We propose to address these problems via \cqda, a parameter-efficient score \emph{adaptation} model optimised to re-calibrate neural link prediction scores for the complex query answering task.
While the neural link predictor is frozen, the adaptation component -- which only increases the number of model parameters by $0.03\%$ -- is trained on the downstream complex query answering task.
Furthermore, the calibration component enables us to support reasoning over queries that include atomic negations, which was previously impossible with link predictors.
In our experiments, \cqda produces significantly more accurate results than current state-of-the-art methods, improving from $34.4$ to $35.1$ Mean Reciprocal Rank values averaged across all datasets and query types while using $\leq 30\%$ of the available training query types.
We further show that \cqda is data-efficient, achieving competitive results with only $1\%$ of the complex training queries, and robust in out-of-domain evaluations.

\subsubsection{Paper 6: FLARE: Faithful Logic-Aided Reasoning and Exploration}

\begin{figure}[!t]
    \centering
    \includegraphics[width=\textwidth, clip]{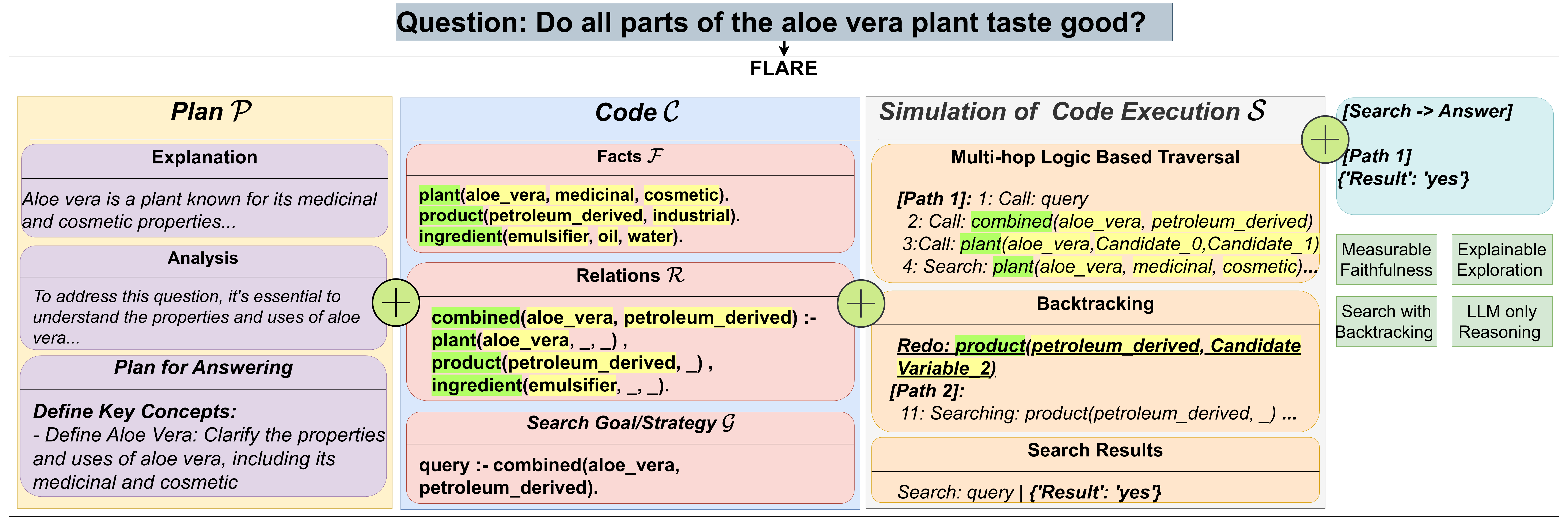}
    \caption{A depiction of the \emph{plan}, \emph{code} and simulated \emph{search} in {\flare}. Each module is generated separately and iteratively, allowing us to obtain the final answer. The green and yellow highlighted text shows the overlap between the facts and the relations between the code and the simulated search.}
    \label{fig:flare}
\end{figure}
Modern Question Answering (QA) and Reasoning approaches based on Large Language Models (LLMs) commonly use prompting techniques, such as Chain-of-Thought (CoT), assuming the resulting generation will have a more granular exploration and reasoning over the question space and scope.
However, such methods struggle with generating outputs that are faithful to the intermediate chain of reasoning produced by the model.
On the other end of the spectrum, neuro-symbolic methods such as Faithful CoT (F-CoT) and Logic-LM propose to combine LLMs with external symbolic solvers.
While such approaches boast a high degree of faithfulness, they usually require a model trained for code generation and struggle with tasks that are ambiguous or hard to formalise strictly.
I introduce \textbf{F}aithful \textbf{L}ogic-\textbf{A}ided \textbf{R}easoning and \textbf{E}xploration (\textbf{\flare}), a novel interpretable approach for traversing the problem space using task decompositions, seen in \cref{fig:flare}.
I use the LLM to plan a solution, formalise the query into facts and predicates, which form the problem space, using a logic programming code and simulate that code execution using an exhaustive multi-hop search over the defined space.
%
%
Our method allows us to compute the faithfulness of the reasoning process w.r.t. the generated code and explicitly trace the steps of the multi-hop search without relying on external solvers.  Our methods achieve SOTA results on $\mathbf{7}$ out of $\mathbf{9}$ diverse reasoning benchmarks.
%
%
I also show that model faithfulness positively correlates with overall performance and further demonstrate that {\textbf{\flare}} allows pinpointing the decisive factors sufficient for and leading to the correct answer with optimal reasoning during the multi-hop search. 
Our findings reveal that successful traces exhibit, on average, a $18.1\%$ increase in unique emergent facts, a $8.6\%$ higher overlap between code-defined and execution-trace relations, and a $3.6\%$ reduction in unused code relations.

\section{Discussion and Future Work}
\label{sec:discuss}

\begin{table}[!t]
\renewcommand{\arraystretch}{1.5} 
\setlength{\tabcolsep}{20pt}      
\resizebox{\textwidth}{!}{%
\begin{tabular}{lccc|ccc|ccc}
\toprule
\multicolumn{1}{c}{} & \multicolumn{3}{c}{\textbf{Internal Procedures}} & \multicolumn{3}{c}{\textbf{Data Imbalances}} & \multicolumn{3}{c}{\textbf{Complex Reasoning}} \\ \midrule
\multicolumn{1}{c}{} & \textbf{M}   & \textbf{D}   & \textbf{A}   & \textbf{M}   & \textbf{D}   & \textbf{A}   & \textbf{M}   & \textbf{D}   & \textbf{A}   \\ \midrule
\rowcolor{gray!10} 1. \citet{DBLP:conf/eacl/ArakelyanLA24}       & \cellcolor{green!25}$\checkmark$ & \cellcolor{green!25}$\checkmark$ & \cellcolor{green!25}$\checkmark$ & \cellcolor{white} & \cellcolor{white} & \cellcolor{white} & \cellcolor{white} & \cellcolor{white} & \cellcolor{white} \\
\rowcolor{gray!5}  2. \citet{arakelyan2024greatbackbones}       & \cellcolor{green!25}$\checkmark$ & \cellcolor{white} & \cellcolor{green!25}$\checkmark$ & \cellcolor{white} & \cellcolor{white} & \cellcolor{white} & \cellcolor{white} & \cellcolor{white} & \cellcolor{white} \\
\rowcolor{gray!10} 3. \citet{DBLP:conf/acl/ArakelyanAA23}       & \cellcolor{white} & \cellcolor{white} & \cellcolor{white} & \cellcolor{green!25}$\checkmark$ & \cellcolor{white} & \cellcolor{green!25}$\checkmark$ & \cellcolor{white} & \cellcolor{white} & \cellcolor{white} \\
\rowcolor{gray!5}  4. \citet{DBLP:journals/coling/abs-2406-14425}       & \cellcolor{white} & \cellcolor{white} & \cellcolor{white} & \cellcolor{green!25}$\checkmark$ & \cellcolor{green!25}$\checkmark$ & \cellcolor{green!25}$\checkmark$ & \cellcolor{white} & \cellcolor{green!25}$\checkmark$ & \cellcolor{white} \\
\rowcolor{gray!10} 5. \citet{DBLP:conf/nips/ArakelyanMDCA23}       & \cellcolor{white} & \cellcolor{white} & \cellcolor{white} & \cellcolor{white} & \cellcolor{white} & \cellcolor{white} & \cellcolor{green!25}$\checkmark$ & \cellcolor{white} & \cellcolor{green!25}$\checkmark$ \\
\rowcolor{gray!5}  6. \citet{DBLP:journals/corr/arakelyan-flare}      & \cellcolor{white} & \cellcolor{white} & \cellcolor{white} & \cellcolor{white} & \cellcolor{white} & \cellcolor{white} & \cellcolor{green!25}$\checkmark$ & \cellcolor{white} & \cellcolor{green!25}$\checkmark$ \\
\rowcolor{gray!10} 7. \citet{DBLP:series/faia/CochezAABD0MNR23}       & \cellcolor{white} & \cellcolor{white} & \cellcolor{white} & \cellcolor{white} & \cellcolor{white} & \cellcolor{white} & \cellcolor{white} & \cellcolor{white} & \cellcolor{green!25}$\checkmark$ \\ \bottomrule
\end{tabular}%
}
\caption{Contributions of referenced works across three main reasoning inconsistency categories: Internal Procedures, Data Imbalances, and Complex Reasoning Tasks. Each category is further divided into three contribution subsections: Method (\textbf{M}), Datasets (\textbf{D}), and Analysis Framework For Reasoning (\textbf{A}). A checkmark (\checkmark) indicates the contribution's relevance to the respective area. The table highlights the distribution of efforts, showcasing where each work has made significant contributions.}
\label{tab:contributions}
\end{table}

The research publications within this thesis contribute to the field of deep learning by expanding our understanding of reasoning inconsistencies and suggesting novel ways to mitigate them across various domains such as NLP, image processing and reasoning over KGs. In particular, we identify and analyse three potential causes for inconsistency: internal processes, data imbalances, and task complexity. The contributions can be segmented into novel methods, datasets and analysis frameworks for detecting, measuring and mitigating reasoning inconsistencies in deep learning models, as seen in \cref{tab:contributions}. 

\subsection{Measuring and Mitigating Reasoning Inconsistencies}

The thesis establishes reasoning inconsistencies related to internal representations and transitions that deep learning models learn. The papers suggest novel adversarial setups that directly expose the susceptibility of deep learning models to shallow heuristics, semantic misalignments, and adversarial vulnerabilities. For instance, the semantic sensitivity of NLI models \citep[\textit{Paper 1}]{DBLP:conf/eacl/ArakelyanLA24} highlights a critical limitation in their robustness and generalization across diverse settings. Adversarial transferability studies \citep[\textit{Paper 2}]{arakelyan2024greatbackbones} reveal how shared backbones escalate model vulnerabilities, emphasizing the need for more robust fine-tuning techniques and vigilance in the current model-sharing practices. In both of the publications, we provide a framework for directly detecting and measuring the extent of reasoning inconsistencies that numerous deep learning models possess. 

The thesis also includes a new method for data-efficient topic-based sampling \citep[\textit{Paper 3}]{DBLP:conf/acl/ArakelyanAA23}, which allows to circumvent the complications with biased and inconsistent reasoning in deep learning models, arising from dataset imbalances described in \cref{subsubsec:data_imbalances}. We directly measure the impact of mitigating these imbalances with our method on the predictive capabilities of the model, showing a significant performance boost both for in-domain and out-of-domain evaluations. The models trained using our sampling method are also less susceptible to erroneous behaviour that arises because of a dominating overrepresentation of specific topics or semantic features. We also show that an emergent property of the contrastive learning method we propose is that the internal representations of the model become more segmented w.r.t. different topics, thus overall boosting the model's effective capacity.

The thesis also contributes a method for synthetically generating question-answering datasets in low-resource settings \citep[\textit{Paper 4}]{DBLP:journals/coling/abs-2406-14425} with a mechanism for automated sample verification and diversification. I constructed a human evaluation for each aspect of the method and the final output and showed that the method works for the Armenian language, which has almost no available machine learning resources for training and evaluation. Some would argue that allegorically, the biggest data imbalance is the availability of no data for even evaluating the reasoning capabilities of the deep learning models, which this publication addresses. Maybe some would be wrong in this assessment, but that, my dear reader, is a discussion for a different thesis.

To tackle the reasoning inconsistencies that emerge because of task complexity and the limited \emph{effective capacity}/expressivity of the model, I dive into two complex query answering tasks over knowledge graphs and natural language. I introduce a new approach for handling complex queries over knowledge graphs \citep[\textit{Paper 5}]{DBLP:conf/nips/ArakelyanMDCA23}, incorporating learnable adaptation layers that optimize intermediate answers and representations during the reasoning process. This enhances the model's ability to generalize to unseen query types, increases its effective capacity, and maintains data efficiency as an added advantage. This allows for circumventing prior limitations present because of task complexity and adds an ingrained tool for verifying and adapting the results of intermediate reasoning answers.

The other aspects I explore in this thesis are the predictive inconsistencies and suboptimal explorations that LLMs have when reasoning in natural language. While LLMs exhibit strong performance on numerous language reasoning tasks, they often lack a structured and faithful inference mechanism when answering complex queries, which does not allow the model to formalise and explore the problem efficiently. Moreover, many prompting paradigms in natural language lack explicit verifiability because the text is inherently freeform. We create a novel method to mitigate these issues. \textbf{F}aithful \textbf{L}ogic-\textbf{A}ided \textbf{R}easoning and \textbf{E}xploration (\textbf{\flare}) \citep[\textit{Paper 6}]{DBLP:journals/corr/arakelyan-flare}, is a novel interpretable approach for traversing the problem space using task decompositions. The method enhances reasoning interpretability and faithfulness by combining task decomposition, Prolog-like logical formalization, and LLM simulated search. Critically, {\flare} addresses task complexity by enhancing the reasoning capacity of LLMs without solely relying on deterministic algorithms and allows for deeper model explorations within the problem. It supports multi-hop reasoning, task decomposition, and logical consistency verification. The results highlight FLARE's state-of-the-art performance on several datasets, achieving significant improvements in reasoning faithfulness and task accuracy.

\subsection{Future Work}





Currently, there is limited work on ingraining LLMs with test time compute capabilities and how it impacts the effective capacity of the model and the verifiability of the suggested reasoning lines. One natural extension for papers \citep[\textit{5 and 6}]{DBLP:conf/nips/ArakelyanMDCA23,DBLP:journals/corr/arakelyan-flare} would be able to directly use test time compute mechanisms, such as self-refinement and others for further enriching the predictive capabilities of the models. An intriguing alternative in a similar vein would involve training a reward model using the formalizations and reasoning paths generated by {\flare}. This approach leverages the ability to directly execute the generated code, allowing us to sample Prolog search paths that yield correct answers as positive examples while treating the LLM-generated incorrect traversal simulations as negative examples. Training this type of model would allow to further tune other LLMs with a differentiable oracle that is capable of assessing the correctness and completeness of the search paths. This would also allow some notion of verifiability to be ingrained into the generated search paths. A direct extension of this can be using a strict logical decomposition and iterative multi-hop reasoning for the query, similar to the approach in \citep[\textit{5}]{DBLP:conf/nips/ArakelyanMDCA23}. This would add the capability to adaptably search over intermediate answers and prune unlikely search directions.

\part{Reasoning Inconsistencies from Internal Processes}
\chapter{Semantic Sensitivities and Inconsistent Predictions: Measuring the Fragility of NLI Models}
\label{chap:semantic_sense}
\section{Introduction}
Transformer-based \citep{vaswani2017attention} Language Models (LMs) have shown solid performance across various NLU tasks \citep{wang2018glue, wang2019superglue}. These advances have led to suggestions regarding the emergent capabilities of the models in terms of syntactic \citep{sinha2020unnatural,hewitt2019structural, jawahar2019does, warstadt2020can}, logic \citep{wei2022emergent,wei2022chain} and semantic \citep{kojima2022large,dasgupta2022language} understanding. However, we present novel evidence that indicates that these models are prone to inconsistent predictions induced by inherent susceptibility towards semantic sensitivities.

To probe the models for these discrepancies, we formalise \emph{semantic comprehension} as the ability to distinguish logical relations within sentences through identifying compositional semantics \citep{jacobson2014compositional,carnap1959introduction}. This means that negligible semantic variations should not impact the inherent relations implied between the texts, e.g. \emph{``There were beads of perspiration on his brow.''} entails both \emph{``Sweat built up upon his face.''} and the slight variation \emph{``The sweat had built up on his face.''}
Authentic comprehension of semantics does allow for such understanding through discovering semantic structures and the inherent relations induced by them \citep{cicourel1991semantics,schiffer1986compositional,rommers2013context}. This means that analysing the emergent semantic understanding within a model should minimally involve testing for sensitivity towards semantics-preserving surface-form variations.

\begin{figure*}[!t]
    \centering
    \includegraphics[width=\textwidth]{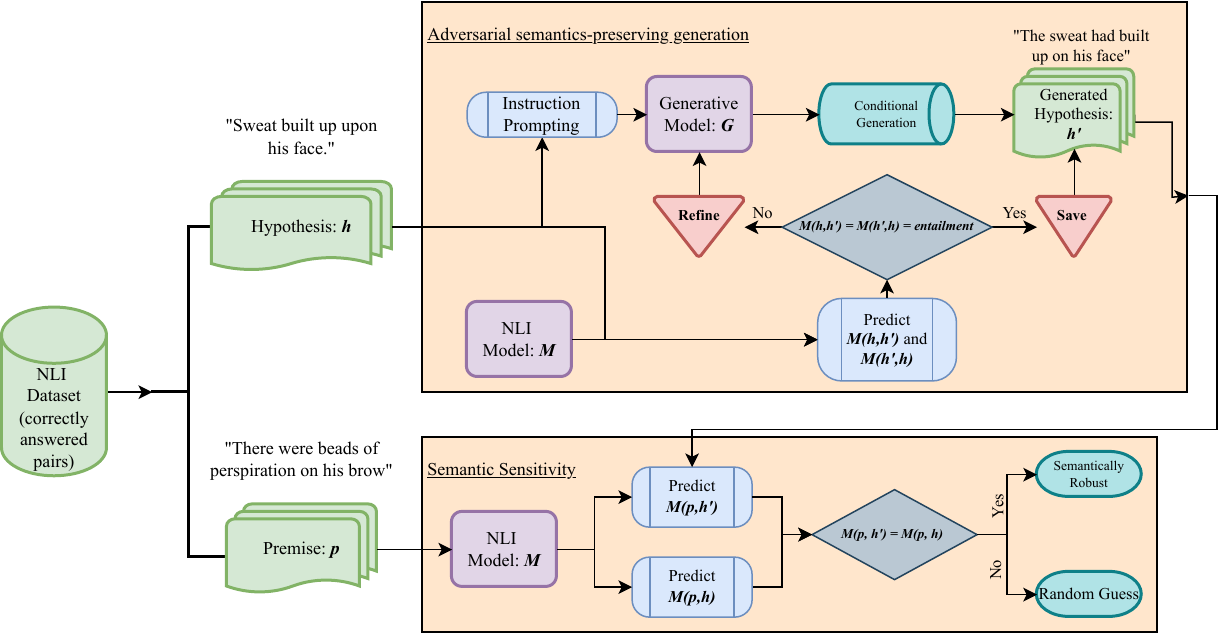}
    \caption{The proposed framework is comprised of two components. (i) a module for generating semantics-preserving surface-form hypothesis variations and (ii) using the generated surface for measuring semantic sensitivity and predictive inconsistency.}
    \label{fig:framework}
\end{figure*}

We particularly focus on the task of textual entailment \citep{dagan2005pascal}, otherwise referred to as Natural Language Inference \citep[NLI]{ bowman2015large}, which has been widely used to probe how well the models understand language \citep{condoravdi2003entailment, williams2017broad, nie2019adversarial}. This is a pairwise input task, where given a premise $p$ and a hypothesis $h$, the objective is to predict if the premise \emph{entails, contradicts} or is \emph{neutral} towards the hypothesis.

We propose a framework for testing semantic sensitivity within transformer-based models trained for NLI, by creating semantics-preserving surface-form variations of the hypothesis (see \autoref{fig:framework}).
These variations are created using conditional generation with Large Language Models (LLMs). We show that proposed candidates do not alter the core meaning or the truth value compared to the original statement. The original and generated sentences maintain denotative equivalence, where two sentences or phrases might be interpreted as having the same truth value or factual content but may carry minor variations of nuances or connotations.
To ensure that the relations are preserved within the candidates during conditional generation, we assert that the NLI model predicts the original and generated hypothesis to symmetrically entail each other. This indicates that the model perceives both the generated and original hypothesis as equivalent. 
After introducing these variations, we evaluate the NLI model by replacing the original hypothesis with the generated candidates. As the candidates are indicated to be equivalent by the same NLI model, this evaluation will indicate whether the model can recover the existent relation between the premise hypothesis pair in the presence of minor semantic-preserving noise. 
We use the samples where the model identifies the existing relation correctly from the original premise hypothesis pair. This ensures that assessing for semantic sensitivity would not be hindered by the discrepancies in model performance.

We systematically study the semantic sensitivity across transformers that achieve state-of-the-art or similar results when trained on NLI datasets, namely RoBERTa \citep{liu2019roberta}, BART \citep{lewis2019bart}, DeBERTa \citep{he2020deberta} and DistilBart \citep{sanh2019distilbert, lewis2019bart} with different parametrizations. To measure the effect of the phenomenon on the inconsistency of the predictions, we use three popular English datasets - MultiNLI \citep[MNLI]{williams2017broad}, SNLI \citep{bowman2015large} and ANLI \citep{nie2019adversarial}. The models are fine-tuned using MNLI, which we choose for \emph{in-domain} testing, as it covers a wide range of topics and is frequently used for zero-shot and few-shot textual classification \citep{yin2019benchmarking}. We use the same models for \emph{out-of-domain} evaluation across the other NLI datasets.

Our contributions are as follows: (i) we propose a novel framework for assessing semantic sensitivity within transformer-based language models (ii) we systematically study the influence of this phenomenon on inconsistent predictions across various transformer variants (iii) we show that the effect is persistent and pronounced across both \emph{in-} and \emph{out-of-domain} evaluations (iv) we further complete ablations to assess the severity of the inconsistent predictions caused by semantic sensitivity.

\section{Related Work}

Semantic comprehension is considered a fundamental building block for language understanding \citep{allen1995natural}. Although attempts have been made to probe language models in terms of compositional semantic capabilities, the conclusions regarding their emergence remain to be discussed.

\subsection{Models appear to understand semantics}

Recently a wide suite of tasks has been proposed for testing models for language understanding \citep{wang2019superglue,zellers2018swag, ribeiro2020beyond} with the credence that a model with strong performance should be able to utilise semantic relations when completing the tasks. In light of these, it has been shown that transformer-based language models can be directly trained \citep{zhang2020semantics, rosset2020knowledge} to utilise semantic structure to gain distributional information within the task. Specifically, NLI models have also been shown to be capable of pragmatic inferences \citep{jeretic2020natural} with a perception of implicature \citep{grice1975logic} and presupposition \citep{stalnaker1977pragmatic,grice1975logic}.

\subsection{Models struggle with semantics}

Directly probing for a specific aspect of semantic understanding has shown that transformer-based language models tend to struggle with semantics \citep{belinkov2022probing}. It has been indicated that pretraining the language models does not exploit semantic information for entity labeling and coreference resolution \citep{liu2019linguistic}. Furthermore, transformer attention heads only minimally capture semantic relations \citep{kovaleva2019revealing} from FrameNet \citep{baker1998berkeley}. Studies have also shown that NLI models, in particular, tend to struggle with lexical variations, including word replacements \citep{glockner-etal-2018-breaking, ivan2018behavior, geiger2020neural}, and sequence permutations \citep{sinha2021unnatural}.

\begin{table*}[t]
\centering
\begin{tabular}{@{}c|cccccc@{}}
\toprule
         & \textbf{bart-l} & \textbf{roberta-l} & \textbf{distilbart} & \textbf{deberta-b} & \textbf{deberta-l} & \textbf{deberta-xl} \\ \midrule
MNLI$_{(n=10000)}$    & 90.10\%         & 90.56\%            & 87.17\%             & 88.77\%            & 91.32\%            & 91.44\%             \\ \midrule
SNLI$_{(n=10000)}$    & 87.55\%         & 86.44\%            & 84.37\%             & 84.39\%            & 88.87\%            & 88.54\%             \\
ANLI\_r1$_{(n=1000)}$& 46.20\%         & 46.40\%            & 41.40\%             & 35.10\%            & 49.70\%            & 53.00\%             \\
ANLI\_r2$_{(n=1000)}$& 31.60\%         & 27.00\%            & 32.80\%             & 29.80\%            & 32.70\%            & 35.40\%             \\
ANLI\_r3$_{(n=1200)}$& 33.08\%         & 26.75\%            & 32.75\%             & 30.50\%            & 35.92\%            & 38.75\%             \\ \bottomrule
\end{tabular}
\caption{The original accuracy on testing/dev sets for various transformers (b-base, l-large, xl-extra large) on \emph{in-domain} MNLI experiments and zero-shot transfers to \emph{out-of-domain} SNLI and ANLI. The number near the dataset name designates the exact amount of original samples in the testing set. }
\label{tab:nli_eval}
\end{table*}

\subsection{Sensitivity in NLI models}

Probing NLI models for language understanding has been a hallmark testing ground for measuring their emerging capabilities \citep{naik2018stress, wang2015learning, williams2017broad}. A wide range of tests indicates that models trained for NLI are prone to struggling with syntax and linguistic phenomena \citep{dasgupta2018evaluating, naik-etal-2018-stress, an-etal-2019-representation, ravichander-etal-2019-equate, jeretic-etal-2020-natural}. It has also been shown that NLI models heavily rely on lexical overlaps \citep{ivan2018behavior, mccoy-etal-2019-right, naik-etal-2018-stress} and are susceptible to over-attending to particular words for prediction \citep{gururangan-etal-2018-annotation, clark-etal-2019-bert}.
Our line of work is associated with evaluating NLI models for monotonicity reasoning \citep{yanaka2019can} and sensitivity towards specific semantic phenomenon \citep{richardson2020probing}, such as boolean coordination, quantification, etc. However, we systematically test NLI models for their compositional semantic abilities and measuring the degree of inconsistence of their predictions influenced by the phenomenon.

\section{Methodology}

We aim to create a framework for assessing semantic sensitivity within NLI models and measure its impact on the inconsistence of model predictions. The first part of the pipeline we propose is an adversarial semantics-preserving generation for introducing variations within the original samples. The second part of the pipeline involves assessment using the acquired generations.

\subsection{Semantics Preserving Surface-Form Variations}

We formalise NLI as a pairwise input classification task. Given a dataset of premise hypothesis pairs $\mathcal{D} = {(p_1,h_1), \dots (p_n,h_n)}$, where ${ \forall p_i \in P \And h_i \in H}$ are a set of textual tokens $P,H \subseteq \mathcal{T}$, the goal is to classify the pairs as \emph{entailment, contradiction} or \emph{neutrality}, i.e. $\mathcal{C} = \{ E,C,N\}$. We are also given a pre-trained language model (PLM) $\mathcal{M}$ that is trained for textual entailment.
Before introducing semantic variations, only the samples where model $\mathcal{M}$ predicted the label correctly are filtered, i.e. $D_{correct} = \{ \forall (p_i,h_i)\in \mathcal{D} : \mathcal{M}(p_i,h_i) = \hat{y} = y\}$, where $\hat{y}$ is the prediction and $y$ is the original label. This is completed to ensure that the evaluation of semantic sensitivity is not hindered or inflated by the predictive performance and confidence of the model $\mathcal{M}$. This type of filtering is used when probing for emergent syntactic \citep{sinha2021unnatural}, lexical \citep{jeretic-etal-2020-natural}, and numerical \citep{wallace2019nlp} reasoning capabilities. We can see the original accuracy of NLI models and the number of samples used in the study in \autoref{tab:nli_eval}.

To introduce semantics preserving noise within chosen samples, we complete a two-fold refinement process. We utilise a generative LLM $\mathcal{G}$, which has been fine-tuned on natural language instructions \citep{wei2021finetuned, chung2022scaling}, and prompt it to paraphrase the original hypothesis $h_i$, with the following prompt: \emph{Rephrase the following sentence while preserving its original meaning: <$h_i$>. }
This is not sufficient to produce semantics-preserving variations as generative models are prone to hallucinations \citep{ji2023survey} and not assured to produce an equivalent paraphrase. To ensure that the generation $h_i^{\prime}$ is logically equivalent to the original sample and thus semantics-preserving, we impose the condition that the NLI model should infer the relation between the original and generated hypothesis as a symmetric entailment:
\begin{align}\label{eq:label_change}
    \mathcal{M}(h_i, h_i^{\prime}) = \hat{y}_{\mathcal{C}=E} = \mathcal{M}(h_i^{\prime}, h) 
\end{align}

The bidirectional nature of entailment allows us to claim that sentences are logically equivalent \citep{angell1989deducibility,clark1967general}. We refine the proposed variation candidates using the generator $\mathcal{G}$ until $k$ candidates that satisfy the condition are produced.

\subsection{Human Evaluation of Surface-Form Variations}

To further ensure the validity of this variation generation method, we conduct a human evaluation of the generated samples. We randomly sample $100$ examples of generated and original hypothesis pairs across all datasets and employ two annotators to assess whether the sentences are semantically and logically equivalent within the pair. Our results show that in $99\%$ of the cases, the annotators marked the samples as equivalent with an inter-annotator agreement measure of Cohen's $\kappa = 0.94$. This further shows the reliability of the method for generating semantics-preserving surface form variations. We provide further token overlap level analysis in \autoref{sec:appendix}.



\begin{table*}[t]
\centering
\begin{adjustbox}{max width=\textwidth}
\begin{tabular}{@{}c|cccccc@{}}
\toprule
$r_s/r_r$   & \textbf{bart-large} & \textbf{roberta-large} & \textbf{distilbart} & \textbf{deberta-base} & \textbf{deberta-large} & \textbf{deberta-xlarge} \\ \midrule
MNLI      & 6.64\%/12.35\%                            & 5.71\%/11.56\%                               & 9.20\%/ \bf{16.80}\%                                  & 6.66\%/13.81\%                               & 5.38\%/11.54\%                               & 5.89\%/11.49\%                                \\ \midrule
SNLI      & 10.11\%/15.52\%                           & 8.38\%/14.98\%                               & 15.67\%/\bf{23.68}\%                               & 9.96\%/17.01\%                               & 7.83\%/13.39\%                               & 9.50\%/14.69\%                                 \\
ANLI\_r1  & 31.51\%/42.89\%                           & 28.45\%/35.01\%                              & 31.48\%/\bf{52.30}\%                                & 40.0\%/48.99\%                               & 25.66\%/37.88\%                              & 22.71\%/30.73\%                               \\
ANLI\_r2  & 34.39\%/51.91\%                           & 24.62\%/42.80\%                               & 36.09\%/\bf{57.49}\%                               & 34.92\%/48.47\%                              & 28.44\%/44.04\%                              & 29.46\%/46.46\%                               \\
ANLI\_r3  & 29.11\%/51.39\%                           & 21.88\%/45.00\%                               & 29.26\%/52.42\%                               & 33.88\%/\bf{53.17}\%                              & 24.88\%/44.65\%                              & 23.23\%/42.37\%                               \\ \bottomrule
\end{tabular}
\end{adjustbox}
\caption{The strict and relaxed fooling rates of different transformer models across \emph{in-domain} (MNLI) and \emph{out-of-domain} (SNLI, ANLI) evaluations. On average more than half of the labels change towards their logically contrasting counterpart.}
\label{tab:foolrate1}
\end{table*}

\subsection{Evaluating Semantic Sensitivity}

After obtaining $k$ semantic variations for each hypothesis, we test the semantic sensitivity of the model by replacing the original hypothesis $h_i$ with the candidates $\{h_i^{\prime 1}, \dots h_i^{\prime k} \}$ and making a prediction with the NLI model $\mathcal{M}$. As the proposed variations are logically equivalent to the original, we want to test if the new model prediction would vary compared to the original.

\begin{multline}
\mathcal{R}(p_i, h_i, h_i^{\prime j}, \mathcal{O}) 
    = \begin{cases}
       1, \mathcal{O}(\mathcal{M}(p_i, h_i), \mathcal{M}(p_i, h_i^{\prime j})) = 0 \\
       0, \mathcal{O}(\mathcal{M}(p_i, h_i), \mathcal{M}(p_i, h_i^{\prime j})) = 1
   \end{cases}
\end{multline}

Here $\mathcal{O}: \mathcal{C} \times \mathcal{C} \rightarrow \{0, 1\}$ is a boolean matching operator between the labels predicted with original hypothesis $h_i$ and the surface-form variations $h_i^{\prime j}$.
A change in the label would imply that the model is semantically sensitive and the original correct prediction is inconsistent with the label produced for the semantics preserving surface-form variation.  
A graphical representation can be seen in \autoref{fig:semantic_triangle}.
We use two metrics to measure semantic sensitivity within NLI models, both of which are derivative formulations of a Fooling Rate \citep{moosavi2017universal}, which is used for assessing the success of adversarial attacks \citep{chakraborty2018adversarial}. Given $k$ possible surface-form variations for the hypothesis, we test if at least one of the candidates would be able to cause a label change compared to the original prediction, which can be formalised as:

\begin{align}\label{eq:soft_f_r}
    r_r=\frac{\sum_i^{n^\prime}\mathbbm{1}\left[\exists j \in [1,k], \mathcal{R}(p_i, h_i, h_i^{\prime j}, =) \neq 1\right]}{n^\prime}.
\end{align}

Here $n^\prime$ is the number of correctly answered original samples, and the matching operator $\mathcal{O}$ is a simple equality checking operator "$=$". We refer to this metric as a relaxed Fooling Rate. To measure more drastic label changes, i.e. \emph{entailment} to \emph{contradiction} and vice versa, we also define a stricter version of \autoref{eq:soft_f_r}.

\begin{align}\label{eq:strict_f_r}
    r_s=\frac{\sum_i^{n^\prime}\mathbbm{1}\left[\exists j \in [1,k], \mathcal{R}(p_i, h_i, h_i^{\prime j}, =^s) \neq 1\right]}{n^\prime}.
\end{align}

 We replace standard equality for the operator $\mathcal{O}$ in \autoref{eq:soft_f_r} with a strict counterpart that matches only if the predictions are direct opposites, i.e. \emph{entailment} $\leftrightarrow$ \emph{contradiction}. It must be noted that the \emph{neutral} class does not have a direct opposite; thus, the metric for this label remains unchanged. It can be concluded that the inequality $r_s \leq r_r \leq 1$ trivially holds when using these metrics.

\section{Experimental Setup}

\subsection{Model Details}

\subsection{Semantics preserving Generation}

To generate and refine semantic variations of the original hypothesis, we chose \emph{flan-t5-xl} as the generation model $\mathcal{G}$. It is an instruction-tuned LLM that has shown close state-of-the-art performance in tasks such as paraphrasing, zero and few shot generation, chain of thought reasoning (CoT), and multi-task language understanding \citep{chung2022scaling}. For each of the selected hypotheses, we produce $k=5$ unique semantics-preserving variations. To ensure diversity and consistency of the generated text while avoiding computationally expensive exhaustive search, we use a group beam search \citep{vijayakumar2016diverse}  with a temperature $t\in [0.3, 0.6]$ and a maximum output of 40 tokens throughout the generation and refinement procedure. We also further diversify the generation by using the recipe from \citet{li2016simple}.

\subsection{NLI models}

We systematically experiment with transformer architectures that are fine-tuned on MNLI, which exhibit state-of-the-art or close predictive accuracy on the dataset. We specifically choose \emph{bart-large} \citep{lewis2019bart}, \emph{roberta-large} \citep{liu2019roberta}, \emph{deberta-base, deberta-large, deberta-xlarge} \citep{he2020deberta} and \emph{distilbart} \citep{sanh2019distilbert}. These PLMs are taken without change from their original studies through the Transformers library \citep{wolf-etal-2020-transformers}, ensuring the complete reproducibility of the results. To observe the effect in an \emph{out-of-domain} setup, we also evaluate these models on SNLI and ANLI in a zero-shot transfer setting.

\section{Results and Analysis}

This section presents the results and analyses of our semantic sensitivity evaluation framework along with a suite of ablations analysing the phenomenon across various transformer sizes, domains, and label space. Furthermore, we measure the impact of the phenomenon on the inconsistent predictive behaviour of NLI models. 

\subsection{Semantic Sensitivity}

\subsection{In-domain}

We evaluate several PLMs trained on MNLI using our experiments presented in \autoref{tab:foolrate1}. The results show that models are limited in their comprehension of compositional semantics as the relaxed fooling rate on \emph{in-domain} experimentation averages at $r_r = 12.9\%$. This is further reinforced by the fact that more than half, $r_s = 6.58\%$  of the label changes occur with strict inequality. This means that minor semantics-preserving changes lead to a sizable shift in model predictions, even prompting towards the opposite decision edge half the time. The behaviour is consistent across all the transformers and leads us to believe that samples that changed labels after surface-form variations showcase the inconsistent predictive nature of the models. We further elaborate on this in the next section.
Consequently, semantically equivalent variations evidently hinder the decision-making of the NLI models, prompting us to believe that models have limited understanding w.r.t. semantic structure and logical relation, even when the model is trained on texts from the same distribution.

\begin{figure*}[t]
    \centering
    \includegraphics[width=0.8\textwidth]{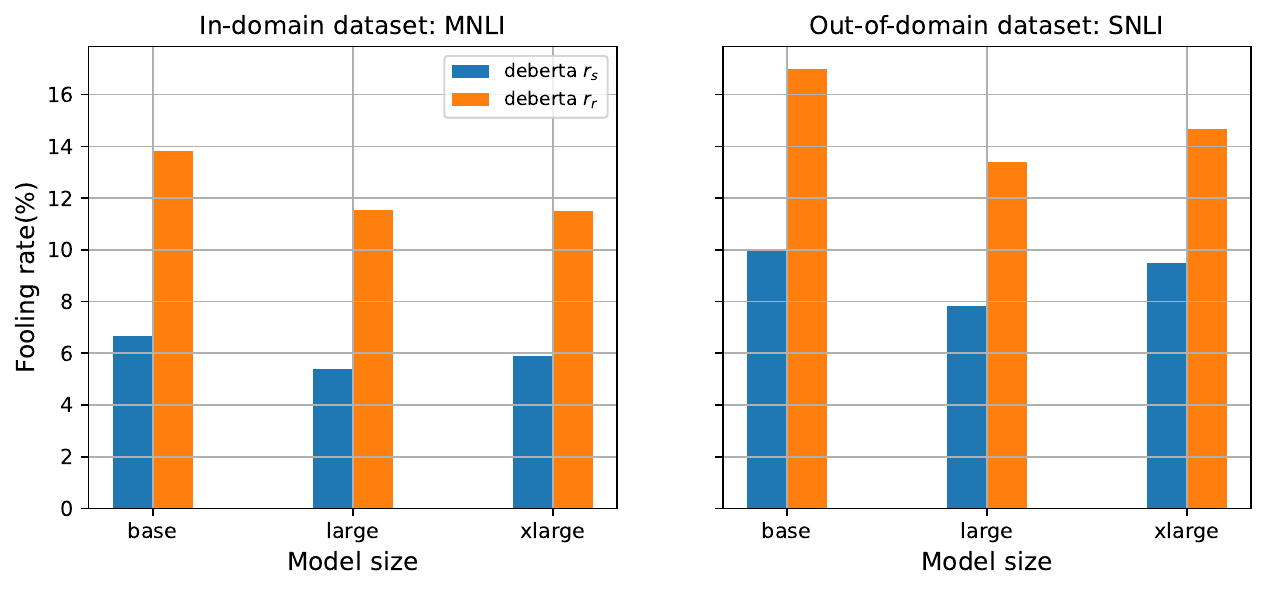}
    \caption{In- and out-of-domain fooling rate of DeBERTa of varied sizes, which are measured on MNLI (left) and SNLI (right). Similarly, $r_s$ and $r_r$ represent the strict and relaxed fooling rates, respectively.}
    \label{fig:frvary}
\end{figure*}

\subsection{Out-of-domain}

We also probe the NLI models in an \emph{out-of-domain} zero-shot setting to assess the transferability of compositional semantic knowledge. Our results in \autoref{tab:foolrate1} show that the discrepancies and limitations in semantic comprehension are even more pronounced in this setting. We see an averaged relaxed fooling rate of $r_r = 23.7\%$, with the maximum at $57.49\%$, which is only marginally better than a majority voting baseline.
It must be noted that because different datasets have varying numbers of samples, the average is weighted w.r.t. the number of sampled instances from the particular dataset in the experiment.
The results on \emph{out-of-domain} evaluation once again follow the pattern that more than half, $r_s = 15.8\%$ of the samples switch the labels to their logically contrasting counterparts. This shows that zero-shot transfer further amplifies the limitations that NLI models have for using semantic structures and preserving logical relations. This further suggests that the semantic variations where a label change occurs are likely to be originally predicted correctly as an inconsistent guess. It follows, that although PLMs fine-tuned on MNLI are widely used for zero-shot classification, their effectiveness diminishes if the classification tasks require syntactic understanding. Indeed, model effectiveness declines and the fooling rates rise as the tasks become more challenging, requiring greater syntactic knowledge, as we can see from the comparison of the results from SNLI to ANLI.

\subsection{Effects of distillation}

Next, we want to probe if the susceptibility towards semantic noise is transferred during model distillation. Thus, we use \emph{DistilBart} that is distilled from a larger pre-trained BART model. While model accuracy remains comparable to the original model in \autoref{tab:nli_eval}, the distilled version struggles sizeably more with surface-form variations. On average, across \emph{in-} and \emph{out-of-} domain evaluation, the distilled NLI model is more sensitive than the original in terms of relaxed fooling rate by $\triangle r_r = 18.4\%$. The effect of supposed inconsistence is amplified when observing the strict fooling rate, where on average $\frac{r_r}{r_s} \leq 1.5$. This indicates that during distillation, models are bound to forget the knowledge regarding compositional semantics making it harder to preserve the logical equivalence during inference.

\subsection{Effects of model size}

We also test how semantics-preserving noise affects models of different sizes and parametrization (see \autoref{fig:frvary}). Although for \emph{in-domain} setup, the relaxed fooling rate metrics marginally drop as the models get bigger, the same cannot be observed in \emph{out-of-domain} setup. It is evident that bigger PLMs from our study are almost as restricted in semantic comprehension as their smaller counterparts. This indicates that emergent semantic capabilities are not only tied to model size, but also widely depend upon the choice of the training dataset.


\begin{table*}[t]
\centering
\begin{adjustbox}{max width=\textwidth}
\begin{tabular}{@{}c|cccc@{}}
\toprule
\textbf{} & $r_s/r_r\left(y=E\right)$ & $r_s/r_r\left(y=N\right)$ & $r_s/r_r\left(y=C\right)$ & $r_s/r_r$       \\ \midrule
MNLI                       & 2.78\%/13.41\%                     & 14.33\%/14.33\%                    & 3.69\%/11.17\%                     & 6.58\%/12.92\%     \\ \midrule
SNLI                       & 9.54\%/18.73\%                     & 19.42\%/19.42\%                    & 2.92\%/11.82\%                     & 10.24\%/16.54\%    \\
ANLI\_r1                   & 21.64\%/41.97\%                    & 38.62\%/38.62\%                    & 29.17\%/44.57\%                    & 29.97\%/41.30\%     \\
ANLI\_r2                   & 20.84\%/46.28\%                    & 49.41\%/49.41\%                    & 21.89\%/50.80\%                     & 31.32\%/48.53\%    \\
ANLI\_r3                   & 11.65\%/52.00\%                     & 47.18\%/47.18\%                    & 16.42\%/46.50\%                     & 27.04\%/48.17\%    \\ \bottomrule
\end{tabular}
\end{adjustbox}
\caption{Fooling rate averaged over all models. $r_s$ represents the strict fooling rate, in which case the predicted label of the evaluation pair is opposite to the original label $y$. $r_r$ measures the proportion of label change. $y\in\{E,N,C\}$ group the $(p,h)$ pairs by their semantic relation, representing entailment, neutrality, and contradiction, respectively.}
\label{tab:foolrate_class}
\end{table*}

\subsection{Severity of Inconsistent Predictions}

\subsection{Consistency across label space}

To analyse the extent of semantic sensitivities within NLI models we test the effect across all the classes in the label spaces, presented in \autoref{tab:foolrate_class}. The per-class breakdown of the strict and relaxed fooling rate indicates that the effect is consistent across the whole label space. This allows us to conclude that the observed limitations in compositional semantic understanding are not caused by class imbalances and are not specific to a particular set of examples. We see the increased fooling rate across all of the labels when comparing \emph{in-domain} and \emph{out-of-domain} experiments. This reinforces the prior indications regarding models' inability to use semantic structure to preserve inherent relations within the data, as all logical relations attain rather similar amounts of fooling rate during direct evaluation.

\begin{figure*}[t]
    \centering
    \includegraphics[width=0.85\textwidth]{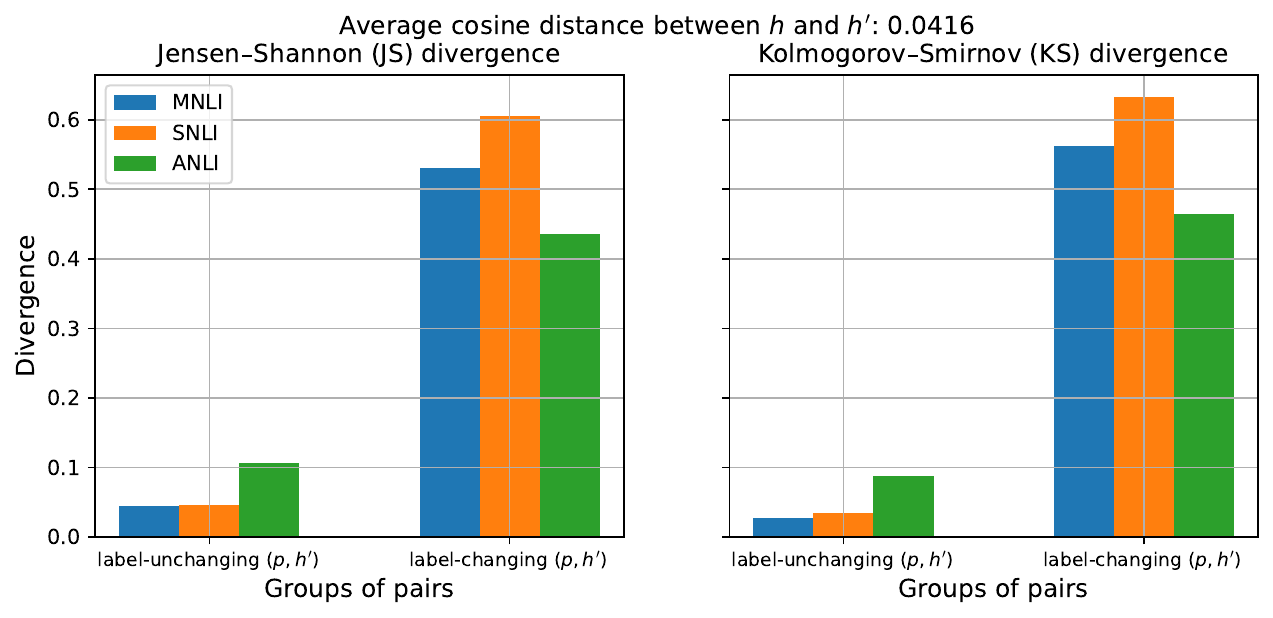}
    \caption{Divergence of predictive probability distribution between $(p,h)$ and $(p,h^\prime)$ measured across the datasets (ANLI is averaged over the rounds) and averaged over all models. All evaluation pairs are split into two groups based on whether they manage to flip the original label. Two divergence metrics are shown -- JS divergence (left) and KS divergence (right).}
    \label{fig:div}
\end{figure*}

\begin{figure}[t]
    \centering
    \includegraphics[width=\textwidth]{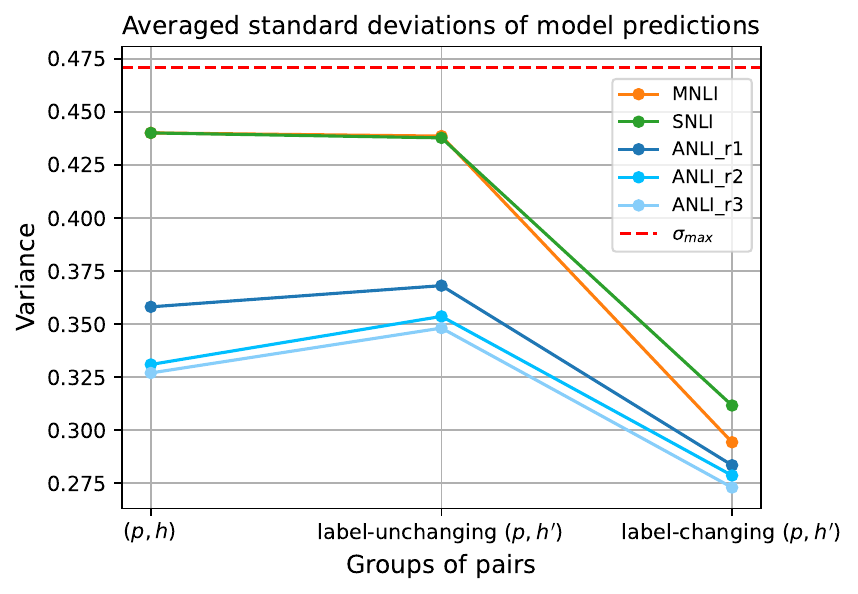}
    \caption{Standard deviation $\sigma$ of predicted label probabilities (obtained from the final softmax layer of the model) averaged for original premise-hypothesis pair (left), surface-form variations that did not cause label changes (mid) and did induce label change (right). The bigger $\sigma$, the more confident the model is w.r.t. the predictions. The results are averaged over all models.}
    \label{fig:var}
\end{figure}

\subsection{Distribution shift in decision making}

Recall that we want to measure the impact of semantics-preserving surface-form variations on NLI models. We study the predictive distributional shift within the samples that cause a changed model prediction. To do this, we initially split the samples into two categories considering whether the sample induced a change of the original prediction within the NLI model. We further average the probability distribution of labels obtained from the final softmax layer of the model for these two categories.
We measure the differences between the two distributions with two statistical tests. To evaluate the relative entropy between them, we use Jensen-Shanon Divergence \citep{fuglede2004jensen}, a symmetric, non-negative, and bounded metric for assessing the similarity between two distributions, $\operatorname{JSD}(P \| Q)=\frac{1}{2} D(P \| M)+\frac{1}{2} D(Q \| M)$, where $D$ is the Kullback–Leibler divergence \citep{joyce2011kullback}. We verify the statistical significance of our findings with the Kolmogorov–Smirnov test \citep{berger2014kolmogorov}, which shows if the two sets of samples are likely to come from the same distribution.

Our results in \autoref{fig:div} show a significant distribution shift when assessing semantics-preserving surface-form variations. The cosine distance in the sentence embedding space between the generated and original samples is negligible at $0.04$. As the absolute cosine similarity values possess limited interpretable meaning, we further explore the distributions of cosine distances towards original samples for the examples that do and do not induce label changes. We measure the Jansen-Shannon divergence of these two distributions at $0.001$, implying they are strongly similar. This reinforces the hypothesis that surface-form variations produce logically equivalent samples with minor distance in the embedding space regardless of the induced label changes. 
However, despite minor changes in the semantic composition, we see a sizable change in the final predictive distribution of the NLI models. We see a significant rise both in Jensen-Shannon divergence and Kalmogorov-Smirnov metric, $\triangle \text{JSD} = 0.51$ and $\triangle \text{K-S} = 0.54$, when comparing the examples where the model prediction has changed compared to the original. This indicates that the generated variations do not cause negligible change within model prediction, but rather can be considered adversarial for the model. It shows that the limited capabilities to utilise syntactic information cause the model to significantly change the final prediction given minuscule variations, which is an inconsistent predictive behaviour. Given that we initially sampled examples that the models answered correctly, these results assert our belief that the models do not display consistent predictive behaviour despite having equivalent inputs. This shows that albeit the strong model performance presented in \autoref{tab:nli_eval}, there is masked degeneration and discrepancies within the NLI models stemming from semantic sensitivity. Our method allows for explicitly quantifying the degree of semantic sensitivity within PLMs and allows to measure the impact of that sensitivity on the decision-making process of the model.

\subsection{Semantic-Sensitivity and decision variations}

We 
lastly analyse the standard deviation within the predicted label distribution produced from the softmax of the model. We compute the standard deviation for the distribution of original premise hypothesis predictions and compare it with a replacement that does not and does cause label changes in PLM classification, see \autoref{fig:var}.
For reference, the upper bound for standard deviation in this 3 class setting happens when the model is greatly confident in one of the classes, i.e. ${softamx} = [1, 0, 0] \rightarrow \sigma_{max} = 0.471$. Bigger $\sigma$ on average implies more confident answers by the PLM. It can be observed that the average predictions with the original samples have a great degree of confidence. We see an interesting phenomenon where the predictive confidence slightly rises across most of the datasets for the cases where the model is able to recover the inherent textual relations. However, when faced with examples that cause label changes, there is a significant drop of $\triangle \sigma = 0.1$ in the standard deviation averaged across the datasets. This signifies that predictive confidence sizably degrades when the model struggles to recover the existent relations because of slight semantics-preserving variations. That further indicates that NLI models are susceptible to semantic sensitivity and have limited knowledge of compositional semantics, which can lead to the degradation of predictive confidence and incidentally inconsistent predictions.

\section{Conclusion}

We present a novel framework for assessing semantic sensitivity in NLI models through generating semantics-preserving variations. Our systematic study of the phenomenon across various datasets and transformer-based PLMs shows that the models consistently struggle with variations requiring knowledge of compositional semantics. This performance deterioration happens across the whole label space, almost regardless of model size. We measure the impact of semantic-sensitivity and show that it diminishes models' predictive confidence and can lead to predictive inconsistency.

\section*{Limitations}

In our work, we cover the semantic-sensitivity that can be found within NLI models. However, the framework can be applied to a wider range of classification tasks. The benchmark can be extended with more datasets and further enhanced with larger human evaluation. Also, we covered PLMs specifically trained for NLI; however, it would be great to cover bigger LLMs, in particular w.r.t. their emergent zero-shot capabilities. Another limitation is that we only cover English-based language models and do not test in multi-lingual or cross-lingual settings.

\section*{Ethics Statement}

Our work completes an analysis of numerous models w.r.t. their decision inconsistency induced by semantic surface form variations. We show that models are somewhat unable to handle logically and semantically equivalent sentences, which would lead to an inconsistent use across various domains and applications. Our generation method does not induce any further exploitation threat and can only be used for measuring the above-mentioned inconsistencies. We exclusively use open source publicly accessible data and models within our experimentations.

\section*{Acknowledgements}
Erik is partially funded by a DFF Sapere Aude research leader grant under grant agreement No 0171-00034B, as well as by a NEC PhD fellowship.
This work is further supported by the Pioneer Centre for AI, DNRF grant number P1.
%


\clearpage
\section{Appendices}
\label{sec:appendix}

\begin{table}[!t]
\begin{tabular}{@{}lrrrr@{}}
\toprule
Dataset & \multicolumn{1}{l}{Fuzzy token match \%} & \multicolumn{1}{l}{average length $h$} & \multicolumn{1}{l}{average length $h\prime$} & \multicolumn{1}{l}{average token overlap} \\ \midrule
mnli & 84.83 & 14.31 & 14.14 & 13.25 \\
snli & 81.55 & 10.81 & 11.21 & 10.38 \\
anli\_r1 & 87.59 & 17.3 & 17.02 & 13.73 \\
anli\_r2 & 86.49 & 15.99 & 15.84 & 12.8 \\
anli\_r3 & 85.17 & 14.32 & 14.29 & 11.27 \\ \bottomrule
\end{tabular}
\caption{Percentages of token matches and other statistics.}
\label{tab:token_anal}
\end{table}

\subsection{Evaluation under Label change}

\begin{figure}[t!]
\includegraphics[width=\textwidth]{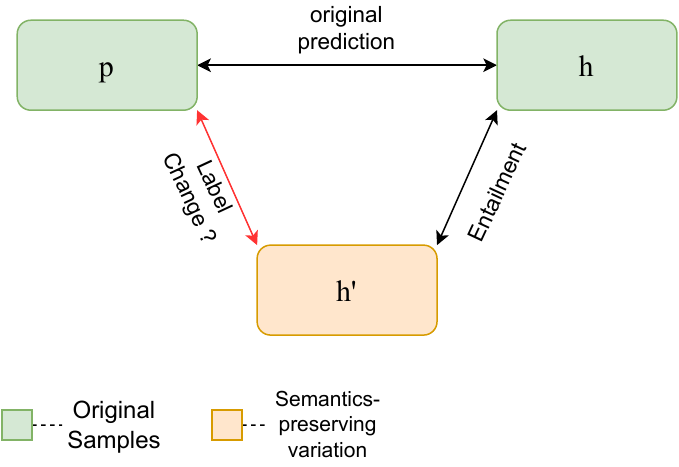}
    \caption{A diagram for assessing semantic similarity. Given the generated semantics-preserving surface-form variation $h^\prime$, we evaluate if a label change occurs when replacing the hypothesis in accordance with \autoref{eq:label_change}}.
    \label{fig:semantic_triangle}
\end{figure}

To assess the extent of the impact of semantic sensitivity, we employ an evaluation under label change. This means we consider the examples that changed the original prediction of the model after a surface-form variation replaced the original hypothesis. A graphical representation of this can be seen in \autoref{fig:semantic_triangle}. It must be noted that we use only the samples that the model originally predicted correctly to avoid incorrect assessment regarding the reasoning behind the false predictions. Our primary aim is to measure the semantic sensitivity within the model predictions and the extent of inconsistency it causes.

\subsection{Token Level-Differences of the generated variations}

We further explore the difference between surface-form variations and original examples by conducting a token-level analysis for each pair $(h, h\prime)$. We compute the average amount of tokens present for the original and generated hypothesis and use fuzzy and exact matching to assess the overlap of tokens on average for each dataset. The results can be seen in \autoref{tab:token_anal}. The results show that the generated and original examples have a high token level overlap which further reinforces the idea that surface form variations are close both syntactically, in the embedding space and logically.

\chapter{With Great Backbones Comes Great Adversarial Transferability}
\label{chap:adversarial}

\section{Introduction}

Machine vision models pre-trained with massive amounts of data and using self-supervised techniques \citep{newell2020useful} are shown to be robust and highly performing\citep{goyal2021self,goldblum2024battle} feature-extracting backbones \citep{elharrouss2022backbones,han2022survey}, which are further used in a variety of tasks, from classification \citep{atito2021sit,chen2020big} to semantic segmentation \citep{ziegler2022self}. However, creating such backbones incurs substantial data annotation \citep{jing2020self} and computational costs \citep{han2022survey}, consequently rendering the use of such publicly available pre-trained backbones the most common and efficient solution for researchers and engineers alike. 
Prior works have focused on analysing safety and adversarial robustness with complete, i.e. \emph{white-box} \citep{porkodi2018survey} or no, i.e. \emph{black-box} \citep{bhambri2019survey} knowledge of the target model weights, fine-tuning data, fine-tuning techniques and other tuning meta-information. Although, in practice, an attacker can access partial knowledge \citep{DBLP:conf/iclr/LordMB22,DBLP:journals/tip/ZhuCLC00ZCH22,DBLP:conf/sp/CarliniCN0TT22} of how the targeted model was produced, i.e. original backbone weights, tuning recipe, etc., the adversarial robustness of models tuned on a downstream task from a given pre-trained backbone remains largely underexplored. We refer to settings with partial knowledge of target model constructions meta-information as \emph{grey-box}. This is important both for research and production settings because with an increased usage \citep{DBLP:conf/nips/GoldblumSNSPSCI23} of publically available pre-trained backbones for downstream applications, we are incapable of assessing the potential exploitation susceptibility and inherent risks within models tuned on top of them and subsequently enhance future pre-trained backbone sharing practices.
\begin{figure*}[t!]
    \centering
    \includegraphics[clip=true,width=\textwidth]{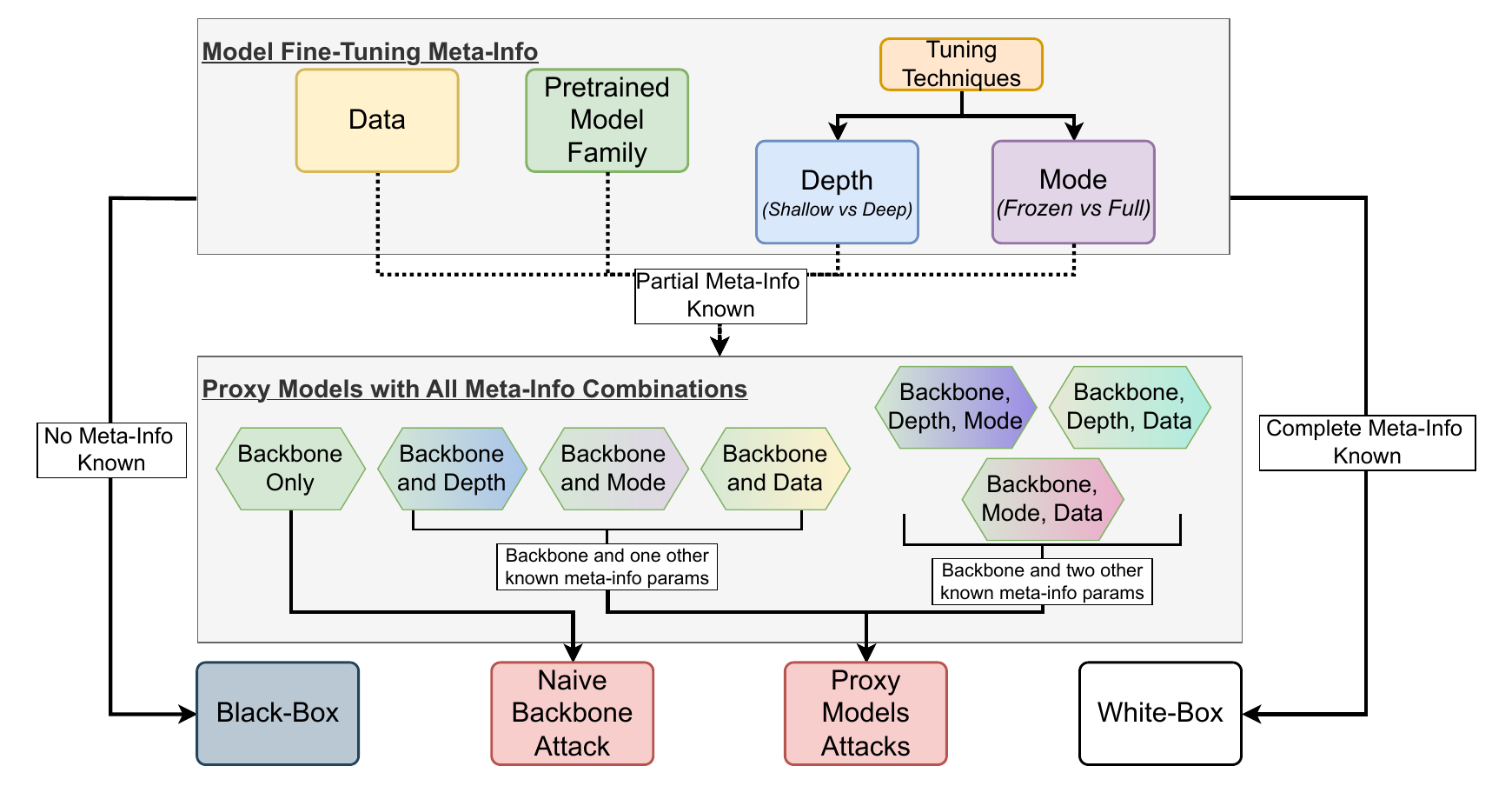}
    \caption{The figure depicts all of the settings used to evaluate adversarial vulnerabilities given different information of the target model construction. From left to right, we simulate exhaustive varying combinations of meta-information available about the target model during adversarial attack construction. All of the created proxy models are used separately to assess adversarial transferability.}
    \label{fig:framework_adv_main}
\end{figure*}

In this work, we systematically explore the safety towards adversarial attacks within the models tuned on a downstream classification task from a known publically available backbone pre-trained with a self-supervised objective. We further explicitly measure the effect of the target model construction meta-information by simulating different levels of its availability during the adversarial attack. 
For this purpose, we initially train $352$ diverse models from $21$ families of commonly used pre-trained backbones using $4$ different fine-tuning techniques and $4$ datasets. We fix each of these networks as a potential target model and transfer adversarial attacks using all of the other models produced from the same backbones as proxy surrogates \citep{DBLP:conf/aaai/QinXYH23,DBLP:conf/iclr/LordMB22} for adversarial attack construction. Each surrogate model simulates varying levels of knowledge availability w.r.t. target model construction on top of the available backbone during adversarial attack construction. This constitutes approximately $20000$ adversarial transferability comparisons between target and proxy pairs across all model families and meta-information variations.
By assessing the adversarial transferability of attacks from these surrogate models, we are able to explicitly measure the impact of the availability of each meta-information combination about the final target model during adversarial sample generation.

We further introduce a naive exploitation method referred to as \emph{backbone attacks} that utilizes only the pre-trained feature extractor for adversarial sample construction. The attack uses projected gradient descent over the representation space to disentangle the features of similar examples. Our results show that both proxy models and even simplistic \emph{backbone attacks} are capable of surpassing strong query-based \emph{black-box} methods and closing to \emph{white-box} performance. The findings indicate that \emph{backbone attacks}, where the attacker lacks meta-information about the target model, are generally more effective than attempts to generate adversarial samples with limited knowledge. This highlights the vulnerability of models built on publicly available backbones.

Our ablations show that \textit{having access to the weights of the pre-trained backbone is functionally equivalent to possessing all other meta-information about the target model when performing adversarial attacks}. We compare these two scenarios and show that both lead to similar vulnerabilities, highlighting the interchangeable nature of these knowledge types in attack effectiveness. Our results emphasize the risks in sharing and deploying pre-trained backbones, particularly concerning the disclosure of meta-information. Our experimental framework can be seen in \cref{fig:framework_adv_main}.

Toward this end, our contributions are as follows:

\begin{itemize} 
\item We introduce, formalize and systematically study the \textbf{grey-box} adversarial setting, which reflects realistic scenarios where attackers have partial knowledge of target model construction, such as access to pre-trained backbone weights and/or fine-tuning meta-information. 
\item We simulate over $20,000$ adversarial transferability comparisons, evaluating the impact of varying levels of meta-information availability about target models during attack construction.
\item We propose a naive attack method, \emph{backbone attacks}, which leverages the pre-trained backbone's representation space for adversarial sample generation, demonstrating that even such a simplistic approach can achieve stronger performance compared to a query-based black-box method and often approaches white-box attack effectiveness. 
\item We show that access to pre-trained backbone weights alone enables adversarial attacks as effectively as access to the full meta-information about the target model, emphasizing the inherent vulnerabilities in publicly available pre-trained backbones.  
\end{itemize}

\section{Related Work}

\subsection{Self Supervised Learning}

With the emergence of massive unannotated datasets in machine vision, such as YFCC100M\citep{DBLP:journals/cacm/ThomeeSFENPBL16}, ImageNet\citep{deng2009imagenet}, CIFAR \citep{krizhevsky2009learning} and others Self Supervised Learning (SSL) techniques \citep{DBLP:journals/pami/JingT21} became increasingly more popular for pre-training the models \citep{newell2020useful}. This prompted the creation of various families of SSL objectives, such as colorization prediction \citep{DBLP:conf/eccv/ZhangIE16}, jigsaw puzzle solving \citep{DBLP:conf/eccv/NorooziF16} with further invariance constraints \citep[PIRL]{DBLP:conf/cvpr/MisraM20}, non-parametric instance discrimination \citep[NPID, NPID++]{DBLP:conf/cvpr/WuXYL18}, unsupervised clustering \citep{DBLP:conf/eccv/CaronBJD18}, rotation prediction \citep[RotNet]{DBLP:conf/iclr/GidarisSK18}, sample clustering with cluster assignment constraints\citep[SwAV]{DBLP:conf/nips/CaronMMGBJ20},  contrastive representation entanglement \citep[SimCLR]{DBLP:conf/icml/ChenK0H20},
self-distillation without labels \citep[DINO]{DBLP:conf/iccv/CaronTMJMBJ21} and others \citep{DBLP:journals/pami/JingT21}. Numerous architectures, like AlexNet \citep{DBLP:conf/nips/KrizhevskySH12}, variants of ResNet\citep{DBLP:conf/cvpr/HeZRS16} and visual transformers \citep{DBLP:conf/iclr/DosovitskiyB0WZ21,DBLP:conf/icml/TouvronCDMSJ21,DBLP:conf/nips/AliTCBDJLNSVJ21} were trained using these SSL methods and shared for public use, thus forming the set of widely used pre-trained backbones.
We obtain all of these models trained with different self-supervised objectives from their original designated studies summarised in VISSL \citep{goyal2021vissl}. An exhaustive list of all models can be seen in \cref{tab:ssl_model_summary}.

\begin{table}[!t]
\centering
\resizebox{\columnwidth}{!}{%
\begin{tabular}{@{}lcc@{}}
\toprule
\textbf{SSL Method} & \textbf{Pretraining Dataset} & \textbf{Architecture} \\ 
\midrule
\multicolumn{3}{l}{\textbf{Colorization} \citep{DBLP:conf/eccv/ZhangIE16}} \\
Colorization & YFCC100M      & AlexNet   \\
Colorization & ImageNet-1K   & AlexNet   \\
Colorization & ImageNet-1K   & ResNet-50 \\
Colorization & ImageNet-21K  & AlexNet   \\
Colorization & ImageNet-21K  & ResNet-50 \\
\midrule
\multicolumn{3}{l}{\textbf{Jigsaw Puzzle}\citep{DBLP:conf/eccv/NorooziF16}} \\
Jigsaw Puzzle & ImageNet-21K & ResNet-50  \\
Jigsaw Puzzle & ImageNet-1K  & ResNet-50  \\
Jigsaw Puzzle & ImageNet-21K & ResNet-50  \\
Jigsaw Puzzle & ImageNet-21K & AlexNet    \\
Jigsaw Puzzle & ImageNet-1K  & AlexNet    \\
Jigsaw Puzzle & ImageNet-1K  & ResNet-50  \\
\midrule
\multicolumn{3}{l}{\textbf{PIRL (Jigsaw-based)}\citep{DBLP:conf/cvpr/MisraM20}} \\
PIRL         & ImageNet-1K   & ResNet-50  \\
\midrule
\multicolumn{3}{l}{\textbf{Rotation Prediction} \citep{DBLP:conf/iclr/GidarisSK18}} \\
RotNet       & ImageNet-1K   & ResNet-50  \\
\midrule
\multicolumn{3}{l}{\textbf{DINO}\citep{DBLP:conf/iccv/CaronTMJMBJ21}} \\
DINO         & ImageNet-1K   & DeiT-Small \\
DINO         & ImageNet-1K   & XCiT-Small \\
\midrule
\multicolumn{3}{l}{\textbf{SimCLR}\citep{DBLP:conf/icml/ChenK0H20}} \\
SimCLR       & ImageNet-1K   & ResNet-50  \\
SimCLR       & ImageNet-1K   & ResNet-101 \\
\midrule
\multicolumn{3}{l}{\textbf{SwAV} \citep{DBLP:conf/nips/CaronMMGBJ20}} \\
SwAV         & ImageNet-1K   & ResNet-50  \\
SwAV         & ImageNet-1K   & ResNet-50  \\
\midrule
\multicolumn{3}{l}{\textbf{DeepCluster V2} \citep{DBLP:conf/eccv/CaronBJD18}} \\
DeepCluster V2 & ImageNet-1K & AlexNet    \\
\midrule
\multicolumn{3}{l}{\textbf{Instance Discrimination (NPID)} \citep{DBLP:conf/cvpr/WuXYL18}} \\
NPID         & ImageNet-1K   & ResNet-50  \\
\bottomrule
\end{tabular}
}
\caption{Summary of Self-Supervised Learning Methods, Pretraining Datasets, and Architectures used in our study.}
\label{tab:ssl_model_summary}
\end{table}

\subsection{Adversarial Attacks}
The availability of pre-trained backbones allows to test them for vulnerabilities towards adversarial attacks, which are learnable imperceptible perturbations generated to mislead models into making incorrect predictions \citep{DBLP:journals/corr/SzegedyZSBEGF13, DBLP:journals/corr/GoodfellowSS14}. Several attack strategies have been studied, including single-step fast gradient descent \citep[FGSM]{DBLP:journals/corr/GoodfellowPMXWOCB14,DBLP:conf/iclr/KurakinGB17a}, and computationally more expensive optimization-based attacks, such as projected gradient descent based attacks \citep[PGD]{DBLP:conf/iclr/MadryMSTV18}, CW \citep{DBLP:conf/ccs/Carlini017}, JSMA \citep{DBLP:conf/ccs/PapernotMGJCS17}, and others \citep{DBLP:conf/cvpr/DongLPS0HL18, DBLP:conf/cvpr/Moosavi-Dezfooli16,DBLP:conf/iclr/MadryMSTV18}. All of these attacks assume complete access to the target model, which is known as the \emph{white-box} \citep{DBLP:conf/ccs/PapernotMGJCS17} setting. These attacks can be \emph{targeted} toward confusing the model to infer a specific wrong class or \emph{untargeted} with the desire that it infers any incorrect label. However, an opposite setting with no information, referred to as \emph{black-box} \citep{DBLP:conf/ccs/PapernotMGJCS17}, has also been explored as a more practical setting. The methods involve attempts at gradient estimation \citep{DBLP:conf/ccs/ChenZSYH17,DBLP:conf/icml/IlyasEAL18,DBLP:conf/eccv/BhagojiHLS18}, adversarial transferability \citep{DBLP:conf/ccs/PapernotMGJCS17, DBLP:conf/eccv/ChenZHW20}, local search \citep{DBLP:journals/corr/NarodytskaK16, DBLP:conf/iclr/BrendelRB18,DBLP:conf/icml/LiLWZG19,DBLP:conf/icml/MoonAS19}, combinatorial perturbations \citep{DBLP:conf/icml/MoonAS19} and others \citep{bhambri2019survey}. However, these methods also require massive sample query budgets ranging from $\left[10^3, 10^5 \right]$ queries or computational resources creating each adversarial sample \citep{bhambri2019survey}. Compared to these, we introduce a novel setup with the knowledge of the pre-trained backbone and varying levels of partially known target model tuning meta-information during adversarial attack construction, which we call \emph{grey-box}. We show that even simple naive attacks are capable of exploiting better than black-box attacks without the need for significantly querying the target model.

\subsection{Adversarial Transferability}

Our work is also aligned with adversarial transferability, where adversarial examples generated for one model can mislead other models, even without access to the target model weights or training data. This property poses significant security concerns, as it allows for effective black-box attacks on systems with no direct access \citep{DBLP:conf/ccs/PapernotMGJCS17, DBLP:conf/icml/IlyasEAL18}. Efforts can be divided into \emph{generation-based} and \emph{optimisation} methods. Generative methods have emerged as an alternative approach to iterative attacks, where adversarial generators are trained to produce transferable perturbations. For instance, \citet{DBLP:conf/cvpr/PoursaeedKGB18} employed autoencoders trained on white-box models to generate adversarial examples. Most of the attacks aiming for adversarial transferability strongly depend on the availability of data from the target domain \citep{DBLP:conf/ccs/Carlini017, DBLP:conf/ccs/PapernotMGJCS17}. However, although current adversarial transferability methods claim to produce massive vulnerabilities in machine vision models, \citet{DBLP:journals/corr/abs-2105-00433} examines the practical implications of adversarial transferability, which are frequently overstated. That study demonstrates that it is nearly impossible to reliably predict whether a specific adversarial example will transfer to an unseen target model in a black-box setting.  This perspective underscores the importance of systematically evaluating transferability in realistic settings, including scenarios where attackers are sensitive to the cost of failed attempts.
In our study, we offer a novel systematic approach to explicitly assess the adversarial transferability with varying levels of meta-information knowledge.

\section{Methodology}

\subsection{Preliminaries}
For consistency, we employ the following notation. We denote each Dataset $\mathcal{D}=\{\mathcal{X},\mathcal{Y}\}$. Where $\mathcal{X}=\{x_1, \dots ,x_{|\mathcal{D}|}\}$ is a set of images, with $x_i \in \mathcal{R}^{H \times W \times C}$, where $H$,$W$ and $C$ are the height, width and the channels of the image accordingly and $\mathcal{Y} = \{y_1 \dots y_n\}$ is used as the set of ground truth labels. We denote the training, validation and testing splits per task as $\mathcal{D}=\{\mathcal{D}_{train}, \mathcal{D}_{val}, \mathcal{D}_{test}\}$. 
A \textit{model} is defined as the following tuple $\mathcal{M} = \mathcal{M}(\mathcal{D}, \mathcal{W}, \mathcal{B}, \mathcal{F})$, where $\mathcal{D}$ contains the dataset used for training, $\mathcal{W}$ are the weights of the trained model and $\mathcal{B}$ is the pre-trained back-bone $\mathcal{B}( \mathcal{W}_{\mathcal{B}})$ with available weights $\mathcal{W}_{\mathcal{B}}$. The notation $\mathcal{F}(\mathcal{T}, \mathcal{Z})$, where $\mathcal{T}$ encodes the \emph{mode} of tuning (e.g., full fine-tuning, partial fine-tuning, etc.) and $\mathcal{Z}$ the \emph{depth} of tuning of the final classifier on top of the backbone.

\subsection{Meta-Information variations}
We define the variations of the available meta-information about the target model $\mathcal{M}$ during an adversarial attack as a \textit{unit of release} $\mathcal{R}=\mathcal{R}(\mathcal{M}(\mathcal{D}, \mathcal{W}, \mathcal{B}(\mathcal{W}_{\mathcal{B}}), \mathcal{F}(\mathcal{T}, \mathcal{Z})))$. For example, if the target fine-tuning mode $\mathcal{Z}^{target}$ and dataset $\mathcal{D}^{target}$ are not known, the unit of release will be $\mathcal{R}=\mathcal{R}(\mathcal{M}(*, \mathcal{W}, \mathcal{B}(\mathcal{W}_{\mathcal{B}}), \mathcal{F}(\mathcal{T}, *)))$. Note that the \emph{black-box} setting will correspond to the unit of release $\mathcal{R}(\mathcal{M}(*, *, *, *, *))$ and the \emph{white-box} setting to $\mathcal{R}(\mathcal{M}(\mathcal{D}, \mathcal{W}, \mathcal{B}(\mathcal{W}_{\mathcal{B}}), \mathcal{F}(\mathcal{T}, \mathcal{Z})))$, all the variations between these are considered \emph{grey-box}. When discussing any experiments within the \emph{gery-box} setup, we assume the minimal unit of release contains knowledge about at least the pre-trained backbone i.e. $\mathcal{R}(\mathcal{M}(*, *, \mathcal{B}(\mathcal{W}_{\mathcal{B}}), *)$.

\subsection{Adversarial Attacks with Proxy Models}

To test the adversarial robustness of the models trained from the same pre-trained backbone, we create a set of proxy models $\mathcal{M}^{proxy} = \{\mathcal{M}^{proxy}_{1} \dots \mathcal{M}^{proxy}_{v}\}$ given the pre-trained backbone $\mathcal{B}$, where $v$ is the number of all possible units of release between \emph{black-box} and \emph{white-box} settings that include the backbone. For each proxy model $\mathcal{M}^{proxy}_{i}$ with its designated meta-information unit of release $\mathcal{R}_i$, we use an adversarial attack $\mathcal{A}$ to generate adversarial noise and further transfer it to the target model $\mathcal{M}^{target}$. This means that given an example image $x$ with a label $y$, target and proxy models $\mathcal{M}^{{target}}$, $\mathcal{M}^{{proxy}}$ we want to produce a sample $x'$ that would fool the target model, such that $\arg \max \mathcal{M}^{{target}} (x') \neq y$. If we are using a targeted attack then we want $\mathcal{M}^{{target}} (x') = t$ where $t$ is the targeted class different from the ground truth $t \neq c_{gt}$.
After creating the adversarial attack for each sample in $\mathcal{D}_{test}^{proxy}$ and $\mathcal{D}_{test}^{target}$ we evaluate the success rate of the attack and the success rate of the transferability onto the target model. To measure the success and robustness of the adversarial attack and its transferability, we define the following metrics:

\begin{itemize}
    \item \textbf{Attack Success Rate (ASR):} This is the proportion of adversarial examples successfully fooling the proxy model $\mathcal{M}_i^{{proxy}}$, defined as:
    \begin{equation}
        \text{ASR}_i = \frac{1}{|\mathcal{D}_{test}^{proxy}|} \sum_{x \in \mathcal{D}_{test}^{proxy}} \mathbb{I} \left[ \arg \max \mathcal{M}^{{proxy}}_i(x') \neq y \right],
    \end{equation}
    where $\mathbb{I}[\cdot]$ is the indicator function.
    \item \textbf{Transfer Success Rate (TSR):} To evaluate the transferability of adversarial examples generated using the proxy model $\mathcal{M}_i^{{proxy}}$ to the target model $\mathcal{M}^{{target}}$, we compute the fooling rate on the target model as:
    \begin{equation}
        \text{TSR}_i = \frac{1}{|\mathcal{D}_{test}^{target}|} \sum_{x \in \mathcal{D}_{test}^{target}} \mathbb{I} \left[ \arg \max \mathcal{M}^{{target}}(x') \neq y \right].
    \end{equation}
\end{itemize}

This setup allows us to explicitly quantify how the availability of diverse meta-information combinations explicitly impacts the adversarial transferability of the given model, thus highlighting the risks in the model-sharing practices. A visual depiction of this can be seen in \cref{fig:framework_adv_main}.
  
\subsection{Backbone Attack}

\begin{algorithm}[!t]
\caption{Backbone Attack}
\label{alg:backbone_pgd}
\KwIn{Model backbone $\mathcal{B}$, clean image $x_0$, perturbation bound $\epsilon$, step size $\alpha$, number of steps $T$, distance function $\mathcal{L}_{\text{cosine}}$, random start flag}
\KwOut{Adversarial image $x_{\text{adv}}$}

\textbf{Initialization:} \\
$x_{\text{adv}} \gets x_0$ 

\If{random start}{
$x_{\text{adv}} \gets x_{\text{adv}} + \text{Uniform}(-\epsilon, \epsilon)$\\
$x_{\text{adv}} \gets \text{Clip}(x_{\text{adv}}, 0, 1)$
}

\textbf{Fixed Original Image Representation:} \\
$z_0 \gets StopGrad(\mathcal{B}(x_0))$\\

\For{$t = 1$ \textbf{to} $T$}{
    \textbf{Forward Pass:} \\
    $z_{\text{adv}} \gets \mathcal{B}(x_{\text{adv}})$ \tcp{Adversarial image representation}
    
    \textbf{Compute Loss and Gradient:} \\
    $\mathcal{L} \gets 1 - \text{cos}(z_{\text{adv}}, z_0)$ \tcp{Distance loss}    
    $g \gets \nabla_{x_{\text{adv}}} \mathcal{L}$ \tcp{Gradient w.r.t $x_{\text{adv}}$}
    
    \textbf{Update Adversarial Image:} \\
    $x_{\text{adv}} \gets x_{\text{adv}} + \alpha \cdot \text{sign}(g)$ \tcp{PGD step}
    
    \textbf{Projection:} \\
    $\delta \gets \text{Clip}(x_{\text{adv}} - x_0, -\epsilon, \epsilon)$ \tcp{Project perturbation into $\ell_\infty$-ball}
    $x_{\text{adv}} \gets \text{Clip}(x_0 + \delta, 0, 1)$ \tcp{pixel range}
}

\Return $x_{\text{adv}}$
\end{algorithm}

To test the vulnerabilities associated with publicly available pre-trained feature extractors, we designed a naive \emph{backbone attack}, which only utilises the known backbone $\mathcal{B}$ of the model $\mathcal{M}^{{target}}$. The aim, similar to the prior paragraph, is to create an adversarial attack from the $\mathcal{B}$ to transfer towards the target model $\mathcal{M}^{{target}}$. To do this, we utilise a Projected Gradient Descent \citep[PGD]{}-based method, where the attack iteratively perturbs the input images in order to maximise the distance between the feature representations of the clean input and the adversarial input, as derived from the backbone $\mathcal{B}$. More formally, let $x$ and $\tilde{x}$ represent the clean input and adversarial input, respectively. The attack iteratively refines $\tilde{x}$ such that:
\begin{align}
    \tilde{x}_{t+1} = \text{Proj}_{\mathcal{S}} \left( \tilde{x}_t + \alpha \cdot \text{sign}\left( \nabla_{\tilde{x}_t} \mathcal{L}_{\mathcal{B}}(x, \tilde{x}_t) \right) \right),
\end{align}
where $\mathcal{L}_{\mathcal{B}}$ is the loss function defined to measure the distance between the feature representations of the clean and adversarial inputs. The backbone representations $f_{\mathcal{B}}$ are extracted as $f_{\mathcal{B}}(x) = \mathcal{B}(x)$, and the differentiable loss can be formulated as:
\begin{align}
    \mathcal{L}_{\mathcal{B}}(x, \tilde{x}) = 1 - \text{cos} \left( f_{\mathcal{B}}(x), f_{\mathcal{B}}(\tilde{x}) \right),
\end{align}
where $\text{cos}(\cdot, \cdot)$ represents the cosine similarity between the two feature vectors. To prevent gradient computation from propagating to the clean representation $f_{\mathcal{B}}(x)$, we utilize a stop-gradient operation $\tilde{f}_{\mathcal{B}}(x) = {SG}(f_{\mathcal{B}}(x))$.
The adversarial input $\tilde{x}$ is initialized with a random perturbation within the $\ell_{\infty}$ ball of radius $\epsilon$, and the updates are iteratively projected back onto this ball using the $\text{Proj}_{\mathcal{S}}$ operator:
\begin{align}
    \text{Proj}_{\mathcal{S}}(\tilde{x}) = \text{clip} \left( x + \delta, 0, 1 \right), \quad \\ \nonumber\text{where} \quad \delta = \text{clip} \left( \tilde{x} - x, -\epsilon, \epsilon \right).
\end{align}

The pseudo-code of the complete process can bee seen in \cref{alg:backbone_pgd}. In summary, the backbone attack focuses solely on the backbone $\mathcal{B}$, without requiring any knowledge of the full target model $\mathcal{M}^{{target}}$, thereby revealing vulnerabilities inherent to publicly available feature extractors.

\section{Experimental Setup}

\begin{figure*}[t!]
    \centering
    \includegraphics[clip=true,width=\textwidth]{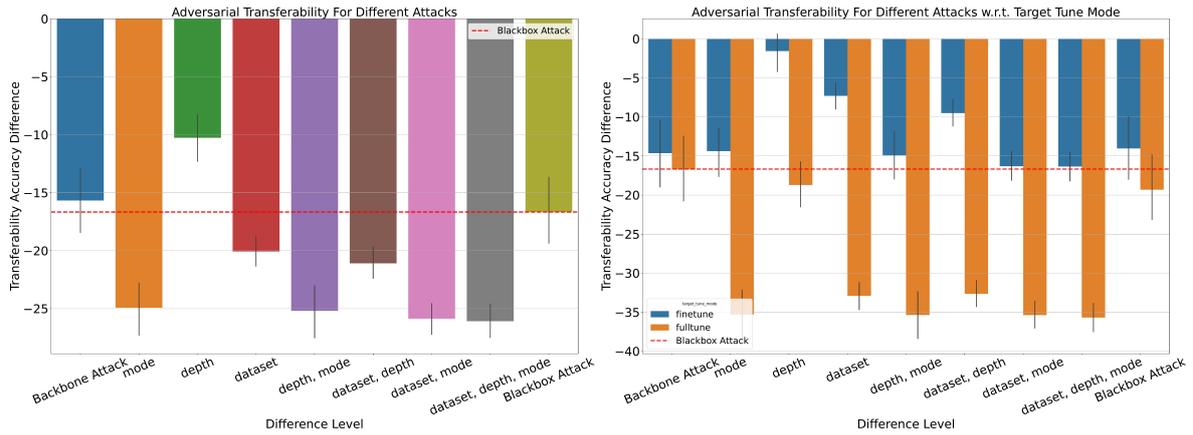}
    \caption{The figure depicts the impact of the \textbf{unavailability}, i.e. difference from the target model, with each possible meta-information combination on adversarial transferability during proxy attack construction and the backbone attack. The results show the average difference from the \emph{white-box} in transferability using PGD with a higher budget (left) and the segmentation w.r.t. in the target training mode (right).}
    \label{fig:whitebox_diff_per_levels}
\end{figure*}

\begin{figure}[t!]
    \centering
    \includegraphics[clip=true,width=\columnwidth]{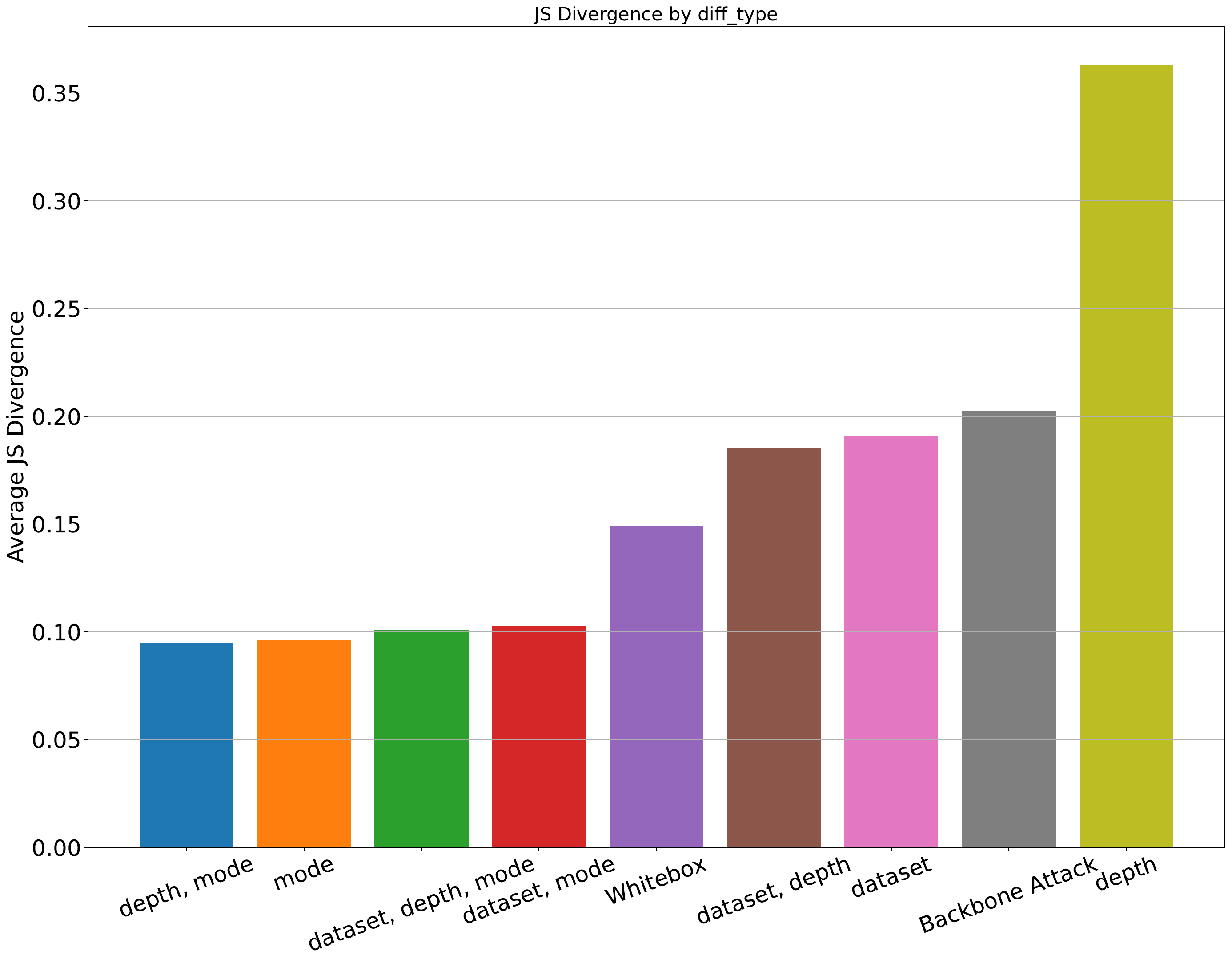}
    \caption{The figure breaks down impact of the \textbf{unavailability}, i.e. difference from the target model, of each possible meta-information combination on the change in the final decision-making of the model. Higher JS divergence implies a bigger change in the final classification of the sample.}
    \label{fig:whitebox_diff_per_levels_tuning}
\end{figure}

\subsection{Image classification datasets}

Through our study, we use $4$ datasets covering both classical and domain-specific classification benchmarks, such as CIFAR-10 and CIFAR-100 \citep{beyer2020we} and  Oxford-IIIT Pets \citep{parkhi2012cats}, Oxford Flowers-102 \citep{nilsback2008automated}. We train the proxy and target model variation on each one of the datasets using the recipe from \citep{kolesnikov2020big}, reproducing the state-of-the-art model performance results \citep{dosovitskiy2020image, yu2022coca,bruno2022efficient,foret2020sharpness}.

\subsection{Model variations}
We use $21$ different models tuned from $5$ architectures, $9$ self-supervised objectives and $3$ pre-training datasets. A detailed overview of these can be seen in \cref{tab:ssl_model_summary}.

\subsection{Model Fintuning Variations}
For training the proxy and target models, we employ two \emph{modes} of training $\mathcal{T}$, with full-tuning of the weights and with fine-tuning only the last added classification layers on top of the pre-trained backbone. We also define the depth of tuning $\mathcal{Z}$ as the number of classification layers added on top of the pre-trained backbone. We use $\{1, 3\}$ final layers corresponding to \emph{shallow} and \emph{deep} tuning settings.

\subsection{Adversarial Attacks}
To assess the \emph{white-box} adversarial attack success rate and the adversarial transferability from the proxy models, we employ FGSM \citep{DBLP:journals/corr/GoodfellowSS14} and PGD \citep{DBLP:conf/iclr/MadryMSTV18}. We use standard attack hyper-parameters introduced in parallel adversarial transferability studies \citep{DBLP:conf/wacv/WasedaNLNE23, DBLP:conf/iclr/NaseerR0KP22}. For a fair comparison, we also use the same values for our \emph{backbone-attack}.
To show that our results are consistent even with a higher computational budget, we report the results of PGD with $4$ times more iterations per sample for \emph{white-box}, proxy and \emph{backbone} attack experiments.
For \emph{black-box} experiments, we use the Square attack \citep{DBLP:conf/eccv/AndriushchenkoC20}, which is a query-efficient method that uses a random search through adversarial sample construction. To standardise the query budget for all architectures and simulate real-world constraints, we allow $10$ queries of the target model per sample.

\begin{figure*}[t!]
    \centering
    \includegraphics[clip=true,width=\textwidth]{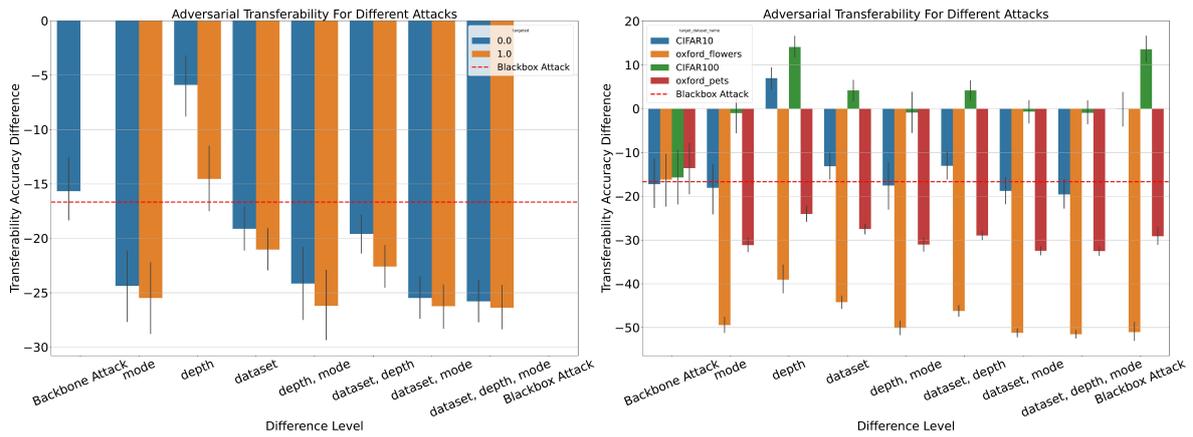}
    \caption{The figure depicts the impact of the \textbf{unavailability}, i.e. difference from the target model, of each possible meta-information combination on adversarial transferability during proxy attack construction and the backbone attack. The results show the average transferability for PGD with a higher budget for targeted vs untargeted attacks (left) and the segmentation w.r.t. the target training dataset (right).}
    \label{fig:whitebox_diff_per_levels_data_target}
\end{figure*}
\begin{table}[!t]
\centering
\resizebox{\columnwidth}{!}{%
\begin{tabular}{@{}lcccc@{}}
\toprule
                       & \multicolumn{2}{c}{\textbf{Original Entropy}} & \multicolumn{2}{c}{\textbf{Adversarial Entropy}} \\ \midrule
\textbf{Metadata type} & \textbf{F-Statistic}    & \textbf{P-Value}    & \textbf{F-Statistic} & \textbf{P-Value} \\
\textit{Target Tune Mode}  & 0.00    & 0.96 & 1238.7  & 0.0 \\
\textit{Proxy Tune Mode}   & 0.02    & 0.88 & 0.5     & 0.4 \\
\textit{Target Dataset}    & 2812.25 & 0.00 & 1184.1  & 0.0 \\
\textit{Proxy Dataset}     & 8.31    & 0.00 & 5.0     & 0.0 \\
\textit{Target Tune Depth} & 5.64    & 0.01 & 0.36    & 0   \\
\textit{Proxy Tune Depth}  & 0.08    & 0.77 & 0.00    & 0   \\ \bottomrule
\end{tabular}%
}
\caption{Variance analysis of entropy values across categorical variables. The table shows F-statistics and p-values for both original and adversarial entropy means. Significant p-values (p \textless 0.05) show notable variations in entropy across meta-information.}
\label{tab:variance_analysis}
\end{table}
\vspace{-10pt}

\section{Results}

\subsection{What meta-information matters}

To quantify the impact of each possible meta-information availability along with the backbone knowledge during adversarial attack construction, we compute the difference between the adversarial attack success rate (ASR) for the target model and the transferability success rate (TSR) from a proxy model, trained from the same backbone, with partial information. We report the results obtained with the PGD attack trained with higher iteration steps per sample as that is more representative for measuring the adversarial attack success in \emph{white-box} and \emph{grey-box} settings. Our results are consistent across the other attack types and variations.
\subsection{Which meta-information is important?}
Our results in \cref{fig:whitebox_diff_per_levels} show that the most significant performance decay compared to a \emph{white-box} attack performance occurs when the attacker is unaware of the \emph{mode} of the training of the target model, i.e. if it is trained with complete parameters or only tunes the last classification layers. The second most impactful knowledge for attack construction is the availability of the target tuning \emph{dataset}. The \emph{depth} of the tuning is the least important knowledge for obtaining a transferable attack. We further show in the right part of \cref{fig:whitebox_diff_per_levels} that models that finetune the last classification layers can be trivially exploited with transferable attacks, achieving results significantly better than strong black-box exploitation and closing white-box attack performance. It is, however, apparent that training all of the model weights substantially decreases the efficiency of proxy attacks, with almost no correlation towards meta-information availability.
We further show that our results remain consistent w.r.t. the choice of the dataset, and regardless if the adversarial attack is targeted or untargeted as seen in \cref{fig:whitebox_diff_per_levels_data_target}. It is interesting to note that for datasets with more domain-specific content, such as Oxford-IIIT Pets and Oxford Flowers-102, the effectiveness of the proxy attack dwindles, although these datasets are much less diverse compared to CIFAR-100. 

\subsection{Meta-information impacts the quality of adversarial attacks}

We also want to measure the effectiveness of the adversarial attack and the impact of meta-information on it by quantifying how the generated adversarial sample has sifted the decision-making of the model. To do this, we compute the entropy of the final softmax layer for each original sample and its adversarial counterpart and complete ANOVA variance analysis \citep{st1989analysis} of entropy distribution. This analysis, presented in \cref{tab:variance_analysis}, tests whether the means of entropies from original and adversarial images differ significantly across the groups of available meta-information. A perfect attack would produce a sample that does not majorly impact the entropy from the model. The analysis reveals that the 
target dataset, and tuning mode significantly influence entropy, particularly in adversarial scenarios. This finding suggests that while this meta-information aids in crafting effective adversarial samples, it also plays a critical role in amplifying entropy shifts, thereby making these adversarial samples more detectable.

To quantify the impact of the meta-information availability during attack construction on the decision-making of the model, we also compute the Jensen-Shannon Divergence \citep{menendez1997jensen} between the output softmax distributions of the model produced for original samples and their adversarial counterparts. High JS divergence suggests a strong attack, as the adversarial example causes a significant shift in the model's predicted probabilities, with minimal changes to the input sample. Our results show that not knowing the \emph{mode} of the target model training causes the most degradation in constructing successful adversarial samples with proxy attacks. The second most important fact is the choice of the target \emph{dataset}, while the \emph{depth} of the final classification layers does not seem to be impactful for creating adversarial samples. This reaffirms our findings from \cref{fig:whitebox_diff_per_levels} and \cref{fig:whitebox_diff_per_levels_tuning}, while also revealing a critical insight: proxy attacks, even when constructed without knowledge of the target model's \emph{dataset} or \emph{depth}, can generate adversarial samples that induce more pronounced distribution shifts than \emph{white-box} attacks. In other words, attackers do not require access to the training dataset or model classification depth to craft adversarial samples capable of significantly disrupting the target model’s decision-making process.

\subsection{Backbone-attacks}

To test the extent of the vulnerabilities that the knowledge of the pre-trained backbone can cause, we evaluate our naive exploitation method, \emph{backbone attack}, that utilizes only the pre-trained feature extractor for adversarial sample construction.
Our results in \cref{fig:whitebox_diff_per_levels} and \cref{fig:whitebox_diff_per_levels_data_target} show that \emph{backbone attacks} are highly effective at producing transferable adversarial samples regardless of the target model tuning \emph{mode}, \emph{dataset} or classification layer \emph{depth}. This naive attack shows significantly higher transferability compared to a strong \emph{black-box} attack with a sizeable query and iteration budget and almost all \emph{proxy attacks}. The results are consistent across all meta-information variations, showing that even a naive attack can exploit the target model vulnerabilities closely to a \emph{white-box} setting, given the knowledge of the pre-trained backbone. Moreover, from \cref{fig:whitebox_diff_per_levels_tuning}, we see that the adversarial samples produced from this attack, on average, cause a bigger shift in the model's decision-making compared to \emph{white-box attacks}. This indicates that backbone attacks amplify the uncertainty in the target model's predictions, making them more disruptive than conventional \emph{white-box} attacks, highlighting the inherent risks of sharing pre-trained backbones for public use. A concerning aspect of backbone attacks is their effectiveness in resource-constrained environments. Unlike black-box attacks, which often require extensive computation or iterative querying, backbone attacks can be executed with minimal resources, leveraging pre-trained models freely available in public repositories. This ease of implementation raises concerns, as it lowers the barrier for malicious actors to exploit adversarial vulnerabilities.

\subsection{Knowing weights vs Knowing everything but the weights}

To isolate the impact of pre-trained backbone knowledge in adversarial transferability, we train two sets of models from the same ResNet-50 SwAV backbone with identical meta-information variations but different batch sizes. This allows the production of two sets of models with matching training meta-information but varying weights; one set is chosen as the target, and the other as the proxy model. We aim to compare the adversarial transferability of the attacks from the set of proxies towards their matching targets with the backbone attacks. This allows us to simulate conditions where adversaries either know all meta-information but lack the weights or have access to the backbone weights alone.
%
\begin{figure}[t!]
    \centering
    \includegraphics[clip=true,width=\columnwidth]{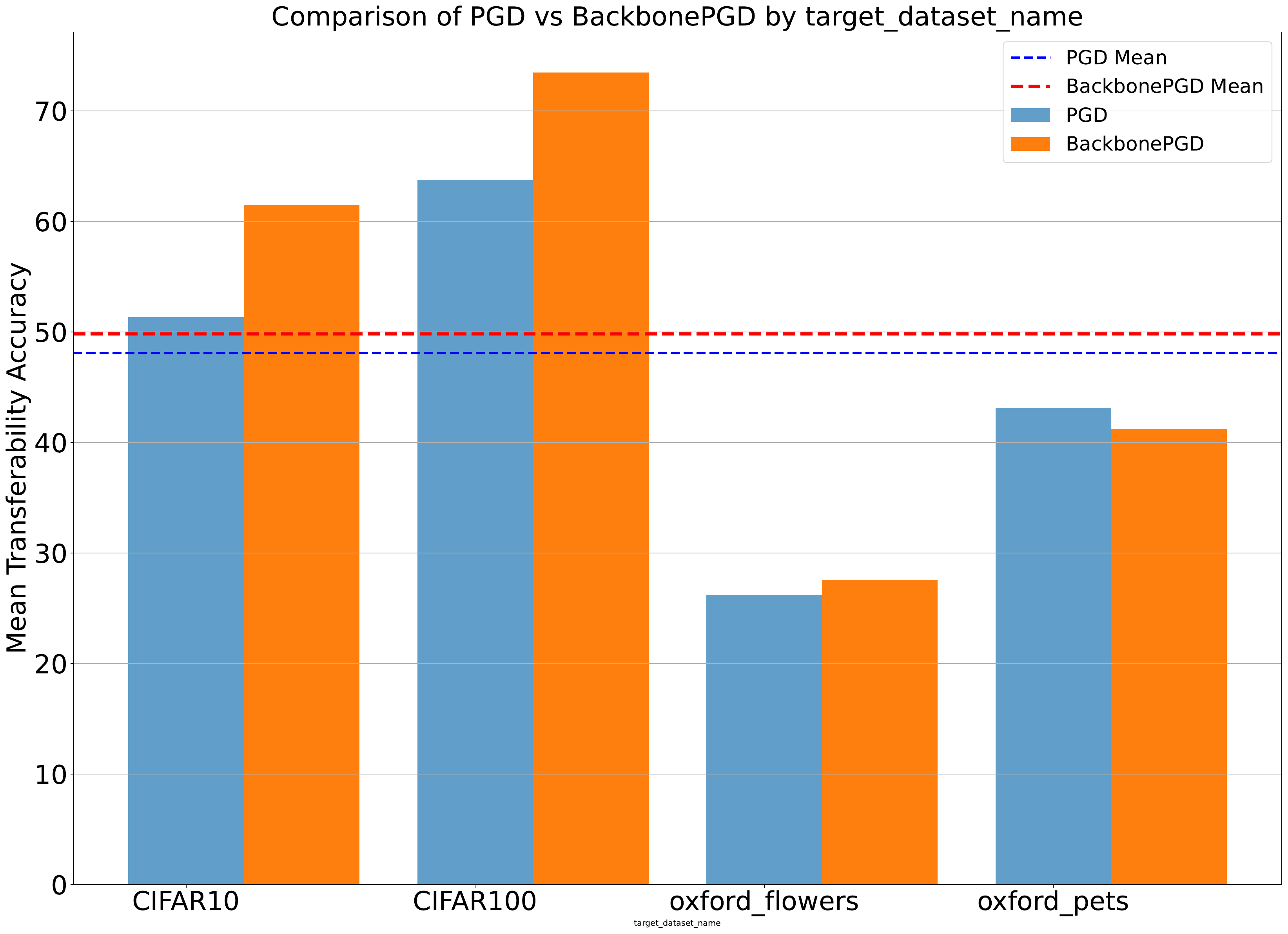}
    \caption{The figure shows scenarios where adversaries either know all meta-information but lack the weights or have access to the backbone weights (SwaV ResNet-50) alone. Knowledge of only the backbone is highlighted as \emph{BackbonePGD}.}
    \label{fig:swav_backbone_meta_info}
\end{figure}
Our results in \cref{fig:swav_backbone_meta_info} show that the knowledge of the pre-trained backbone is, on average, a stronger or at least an equivalent signal for producing adversarially transferable attacks compared to possessing all of the training meta-information without the knowledge of the weights. The results are consistent across all of the datasets, with domain-specific datasets showing marginal differences in adversarial transferability between the two scenarios. This means that possessing information about only the target model backbone is equivalent to knowing all of the training meta-information for constructing transferable adversarial samples.

\section{Conclusions}

In this paper, we investigated the vulnerabilities of machine vision models fine-tuned from publicly available pre-trained backbones under a novel \emph{grey-box} adversarial setting. Through an extensive evaluation framework, including over 20,000 adversarial transferability comparisons, we measured the effect of varying levels of training meta-information availability for constructing transferable adversarial attacks.
We also introduced a naive \emph{backbone attack} method, showing that access to backbone weights is sufficient for obtaining adversarial attacks significantly better than query-based \emph{black-box} settings and approaching white-box performance. We found that attacks crafted using only the backbone weights often induce more substantial shifts in the model's decision-making than traditional white-box attacks. 
We demonstrated that access to backbone weights is equivalent in effectiveness to possessing all meta-information about the target model, making public backbones a critical security concern.
Our results highlight significant security risks associated with sharing pre-trained backbones, as they enable attackers to craft highly effective adversarial samples, even with minimal additional information. These findings underscore the need for stricter practices in sharing and deploying pre-trained backbones to mitigate the inherent vulnerabilities exposed by adversarial transferability.

\part{Reasoning Inconsistencies from Data}
\chapter{Topic-Guided Sampling For Data-Efficient Multi-Domain Stance Detection}
\label{chap:tested}

\section{Introduction}
\label{sec:intro}

The goal of stance detection is to identify the viewpoint expressed by an author within a piece of text towards a designated topic \citep{mohammad2016semeval}. Such analyses can be used in a variety of domains ranging from identifying claims within political or ideological debates~\citep{somasundaran2010recognizing, thomas2006get}, identifying mis- and disinformation \citep{hanselowski2018retrospective,hardalov2021survey}, public health policymaking \citep{glandt2021stance,hossain2020covidlies, osnabrugge2023cross}, news recommendation \citep{reuver2021no} to investigating attitudes voiced on social media  \citep{qazvinian2011rumor,augenstein-etal-2016-stance,conforti2020will}. However, in most domains, and even more so for cross-domain stance detection, the exact formalisation of the task gets blurry, with varying label sets and their corresponding definitions, data collection protocols and available annotations. Furthermore, this is accompanied by significant changes in the topic-specific vocabulary \citep{somasundaran2010recognizing,wei2019modeling}, text style \citep{pomerleau2017fake, ferreira2016emergent} and topics mentioned either explicitly \citep{qazvinian2011rumor, walker2012corpus} or implicitly \citep{hasan2013stance,gorrell2019semeval}. Recently, a benchmark of $16$ datasets \citep{hardalov2021cross} covering a variety of domains and topics has been proposed for testing stance detection models across multiple domains. It must be noted that these datasets are highly imbalanced, with an imbalanced label distribution between the covered topics, i.e. inter-topic and within each topic, i.e. per-topic, as can be seen in \autoref{fig:imbalanced} and \autoref{fig:label_dist}. This further complicates the creation of a robust stance detection classifier.

\begin{figure}[t]
\centering
\includegraphics[width=\textwidth]{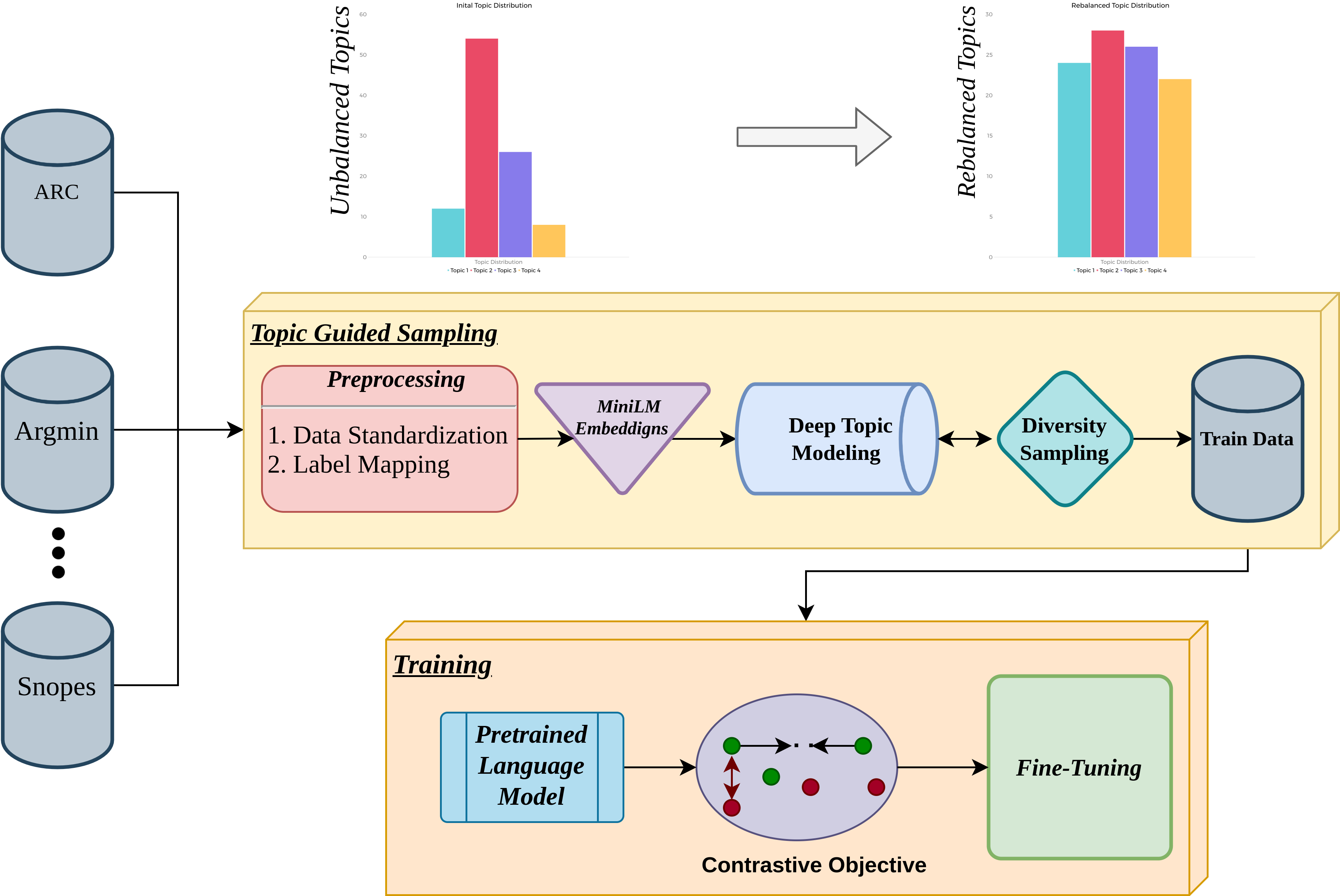}
\caption{The two components of TESTED: Topic Guided Sampling (top) and training with contrastive objective (bottom).}
\label{fig:TESTED}
\end{figure}



Given the inherent skew present within the dataset and variances within each domain, we propose a topic-guided diversity sampling method, which produces a data-efficient representative subset while mitigating label imbalances. These samples are used for fine-tuning a Pre-trained Language Model (PLM), using a contrastive learning objective to create a robust stance detection model. These two components form our \textbf{T}opic \textbf{E}fficient \textbf{St}anc\textbf{E} \textbf{D}etection	(TESTED) framework, as seen in \autoref{fig:TESTED}, and are analysed separately to pinpoint the factors impacting model performance and robustness. We test our method on the multi-domain stance detection benchmark by \citet{hardalov2021cross}, achieving state-of-the-art results with both in-domain, i.e. all topics seen and out-of-domain, i.e. unseen topics evaluations. Note though that TESTED could be applied to any text classification setting.

In summary, our \textbf{contributions} are:
\begin{itemize}[noitemsep]
    \item We propose a novel framework (TESTED) for predicting stances across various domains, with data-efficient sampling and contrastive learning objective;
    \item Our proposed method achieves SOTA results both in-domain and out-of-domain;
    \item Our analysis shows that our topic-guided sampling method mitigates dataset imbalances while accounting for better performance than other sampling techniques; 
    \item The analysis shows that the contrastive learning objective boosts the ability of the classifier to differentiate varying topics and stances.
\end{itemize}

\section{Related Work}
\label{sec:related_work}

\subsection{Stance Detection} is an NLP task which aims to identify an author's attitude towards a particular topic or claim. The task has been widely explored in the context of mis- and disinformation detection \citep{ferreira2016emergent,hanselowski2018retrospective,ZUBIAGA2018273,hardalov2021survey}, sentiment analysis \citep{mohammad2017stance, aldayel2019your} and argument mining \citep{boltuvzic2014back,sobhani2015argumentation,wang2019survey}. Most papers formally define stance detection as a pairwise sequence classification where stance targets are provided \citep{kuccuk2020stance}. However, with the emergence of different data sources, ranging from debating platforms \citep{somasundaran2010recognizing,hasan2014you,aharoni2014benchmark} to social media \citep{mohammad2016semeval,gorrell2019semeval}, and new applications \citep{zubiaga2018detection,hardalov2021survey}, this formal definition has been subject to variations w.r.t. the label dictionary inferred for the task.

Previous research has predominantly focused on a specific dataset or domain of interest, outside of a few exceptions like multi-target \citep{sobhani2017dataset,wei2018multi} and cross-lingual \citep{hardalov2022few} stance detection. In contrast, our work focuses on multi-domain stance detection, while evaluating in- and out-of-domain on a $16$ dataset benchmark with state-of-the-art baselines \citep{hardalov2021cross}.

\subsection{Topic Sampling}

Our line of research is closely associated with diversity \citep{ren2021survey} and importance \citep{beygelzimer2009importance} sampling and their applications in natural language processing \citep{zhu2008active,zhou2020informed}. Clustering-based sampling approaches have been used for  automatic speech recognition \citep{syed2016supervised}, image classification \citep{ranganathan2017deep,yan2022mitigating} and semi-supervised active learning \citep{buchert2022exploiting} with limited use for textual data \citep{yang2014active} through topic modelling \citep{blei2003latent}. This research proposes an importance-weighted topic-guided diversity sampling method that utilises deep topic models, for mitigating inherent imbalances present in the data, while preserving relevant examples.

\subsection{Contrastive Learning}  has been used for tasks where the expected feature representations should be able to differentiate between similar and divergent inputs \citep{liu2021self,10.1145/3561970}. Such methods have been used for image classification \citep{khosla2020supervised}, captioning \citep{dai2017contrastive} and textual representations \citep{giorgi2020declutr,jaiswal2020survey,ostendorff-etal-2022-neighborhood}. The diversity of topics \citep{qazvinian2011rumor,walker2012corpus,hasan2013stance}, vocabulary \citep{somasundaran2010recognizing,wei2019modeling} and expression styles \citep{pomerleau2017fake} common for stance detection can be tackled with contrastive objectives, as seen for similar sentence embedding and classification tasks \citep{gao2021simcse, yan2021consert}. 

\section{Datasets}
\label{sec:datasets}
Our study uses an existing multi-domain dataset benchmark \citep{hardalov2021cross}, consisting of $16$ individual datasets split into four source groups: \textit{Debates, News, Social Media, Various}. The categories include datasets about debating and political claims including arc \citep{hanselowski2018retrospective,habernal2017argument}, iac1 \citep{walker2012corpus}, perspectum \citep{chen2019seeing}, poldeb \citep{somasundaran2010recognizing}, scd \citep{hasan2013stance}, news like emergent \citep{ferreira2016emergent}, fnc1 \citep{pomerleau2017fake}, snopes \citep{hanselowski-etal-2019-richly}, social media like mtsd \cite{sobhani2017dataset}, rumour \citep{qazvinian2011rumor}, semeval2016t6 \citep{mohammad2016semeval}, semeval2019t7 \citep{gorrell2019semeval}, wtwt \citep{conforti2020will} and datasets that cover a variety of diverse topics like argmin \citep{stab-etal-2018-cross}, ibmcs \citep{bar-haim_stance_2017} and vast \citep{allaway-mckeown-2020-zero}. Overall statistics for all of the datasets can be seen in \autoref{Appendix:data}.

\subsection{Data Standardisation}

As the above-mentioned stance datasets from different domains possess different label inventories, the stance detection benchmark by \citet{hardalov2021cross} introduce a mapping strategy to make the class inventory homogeneous. We adopt that same mapping for a fair comparison with prior work, shown in Appendix \ref{Appendix:data}.

\section{Methods}

Our goal is to create a stance detection method that performs strongly on the topics known during training and can generalize to unseen topics. The benchmark by \citet{hardalov2021cross} consisting of $16$ datasets is highly imbalanced w.r.t the inter-topic frequency and per-topic label distribution, as seen in \autoref{fig:imbalanced}. 

These limitations necessitate a novel experimental pipeline. The first component of the pipeline we propose is an importance-weighted topic-guided diversity sampling method that allows the creation of supervised training sets while mitigating the inherent imbalances in the data. We then create a stance detection model by fine-tuning a Pre-trained Language Model (PLM) using a contrastive objective.


\label{sec:methods}

\subsection{Topic-Efficient Sampling}
\label{sec:methods:topic}

We follow the setting in prior work on data-efficient sampling \cite{buchert2022exploiting,yan2022mitigating}, framing the task as a selection process between multi-domain examples w.r.t the theme discussed within the text and its stance. This means that given a set of datasets $\mathcal{D} = (\mathcal{D}_1, \dots \mathcal{D}_n)$ with their designated documents $\mathcal{D}_i = (d_{i}^1, \dots d_{i}^m)$, we wish to select a set of diverse representative examples $\mathcal{D}_{{train}}$, that are balanced w.r.t the provided topics $\mathcal{T} = (t_1, \dots t_q)$ and stance labels $L = (l_1, \dots l_k)$. 


\subsection{Diversity Sampling via Topic Modeling}

We thus opt for using topic modelling to produce a supervised subset from all multi-domain datasets. Selecting annotated examples during task-specific fine-tuning is a challenging task \cite{shao2019learning}, explored extensively within active learning research \cite{hino2020active, konyushkova2017learning}. Random sampling can lead to poor generalization and knowledge transfer within the novel problem domain \cite{das2021importance, perez2021true}. To mitigate the inconsistency caused by choosing suboptimal examples, we propose using deep unsupervised topic models, which allow us to sample relevant examples for each topic of interest. We further enhance the model with an importance-weighted diverse example selection process \cite{shao2019learning,yang2015multi} within the relevant examples generated by the topic model. The diversity maximisation sampling is modeled similarly to \citet{yang2015multi}.

The topic model we train is based on the technique proposed by \citet{angelov2020top2vec} that tries to find topic vectors while jointly learning document and word semantic embeddings. The topic model is initialized with weights from the \textit{all-MiniLM-L6} PLM, which has a strong performance on sentence embedding benchmarks \citep{wang2020minilm}. It is shown that learning unsupervised topics in this fashion maximizes the total information gained, about all texts $\mathcal{D}$ when described by all words $\mathcal{W}$.

    \[\mathcal{I}(\mathcal{D}, \mathcal{W})=\sum_{d \in \mathcal{D}} \sum_{w \in \mathcal{W}} P(d, w) \log \left(\frac{P(d, w)}{P(d) P(w)}\right)\]

This characteristic is handy for finding relevant samples across varying topics, allowing us to search within the learned documents $d_i$. We train a deep topic model $\mathcal{M}_{topic}$ using multi-domain data $\mathcal{D}$ and obtain topic clusters $\mathcal{C} = (\mathcal{C}_i, \dots \mathcal{C}_t)$, where $\lvert \mathcal{C} \rvert = t$ is the number of topic clusters. We obtain the vector representation for $\forall d_i$ from the tuned PLM embeddings $\mathcal{E} = (e_1, \dots e_m)$ in $\mathcal{M}_{topic}$, while iteratively traversing through the clusters $\mathcal{C}_i \in \mathcal{C}$.

\begin{algorithm}[!t]
\caption{Topic Efficient Sampling}\label{algo:sampling}

\KwIn{$S \geq 0$ \tcp*[h]{Sampling Threshold}}

\KwIn{${Avg} \in \{{moving}, {exp} \}$}

\KwOut{$\lvert \mathcal{C} \rvert > 0$}

$\mathcal{D}_{{train}} \gets \{\}$\;

$I \gets \{\frac{\lvert \mathcal{C}_1 \rvert}{\sum_{\mathcal{C}_i \in \mathcal{C}} \mathcal{C}_i}, \dots, \frac{\lvert \mathcal{C}_t \rvert}{\sum_{\mathcal{C}_i \in \mathcal{C}} \mathcal{C}_i} \}$ \tcp*[h]{Cluster Importances}\;

\For{$\mathcal{C}_i \in \mathcal{C}$}{

    $\mathcal{E}_i \gets \{{PLM}(d_i^1), \dots \} = \{\mathbf{e}_i^1, \dots, \mathbf{e}_i^m \}$\;

    $s_i \gets \max(1, S \cdot I_i)$ \tcp*[h]{Threshold per cluster}\;

    $j \gets 0$\;

    ${{cent}}_0 \gets \frac{\sum_{\mathbf{e}_i \in \mathcal{E}} \mathbf{e}_i}{\lvert \mathcal{E} \rvert}$ \tcp*[h]{Centroid of the cluster}\;

    \While{$j \leq s_i$}{

        ${{sim}} \gets \frac{\langle\mathcal{E}, {cent}\rangle}{\|\mathcal{E}\|\|{cent}\|}$ \tcp*[h]{Similarity Ranking}\;

        ${sample} \gets \argsort({{sim}}, {Ascending})[0]$ \tcp*[h]{Take the sample most diverse from the centroid}\;

        $\mathcal{D}_{{train}} \gets \mathcal{D}_{{train}} \cup {sample}$\;

        $j \gets j + 1$\;

        ${{cent}}_j \gets 
        \begin{cases}  
            \alpha \cdot \mathbf{e}_{sample} + (1-\alpha) \cdot {cent}_{j-1} & \text{if } {exp}\\ 
            \frac{(j-1)}{j} \cdot {{cent}}_{j-1} + \frac{\mathbf{e}_{sample}}{j} & \text{if } {moving}
        \end{cases}$ \tcp*[h]{Centroid update w.r.t. sampled data}\;

    }

}

\Return $\mathcal{D}_{{train}}$\;

\end{algorithm}

Our sampling process selects increasingly more diverse samples after each iteration. This search within the relevant examples is presented in \autoref{algo:sampling}.
This algorithm selects a set of diverse samples from the given multi-domain datasets $\mathcal{D}$, using the clusters from a deep topic model $\mathcal{M}_{topic}$ and the sentence embeddings $\mathcal{E}$ of the sentences as a basis for comparison. The algorithm starts by selecting a random sentence as the first diverse sample and uses this sentence to calculate a ``centroid'' embedding. It then iteratively selects the next most dissimilar sentence to the current centroid, until the desired number of diverse samples is obtained.


\subsection{Topic-Guided Stance Detection}\label{sec:methods:stance}

\subsection{Task Formalization}

Given the topic, $t_i$ for each document $d_i$ in the generated set $\mathcal{D}_{train}$ we aim to classify the stance expressed within that text towards the topic. For a fair comparison with prior work, we use the label mapping from the previous multi-domain benchmark \citep{hardalov2021cross} and standardise the original labels $L$ into a five-way stance classification setting, $S = \{\text{Positive, Negative, Discuss, Other, Neutral}\}$. Stance detection can be generalized as pairwise sequence classification, where a model learns a mapping $f:(d_i, t_i) \to S$. We combine the textual sequences with the stance labels to learn this mapping. 
The combination is implemented using a simple prompt commonly used for NLI tasks \cite{lan2019albert, raffel2020exploring,hambardzumyan2021warp}, where the textual sequence becomes the premise and the topic the hypothesis.

\begin{equation*} \label{prompt:combination}
\begin{split}
& \text{[CLS] premise: {premise}} \\ 
& \text{hypothesis: {topic} [EOS]}
\end{split}
\end{equation*}

The result of this process is a supervised dataset for stance prediction $\mathcal{D}_{train} = ((Prompt(d_1,t_1), s_1) \dots (Prompt(d_n,t_n), s_n))$ where $\forall s_i \in S$. This method allows for data-efficient sampling, as we at most sample $10\%$ of the data while preserving the diversity and relevance of the selected samples. The versatility of the method allows \emph{TESTED} to be applied to any text classification setting.

\subsection{Tuning with a Contrastive Objective}

After obtaining the multi-domain supervised training set $\mathcal{D}_{train}$, we decided to leverage the robustness of PLMs, based on a transformer architecture \cite{vaswani2017attention} and fine-tune on $\mathcal{D}_{train}$ with a single classification head. This effectively allows us to transfer the knowledge embedded within the PLM onto our problem domain. For standard fine-tuning of the stance detection model $\mathcal{M}_{stance}$ we use cross-entropy as our initial loss:
\begin{equation}
\mathcal{L}_{CE} = -\sum \limits_{i \in S}  y_i \log \left(\mathcal{M}_{stance}(d_i)\right)
\end{equation}

Here $y_i$ is the ground truth label. However, as we operate in a multi-domain setting, with variations in writing vocabulary, style and covered topics, it is necessary to train a model where similar sentences have a homogeneous representation within the embedding space while keeping contrastive pairs distant. We propose a new contrastive objective based on the \textit{cosine} distance between the samples to accomplish this. In each training batch $B=(d_1, \dots d_b)$, we create a matrix of contrastive pairs $\mathcal{P} \in \mathcal{R}^{b \times b}$, where $\forall i,j =\overline{1,b}$, $\mathcal{P}_{ij} =1$ if $i$-th and $j$-th examples share the same label and $-1$ otherwise. The matrices can be precomputed during dataset creation, thus not adding to the computational complexity of the training process. We formulate our pairwise contrastive objective $\mathcal{L}_{CL}(x_i, x_j, \mathcal{P}_{ij})$ using matrix $\mathcal{P}$.

\begin{align}
\mathcal{L}_{CL}=\resizebox{.72\hsize}{!}{$ \begin{cases} e(1- e^{\cos \left(x_i, x_j\right) - 1)}, \mathcal{P}_{ij}=1 \\ e^{\max \left(0, \cos \left(x_i, x_j\right)-\beta\right)}-1, \mathcal{P}_{ij}=-1\end{cases}$
}
\end{align}

Here $x_i,x_j$ are the vector representations of examples $d_i,d_j$. The loss is similar to cosine embedding loss and soft triplet loss
 \citep{barz2020deep,qian2019softtriple}; however, it penalizes the opposing pairs harsher because of the exponential nature, but does not suffer from computational instability as the values are bounded in the range $[0,e - \frac{1}{e}]$. 
 The final loss is:  
 \begin{align}
     \mathcal{L} = \mathcal{L}_{CE} + \mathcal{L}_{CL}
 \end{align}

We use the fine-tuning method from \citet{mosbach2020stability, liu2019roberta} to avoid the instability caused by catastrophic forgetting, small-sized fine-tuning datasets or optimization difficulties.

\section{Experimental Setup}
\label{sec:experiments}
\setlength{\tabcolsep}{3pt}
\begin{table*}
\centering
\begin{adjustbox}{max width=\textwidth}
\begin{tabular}{@{}lc|ccccc|ccc|ccccc|ccc@{}}
\toprule
 &
  $\mathrm{F}_1$ avg. &
  \rotatebox{45}{arc} &
  \rotatebox{45}{iac1} &
  \rotatebox{45}{perspectrum} &
  \rotatebox{45}{poldeb} &
  \rotatebox{45}{scd} &
  \rotatebox{45}{emergent} &
  \rotatebox{45}{fnc1} &
  \rotatebox{45}{snopes} &
  \rotatebox{45}{mtsd} &
  \rotatebox{45}{rumor} &
  \rotatebox{45}{semeval16} &
  \rotatebox{45}{semeval19} &
  \rotatebox{45}{wtwt} &
  \rotatebox{45}{argmin} &
  \rotatebox{45}{ibmcs} &
  \rotatebox{45}{vast} \\ \midrule
Majority class baseline &
  27.60 &
  21.45 &
  21.27 &
  34.66 &
  39.38 &
  35.30 &
  21.30 &
  20.96 &
  43.98 &
  19.49 &
  25.15 &
  24.27 &
  22.34 &
  15.91 &
  33.83 &
  34.06 &
  17.19 \\
Random baseline &
  35.19 &
  18.50 &
  30.66 &
  50.06 &
  48.67 &
  50.08 &
  31.83 &
  18.64 &
  45.49 &
  33.15 &
  20.43 &
  31.11 &
  17.02 &
  20.01 &
  49.94 &
  50.08 &
  33.25 \\
MoLE &
  65.55 &
  63.17 &
  38.50 &
  85.27 &
  50.76 &
  \textbf{65.91} &
  \textbf{83.74} &
  75.82 &
  75.07 &
  \textbf{65.08} &
  \textbf{67.24} &
  \textbf{70.05} &
  57.78 &
  68.37 &
  \textbf{63.73} &
  79.38 &
  38.92 \\ \midrule
TESTED (Our Model) &
  \textbf{69.12} &
  \textbf{64.82} &
  \textbf{56.97} &
  \textbf{83.11} &
  \textbf{52.76} &
  64.71 &
  82.10 &
  \textbf{83.17} &
  \textbf{78.61} &
  63.96 &
  66.58 &
  69.91 &
  \textbf{58.72} &
  \textbf{70.98} &
  62.79 &
  \textbf{88.06} &
  \textbf{57.47} \\
Topic $\rightarrow$ Random Sampling &
  61.14 &
  53.92 &
  42.59 &
  77.68 &
  44.08 &
  52.54 &
  67.55 &
  75.60 &
  72.67 &
  56.35 &
  59.08 &
  66.88 &
  57.28 &
  69.32 &
  52.02 &
  76.93 &
  53.80 \\
Topic $\rightarrow$ Stratified Sampling &
  64.01 &
  50.27 &
  51.57 &
  77.78 &
  46.67 &
  62.13 &
  79.00 &
  77.90 &
  76.44 &
  61.50 &
  64.92 &
  68.45 &
  51.96 &
  69.47 &
  56.76 &
  78.30 &
  51.16 \\
- Contrastive Objective &
  65.63 &
  61.11 &
  55.50 &
  81.85 &
  43.81 &
  63.04 &
  80.84 &
  79.05 &
  73.43 &
  62.18 &
  61.57 &
  60.17 &
  56.06 &
  68.79 &
  59.51 &
  86.94 &
  56.35 \\
\begin{tabular}[c]{@{}l@{}}Topic Sampling $\rightarrow$ Stratified \\ - Contrastive Loss\end{tabular} &
  63.24 &
  60.98 &
  49.17 &
  77.85 &
  45.54 &
  58.23 &
  77.36 &
  75.80 &
  74.77 &
  60.85 &
  63.69 &
  62.59 &
  54.74 &
  62.85 &
  53.67 &
  86.04 &
  47.72 \\ \bottomrule
\end{tabular}%
\end{adjustbox}
\caption{In-domain results reported with macro averaged F1, averaged over experiments. In lines under {TESTED}, we replace (for Sampling) $(\rightarrow)$ or remove (for loss) $(-)$, the comprising components. }
\label{tab:in_domiain_results}
\end{table*}
\begin{table*}
\centering
\begin{adjustbox}{max width=\textwidth}
\begin{tabular}{@{}lc|ccccc|ccc|ccccc|ccc@{}}
\toprule
 &
  $\mathrm{F_1}$ avg. &
  \rotatebox{45}{arc} &
  \rotatebox{45}{iac1} &
  \rotatebox{45}{perspectrum} &
  \rotatebox{45}{poldeb} &
  \rotatebox{45}{scd} &
  \rotatebox{45}{emergent} &
  \rotatebox{45}{fnc1} &
  \rotatebox{45}{snopes} &
  \rotatebox{45}{mtsd} &
  \rotatebox{45}{rumor} &
  \rotatebox{45}{semeval16} &
  \rotatebox{45}{semeval19} &
  \rotatebox{45}{wtwt} &
  \rotatebox{45}{argmin} &
  \rotatebox{45}{ibmcs} &
  \rotatebox{45}{vast} \\ \midrule
MoLE w/ Hard Mapping &
  32.78 &
  25.29 &
  35.15 &
  29.55 &
  22.80 &
  16.13 &
  58.49 &
  47.05 &
  29.28 &
  23.34 &
  32.93 &
  37.01 &
  21.85 &
  16.10 &
  34.16 &
  72.93 &
  22.89 \\
MoLE w/ Weak Mapping &
  49.20 &
  \textbf{51.81} &
  38.97 &
  58.48 &
  47.23 &
  53.96 &
  \textbf{82.07} &
  51.57 &
  56.97 &
  40.13 &
  \textbf{51.29} &
  36.31 &
  31.75 &
  22.75 &
  50.71 &
  75.69 &
  37.15 \\
MoLE w/Soft Mapping &
  46.56 &
  48.31 &
  32.21 &
  62.73 &
  54.19 &
  51.97 &
  46.86 &
  57.31 &
  53.58 &
  37.88 &
  44.46 &
  36.77 &
  28.92 &
  28.97 &
  57.78 &
  72.11 &
  30.96 \\ \midrule
TESTED &
  \textbf{59.41} &
  50.80 &
  \textbf{57.95} &
  \textbf{78.95} &
  \textbf{55.62} &
  \textbf{55.23} &
  80.80 &
  \textbf{72.51} &
  \textbf{61.70} &
  \textbf{55.49} &
  39.44 &
  \textbf{40.54} &
  \textbf{46.28} &
  \textbf{42.77} &
  \textbf{72.07} &
  \textbf{86.19} &
  \textbf{54.33} \\ \midrule
Topic Sampling $\rightarrow$ Stratified &
  50.38 &
  38.47 &
  46.54 &
  69.75 &
  50.54 &
  51.37 &
  68.25 &
  59.41 &
  51.64 &
  48.24 &
  28.04 &
  29.69 &
  34.97 &
  38.13 &
  63.83 &
  83.20 &
  44.06 \\
- Contrastive Loss &
  54.63 &
  47.96 &
  50.09 &
  76.51 &
  47.49 &
  51.93 &
  75.22 &
  68.69 &
  56.53 &
  49.47 &
  33.95 &
  37.96 &
  44.10 &
  39.56 &
  63.09 &
  83.59 &
  48.03 \\ \bottomrule
\end{tabular}%
\end{adjustbox}
\caption{Out-of-domain results with macro averaged F1. In lines under {TESTED}, we replace (for Sampling) $(\rightarrow)$ or remove (for loss) $(-)$, the comprising components. Results for MoLE w/Soft Mapping are aggregated across with best per-embedding results present in the study \citep{hardalov2021cross}.}
\label{tab:ood_results}
\end{table*}
\setlength{\tabcolsep}{6pt}

\subsection{Evaluation}
We evaluate our method on the $16$ dataset multi-domain benchmark and the baselines proposed by \citet{hardalov2021cross}. To directly compare with prior work, we use the same set of evaluation metrics: macro averaged F1, precision, recall and accuracy.

\subsection{Model Details}

We explore several PLM transformer architectures within our training and classification pipelines in order to evaluate the stability of the proposed technique. We opt to finetune a pre-trained \textit{roberta-large} architecture \cite{liu2019roberta, conneau2019unsupervised}. For fine-tuning, we use the method introduced by \citet{mosbach2020stability}, by adding a linear warmup on the initial $10\%$ of the iteration raising the learning rate to $2e^{-5}$ and decreasing it to $0$ afterwards. We use a weight decay of $\lambda = 0.01$ and train for $3$ epochs with global gradient clipping on the stance detection task. We further show that learning for longer epochs does not yield sizeable improvement over the initial fine-tuning. The optimizer used for experimentation is an AdamW \citep{loshchilov2017decoupled} with a bias correction component added to stabilise the experimentation \citep{mosbach2020stability}. 

\subsection{Topic Efficiency}

Recall that we introduce a topic-guided diversity sampling method within \textbf{\textit{TESTED}}, which allows us to pick relevant samples per topic and class for further fine-tuning. We evaluate its effectiveness by fine-tuning PLMs on the examples it generates and comparing it with training on a random stratified sample of the same size. 

\section{Results and Analysis}

In this section, we discuss and analyze our results, while comparing the performance of the method against the current state-of-the-art \citep{hardalov2021cross} and providing an analysis of the topic efficient sampling and the contrastive objective.

\subsection{Stance Detection}

\subsection{In-domain}

We train on our topic-efficient subset $\mathcal{D}_{train}$ and test the method on all datasets $\mathcal{D}$ in the multi-domain benchmark. Our method TESTED is compared to MoLE \citep{hardalov2021cross}, a strong baseline and the current state-of-the-art on the benchmark. The results, presented in \autoref{tab:in_domiain_results}, show that TESTED has the highest average performance on in-domain experiments with an increase of $3.5$ F1 points over MoLE, all while using $\leq10\%$ of the amount of training data in our subset $\mathcal{D}_{train}$ sampled from the whole dataset $\mathcal{D}$. Our method is able to outperform all the baselines on $10$ out of $16$ datasets. On the remaining $6$ datasets the maximum absolute difference between TESTED and MoLE is $1.1$ points in F1. 
We also present ablations for TESTED, by replacing the proposed sampling method with other alternatives, removing the contrastive objective or both simultaneously. Replacing Topic Efficient sampling with either \emph{Random} or \emph{Stratified} selections deteriorates the results for all datasets with an average decrease of $8$ and $5$ F1 points, respectively. We attribute this to the inability of other sampling techniques to maintain inter-topic distribution and per-topic label distributions balanced while selecting diverse samples. We further analyse how our sampling technique tackles these tasks in \autoref{subsec:imbalance}. We also see that removing the contrastive loss also results in a deteriorated performance across all the datasets with an average decrease of $3$ F1 points. In particular, we see a more significant decrease in datasets with similar topics and textual expressions, i.e. \emph{poldeb} and \emph{semeval16}, meaning that learning to differentiate between contrastive pairs is essential within this task. We analyse the effect of the contrastive training objective further in \autoref{subsec:contrastive}.


\subsection{Out-of-domain}

In the out-of-domain evaluation, we leave one dataset out of the training process for subsequent testing. We present the results of TESTED in \autoref{tab:ood_results}, showing that it is able to overperform over the previous state-of-the-art significantly. The metrics in each column of \autoref{tab:ood_results} show the results for each dataset held out from training and only evaluated on. Our method records an increased performance on $13$ of $16$ datasets, with an averaged increase of $10.2$ F1 points over MoLE, which is a significantly more pronounced increase than for the in-domain setting, demonstrating that the strength of TESTED lies in better out-of-domain generalisation. We can also confirm that replacing the sampling technique or removing the contrastive loss results in lower performance across all datasets, with decreases of $9$ and $5$ F1 points respectively. This effect is even more pronounced compared to the in-domain experiments, as adapting to unseen domains and topics is facilitated by diverse samples with a balanced label distribution.

\subsection{Imbalance Mitigation Through Sampling}
\label{subsec:imbalance}
\subsection{Inter-Topic}
\begin{figure*}
\centering
\includegraphics[width=\textwidth]{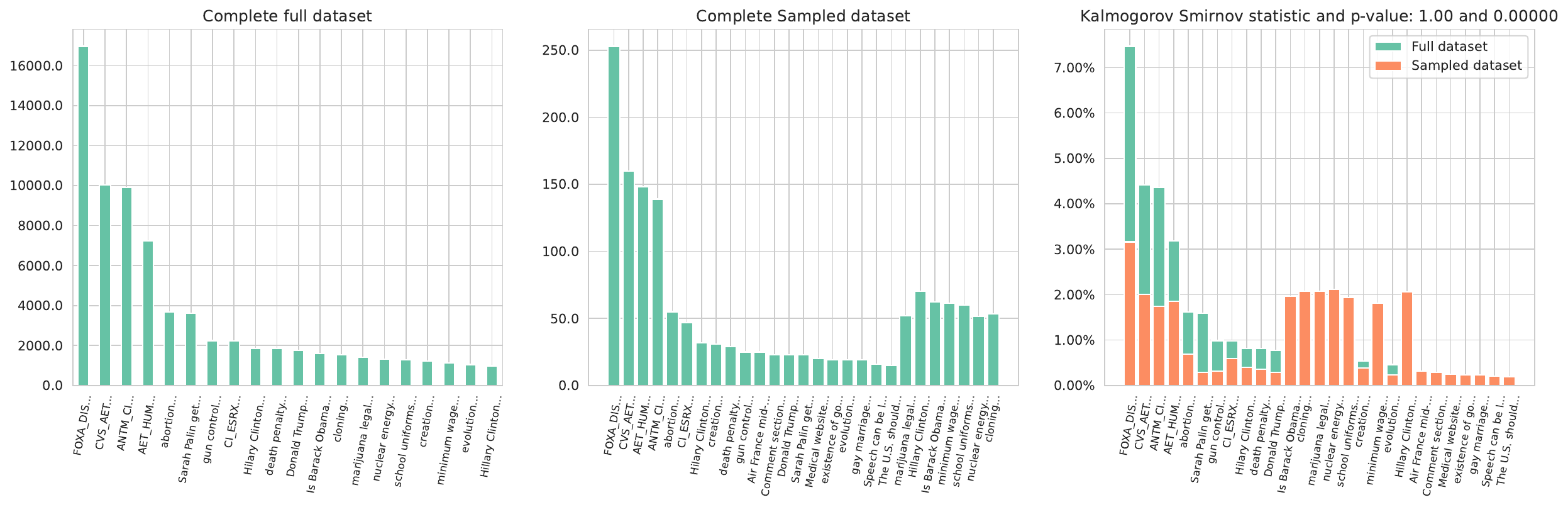}
\caption{Distributions of top 20 most frequent topics in complete dataset $\mathcal{D}$ (left), Sampled dataset $\mathcal{D}_{train}$ (mid) and their aggregated comparison (right). The distribution of top $20$ topics in $\{\mathcal{D}\} - \{\mathcal{D}_{train}\}$ is added to the tail of the figure (mid). 
}
\label{fig:imbalanced}
\end{figure*}

To investigate the inter-topic imbalances, we look at the topic distribution for the top $20$ most frequent topics covered in the complete multi-domain dataset $\mathcal{D}$, which accounts for $\geq 40 \%$ of the overall data. As we can see in  \autoref{fig:imbalanced}, even the most frequent topics greatly vary in their representation frequency, with $\sigma = 4093.55$, where $\sigma$ is the standard deviation between represented amounts. For the training dataset $\mathcal{D}_{train}$, by contrast, the standard deviation between the topics is much smaller $\sigma = 63.59$. This can be attributed to the fact that $\mathcal{D}_{train}$ constitutes  $\leq10\%$ of $\mathcal{D}$, thus we also show the aggregated data distributions in \autoref{fig:imbalanced}. 
For a more systematic analysis, we employ the two sample Kolmogorov-Smirnov (KS) test \citep{kalmagorov}, to compare topic distributions in $\mathcal{D}$ and $\mathcal{D}_{train}$  for each dataset present in $\mathcal{D}$. The test compares the cumulative distributions (CDF) of the two groups, in terms of their maximum-absolute difference, $\text { stat }=\sup _x\left|F_1(x)-F_2(x)\right|$.

\begin{table}
\centering
\fontsize{10}{10}\selectfont
\begin{tabular}{@{}lrr@{}}
\toprule
 dataset                    & stat  & p-value  \\ \midrule
\textbf{fnc-1-ours}         & 1.00 & 0.007937 \\
\textbf{arc}                & 0.40 & 0.873016 \\
\textbf{emergent}           & 0.80 & 0.079365 \\
wtwt                        & 0.20 & 1.000000 \\
\textbf{rumor}              & 0.40 & 0.873016 \\
\textbf{snopes}             & 0.40 & 0.873016 \\
\textbf{perspectrum}        & 0.60 & 0.357143 \\
\textbf{vast}               & 0.60 & 0.357143 \\
\textbf{semeval2016task6}   & 0.40 & 0.873016 \\
\textbf{iac}                & 0.40 & 0.873016 \\
mtsd                        & 0.25 & 1.000000 \\
\textbf{argmin}             & 0.40 & 0.873016 \\
\textbf{scd}                & 1.00 & 0.007937 \\
\textbf{ibm\_claim\_stance} & 0.80 & 0.079365 \\
\textbf{politicaldebates}   & 0.50 & 1.000000 \\ \bottomrule
\end{tabular}%
\caption{KS test for topic distributions. The topics in bold designate a rejected null-hypothesis (criteria: $p\leq0.05$ or ${stat} \geq 0.4$), that the topics in $\mathcal{D}$ and $\mathcal{D}_{train}$ come from the same distribution.}
\label{tab:kalmagorov}
\end{table}

The results in \autoref{tab:kalmagorov} show that the topic distribution within the full and sampled data $\mathcal{D}$, $\mathcal{D}_{train}$, cannot be the same for most of the datasets. The results for the maximum-absolute difference also show that with at least $0.4$ difference in CDF, the sampled dataset $\mathcal{D}_{train}$ on average has a more balanced topic distribution. The analysis in \autoref{fig:imbalanced} and \autoref{tab:kalmagorov}, show that the sampling technique is able to mitigate the inter-topic imbalances present in $\mathcal{D}$. A more in-depth analysis for each dataset is provided in \autoref{Appendix:imbalances}.

\subsection{Per-topic}

\begin{figure}
\centering
\includegraphics[width=\textwidth]{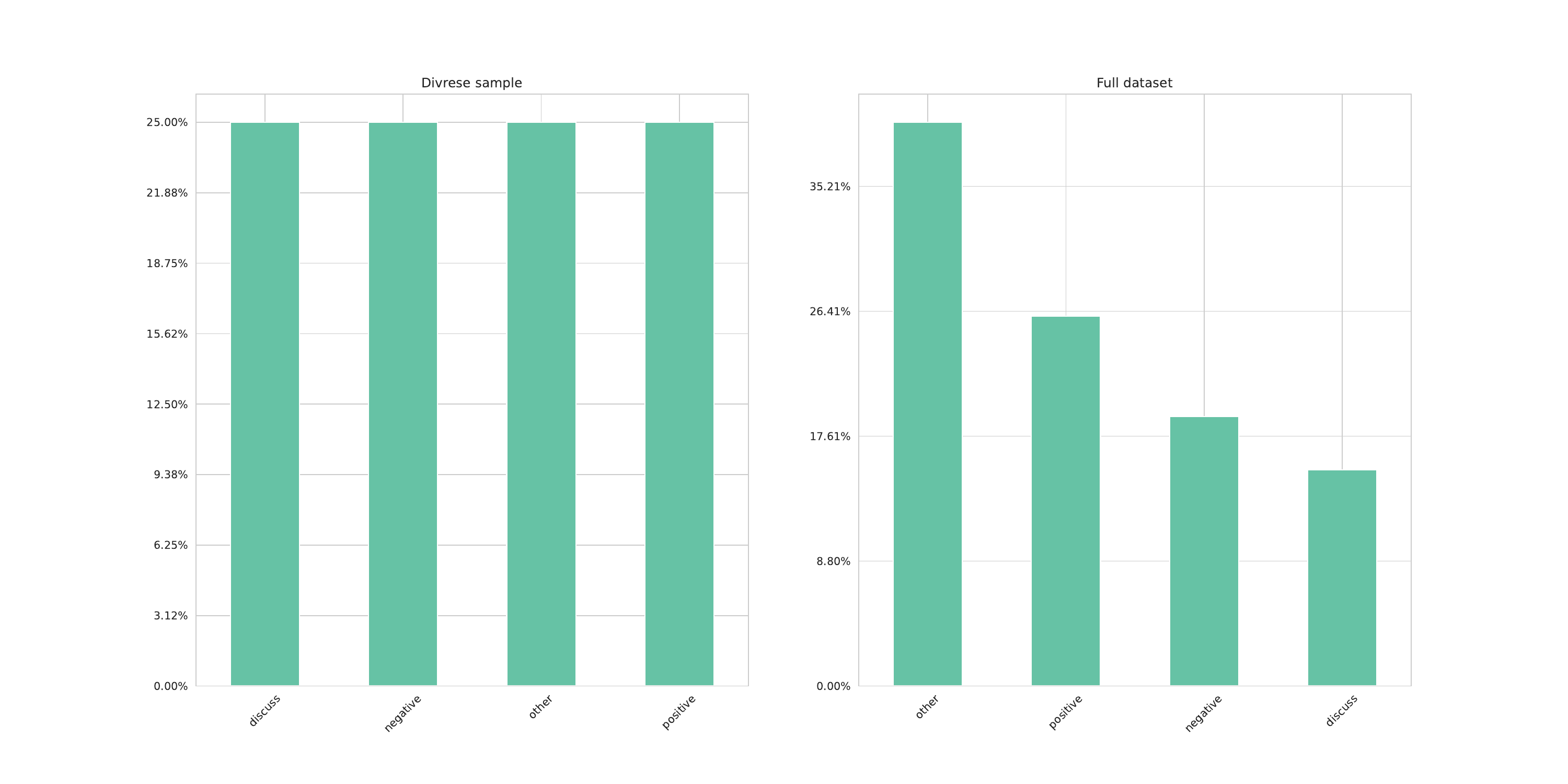}
\caption{Label distribution in $\mathcal{D}$ (right) and $\mathcal{D}_{train}$ (left).}
\label{fig:label_dist}
\end{figure}

For the per-topic imbalance analysis, we complete similar steps to the inter-topic analysis, with the difference that we iterate over the top $20$ frequent topics looking at \emph{label} imbalances within each topic. We examine the label distribution for the top $20$ topics for a per-topic comparison. The standard deviation in label distributions averaged across those 20 topics is $\sigma=591.05$ for the whole dataset $\mathcal{D}$ and the sampled set $\mathcal{D}_{train}$ $\sigma=11.7$. This can be attributed to the stratified manner of our sampling technique. This is also evident from \autoref{fig:label_dist}, which portrays the overall label distribution in $\mathcal{D}$ and $\mathcal{D}_{train}$. 

To investigate the difference in label distribution for each of the top 20 topics in $\mathcal{D}$, we use the KS test, presented in \autoref{tab:per_topic_imbalance}. For most topics, we see that the label samples in $\mathcal{D}$ and $\mathcal{D}_{train}$ cannot come from the same distribution. This means that the per-topic label distribution in the sampled dataset $\mathcal{D}_{train}$, does not possess the same imbalances present in $\mathcal{D}$.

\begin{table}[t]
\centering
\fontsize{8}{9}\selectfont
\begin{tabular}{@{}lr@{}}
\toprule
topic                                  & p-values \\ \midrule
\textbf{FOXA\_DIS}                     & 0.028571 \\
\textbf{CVS\_AET}                      & 0.028571 \\
\textbf{ANTM\_CI}                      & 0.028571 \\
\textbf{AET\_HUM}                      & 0.047143 \\
abortion                               & 0.100000 \\
\textbf{Sarah Palin getting divorced?} & 0.028571 \\
\textbf{gun control}                   & 0.001879 \\
\textbf{CI\_ESRX}                      & 0.028571 \\
\textbf{Hilary Clinton}                & 0.001468 \\
death penalty                          & 0.100000 \\
\textbf{Donald Trump}                  & 0.002494 \\
\textbf{Is Barack Obama muslim?}       & 0.028571 \\
cloning                                & 0.333333 \\
\textbf{marijuana legalization}        & 0.032178 \\
nuclear energy                         & 0.333333 \\
school uniforms                        & 0.333333 \\
\textbf{creation}                      & 0.003333 \\
minimum wage                           & 0.333333 \\
evolution                              & 0.100000 \\
\textbf{lockdowns}                     & 0.000491 \\ \bottomrule
\end{tabular}
\caption{KS test for label distributions. The topics in bold designate a rejected null-hypothesis (criteria: $p\leq0.05$), that the label samples in $\mathcal{D}$ and $\mathcal{D}_{train}$ averaged per top $20$ topics come from the same distribution.}
\label{tab:per_topic_imbalance}
\end{table}

\begin{figure}[t]
\centering
\includegraphics[width=\textwidth]{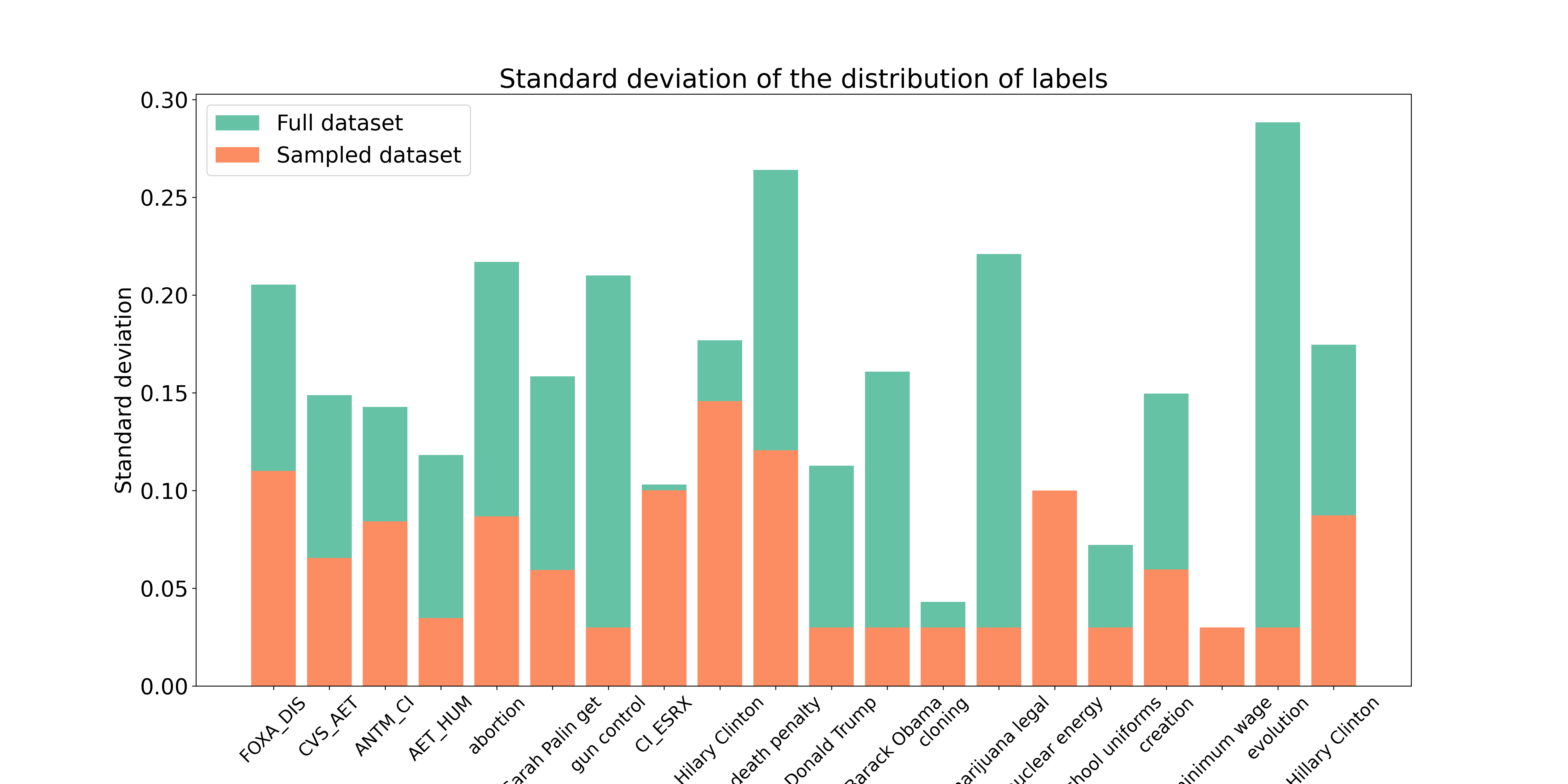}
\caption{Normalized Standard Deviation in label distribution for top 20 topics.
}
\label{fig:per_topic_dist_std}
\end{figure}

We can also see the normalized standard deviation for the label distribution within $\mathcal{D}_{train}$ is lower than in $\mathcal{D}$, as shown in \autoref{fig:per_topic_dist_std}. This reinforces the finding that per-topic label distributions in the sampled dataset are more uniform. For complete per-topic results, we refer the reader to  \autoref{Appendix:imbalances}.

\subsection{Performance} 

Using our topic-efficient sampling method is highly beneficial for in- and out-of-domain experiments, presented in \autoref{tab:in_domiain_results} and \autoref{tab:ood_results}. Our sampling method can select diverse and representative examples while outperforming \textit{Random} and \textit{Stratified} sampling techniques by $8$ and $5$ F1 points on average. This performance can be attributed to the mitigated inter- and per-topic imbalance in $\mathcal{D}_{train}$.

\subsection{Data Efficiency}
\label{subsec:data_efficiency}

\begin{figure}[t]
\centering
\includegraphics[width=\textwidth]{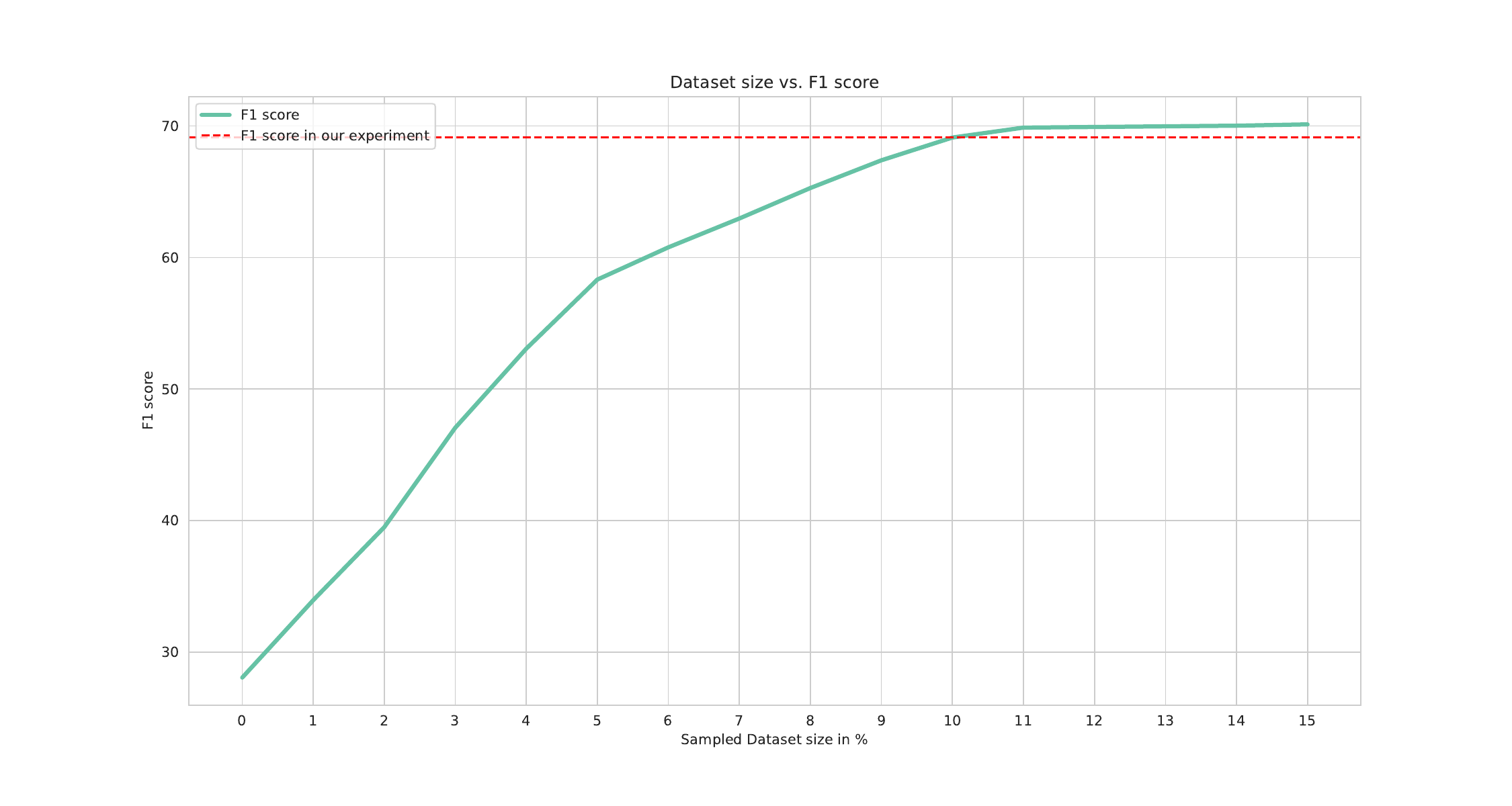}
\caption{Sampled Data size vs Performance. Performance increases with a bigger sampled selection.}
\label{fig:data_efficient}
\end{figure}

TESTED allows for sampling topic-efficient, diverse and representative samples while preserving the balance of topics and labels. This enables the training of data-efficient models for stance detection while avoiding redundant or noisy samples. We analyse the data efficiency of our method by training on datasets with sizes $[1\%,15\%]$ compared to the overall data size $\lvert \mathcal{D}\rvert$, sampled using our technique. Results for the in-domain setting in terms of averaged F1 scores for each sampled dataset size are shown in \autoref{fig:data_efficient}. One can observe a steady performance increase with the more selected samples, but diminishing returns from the $10\%$ point onwards. This leads us to use $~10\%$ as the optimal threshold for our sampling process, reinforcing the data-efficient nature of TESTED.

\subsection{Contrastive Objective Analysis}
\label{subsec:contrastive}

\begin{figure}
\centering
\includegraphics[width=\textwidth]{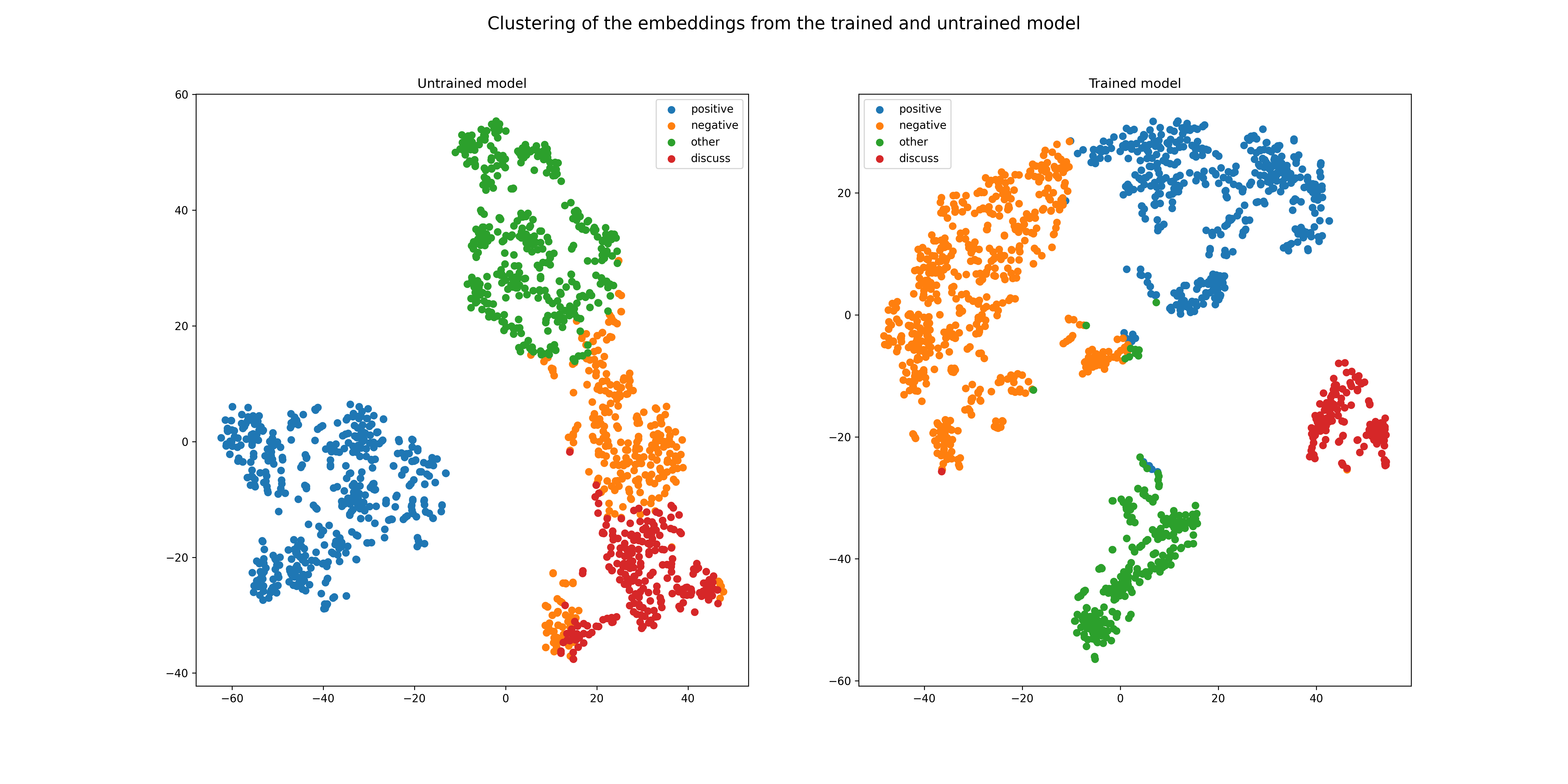}
\caption{Sample Representation before (left) and after (right) contrastive training.}
\label{fig:contrastive}
\end{figure}

To analyse the effect of the contrastive loss, we sample $200$ unseen instances stratified across each dataset and compare the sentence representations before and after training. To compare the representations, we reduce the dimension of the embeddings with t-SNE and cluster them with standard K-means. We see in \autoref{fig:contrastive} that using the objective allows for segmenting contrastive examples in a more pronounced way. The cluster purity also massively rises from $0.312$ to $0.776$ after training with the contrastive loss. This allows the stance detection model to differentiate and reason over the contrastive samples with greater confidence.

\section{Conclusions}
\label{sec:conclusions}

We proposed TESTED, a novel end-to-end framework for multi-domain stance detection. The method consists of a data-efficient topic-guided sampling module, that mitigates the imbalances inherent in the data while selecting diverse examples, and a stance detection model with a contrastive training objective. TESTED yields significant performance gains compared to strong baselines on in-domain experiments, but in particular generalises well on out-of-domain topics, achieving a $10.2$ F1 point improvement over the state of the art, all while using $\leq10\%$ of the training data. While in this paper, we have evaluated TESTED on stance detection, the method is applicable to text classification more broadly, which we plan to investigate in more depth in future work. 

\section*{Limitations}

Our framework currently only supports English, thus not allowing us to complete a cross-lingual study. Future work should focus on extending this study to a multilingual setup. Our method is evaluated on a $16$ dataset stance benchmark, where some domains bear similarities. The benchmark should be extended and analyzed further to find independent datasets with varying domains and minimal similarities, allowing for a more granular out-of-domain evaluation.

\label{sec:limitations}

\section*{Acknowledgements}

This research is funded by a DFF Sapere Aude research leader grant under grant agreement No 0171-00034B, as well as supported by the Pioneer Centre for AI, DNRF
grant number P1.

\section{Appendix}
\label{sec:Appendix}


\section{Imbalance analysis}
\label{Appendix:imbalances}

\subsection{Inter-topic}
\label{Appendix:inter_topic_analysis}

\begin{figure*}
\centering
\includegraphics[width=\textwidth]{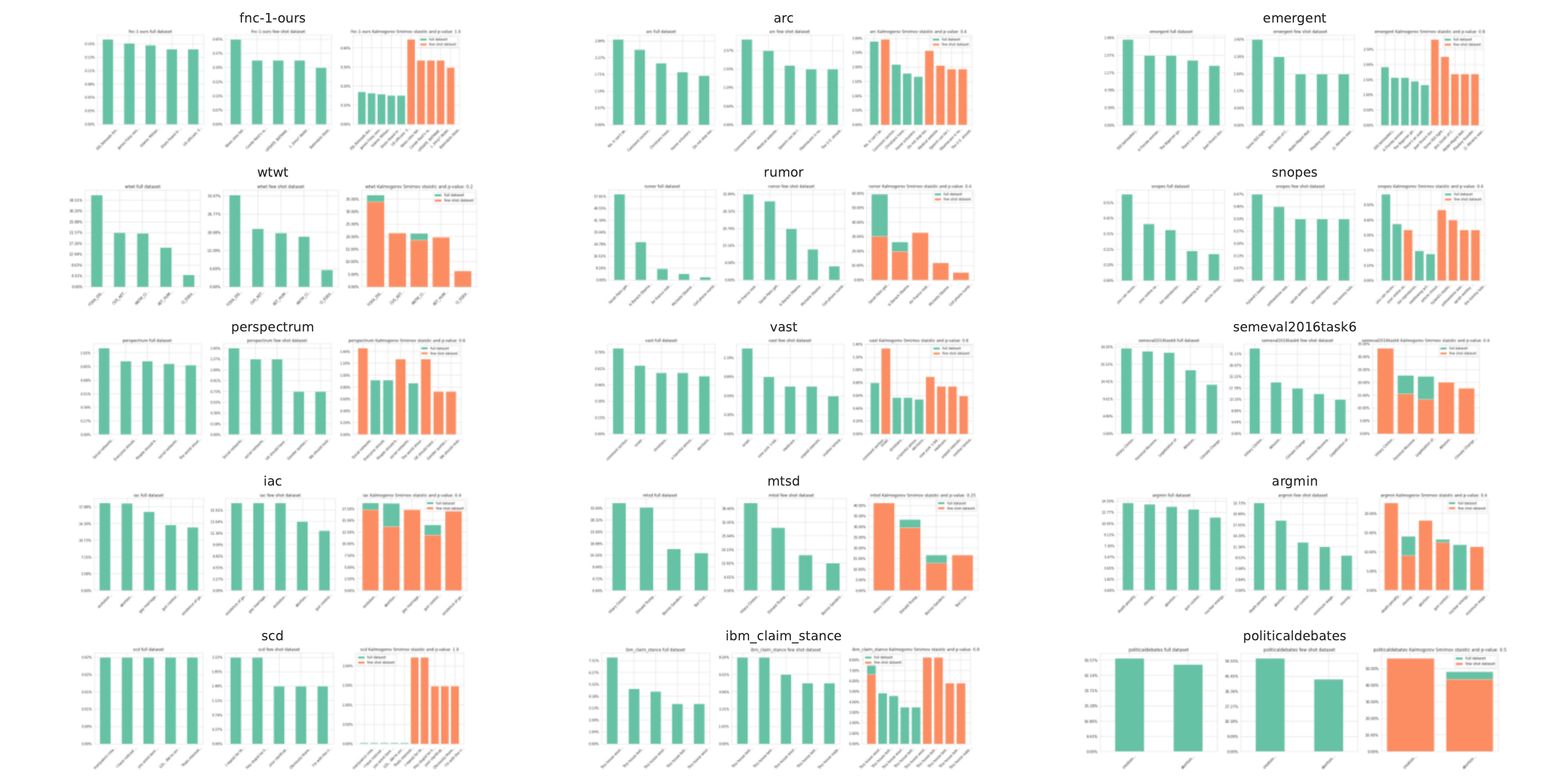}
\caption{Distributions of top 20 most frequent topics for each dataset (left), Sampled dataset $\mathcal{D}_{{train} = {dataset}}$ (mid) and their aggregated comparison (right).
}
\label{fig:imbalanced_big}
\end{figure*}

To complement our inter-topic imbalance mitigation study, we complete an ablation on all topics in $\mathcal{D}$ and report them on a per-domain basis in \autoref{fig:imbalanced_big}. The trend is similar to the one in \autoref{fig:imbalanced}, where the dataset with imbalanced distributions is rebalanced, and balanced datasets are not corrupted.

\subsection{Per-topic}
\label{Appendix:per_topic_analysis}

\begin{figure*}
\centering
\includegraphics[width=\textwidth]{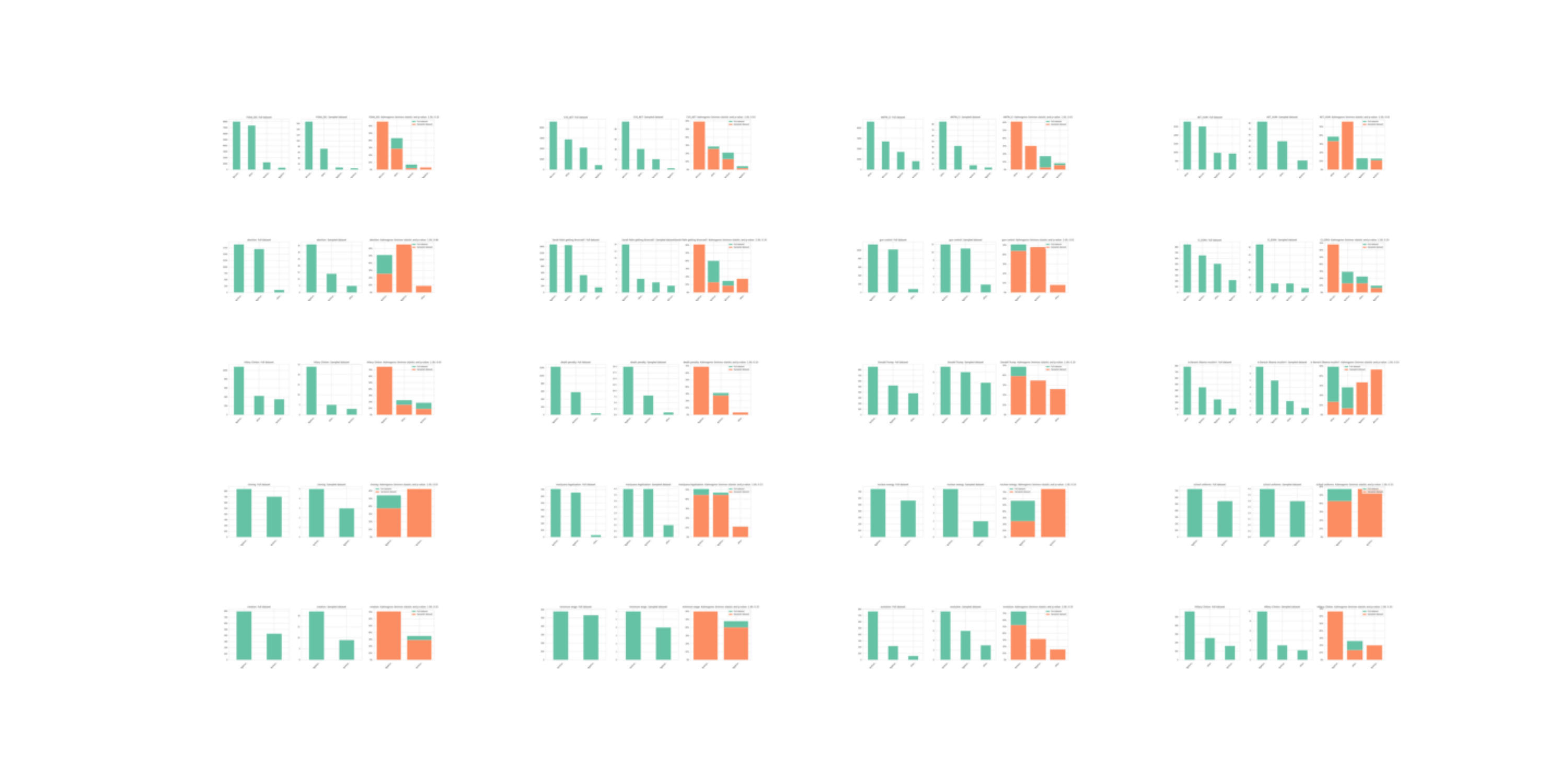}
\caption{Distributions of labels for top 20 most frequent topics for $\mathcal{D}$ (left), Sampled dataset $\mathcal{D}_{{train} = {dataset}}$ (mid) and their aggregated comparison (right).
}
\label{fig:per_topic_big}
\end{figure*}

We show that our topic-efficient sampling method allows us to balance the label distribution for unbalanced topics, while not corrupting the ones distributed almost uniformly. To do this, we investigate each of the per-topic label distributions for the top $20$ most frequent topics while comparing  the label distributions for $\mathcal{\mathcal{D}}$ and $\mathcal{D}_{train}$, presented in \autoref{fig:per_topic_big}.

\section{Evaluation Metrics}

To evaluate our models and have a fair comparison with the introduced benchmarks we use a standard set of metrics for classification tasks such as macro-averaged F1, precision, recall and accuracy.
\begin{gather}
Acc = \frac{TP+TN}{TP+TN+FP+FN} \\
Prec = \frac{TP}{TP+FP} \\
Recall = \frac{TP}{TP+FN} \\
F1 = \frac{2*Prec*Recall}{Prec+Recall} = \frac{2*TP}{2*TP+FP+FN}
\end{gather}

\section{Dataset Statistics}
\label{Appendix:data}
We use a stance detection benchmark \citep{hardalov2021cross} whose data statistics are shown in Table \ref{tab:data_stat}. The label mapping employed is shown in Table \ref{tab:label_mapping}.

\begin{table}[t]
\centering
\resizebox{\textwidth}{!}{
\begin{tabular}{lrrrr}
\toprule Dataset & Train & Dev & Test & Total \\
\midrule arc & 12,382 & 1,851 & 3,559 & 17,792 \\
argmin & 6,845 & 1,568 & 2,726 & 11,139 \\
emergent & 1,770 & 301 & 524 & 2,595 \\
fnc1 & 42,476 & 7,496 & 25,413 & 75,385 \\
iac1 & 4,227 & 454 & 924 & 5,605 \\
ibmcs & 935 & 104 & 1,355 & 2,394 \\
mtsd & 3,718 & 520 & 1,092 & $5,330$ \\
perspectrum & 6,978 & 2,071 & 2,773 & 11,822 \\
poldeb & 4,753 & 1,151 & 1,230 & 7,134 \\
rumor & 6,093 & 471 & 505 & $7,276$ \\
scd & 3,251 & 624 & 964 & 4,839 \\
semeval2016t6 & 2,497 & 417 & 1,249 & 4,163 \\
semeval2019t7 & 5,217 & 1,485 & 1,827 & 8,529 \\
snopes & 14,416 & 1,868 & 3,154 & 19,438 \\
vast & 13,477 & 2,062 & 3,006 & 18,545 \\
wtwt & 25,193 & 7,897 & 18,194 & 51,284 \\
\midrule Total & 154,228 & 30,547 & 68,495 & 253,270 \\
\bottomrule
\end{tabular}
}
\caption{Dataset statistics of the stance detection benchmark by \citet{hardalov2021cross} also used in this paper. Note that the rumour and mtsd datasets are altered in that benchmark as some of the data was unavailable.}
\label{tab:data_stat}
\end{table}

\begin{table}[t]
\centering
\resizebox{\textwidth}{!}{%
\begin{tabular}{@{}ll@{}}
\toprule
Label    & Description                                                             \\ \midrule
Positive & agree, argument for, for, pro, favor, support, endorse                  \\
Negative & disagree, argument against, against, anti, con, undermine, deny, refute \\
Discuss  & discuss, observing, question, query, comment                            \\
Other    & unrelated, none, comment                                                \\
Neutral  & neutral                                                                 \\ \bottomrule
\end{tabular}%
}
\caption{Hard stance label mapping employed in this paper, following the stance detection benchmark by \citet{hardalov2021cross}.}
\label{tab:label_mapping}
\end{table}

\section{TESTED with different backbones}
\label{sec:transformers}

We chose to employ different PLM's as the backbone for TESTED and report the results in the \autoref{tab:plm_effect}. The PLMs are taken from the set of \textit{roberta-base, roberta-large, xlm-roberta-base, xlm-roberta-large. }The differences between models with a similar number of parameters are marginal. We can see a degradation of the F1 score between the \textit{base} and \textit{large}  versions of the models, which can be attributed to the expressiveness the models possess. We also experiment with the distilled version of the model and can confirm that in terms of the final F1 score, it works on par with the larger models. This shows that we can utilise smaller and more computationally efficient models within the task with marginal degradation in overall performance.

\setlength{\tabcolsep}{3pt}
\begin{table*}
\centering
\begin{adjustbox}{max width=\textwidth}
\begin{tabular}{@{}lc|ccccc|ccc|ccccc|ccc@{}}
\toprule
 &
  $\mathrm{F}_1$ avg. &
  \rotatebox{45}{arc} &
  \rotatebox{45}{iac1} &
  \rotatebox{45}{perspectrum} &
  \rotatebox{45}{poldeb} &
  \rotatebox{45}{scd} &
  \rotatebox{45}{emergent} &
  \rotatebox{45}{fnc1} &
  \rotatebox{45}{snopes} &
  \rotatebox{45}{mtsd} &
  \rotatebox{45}{rumor} &
  \rotatebox{45}{semeval16} &
  \rotatebox{45}{semeval19} &
  \rotatebox{45}{wtwt} &
  \rotatebox{45}{argmin} &
  \rotatebox{45}{ibmcs} &
  \rotatebox{45}{vast} \\ \midrule
  TESTED$_{\textit{reberta-large}}$ &
    69.12 &
    64.82 &
    56.97 &
    83.11 &
    52.76 &
    64.71 &
    82.10 &
    83.17 &
    78.61 &
    63.96 &
    66.58 &
    69.91 &
    58.72 &
    70.98 &
    62.79 &
    88.06 &
    57.47 \\
  TESTED$_{\textit{xlm-reberta-large}}$ &
    68.86 &
    64.35 &
    57.0 &
    82.71 &
    52.93 &
    64.75 &
    81.72 &
    82.71 &
    78.38 &
    63.66 &
    66.71 &
    69.76 &
    58.27 &
    71.29 &
    62.73 &
    87.75 &
    57.2 \\
  TESTED$_{\textit{reberta-base}}$ &
    65.32 &
    59.71 &
    51.86 &
    76.75 &
    50.23 &
    61.35 &
    78.84 &
    82.09 &
    73.31 &
    62.87 &
    65.46 &
    63.89 &
    58.3 &
    67.28 &
    58.28 &
    83.81 &
    51.09 \\
  TESTED$_{\textit{xlm-reberta-base}}$ &
    65.05 &
    60.26 &
    51.96 &
    76.2 &
    51.82 &
    58.74 &
    74.68 &
    77.9 &
    72.61 &
    62.71 &
    66.08 &
    69.74 &
    53.27 &
    65.83 &
    59.09 &
    87.92 &
    52.08 \\
    \midrule
  TESTED$_{\textit{distilroberta-base}}$ &
    68.86 &
    61.78 &
    56.94 &
    80.36 &
    46.29 &
    64.1 &
    79.26 &
    81.37 &
    73.44 &
    62.6 &
    63.4 &
    63.75 &
    56.53 &
    68.35 &
    57.27 &
    81.93 &
    56.3 \\
    \bottomrule
  \end{tabular}%
\end{adjustbox}
\caption{In-domain results reported with macro averaged F1, with varying backbones when using TESTED.}
\label{tab:plm_effect}
\end{table*}
\setlength{\tabcolsep}{6pt}

\chapter{SynDARin: Synthesising Datasets for Automated Reasoning in Low-Resource Languages}
\label{chap:syndarin}

\section{Introduction}

Question Answering (QA) has been a hallmark task for testing reading comprehension and reasoning capabilities in NLP systems.
The availability of numerous English benchmarks that frame the problem as extractive, cloze-style or open-domain \citep{yang2015wikiqa,rajpurkar2016squad,chen2017reading} reasoning tasks, along with novel pre-trained language models (PLMs) \citep{devlin2018bert,lewis2019bart} and LLMs \citep{touvron2023llama,jiang2023mistral,achiam2023gpt} allowed for the development and granular evaluation of QA systems that occasionally boast human-like or better performance \citep{devlin2018bert,min2023recent,rogers2023qa}.
 Cross-lingual alignment through translation-following has improved performance on multilingual benchmarks like XQUAD and MLQA 
 \citep{ranaldi-pucci-2023-english}.
Although some concentrated effort has been made to create multilingual QA resources \citep{lewis2019mlqa,asai2018multilingual,liu2019xqa}, the datasets remain rather scarce and usually cover a small selected set of languages due to the labour-intensive annotation costs. 
The proposed methods suggest using direct machine translation \citep{lewis2019mlqa,carrino2019automatic} or multilingual synthetic data generation \citep{riabi2020synthetic,agrawal2023qameleon,shakeri2020towards}. However, these approaches are directly bound to introduce biases and hallucinations during translation \citep{artetxe2020translation}, cross-lingual transfer \citep{lauscher2020zero, guerreiro2023hallucinations} or generation \citep{ahuja2023mega}. 
%
%
\begin{figure*}[t!]
    \centering
    \includegraphics[trim=0.5cm 0.0cm 0.5cm 0.5cm,clip=true,width=0.95\textwidth]{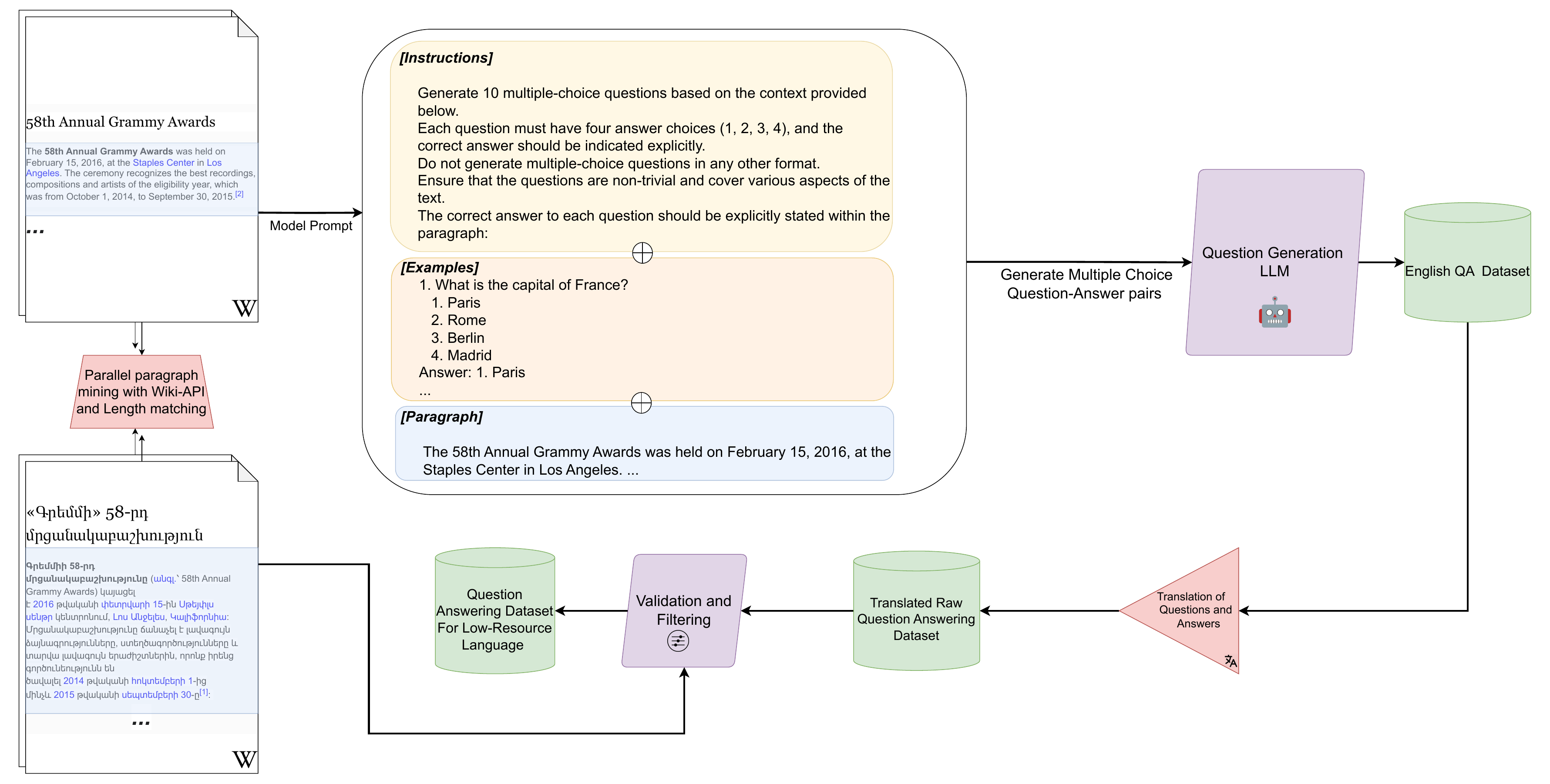}
    \caption{The proposed framework is comprised of three components: (i) a module for mining parallel paragraphs using wiki-API and length matching; (ii) generating a synthetic question-answering dataset with an LLM using the mined English paragraphs; (iii) translating the question-answer pairs and Filtering/Validating them for obtaining a high-quality synthetic QA dataset in the low-resource language.}
    \label{fig:framework_syndarin_main}
\end{figure*}
These limitations directly hinder the possibility to \emph{develop} and \emph{evaluate} the multilingual QA capabilities of language models in low-resource languages.

In this work, we propose \textbf{S}yn\textbf{DAR}in, a novel method for synthesising datasets for automated reasoning in low-resource languages that circumvents the above-mentioned obstacles and test it by creating a QA dataset for the Armenian language, which has virtually no presence of structured NLP datasets \citep{avetisyan2023large}.
We mine parallel English and Armenian introductory paragraphs from the same diverse set of Wikipedia articles, ensuring that the contents match by comparing their relative length. Similar mining approaches have been shown to be efficient for this task \citep{lewis2021paq,artetxe2019massively}.
This allows us to obtain human-curated text from diverse topics while bypassing a wide chunk of direct content translation and annotation.
Given the English subset of this data, we generate MC question-answer pairs by prompting an LLM to produce queries with an answer explicitly mentioned within the paragraph.
Following \citet{lewis2019mlqa}, we filter out examples that do not contain the answer substring verbatim in the paragraph and additionally perform a human evaluation on a subset of $50$ examples and show that $98\%$ of these question-answer pairs are answerable and maintain quality.
The produced question-answers are subsequently translated using an automated tool and further validated by answer substring and semantic matching in the parallel Armenian paragraph.
This allows us to mitigate the likelihood of hallucinated, biased, and inconsistent entries in the final QA dataset. Our human evaluation with native Armenian speakers shows that ~$70\%$ of such corrupted examples are removed.
We use the dataset as a reasoning benchmark for Armenian and evaluate several LLMs in zero-shot, few-shot, and fine-tuned modes.
We show that the dataset cannot be trivially solved, thus highlighting it as a useful resource for measuring model performance.
In sum, our contributions are as follows: (i) a novel method for QA dataset construction in low-resource languages, (ii) a QA dataset in Armenian, (iii) ablations showing the quality of the generated samples, and (iv) an evaluation of several LLM families on the QA dataset.

\section{Methodology}

An outline of \textbf{S}yn\textbf{DAR}in can be seen in \cref{fig:framework_syndarin_main}.

\subsection{Parallel Data Mining}

Given parallel English and Armenian introductory paragraph tokens $\mathcal{P}_{\text{En}} = (T_1, \dots T_n)$, $\mathcal{P}_{\text{Arm}} = (T_1,\dots T_m)$ obtained from a diverse set of Wiki articles, we want to save the segments that contain the same content. As the introductory paragraphs in Wikipedia contain highly similar information \citep{lewis2019mlqa}, we found that filtering out the paragraph pairs based on their relative view count and the number of tokens, i.e. length, is sufficient. To do this, we simply define a conditional rejection process on Wikipedia pages that have been viewed more than $1000$ and edited more than 5 times $| \lVert \mathcal{P}_{\text{En}}\rVert-\lVert\mathcal{P}_{\text{Arm}}\rVert | \leq K_{\text{DM}}$, where $K_{\text{DM}}$ is the threshold for the length difference. A higher length difference would imply that the contents of the paragraphs are misaligned, thus making us reject such samples. Consequently, we are able to obtain naturally written human-curated parallel paragraphs that cover a diverse set of topics.

\subsection{QA Generation}

After obtaining the parallel data, we prompt an LLM $\mathcal{M}$ with instructions $\mathcal{I}=(T_1, \dots T_{|\mathcal{I}|})$ and $10$ in-context example demonstrations $\mathcal{E} = (E_1,\dots E_{10})$, where $\forall i, E_i=(T_1,\dots T_{|E_i|})$, to generate diverse English MC question-answer pairs $\mathcal{K_{\text{Eng}}}=\left\{\left(q_1, a_1\right) \ldots\left(q_N, a_N\right)\right\}$ given an English context paragraph $\mathcal{P}_{\text{En}}$:
\newcommand\shortdots{\makebox[0.75em][c]{.\hfil.\hfil.}}
\small
\begin{align}
    q_i, a_i \sim \prod_{t=1}^{|\mathcal{K}_i|} P_{\mathcal{M}}\left(T_t^{(i)} \mid T_1^{(i)}, \shortdots, T_{t-1}^{(i)}, \mathcal{I}, \mathcal{E}, \mathcal{P}_{\text{En}}\right)
\end{align}
\normalsize

\begin{table}[t!]
\centering
\begin{adjustbox}{max width=\textwidth}
\begin{tabular}{lccccccccc}
\toprule
  \textbf{Who} & \textbf{Where} & \textbf{What} & \textbf{When} & \textbf{Which} & \textbf{How} & \textbf{General} & \textbf{Why} \\
304 & 128 & 1536 & 215 & 473 & 244 & 76 & 16 \\
\bottomrule
\end{tabular}
\end{adjustbox}
\caption{Frequency of Question Types in the generated English question-answer pairs.}
\label{tab:qa_types}
\end{table}

We filter out all repeating questions, $\forall \{i,j: i\neq j\}, q_i\neq q_j$, and question-answer pairs where the answer span is not exactly mentioned within the text, i.e. $a_i \not\subset  \mathcal{P}_{\text{En}}$. An example input used for generation can be seen in \cref{fig:framework_syndarin_main}. This generation and validation pipeline resembles the ones in \citet{lewis2021paq, agrawal2023qameleon}, which have shown successful question-generation results for the English language. Several examples of produced questions are available in \cref{sec:appendix_syndarin}.

\subsection{Translation and Validation}

We transfer the generated question-answer pairs $\mathcal{K_{\text{Eng}}}$ into Armenian by using the Google Translate API to obtain $\mathcal{K_{\text{Arm}}}$. To mitigate the inconsistencies introduced during the translation process, we save only the samples where the translated answer $a_i \in \mathcal{K_{\text{Arm}}}$ is contained within and semantically related to the paragraph $\mathcal{P}_{\text{Arm}}$. To do this, we use a fuzzy substring matching function $\mathcal{F}: \mathcal{T} \times \mathcal{T} \rightarrow [0,1] $, along with a multilingual language model $\mathcal{M}_{\text{sim}}:\mathcal{T} \rightarrow \mathcal{R}^d$ to measure semantic similarity, where $\mathcal{T}$ is an arbitrary set of tokens and $d$ is the dimensionality of the embedding space of the model. Samples below a certain threshold, $\mathcal{F}(a_i, \mathcal{P}_{\text{Arm}}) \leq K_{\text{Fuzz}} \text{ and } \cos(\mathcal{M}(a_i), \mathcal{M}(\mathcal{P}_{\text{Arm}}))\leq K_{\text{Sim}}$ are filtered out. Note that exact matching is insufficient, as the morphology of the translated answer tokens can vary in the low-resource language. The multiple-choice answers are balanced uniformly in the final dataset so as not to introduce a bias toward any particular answer ordering.

\begin{table}[t!]
\centering
\begin{adjustbox}{max width=\textwidth}
{
\begin{tabular}{@{}lll@{}}
\toprule
\textit{\textbf{Problem type(\%)}} & \textit{Filtered} & \textit{Unfiltered} \\ \midrule
Partially Missing Info & 38 & 77 \\
Bad Translation     & 5  & 51 \\
Partially Correct Answers     & 22 & 31 \\
Several Correct Answers       & 27 & 45 \\
Date Mismatch                 & 13 & 17 \\
Other                         & 8  & 22 \\ \bottomrule
\end{tabular}
}
\end{adjustbox}
\caption{Unanswerable sample analysis before(Unfiltered) and after(Filtered) the validation. Annotators can choose multiple reasons per sample.}
\label{tab:human_annot}
\end{table}

\section{Experimental Setup}

\begin{figure}[t!]
    \centering
    \includegraphics[width=\textwidth]{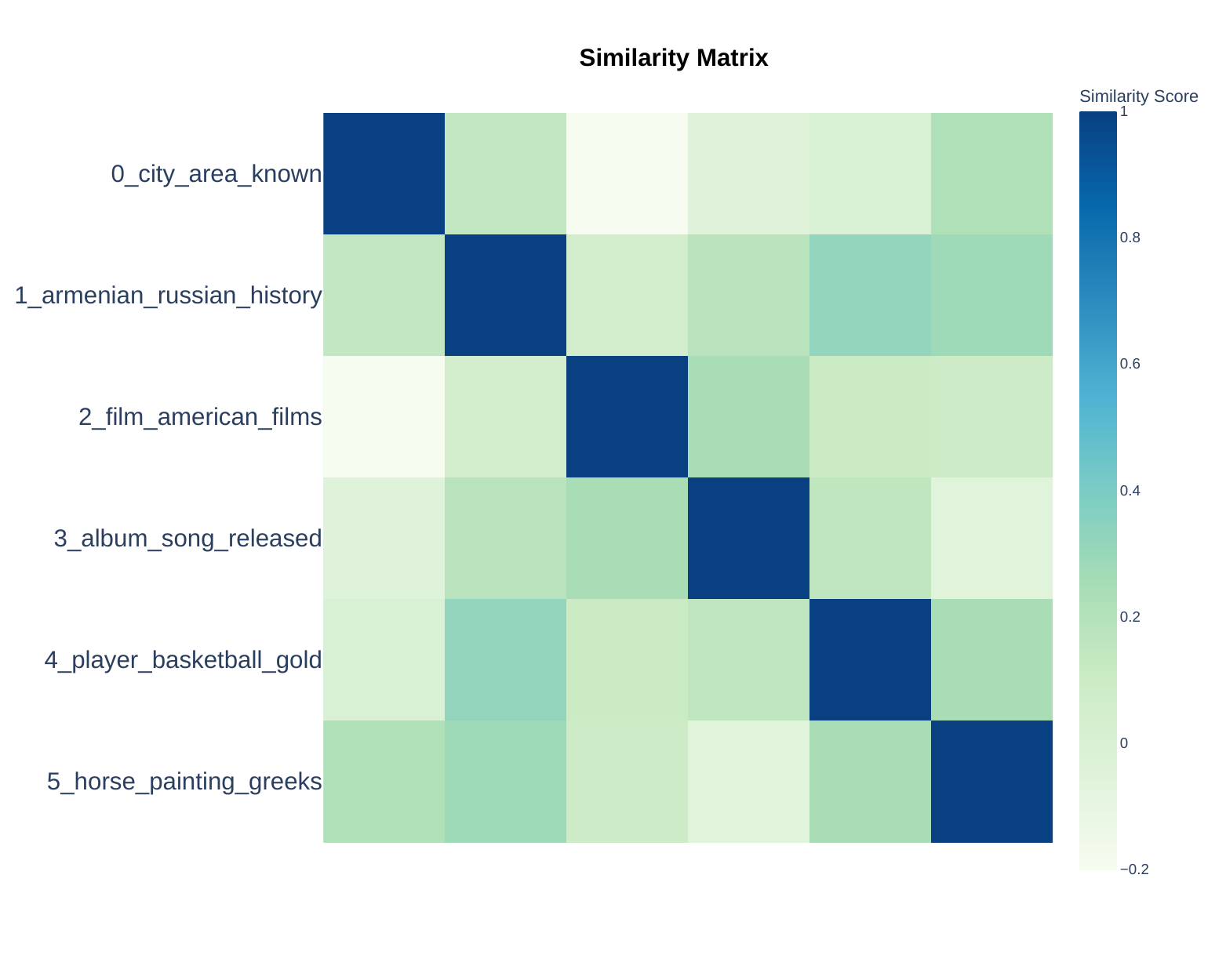}
    \caption{BERTopic embeddings similarity heatmap for the top 6 frequent topics in the mined English paragraphs.}
    \label{fig:topic_dist}
\end{figure}

\subsection{QA Generation}
Our QA generation uses GPT-4 \cite{achiam2023gpt}, known for generating high-quality text \citep{zhou2023synthetic} and synthetic data \citep{hamalainen2023evaluating,li2023synthetic}.

\subsection{Substring Matching and Semantic Similarity} 
We employ Levenshtein distance for fuzzy substring matching ($\mathcal{F}$) and multilingual sentence embeddings \citep{reimers-2019-sentence-bert} ($\mathcal{M}_{\text{sim}}$) for semantic similarity using cosine distance.

\subsection{Armenian QA Benchmarking}
We benchmark GPT-3.5 \cite{achiam2023gpt}, CMD-R, and CMD-R+ \cite{cohere2024commandr} using $\{0,2,4,6\}$ in-context examples with few-shot prompting \cite{brown2020language} on the Armenian QA dataset. We further frame the task as classification with multiple-choice answers and perform supervised fine-tuning with a recipe \citep{mosbach2020stability} on XLM-RoBERTa-base \cite{conneau2019unsupervised}, with $\{32,64, \dots ,980\}$ training samples and benchmark it on the same testing set. Following \citet{poliak2018hypothesis}, we analyze model performance on \emph{question-only} and \emph{paragraph-only} inputs for bias detection.

\section{Results}

\subsection{English QA Dataset Generation}
We mined $300$ parallel English-Armenian Wikipedia paragraphs and generated $10$ diverse questions with $4$ MC answers each, resulting in $3000$ English QA pairs.

\subsection{Dataset Diversity}
We assessed question diversity (\cref{tab:qa_types}) and found meaningful variation consistent with prior human-curated datasets \citep{lewis2019mlqa, rajpurkar2016squad}. Topic modeling using BERTopic \citep{grootendorst2022bertopic} validated the subject diversity (\cref{fig:topic_dist}). A granular diversity analysis within the dataset is presented in \cref{sec:appendix_syndarin}.

\subsection{Human Evaluation}
To assess the data quality, we follow \citet{lewis2021paq} and ask two English-speaking human annotators to manually inspect $50$ randomly chosen samples from the English QA dataset regarding the captured contextual information and answerability of the sample question. The results show, with an inter-annotator agreement score of Cohen's $\kappa = 0.99$, that $98\%$ of examples contain sufficient details to answer the question while accurately capturing contextual information.

\subsection{Automatic Translation and Validation}
We translate the obtained $3000$ QA samples and pass the results through our validation pipeline to produce $1235$ filtered Armenian examples.  

\begin{table}[t!]
\centering
\begin{adjustbox}{max width=\textwidth}
\begin{tabular}{@{}lcccc@{}}
\toprule
& \multicolumn{4}{c}{Accuracy} \\
\cmidrule{2-5}
Filter & \texttt{128} & \texttt{256} & \texttt{512} & \texttt{987} \\ \midrule
\emph{Complete}  &30.1\%          & 33.5\%       & 38.7\%       & 39.5\%       \\
\emph{paragraph-only} & \underline{26.7\%} & \underline{28.3\%}       & \underline{23.9\%}       & \underline{28.3\%}       \\
\emph{question-only} & \underline{22.1\%}  & \underline{22.7\%}       & \underline{19.4\%}       & \underline{23.5\%}       \\
\emph{Random performance} & \multicolumn{4}{c}{\textbf{25.0\%}} \\ \bottomrule
\end{tabular}
\end{adjustbox}
\caption{The results of fine-tuning XLM-Roberta on the Armenian QA dataset with a varying number of training samples in different degeneracy testing scenarios.}
\label{tab:xlm_roberta_short}
\end{table}

\subsection{Armenian QA dataset}

We use these samples and their designated Armenian paragraphs to form the QA dataset. We split the data into $80/20$ \emph{train/test} buckets with $987$ samples in training and $247$ in testing. We ensure that the paragraphs in the testing set are not contained in the train set to avoid any data leakage. We maintain a uniform distribution of MC questions within the answers, avoiding bias towards any answer ordering.

\subsection{Human Evaluation}
We assessed the translation validation pipeline and datasets using two native-speaking annotators. They reviewed the \emph{test} set, which was mixed with 100 randomly flagged poor samples from automatic validation. Annotators either answered the samples or marked them as unanswerable, citing reasons from a predefined set, see in \cref{tab:human_annot}. Results showed that $87\%$ of the flagged examples were unanswerable due to insufficient context, translation errors, or hallucinations. The error breakdown in \cref{tab:human_annot} highlights the quality improvement in filtered samples w.r.t. to the abovementioned discrepancies, where annotators answered correctly in $75\%$ of cases. We measure the inter-annotator agreement using Cohen's $\kappa=0.8$. These confirm the ability of our validation pipeline to maintain the dataset quality. 

\subsection{Benchmarks}
\begin{table}[t!]
\centering
\begin{adjustbox}{max width=\textwidth }
\begin{tabular}{@{}lcccc@{}}
\toprule
& \multicolumn{4}{c}{Accuracy} \\
\cmidrule{2-5}
Model Name    & \texttt{0} & \texttt{2} & \texttt{4} & \texttt{6} \\ \midrule
Command-R          & 58.7\%       & \textbf{68.4}\%       & \underline{64.8}\%       & 64.0\%       \\
Command-R+      & 59.3\%       & 67.2\%       & \underline{69.6}\%       & \textbf{70.9}\%       \\
GPT-3.5           & 56.3\%       & \underline{56.3}\%       & \textbf{59.1}\%       & 54.3\%       \\ \bottomrule
\end{tabular}
\end{adjustbox}
\caption{Model Accuracy with a varying number of provided in-context samples before generation.}
\label{tab:model_accuracy}
\end{table}

To show the value of the created dataset, we investigate if it suffers from statistical biases or degenerate solutions by training an XLM-RoBERTa model on inputs that contain only the paragraph or the question, excluding everything else from the sample. The results in \cref{tab:xlm_roberta_short} show that regardless of the number of training samples, the models trained with question and paragraph-only samples behave similarly to random chance, while training with complete data gradually increases the performance, highlighting that the dataset is unlikely to suffer from inconsistencies and degenerate solutions and can be used for developing QA capabilities for Armenian. We further benchmark several state-of-the-art LLMs on this dataset in supervised fine-tuning, \emph{zero-shot} and \emph{few-shot} settings. We see in \cref{tab:model_accuracy} that even the largest models do not trivially solve the dataset, showing its utility as a benchmarking tool. 

\section{Conclusion}

We propose \textbf{S}yn\textbf{DAR}in, a novel method for constructing QA datasets for low-resource languages and producing a dataset for the Armenian language. Systematic studies of the reliability of the individual modules to produce diverse QA samples that maintain answerability and quality show the effectiveness of the method. We further use the produced Armenian QA dataset to benchmark state-of-the-art LLMs and show the value of the proposed resource in evaluating QA reasoning capabilities in the low-resource language.

\section*{Limitations}

The proposed methods have currently been tested only for a smaller-scale QA dataset creation in Armenian, thus not allowing us to complete a wider cross-lingual study. The study benchmarks should be extended and analyzed further in more multilingual, low-resource languages. In the case of extremely rare low-resource languages, the automatic translation part within our pipeline would require either the development of such a translation method, robust cross-lingual transfer from a similar language, or direct manual effort, all of which are bound to introduce either qualitative or logistic complications while creating the final QA resource.

\section*{Acknowledgments}
$\begin{array}{l}\includegraphics[width=1cm]{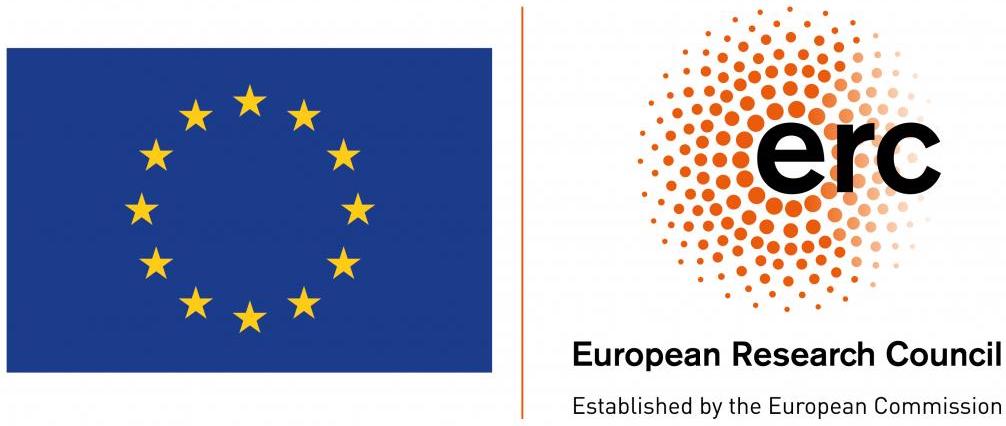} \end{array}$ 
Erik is partially funded by a DFF Sapere Aude research leader grant under grant agreement No 0171-00034B, as well as by an NEC PhD fellowship, and is supported by the Pioneer Centre for AI, DNRF grant number P1.
Pasquale was partially funded by ELIAI (The Edinburgh Laboratory for Integrated Artificial Intelligence), EPSRC (grant no.\ EP/W002876/1), an industry grant from Cisco, and a donation from Accenture LLP.
Isabelle's research is partially funded by the European Union (ERC, ExplainYourself, 101077481), and is supported by the Pioneer Centre for AI, DNRF grant number P1.
This work was supported by the Edinburgh International Data Facility (EIDF) and the Data-Driven Innovation Programme at the University of Edinburgh.

\begin{figure}[t!]
    \centering
    \includegraphics[width=\textwidth]{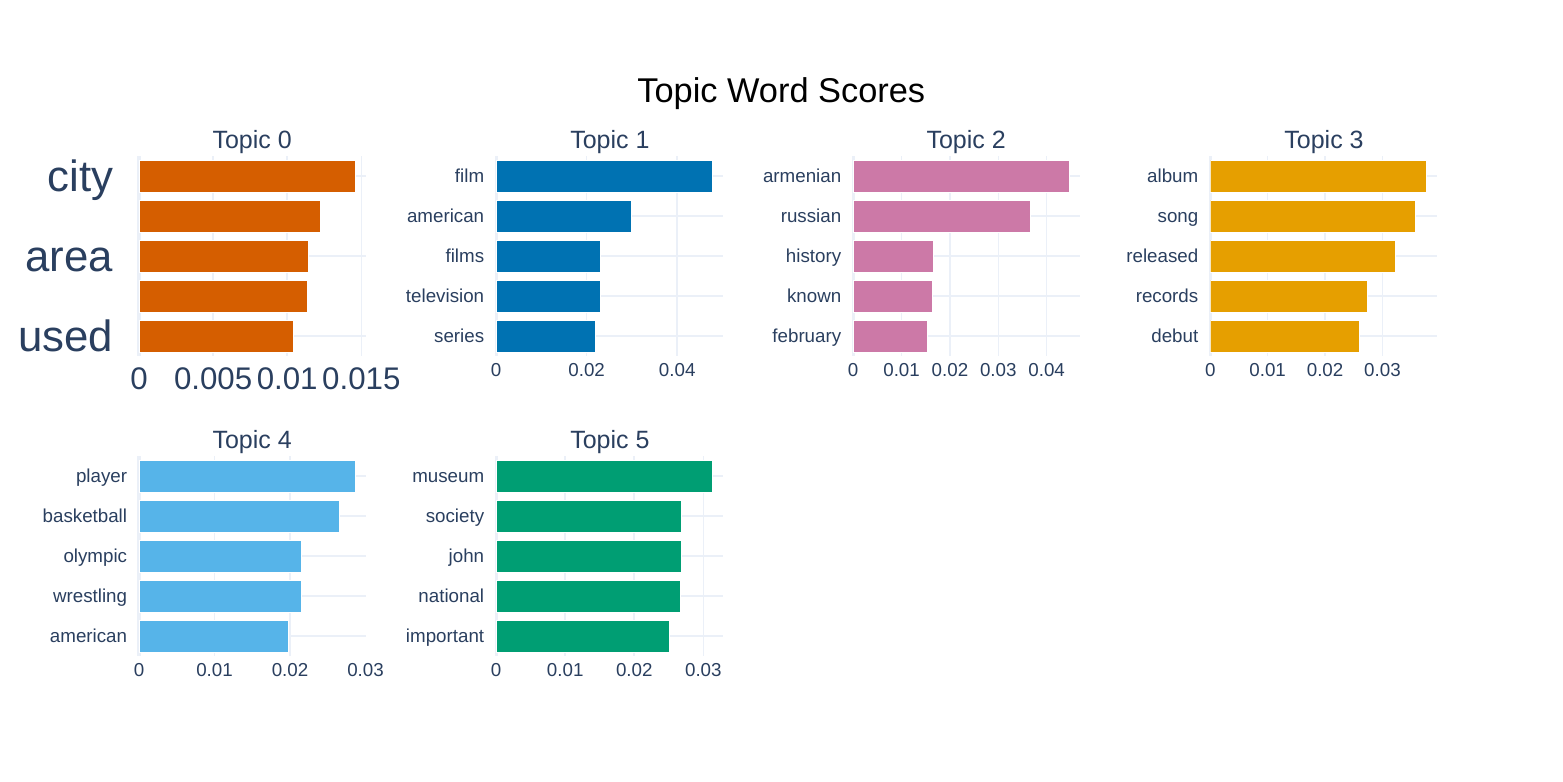}
    \caption{The usage of frequent words in the top 6 frequent topics present within the mined English paragraphs.}
    \label{fig:topic_word_distrib}
\end{figure}

\section{Appendix}
\label{sec:appendix_syndarin}

\begin{table*}[hbt!]
\centering
\begin{adjustbox}{max width=\textwidth}
\begin{tabular}{lcccccccccccccccc}
\toprule
 \textbf{OTHER} & \textbf{NORP} & \textbf{GPE} & \textbf{PERCENT} & \textbf{PERSON} & \textbf{DATE} & \textbf{ORG} & \textbf{WORK OF ART} & \textbf{LANGUAGE} & \textbf{QUANTITY} & \textbf{EVENT} & \textbf{MONEY} & \textbf{LOC} & \textbf{ORDINAL} & \textbf{TIME} & \textbf{FAC} & \textbf{PRODUCT} \\
 3178 & 172 & 223 & 8 & 397 & 335 & 327 & 14 & 10 & 25 & 21 & 9 & 52 & 38 & 9 & 9 & 3 \\
\bottomrule
\end{tabular}
\end{adjustbox}
\caption{Distribution of Entities within question-answer pairs in the generated English QA dataset. The Entity labelling scheme follows \citeauthor{honnibal2020spacy}}
\label{tab:qa_ner}
\end{table*}

\subsection{Generated Question-Answer pairs}

We showcase examples of generated and validated question-answer pairs along with their designated English paragraph $\mathcal{P}_{\text{Eng}}$ in \cref{tab:qa_gen}. These are representative samples of the generation process, further reinforced by the fact that human evaluation of the quality of the generation showed that $98\%$ of the examples are answerable and maintain quality.

\subsection{What are the questions about?}

To understand the type of inquiries asked within the questions, we employ a pre-trained model for Named Entity Recognition (NER) from spaCy\footnote{\url{https://spacy.io/api/entityrecognizer}} and detect all the entity types mentioned within the question-answer pairs. The results can be seen in \cref{tab:qa_ner}, showing that the object of the inquiries can vary massively from people (PERSON) and locations (LOC) to organization (ORG), numeric values (DATE, ORDINAL, TIME), etc. This further ensures that we are able to generate high-quality questions with diverse compositions and object of inquiry types.

\subsection{Topic Distribution the parallel paragraphs}

To estimate the overlap within the topics found in the mined paragraphs, we use unsupervised topic modeling BERTopic \citep{grootendorst2022bertopic} to segment the $5$ most frequently occurring segments. We measure the overlap between these by calculating the averaged cosine distance of the topic embeddings obtained from BERTopic. The results can be seen in \cref{fig:topic_dist} and \cref{fig:topic_word_distrib}, validating our hypothesis that we are able to cover diverse themes using our parallel paragraph mining method.

\begin{table*}[ht!]
\centering
\begin{tabular}{p{\textwidth}}
\toprule
\textbf{Example 1: UEFA Champions League} \\
\midrule
Since the rebranding of the European Champion Clubs' Cup as the UEFA Champions League in 1992, 107 different players from 37 countries have scored three goals or more in a single match (a hat-trick) on 152 occasions, representing 53 clubs from 17 leagues. The first player to achieve the feat was Juul Ellerman, who scored three times for PSV Eindhoven in a 6–0 victory over Žalgiris on 16 September 1992. Lionel Messi and Cristiano Ronaldo have scored three or more goals in a match eight times each in the Champions League, more than any other player, followed by Robert Lewandowski with six, and Karim Benzema with four. \\
\textbf{Question:} What was the original name of the UEFA Champions League? \\
\textbf{Answers:} 1. European Champion Clubs' Cup, 2. European Premier League, 3. UEFA Football Cup, 4. European Soccer Championship \\
\textbf{Correct Answer:} 1. European Champion Clubs' Cup \\
\midrule
\textbf{Example 2: Sign Languages} \\
\midrule
Sign languages (also known as signed languages) are languages that use the visual-manual modality to convey meaning, instead of spoken words. Sign languages are expressed through manual articulation in combination with non-manual markers. Sign languages are full-fledged natural languages with their own grammar and lexicon. Sign languages are not universal and are usually not mutually intelligible, although there are also similarities among different sign languages. \\
\textbf{Question:} What is the primary modality used to convey meaning in sign languages? \\
\textbf{Answers:} 1. Auditory-vocal, 2. Visual-manual, 3. Tactile-kinesthetic, 4. Olfactory-gustatory \\
\textbf{Correct Answer:} 2. Visual-manual \\
\bottomrule
\end{tabular}
\caption{Examples of English paragraphs along with their generated question-answer pairs}
\label{tab:qa_gen}
\end{table*}
\begin{figure}[t!]
    \centering
    \includegraphics[width=\textwidth]{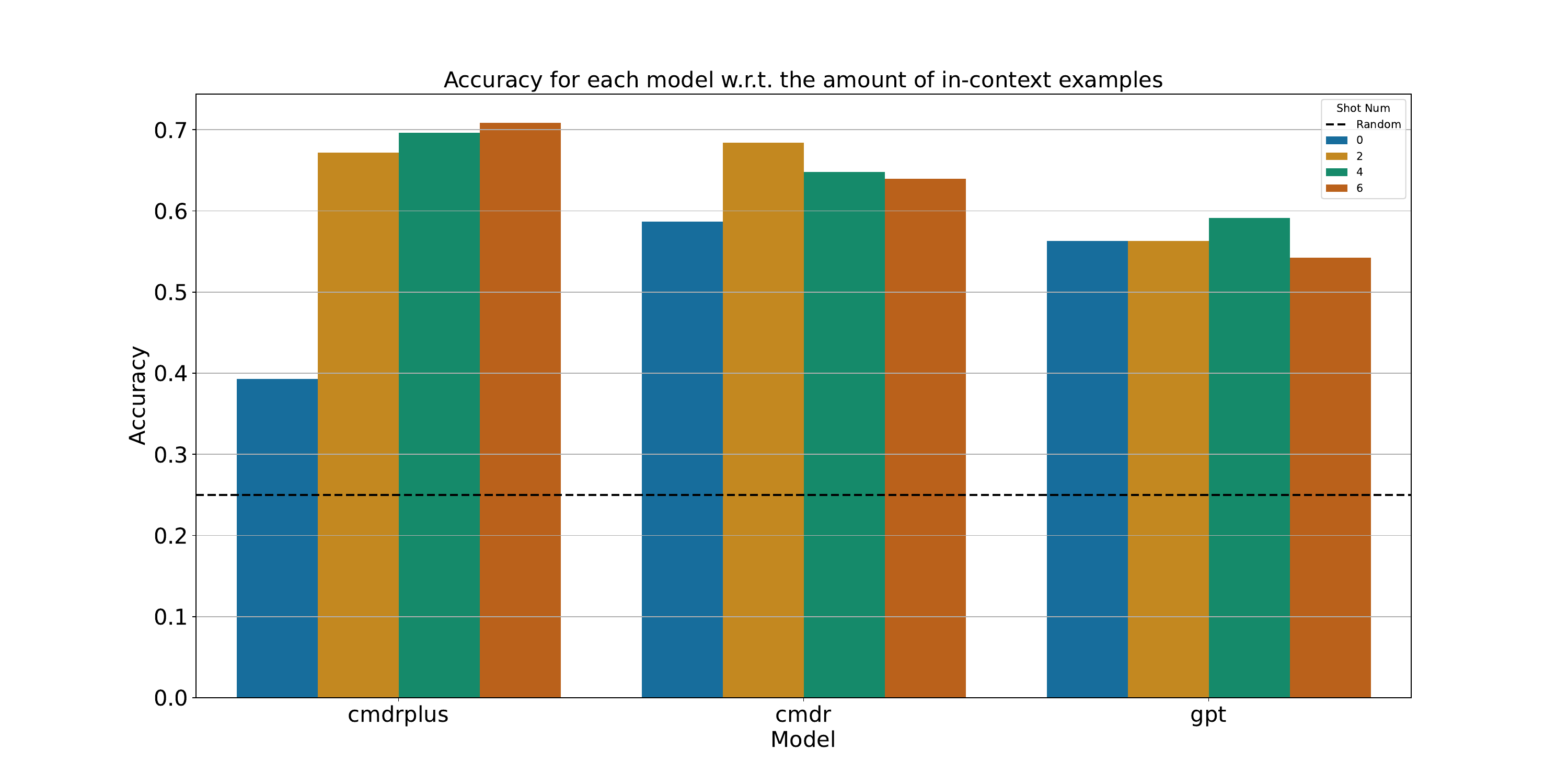}
    \caption{Accuracy of each model with a varying number of in-context examples given before generation.}
    \label{fig:topic_llm_acc}
\end{figure}

\begin{figure}[t!]
    \centering
    \includegraphics[width=\columnwidth]{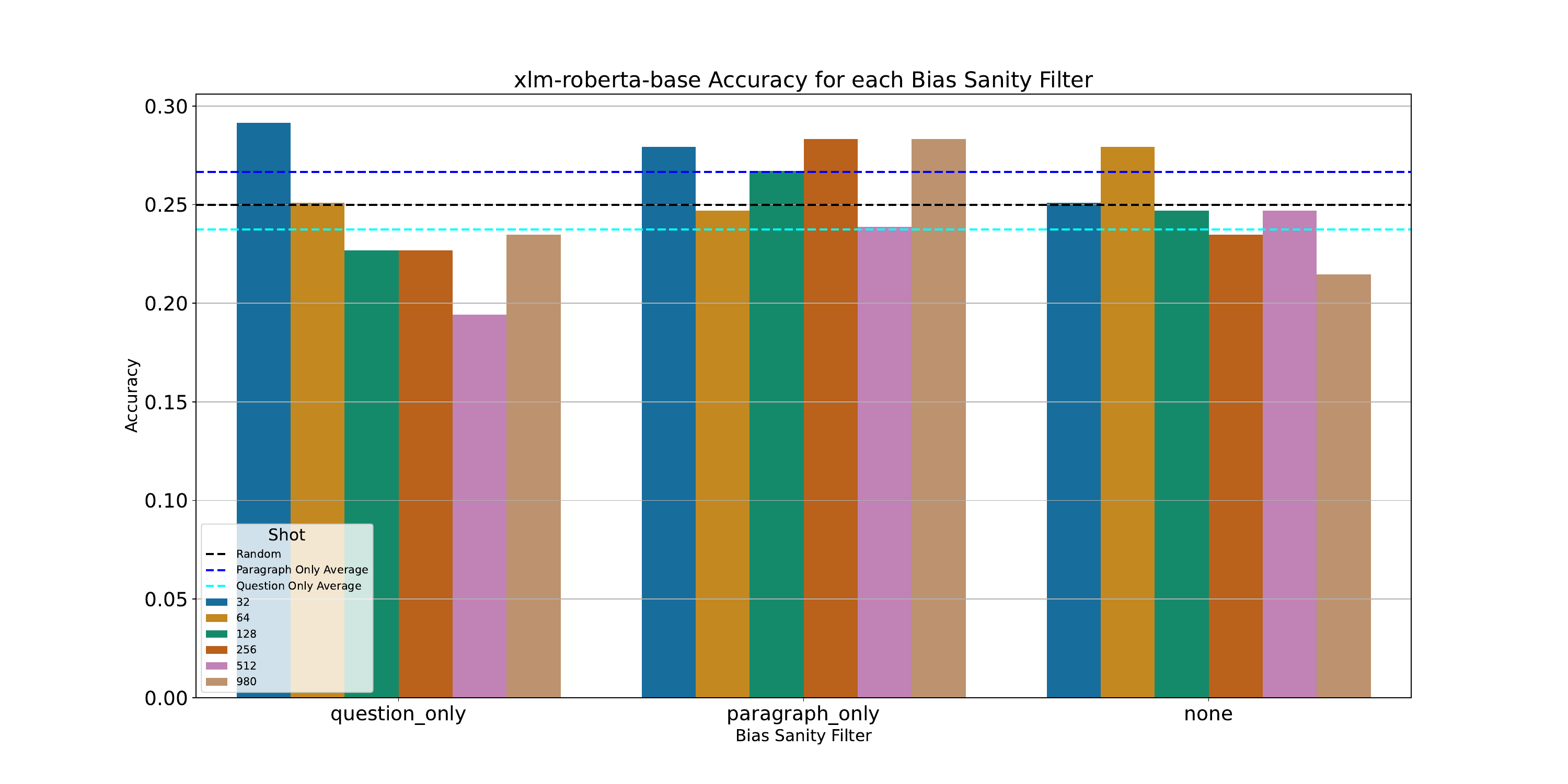}
    \caption{The results of fine-tuning XLM-Roberta on the Armenian QA dataset with a varying number of training samples while using only paragraphs, questions or random data.}
    \label{fig:topic_xlm_acc}
\end{figure}

\subsection{Benchmarking with Armenian QA dataset}

To show the usefulness of the created dataset, we benchmark several SOTA LLMs on it in supervised fine-tuning, \emph{zero-shot} and \emph{few-shot} settings. We further investigate if the dataset suffers from statistical biases or degenerate solutions by training an XLM-RoBERTa model on inputs that contain only the paragraph or the question, excluding everything else from the sample. The results in \cref{fig:topic_xlm_acc} show us that regardless of the amount of provided training samples, the question, and paragraph-only evaluations behave similarly to random chance, highlighting that the dataset is unlikely to suffer from inconsistencies and degenerate solutions.

We benchmark several LLMs, shown in \cref{fig:topic_llm_acc}, using produced Armenian QA benchmark and show that while increasing the number of model parameters and in-context samples helps the overall model performance, still even very large models are unable to solve the dataset trivially, thus showing its value as a benchmarking resource.

\part{Reasoning Inconsistencies from Task Complexity}

\chapter{Adapting Neural Link Predictors for Data-Efficient Complex Query Answering}
\label{chap:cqda}

\section{Introduction}

A Knowledge Graph (KG) is a knowledge base representing the relationships between entities in a relational graph structure.
The flexibility of this knowledge representation formalism allows KGs to be widely used in various domains.
Examples of KGs include general-purpose knowledge bases such as Wikidata~\citep{wikidata}, DBpedia~\citep{auer2007dbpedia}, 
Freebase~\citep{bollacker2008freebase},
and YAGO~\citep{suchanek2007yago}; application-driven graphs such as the Google Knowledge Graph, Microsoft's Bing Knowledge Graph, and Facebook's Social Graph~\citep{DBLP:journals/cacm/NoyGJNPT19}; and domain-specific ones such as SNOMED CT~\citep{loinc-585}, MeSH~\citep{mesh}, and Hetionet~\citep{Himmelstein087619} for life sciences; and WordNet~\citep{DBLP:conf/naacl/Miller92} for linguistics. 
Answering complex queries over Knowledge Graphs involves a logical reasoning process where a conclusion should be inferred from the available knowledge. 
Neural link predictors~\citep{nickel2015review} tackle the problem of identifying missing edges in large KGs.
However, in many domains, it is a challenge to develop techniques for answering complex queries involving multiple and potentially unobserved edges, entities, and variables rather than just single edges.

\begin{wrapfigure}{O}{0.5\textwidth}
    \centering
    \includegraphics[width=\textwidth]{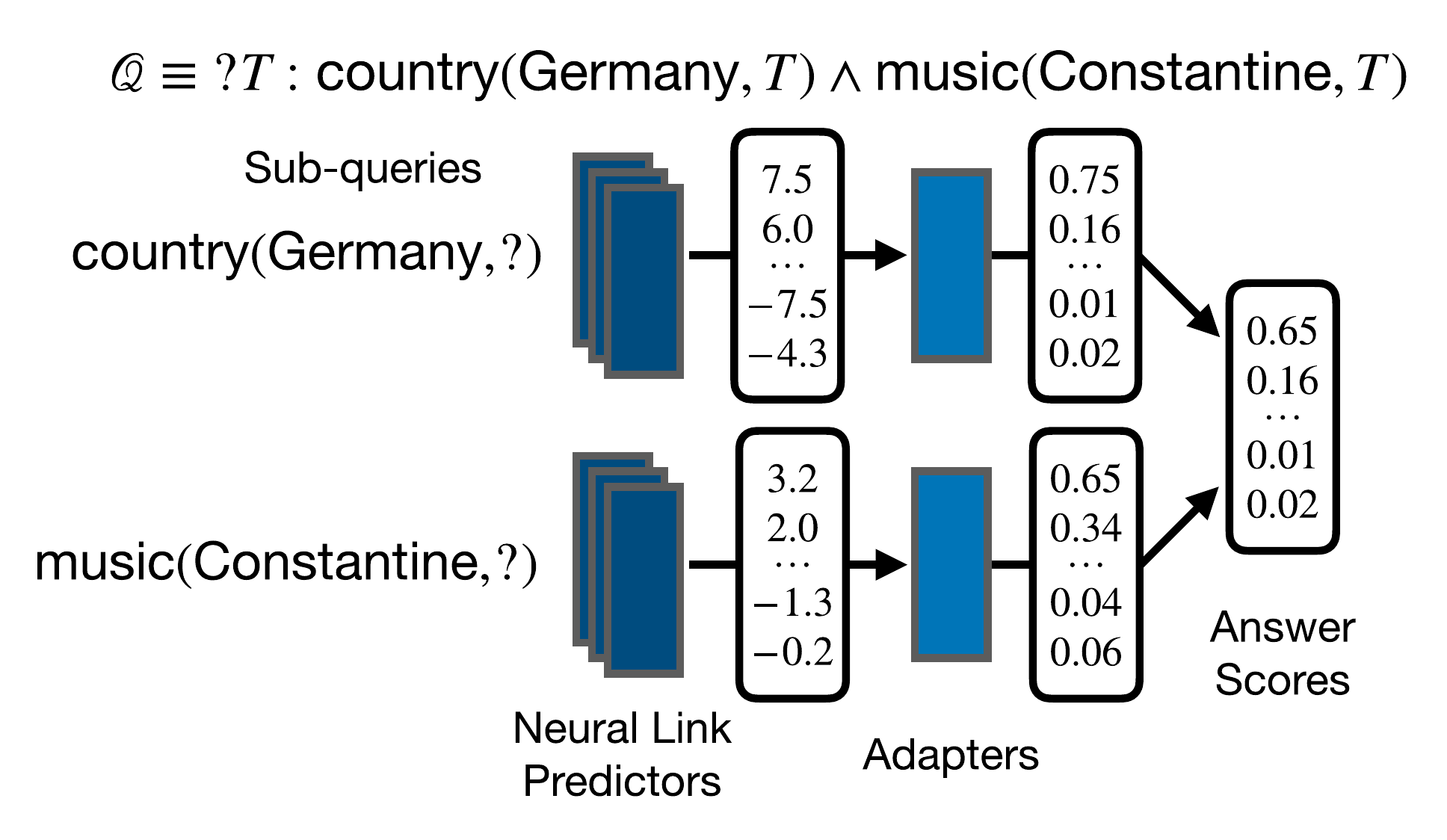}
\caption{Given a complex query $\mathcal{Q}$, \cqda adapts the neural link prediction scores for the sub-queries to improve the interactions between them.} 
\label{fig:query}
\end{wrapfigure}
%

%
Prior work proposed to address this problem using specialised neural networks trained end-to-end for the query answering task~\citep{hamilton2018embedding,daza2020message,hongyu2020query2box,ren2020beta,zhu2022neural}, which offer little interpretability and require training with large and diverse datasets of query-answer pairs.
These methods stand in contrast with Complex Query Decomposition~\citep[CQD,][]{DBLP:conf/iclr/ArakelyanDMC21,DBLP:conf/ijcai/MinerviniADC22}, which showed that it is sufficient to re-use a simple link prediction model to answer complex queries, thus reducing the amount of training data required by orders of magnitude while allowing the possibility to explain intermediate answers.
While effective, CQD does not support negations, and fundamentally, it relies on a link predictor whose scores are not necessarily calibrated for the complex query answering task.
Adapting a neural link predictor for the query answering task while maintaining the data and parameter efficiency of CQD, as well as its interpretable nature, is the open challenge we take on in this paper.
We propose \cqda, a lightweight \emph{adaptation} model trained to calibrate link prediction scores, using complex query answering as the optimisation objective.
We define the adaptation function as an affine transformation of the original score with a few learnable parameters.
The low parameter count and the fact that the adaptation function is independent of the query structure allow us to maintain the efficiency properties of CQD.
Besides, the calibration enables a natural extension of CQD to queries with atomic negations.

An evaluation of \cqda on three benchmark datasets for complex query answering shows an increase from $34.4$ to $35.1$ MRR over the current state-of-the-art averaged across all datasets while using $\leq 30\%$ of the available training query types.
In ablation experiments, we show that the method is data-efficient; it achieves results comparable to the state-of-the-art while using only $1\%$ of the complex queries.
Our experiments reveal that \cqda can generalise across unseen query types while using only $1\%$ of the instances from a single complex query type during training.
%

%
\section{Related Work}
\label{sec:related_work_cqda}
\subsection{Link Predictors in Knowledge Graphs}
\label{para:link_pred}

Reasoning over KGs with missing nodes has been widely explored throughout the last few years. One can approach the task using latent feature models, such as neural link predictors~\citep{bordes2013translating,trouillon2016,yang2014embedding,dettmers2018convolutional,sun2019rotate,balavzevic2019tucker,amin2020lowfer} which learn continuous representations for the entities and relation types in the graph and can answer atomic queries over incomplete KGs.
Other research lines tackle the link prediction problem through graph feature models~\citep{xiong2017deeppath,das2017go,hildebrandt2020reasoning,yang2017differentiable,sadeghian2019drum}, and Graph Neural Networks~\citep[GNNs,][]{schlichtkrull2018modeling, vashishth2019composition,teru2020inductive}.

\subsection{Complex Query Answering}

Complex queries over knowledge graphs can be formalised by extending one-hop atomic queries with First Order Logic (FOL) operators, such as the existential quantifier ($\exists$), conjunctions ($\land$), disjunctions ($\lor$) and negations ($\lnot$).
These FOL constructs can be represented as directed acyclic graphs, which are used by embedding-based methods that represent the queries using geometric objects \citep{hongyu2020query2box,hamilton2018embedding} or probabilistic distributions \citep{ren2020beta,zhang2021cone,choudhary2021self} and search the embedding space for the answer set.
It is also possible to enhance the properties of the embedding space using GNNs and Fuzzy Logic \citep{zhu2022neural,chen2022fuzzy}.
A recent survey \citep{ren2023neural} provides a broad overview of different approaches.
Recent work~\citep{daza2020message,hamilton2018embedding,ren2020beta} suggests that such methods require a large dataset with millions of diverse queries during the training, and it can be hard to explain their predictions.
Our work is closely related to CQD~\citep{DBLP:conf/iclr/ArakelyanDMC21,DBLP:conf/ijcai/MinerviniADC22}, which uses a pre-trained neural link predictor along with fuzzy logical t-norms and t-conorms for complex query answering.
A core limitation of CQD is that the pre-trained neural link predictor produces scores not calibrated to interact during the complex query-answering process.
This implies that the final scores of the model are highly dependent on the choice of the particular t-(co)norm aggregation functions, which, in turn, leads to discrepancies within the intermediate reasoning process and final predictions.
As a side effect, the lack of calibration also means that the equivalent of logical negation in fuzzy logic does not work as expected.

With \cqda, we propose a solution to these limitations by introducing a scalable adaptation function that calibrates link prediction scores for query answering.
Furthermore, we extend the formulation of CQD to support a broader class of FOL queries, such as queries with atomic negation.

\section{Background} \label{sec:background}
A Knowledge Graph $\mathcal{G} \subseteq \mathcal{E} \times \mathcal{R} \times \mathcal{E}$ can be defined as a set of subject-predicate-object $\langle s, p, o \rangle$ triples, where each triple encodes a relationship of type $p \in \mathcal{R}$ between the subject $s \in \mathcal{E}$ and the object $o \in \mathcal{E}$ of the triple, where $\mathcal{E}$ and $\mathcal{R}$ denote the set of all entities and relation types, respectively.
A Knowledge Graph can be represented as a First-Order Logic Knowledge Base, where each triple $\langle s, p, o \rangle$ denotes an atomic formula $p(s, o)$, with $p \in \mathcal{R}$ a binary predicate and $s, o \in \mathcal{E}$ its arguments.

\subsection{First-Order Logical Queries}
We are concerned with answering logical queries over incomplete knowledge graphs.
We consider queries that use existential quantification ($\exists$) and conjunction ($\land$) operations.
Furthermore, we include disjunctions ($\lor$) and atomic negations ($\neg$).
We follow \citet{hongyu2020query2box} by transforming a logical query into Disjunctive Normal Form~\citep[DNF,][]{DBLP:books/daglib/0023601}, \ie a disjunction of conjunctive queries, along with the subsequent extension with atomic negations in \citep{ren2020beta}. We denote such queries as follows:
\begin{equation} \label{eq:dnf-query}
\begin{aligned}
& \mathcal{Q}[A] \triangleq ?A : \exists V_{1}, \ldots, V_{m}.\left( e^{1}_{1} \land \ldots \land e^{1}_{n_{1}} \right) \lor \ldots \lor \left( e^{d}_{1} \land \ldots \land e^{d}_{n_{d}} \right), \\
& \qquad \text{where} \; e^{j}_{i} = p(c, V), \; \text{with} \; V \in \{ A, V_{1}, \ldots, V_{m} \}, c \in \mathcal{E}, p \in \mathcal{R}, \\
& \qquad \text{or} \; e^{j}_{i} = p(V, V^{\prime}), \; \text{with} \; V, V^{\prime} \in \{ A, V_{1}, \ldots, V_{m} \}, V \neq V^{\prime}, p \in \mathcal{R}.
\end{aligned}
\end{equation}

In \cref{eq:dnf-query}, the variable $A$ is the \emph{target} of the query, $V_{1}, \ldots, V_{m}$ denote the \emph{bound variable nodes}, while $c \in \mathcal{E}$ represent the \emph{input anchor nodes}, which correspond to known entities in the query.
Each $e_{i}$ denotes a logical atom, with either one ($p(c, V)$) or two variables ($p(V, V^{\prime})$).
The goal of answering the logical query $\mathcal{Q}$ consists in finding the answer set $\llbracket \mathcal{Q} \rrbracket \subseteq \mathcal{E}$ such that $a \in \llbracket \mathcal{Q} \rrbracket$ iff $\mathcal{Q}[a]$ holds true.
As illustrated in \cref{fig:query}, the \emph{dependency graph} of a conjunctive query $\mathcal{Q}$ is a graph where nodes correspond to variable or non-variable atom arguments in $\mathcal{Q}$ and edges correspond to atom predicates.
We follow \citet{hamilton2018embedding} and focus on queries whose dependency graph is a directed acyclic graph, where anchor entities correspond to source nodes, and the query target $A$ is the unique sink node.
\begin{example}[Complex Query]
Consider the question ``\emph{Which people are German and produced the music for the film Constantine?}''.
It can be formalised as a complex query $\mathcal{Q} \equiv\ ?T: \text{country}(\text{Germany}, T) \land \text{producerOf}(\text{Constantine}, T)$, where \emph{Germany} and \emph{Constantine} are anchor nodes, and $T$ is the target of the query, as presented in \cref{fig:query}.
The answer $\llbracket \mathcal{Q} \rrbracket$ corresponds to all the entities in the knowledge graph that are German composers for the film Constantine.
\end{example}

\subsection{Continuous Query Decomposition} 
CQD is a framework for answering EPFO logical queries in the presence of missing edges~\citep{DBLP:conf/iclr/ArakelyanDMC21,DBLP:conf/ijcai/MinerviniADC22}.
Given a query $\mathcal{Q}$, CQD defines the score of a target node $a \in \mathcal{E}$ as a candidate answer for a query as a function of the score of all atomic queries in $\mathcal{Q}$, given a variable-to-entity substitution for all variables in $\mathcal{Q}$.
Each variable is mapped to an \emph{embedding vector} that can either correspond to an entity $c \in \mathcal{E}$ or to a \emph{virtual entity}.
The score of each of the query atoms is determined individually using a neural link predictor~\citep{nickel2015review}.
Then, the score of the query with respect to a given candidate answer $\mathcal{Q}[a]$ is computed by aggregating all of the atom scores using t-norms and t-conorms -- continuous relaxations of the logical conjunction and disjunction operators.
\subsection{Neural Link Predictors}
A neural link predictor is a differentiable model where atom arguments are first mapped into a $d$-dimensional embedding space and then used to produce a score for the atom.
More formally, given a query atom $p(s, o)$, where $p \in \mathcal{R}$ and $s, o \in \mathcal{E}$, the score for $p(s, o)$ is computed as $\phi_{p}(\mathbf{e}_{s}, \mathbf{e}_{o})$, where $\mathbf{e}_{s}, \mathbf{e}_{o} \in \mathbb{R}^{d}$ are the embedding vectors of $s$ and $o$, and $\phi_{p} : \mathbb{R}^{d} \times \mathbb{R}^{d} \mapsto [0, 1]$ is a \emph{scoring function} computing the likelihood that entities $s$ and $o$ are related by the relationship $p$.
Following \citet{DBLP:conf/iclr/ArakelyanDMC21,DBLP:conf/ijcai/MinerviniADC22}, in our experiments, we use a regularised variant of ComplEx~\citep{trouillon2016,lacroix2018canonical} as the neural link predictor of choice, due to its simplicity, efficiency, and generalisation properties~\citep{DBLP:conf/iclr/RuffinelliBG20}.
To ensure that the output of the neural link predictor is always in $[0, 1]$, following \citet{DBLP:conf/iclr/ArakelyanDMC21,DBLP:conf/ijcai/MinerviniADC22}, we use either a sigmoid function or min-max re-scaling.
\subsection{T-norms and Negations} \label{parag:t_norms}
Fuzzy logic generalises over Boolean logic by relaxing the logic conjunction ($\wedge$), disjunction ($\vee$) and negation ($\neg$) operators through the use of t-norms, t-conorms, and fuzzy negations.
A \emph{t-norm} $\top : [0, 1] \times [0, 1] \mapsto [0, 1]$ is a generalisation of conjunction in fuzzy logic~\citep{DBLP:books/sp/KlementMP00,DBLP:journals/fss/KlementMP04a}.
Some examples include the \emph{Gödel t-norm} $\top_{\text{min}}(x, y) = \min\{ x, y \}$, the \emph{product t-norm} $\top_{\text{prod}}(x, y) = x \times y$, and the \emph{Łukasiewicz t-norm} $\top_{\text{Luk}}(x, y) = \max \{ 0, x + y - 1 \}$.
Analogously, \emph{t-conorms} are dual to t-norms for disjunctions -- given a t-norm $\top$, the complementary t-conorm is defined by $\bot(x, y) = 1 - \top(1 - x, 1 - y)$.
In our experiments, we use the Gödel t-norm and product t-norm with their corresponding t-conorms.

Fuzzy logic also encompasses negations $n : [0, 1] \mapsto [0, 1]$. The \emph{standard} $n_{\text{stand}}(x)=1-x$ and \emph{strict cosine} $n_{\text{cos}}=\frac{1}{2}(1+\cos (\pi x))$ are common examples of fuzzy negations\citep{kruse1993fuzzy}.
To support a broader class of queries, we introduce the \emph{standard} and \emph{strict cosine} functions to model negations in \cqda, which was not considered in the original formulation of CQD.
\subsection{Continuous Query Decomposition}
Given a DNF query $\mathcal{Q}$ as defined in \cref{eq:dnf-query}, CQD aims to find the variable assignments that render $\mathcal{Q}$ true.
To achieve this, CQD casts the problem of query answering as an optimisation problem.
The aim is to find a mapping from variables to entities $S = \lbrace A\leftarrow a, V_1\leftarrow v_1, \ldots, V_m\leftarrow v_m \rbrace$, where $a, v_1, \ldots, v_m\in\mathcal{E}$ are entities and $A, V_1, \ldots, V_m$ are variables, that \emph{maximises} the score of $\mathcal{Q}$:
\begin{equation} \label{eq:cont}
\begin{aligned}
& \argmax_S \text{score}(\mathcal{Q}, S) = \argmax_{A, V_{1, \ldots, m} \in \mathcal{E}} \left( e^{1}_{1} \top \ldots \top e^{1}_{n_{1}} \right) \bot \ldots \bot \left( e^{d}_{1} \top \ldots \top e^{d}_{n_{d}} \right) \\
& \qquad \text{where} \; e^{j}_{i} = \phi_{p}(\mathbf{e}_{c}, \mathbf{e}_{V}), \; \text{with} \; V \in \{ A, V_{1}, \ldots, V_{m} \}, c \in \mathcal{E}, p \in \mathcal{R} \\
& \qquad \text{or} \; e^{j}_{i} = \phi_{p}(\mathbf{e}_{V}, \mathbf{e}_{V^{\prime}}), \; \text{with} \; V, V^{\prime} \in \{ A, V_{1}, \ldots, V_{m} \}, V \neq V^{\prime}, p \in \mathcal{R},
\end{aligned}
\end{equation}
\noindent where $\top$ and $\bot$ denote a t-norm and a t-conorm -- a continuous generalisation of the logical conjunction and disjunction, respectively -- and $\phi_{p}(\mathbf{e}_{s}, \mathbf{e}_{o}) \in [0, 1]$ denotes the neural link prediction score for the atom $p(s, o)$.
\subsection{Complex Query Answering via Combinatorial Optimisation}
Following \citet{DBLP:conf/iclr/ArakelyanDMC21,DBLP:conf/ijcai/MinerviniADC22}, we solve the optimisation problem in \cref{eq:cont} by greedily searching for a set of variable substitutions $S = \{ A \leftarrow a, V_{1} \leftarrow v_{1}, \ldots, V_{m} \leftarrow v_{m} \}$, with $a, v_{1}, \ldots, v_{m} \in \mathcal{E}$, that maximises the complex query score, in a procedure akin to \emph{beam search}.
We do so by traversing the dependency graph of a query $\mathcal{Q}$ and, whenever we find an atom in the form $p(c, V)$, where $p \in \mathcal{R}$, $c$ is either an entity or a variable for which we already have a substitution, and $V$ is a variable for which we do not have a substitution yet, we replace $V$ with all entities in $\mathcal{E}$ and retain the top-$k$ entities $t \in \mathcal{E}$ that maximise $\phi_{p}(\mathbf{e}_{c}, \mathbf{e}_{t})$ -- \ie the most likely entities to appear as a substitution of $V$ according to the neural link predictor.
As we traverse the dependency graph of a query, we keep a beam with the most promising variable-to-entity substitutions identified so far.
\begin{example}[Combinatorial Optimisation]
Consider the query ``\emph{Which musicians $M$ received awards associated with a genre $g$?}, which can be rewritten as $?M : \exists A.\text{assoc}(g, A)\land \text{received}(A, M)  $.
To answer this query using combinatorial optimisation, we must find the top-$k$ awards $a$ that are candidates to substitute the variable $A$ in $\text{assoc}(g,A)$.
This will allow us to understand the awards associated with the genre $g$.
Afterwards, for each candidate substitution for $A$, we search for the top-$k$ musicians $m$ that are most likely to substitute $M$ in $\text{received}(A, M)$, ending up with $k^{2}$ musicians. Finally, we rank the $k^2$ candidates using the final query score produced by a t-norm. $\blacksquare$
\end{example}
\section{Calibrating Link Prediction Scores on Complex Queries}
\label{sec:adapting}

The main limitation in the CQD method outlined in \cref{sec:background} is that neural link predictors $\phi$ are trained to answer simple, atomic queries, and the resulting answer scores are not trained to interact with one another.

\begin{figure}
\begin{floatrow}
\ffigbox{%
    \centering
    \includegraphics[width=\columnwidth]{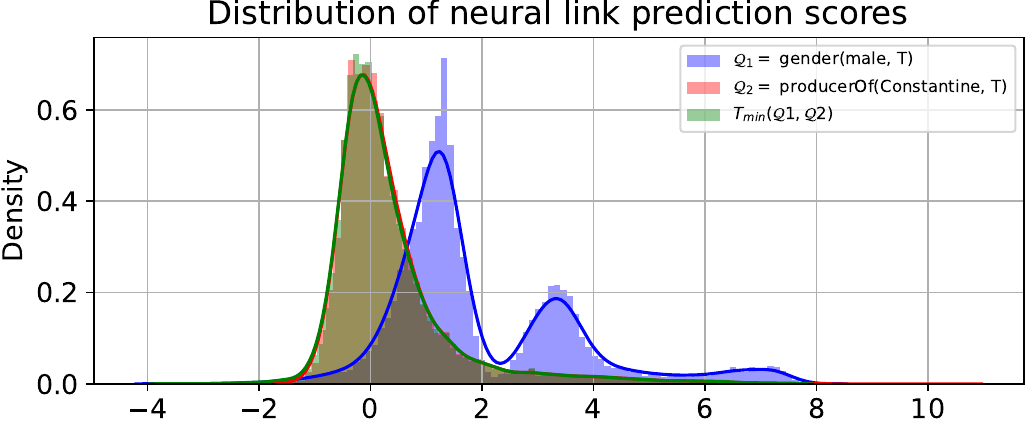}
}{%
    \caption{The distributions of two atomic scores $\mathcal{Q}_{{1}}$ and $\mathcal{Q}_{{2}}$, and the aggregated results via $\top_{min}$ -- the scores from $\mathcal{Q}_{{2}}$ dominate the final scores.}
\label{fig:lp_variation}
}
\capbtabbox{%
  \caption{Statistics on the different types of query structures in FB15K, FB15K-237, and NELL995.} \label{tab:queries}
}{%
    \resizebox{0.5\textwidth}{!}{%
    \begin{tabular}{@{}ccccc@{}}
    \toprule
    {\bf Split}                  & {\bf Query Types}     & {\bf FB15K}  & {\bf FB15K-237} & {\bf NELL995} \\ \midrule
    \multirow{2}{*}{\bf Train} & {1p}, {2p}, {3p}, {2i}, {3i}   & 273,710 & 149,689 & 107,982 \\
                           & {2in}, {3in}, {inp}, {pin}, {pni} & 27,371  & 14,968  & 10,798  \\ \midrule
    \multirow{2}{*}{\bf Valid} & {1p} & 59,078 & 20,094    & 16,910  \\
                           & Others         & 8,000  & 5,000     & 4,000   \\ \midrule
    \multirow{2}{*}{\bf Test}  & {1p} & 66,990 & 22,804    & 17,021  \\
                           & Others         & 8,000  & 5,000     & 4,000   \\ \bottomrule
    \end{tabular}%
    }
}
\end{floatrow}
\end{figure}

\begin{example}
    Consider the running example query ``\emph{Which people are German and produced the music for the film Constantine?}'' which can be rewritten as a complex query $\mathcal{Q} \equiv\ ?T: \text{country}(\text{Germany}, T) \land \text{producerOf}(\text{Constantine}, T)$.
    To answer this complex query, CQD answers the atomic sub-queries $ \mathcal{Q}_1 = \text{country}(\text{Germany}, T)$ and $\mathcal{Q}_2 = \text{producerOf}(\text{Constantine}, T)$ using a neural link predictor, and aggregates the resulting scores using a t-norm. 
    However, the neural link predictor was only trained on answering atomic queries, and the resulting scores are not calibrated to interact with each other.
    For example, the scores for the atomic queries about the relations $\text{country}$ and $\text{producerOf}$ may be on different scales, which causes problems when aggregating such scores via t-norms.
    Let us assume the top candidates for the variable $T$ coming from the atomic queries $\mathcal{Q}_1, \mathcal{Q}_2$ are $\mathcal{A}_1 \gets \emph{Sam Shepard}$ and $\mathcal{A}_2 \gets \emph{Klaus Badelt}$, with their corresponding neural link prediction scores $1.2$ and $8.9$, produced using $\phi_{\textit{country}}$ and $\phi_{\textit{producerOf}}$.
    We must also factor in the neural link prediction score of the candidate $\mathcal{A}_1$ for query $\mathcal{Q}_2$ at $7.4$ and vice versa at $0.5$.
    When using the Gödel t-norm $\top_{\text{min}}(x, y) = \min\{ x, y \}$, the scores associated with the variable assignments $\mathcal{A}_1,\mathcal{A}_2$ are computed as, $\min(8.0,0.5)=0.5$ $\min(7.4, 1.2) = 1.2$.
    For both answers $\mathcal{A}_{1}$ and $\mathcal{A}_{2}$, the scores produced by $\phi_{\textit{country}}$ for $\mathcal{Q}_1$ are always lower than the scores produced with $\phi_{\textit{producerOf}}$ for $\mathcal{Q}_2$, meaning that the scores of the latter are not considered when producing the final answer.
    This phenomenon can be broadly observed in CQD, illustrated in \cref{fig:lp_variation}. $\blacksquare$
\end{example}
To address this problem, we propose a method for adaptively learning to calibrate neural link prediction scores by back-propagating through the complex query-answering process.
More formally, let $\phi_{p}$ denote a neural link predictor.
We learn an additional adaptation function $\rho_{\theta}$, parameterised by $\theta = \{ \alpha, \beta \}$, with $\alpha, \beta \in \mathbb{R}$.
Then, we use the composition of $\rho_{\theta}$ and $\phi_{p}$, $\rho_{\theta} \circ \phi_{p}$, such that:
\begin{equation} \label{eq:adapt}
\rho_{\theta}(\phi_{p}(\mathbf{e}_{V}, \mathbf{e}_{V^{\prime}})) = \phi_{p}(\mathbf{e}_{V}, \mathbf{e}_{V^{\prime}}) (1 + \alpha) + \beta.
\end{equation}
%
%
Here, the function $\rho$ defines an affine transformation of the score and when the parameters $\alpha = \beta = 0$, the transformed score $\rho_{\theta}(\phi_{p}(\mathbf{e}_{V}, \mathbf{e}_{V^{\prime}}))$ recovers the original scoring function.
The parameters $\theta$ can be conditioned on the representation of the predicate $p$ and the entities $V$ and $V^{\prime}$, \ie $\theta = \psi(\mathbf{e}_{V}, \mathbf{e}_{p}, \mathbf{e}_{V^{\prime}})$; here, $\psi$ is an end-to-end differentiable neural module with parameters $\mathbf{W}$. $\mathbf{e}_{V}$, $\mathbf{e}_{p}$, $\mathbf{e}_{V^{\prime}}$ respectively denote the representations of the subject, predicate, and object of the atomic query.
In our experiments, we consider using one or two linear transformation layers with a ReLU non-linearity as options for $\psi$. 

The motivation for our proposed adaptation function is twofold.
Initially, it is monotonic, which is desirable for maintaining the capability to interpret intermediate scores, as in the original formulation of CQD.
Moreover, we draw inspiration from the use of affine transformations in methodologies such as Platt scaling~\citep{platt1999probabilistic}, which also use a linear function for calibrating probabilities and have been applied in the problem of calibration of link prediction models~\citep{tabacof2020probability}.
Parameter-efficient adaptation functions have also been applied effectively in other domains, such as adapter layers~\cite{houlsby2019parameter} used for fine-tuning language models in NLP tasks.

%
%

%
\subsection{Training}
For training the score calibration component in \cref{eq:adapt}, we first compute how likely each entity $a^{\prime} \in \mathcal{E}$ is to be an answer to the query $\mathcal{Q}$.
To this end, for each candidate answer $a^{\prime} \in \mathcal{E}$, we compute the \emph{answer score} as the complex query score assuming that $a^{\prime} \in \mathcal{E}$ is the final answer as:
\begin{equation} \label{eq:cmax}
\begin{aligned}
\text{score}(\mathcal{Q}, A \leftarrow a^{\prime}) = & \max_{S} \text{score}(\mathcal{Q}, S), \; \text{where} \; A \leftarrow a^{\prime} \in S.
\end{aligned}
\end{equation}
\cref{eq:cmax} identifies the variable-to-entity substitution $S$ that 
\begin{inparaenum}[1)]
\item maximises the query score $\text{score}(\mathcal{Q}, S)$, defined in \cref{eq:cont}, and
\item associates the answer variable $A$ with $a^{\prime} \in \mathcal{E}$, \ie $A \leftarrow a^{\prime} \in S$.
\end{inparaenum}
For computing $S$ with the additional constraint that $A \leftarrow a^{\prime} \in S$, we use the complex query answering procedure outlined in \cref{sec:background}.
We optimise the additional parameters $\mathbf{W}$ introduced in \cref{sec:adapting}, by gradient descent on the likelihood of the true answers on a dataset $\mathcal{D} = \{ (\mathcal{Q}_i, a_i) \}_{i=1}^{|\mathcal{D}|}$ of query-answer pairs by using a \emph{1-vs-all} cross-entropy loss, introduced by \citet{lacroix2018canonical}, which was also used to train the neural link prediction model:
\begin{equation} \label{eq:loss}
\begin{aligned}
\mathcal{L}(\mathcal{D}) = & \sum_{\left( \mathcal{Q}_i, a_i \right) \in \mathcal{D}} - \text{score}(\mathcal{Q}_i, A \leftarrow a_i) + \log \left[ \sum_{a^{\prime} \in \mathcal{E}} \exp\left( \text{score}(\mathcal{Q}_i, A \leftarrow a^{\prime})  \right) \right].
\end{aligned}
\end{equation}
In addition to the \emph{1-vs-all}~\citep{DBLP:conf/iclr/RuffinelliBG20} loss in \cref{eq:loss}, we also experiment with the binary cross-entropy loss, using the negative sampling procedure from \citet{ren2020beta}.
%

%
%

\section{Experiments}

\subsection{Datasets}
\begin{figure*}[t]
    \centering
    \includegraphics[width=\textwidth]{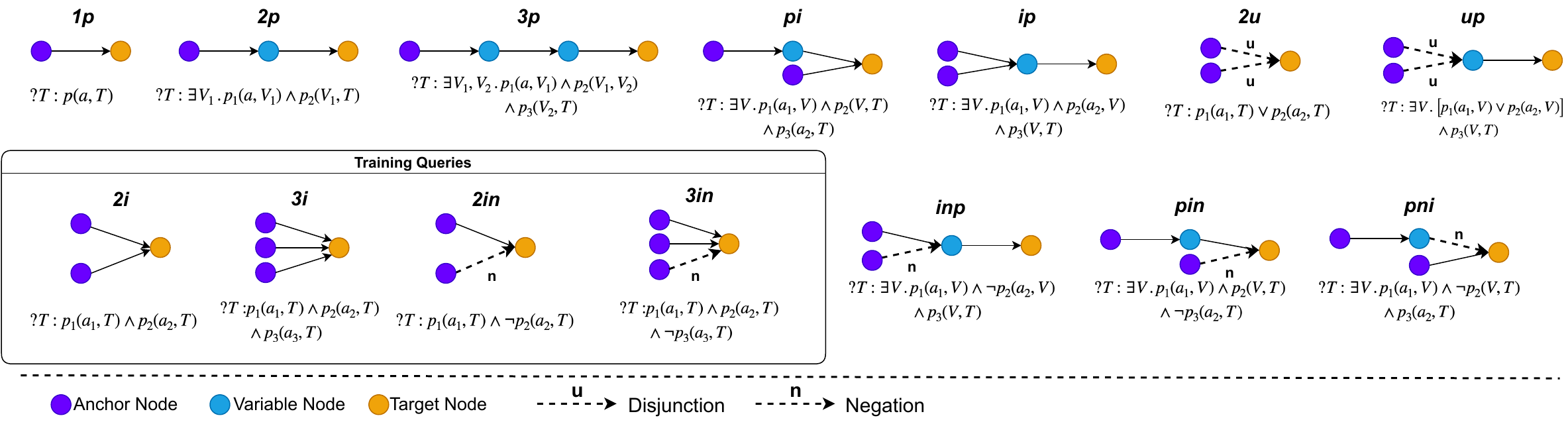}
    \caption{Query structures considered in our experiments, as proposed by \citet{ren2020beta} -- the naming of each query structure corresponds to \emph{projection} ({\bf p}), \emph{intersection} ({\bf i}), \emph{union} ({\bf u}) and \emph{negation} ({\bf n}), reflecting how they were generated in the BetaE paper~\citep{ren2020beta}. An example of a {\bf pin} query is $?T : \exists V . p(a, V), q(V, T), \neg r(b, T)$, where $a$ and $b$ are anchor nodes, $V$ is a variable node, and $T$ is the query target node.}
    \label{fig:query_types}
\end{figure*}

To evaluate the complex query answering capabilities of our method, we use a benchmark comprising of 3 KGs: FB15K \citep{bordes2013translating}, FB15K-237 \citep{toutanova2015observed} and NELL995 \citep{xiong2017deeppath}.
For a fair comparison with previous work, we use the datasets of FOL queries proposed by \citet{ren2020beta}, which includes nine structures of EPFO queries and 5 query types with atomic negations, seen in \cref{fig:query_types}.
The datasets provided by \citet{ren2020beta} introduce queries with \emph{hard} answers, which are the answers that cannot be obtained by direct graph traversal; in addition, this dataset does not include queries with more than 100 answers, 
increasing the difficulty of the complex query answering task.
The statistics for each dataset can be seen in \cref{tab:queries}.
Note that during training, we only use \emph{2i}, \emph{3i}, \emph{2in}, and \emph{3in} queries, corresponding to $\leq30\%$ of the training dataset, for the adaptation of the neural link predictor.
To assess the model's ability to generalise, we evaluate it on all query types.

\begin{table*}[t]
\centering
\caption{MRR results for FOL queries on the testing sets. $\textbf{avg}_p$ designates the averaged results for EPFO queries ($\wedge, \vee$), while $\textbf{avg}_n$ pertains to queries including negations ($\neg$). The results for CQD are taken from \citet{DBLP:conf/ijcai/MinerviniADC22}, while all the remaining come from \citet{zhu2022neural}.}
\resizebox{\textwidth}{!}{
\begin{tabular}{@{}ccccccccccccccccc@{}}
\toprule
\multicolumn{1}{l}{\textbf{Model}} &
  \multicolumn{1}{l}{$\textbf{avg}_p$} &
  \multicolumn{1}{l}{$\textbf{avg}_n$}&
  \multicolumn{1}{l}{\textbf{1p}} &
  \multicolumn{1}{l}{\textbf{2p}} &
  \multicolumn{1}{l}{\textbf{3p}} &
  \multicolumn{1}{l}{\textbf{2i}} &
  \multicolumn{1}{l}{\textbf{3i}} &
  \multicolumn{1}{l}{\textbf{pi}} &
  \multicolumn{1}{l}{\textbf{ip}} &
  \multicolumn{1}{l}{\textbf{2u}} &
  \multicolumn{1}{l}{\textbf{up}} &
  \multicolumn{1}{l}{\textbf{2in}} &
  \multicolumn{1}{l}{\textbf{3in}} &
  \multicolumn{1}{l}{\textbf{inp}} &
  \multicolumn{1}{l}{\textbf{pin}} &
  \multicolumn{1}{l}{\textbf{pni}} \\ \midrule
\multicolumn{17}{c}{\bf FB15K} \\ \midrule
GQE &
  28.0 &
  - &
  54.6 &
  15.3 &
  10.8 &
  39.7 &
  51.4 &
  27.6 &
  19.1 &
  22.1 &
  11.6 &
  - &
  - &
  - &
  - &
  - \\
Q2B &
  38.0 &
  - &
  68.0 &
  21.0 &
  14.2 &
  55.1 &
  66.5 &
  39.4 &
  26.1 &
  35.1 &
  16.7 &
  - &
  - &
  - &
  - &
  - \\
BetaE &
  41.6 &
  11.8 &
  65.1 &
  25.7 &
  24.7 &
  55.8 &
  66.5 &
  43.9 &
  28.1 &
  40.1 &
  25.2 &
  14.3 &
  14.7 &
  11.5 &
  6.5 &
  12.4 \\
CQD-CO &
  46.9 &
  - &
  \textbf{89.2} &
  25.3 &
  13.4 &
  74.4 &
  78.3 &
  44.1 &
  33.2 &
  41.8 &
  21.9 &
  - &
  - &
  - &
  - &
  - \\
CQD-Beam &
  58.2 &
  - &
  \textbf{89.2} &
  54.3 &
  28.6 &
  74.4 &
  78.3 &
  58.2 &
  67.7 &
  42.4 &
  30.9 &
  - &
  - &
  - &
  - &
  - \\
ConE &
  49.8 &
  14.8 &
  73.3 &
  33.8 &
  29.2 &
  64.4 &
  73.7 &
  50.9 &
  35.7 &
  55.7 &
  31.4 &
  17.9 &
  18.7 &
  12.5 &
  9.8 &
  15.1 \\
\multicolumn{1}{l}{GNN-QE} &
  \multicolumn{1}{l}{\textbf{72.8}} &
  \multicolumn{1}{l}{38.6} &
  \multicolumn{1}{l}{88.5} &
  \multicolumn{1}{l}{\textbf{69.3}} &
  \multicolumn{1}{l}{58.7} &
  \multicolumn{1}{l}{\textbf{79.7}} &
  \multicolumn{1}{l}{\textbf{83.5}} &
  \multicolumn{1}{l}{69.9} &
  \multicolumn{1}{l}{70.4} &
  \multicolumn{1}{l}{\textbf{74.1}} &
  \multicolumn{1}{l}{\textbf{61.0}} &
  \multicolumn{1}{l}{44.7} &
  \multicolumn{1}{l}{41.7} &
  \multicolumn{1}{l}{\textbf{42.0}} &
  \multicolumn{1}{l}{30.1} &
  \multicolumn{1}{l}{\textbf{34.3}} \\ 
\multicolumn{1}{l}{\cqda} &
  \multicolumn{1}{l}{70.4} &
  \multicolumn{1}{l}{\textbf{42.8}} &
  \multicolumn{1}{l}{\textbf{89.2}} &
  \multicolumn{1}{l}{64.5} &
  \multicolumn{1}{l}{57.9} &
  \multicolumn{1}{l}{76.1} &
  \multicolumn{1}{l}{79.4} &
  \multicolumn{1}{l}{\textbf{70.0}} &
  \multicolumn{1}{l}{\textbf{70.6}} &
  \multicolumn{1}{l}{68.4} &
  \multicolumn{1}{l}{57.9} &
  \multicolumn{1}{l}{\textbf{54.7}} &
  \multicolumn{1}{l}{\textbf{47.1}} &
  \multicolumn{1}{l}{37.6} &
  \multicolumn{1}{l}{\textbf{35.3}} &
  \multicolumn{1}{l}{24.6} \\ \midrule
\multicolumn{17}{c}{\bf FB15K-237} \\ \midrule
GQE &
  16.3 &
  - &
  35.0 &
  7.2 &
  5.3 &
  23.3 &
  34.6 &
  16.5 &
  10.7 &
  8.2 &
  5.7 &
  - &
  - &
  - &
  - &
  - \\
Q2B &
  20.1 &
  - &
  40.6 &
  9.4 &
  6.8 &
  29.5 &
  42.3 &
  21.2 &
  12.6 &
  11.3 &
  7.6 &
  - &
  - &
  - &
  - &
  - \\
BetaE &
  20.9 &
  5.5 &
  39.0 &
  10.9 &
  10.0 &
  28.8 &
  42.5 &
  22.4 &
  12.6 &
  12.4 &
  9.7 &
  5.1 &
  7.9 &
  7.4 &
  3.5 &
  3.4 \\
CQD-CO &
  21.8 &
  - &
  \textbf{46.7} &
  9.5 &
  6.3 &
  31.2 &
  40.6 &
  23.6 &
  16.0 &
  14.5 &
  8.2 &
  - &
  - &
  - &
  - &
  - \\
CQD-Beam &
  22.3 &
  - &
  \textbf{46.7} &
  11.6 &
  8.0 &
  31.2 &
  40.6 &
  21.2 &
  18.7 &
  14.6 &
  8.4 &
  - &
  - &
  - &
  - &
  - \\
ConE &
  23.4 &
  5.9 &
  41.8 &
  12.8 &
  11.0 &
  32.6 &
  47.3 &
  25.5 &
  14.0 &
  14.5 &
  10.8 &
  5.4 &
  8.6 &
  7.8 &
  4.0 &
  3.6 \\
GNN-QE &
  \textbf{26.8} &
  10.2 &
  42.8 &
  \textbf{14.7} &
  \textbf{11.8} &
  \textbf{38.3} &
  \textbf{54.1} &
  \textbf{31.1} &
  18.9 &
  16.2 &
  \textbf{13.4} &
  10.0 &
  16.8 &
  9.3 &
  7.2 &
  \textbf{7.8} \\ 
\cqda &
   25.3 &
   \textbf{10.9} &
  \textbf{46.7} &
  13.6 &
  11.4 &
  33.1 &
  45.4 &
  26.5 &
  \textbf{20.4} &
  \textbf{17.5} &
  11.4 &
  \textbf{13.6} &
  \textbf{16.8} &
  \textbf{9.5} &
  \textbf{8.9} &
  5.8 \\ \midrule
\multicolumn{17}{c}{\bf NELL995} \\ \midrule
GQE &
  18.6 &
  - &
  32.8 &
  11.9 &
  9.6 &
  27.5 &
  35.2 &
  18.4 &
  14.4 &
  8.5 &
  8.8 &
  - &
  - &
  - &
  - &
  - \\
Q2B &
  22.9 &
  - &
  42.2 &
  14.0 &
  11.2 &
  33.3 &
  44.5 &
  22.4 &
  16.8 &
  11.3 &
  10.3 &
  - &
  - &
  - &
  - &
  - \\
BetaE &
  24.6 &
  5.9 &
  53.0 &
  13.0 &
  11.4 &
  37.6 &
  47.5 &
  24.1 &
  14.3 &
  12.2 &
  8.5 &
  5.1 &
  7.8 &
  10.0 &
  3.1 &
  3.5 \\
CQD-CO &
  28.8 &
  - &
  \textbf{60.4} &
  17.8 &
  12.7 &
  39.3 &
  46.6 &
  30.1 &
  22.0 &
  17.3 &
  13.2 &
  - &
  - &
  - &
  - &
  - \\
CQD-Beam &
  28.6 &
  - &
  \textbf{60.4} &
  20.6 &
  11.6 &
  39.3 &
  46.6 &
  25.4 &
  23.9 &
  17.5 &
  12.2 &
  - &
  - &
  - &
  - &
  - \\
ConE &
  27.2 &
  6.4 &
  53.1 &
  16.1 &
  13.9 &
  40.0 &
  50.8 &
  26.3 &
  17.5 &
  15.3 &
  11.3 &
  5.7 &
  8.1 &
  10.8 &
  3.5 &
  3.9 \\
GNN-QE &
  28.9 &
  9.7 &
  53.3 &
  18.9 &
  14.9 &
  42.4 &
  52.5 &
  30.8 &
  18.9 &
  15.9 &
  12.6 &
  9.9 &
  14.6 &
  11.4 &
  6.3 &
  6.3 \\ 
\cqda &
  \textbf{32.3} &
  \textbf{13.3} &
  \textbf{60.4} &
  \textbf{22.9} &
  \textbf{16.7} &
  \textbf{43.4} &
  \textbf{52.6} &
  \textbf{32.1} &
  \textbf{26.4} &
  \textbf{20.0} &
  \textbf{17.0} &
  \textbf{15.1} &
  \textbf{18.6} &
  \textbf{15.8} &
  \textbf{10.7} &
  \textbf{6.5} \\ \bottomrule
\end{tabular}
}
\label{tab:results_ours}
\end{table*}

\subsection{Evaluation Protocol}
For a fair comparison with prior work, we follow the evaluation scheme in \citet{ren2020beta} by separating the answer of each query into \emph{easy} and \emph{hard} sets. For test and validation splits, we define \emph{hard} queries as those that cannot be answered via direct traversal along the edges of the KG and can only be answered by predicting at least one missing link, meaning \emph{non-trivial} reasoning should be completed. We evaluate the method on non-trivial queries by calculating the rank $r$ for each hard answer against non-answers and computing the Mean Reciprocal Rank (MRR).

\subsection{Baselines}
We compare \cqda with state-of-the-art methods from various solution families in \cref{sec:related_work_cqda}.
In particular, we choose GQE~\citep{hamilton2018embedding}, Query2Box~\citep{hongyu2020query2box}, BetaE~\citep{ren2020beta} and ConE~\citep{zhang2021cone} as strong baselines for query embedding methods.
We also compare with methods based on GNNs and fuzzy logic, such as FuzzQE~\citep{chen2022fuzzy}, GNN-QE~\citep{zhu2022neural}, and the original CQD \citep{DBLP:conf/iclr/ArakelyanDMC21,DBLP:conf/ijcai/MinerviniADC22}, which uses neural link predictors for answering EPFO queries without any fine-tuning on complex queries.

\subsection{Model Details}
Our method can be used with any neural link prediction model.
Following \citet{DBLP:conf/iclr/ArakelyanDMC21,DBLP:conf/ijcai/MinerviniADC22}, we use ComplEx-N3~\citep{lacroix2018canonical}.
We identify the optimal hyper-parameters using the validation MRR.
We train for $50,000$ steps using Adagrad as an optimiser and 0.1 as the learning rate.
The beam-size hyper-parameter $k$ was selected in $k \in \{ 512, 1024, \ldots, 8192 \}$, and the loss was selected across \emph{1-vs-all}~\citep{lacroix2018canonical} and binary cross-entropy with one negative sample.
\subsection{Parameter Efficiency}
\begin{figure}
\begin{floatrow}
\ffigbox{%
    \centering
\includegraphics[clip,width=0.48\textwidth]{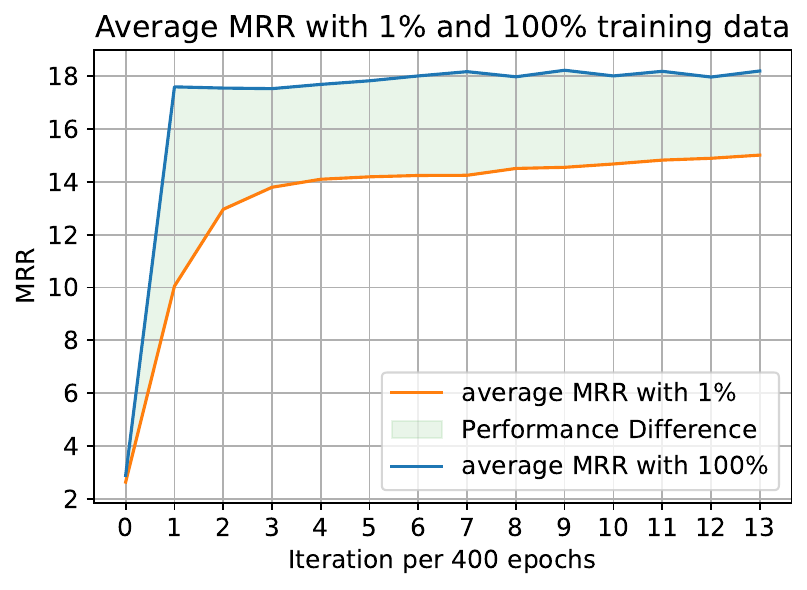}
}{%
    \caption{Average test MRR score ($y$-axis) of \cqda using $1\%$ and $100\%$ of the training queries from FB15K-237 throughout the training iterations ($x$-axis).} \label{fig:adapt_speed}
}
\killfloatstyle
\capbtabbox{%
    \resizebox{0.48\textwidth}{!}{%
    \begin{tabular}{@{}cccc@{}}
    \toprule
    \multicolumn{4}{c}{\bf Number of parameters} \\
    \midrule
    & \multicolumn{1}{c}{FB15K} & \multicolumn{1}{c}{FB15K-237} & NELL \\ \cmidrule(lr){2-2} \cmidrule(lr){3-3} \cmidrule(lr){4-4} 
    \cqda & $\underbrace{1.3 \times 10^7}_{frozen}$                        & $\underbrace{1.3 \times 10^7}_{frozen}$                           & $\underbrace{7.5 \times 10^7}_{frozen}$ \\
            &\textit{{         }}\small \qquad $\mathbf{+ 4 \times 10^3}$ &\small \qquad $\mathbf{+ 4 \times 10^3}$   & \small \qquad $\mathbf{+ 4 \times 10^3}$ \\
    BetaE   & $1.3 \times 10^7$                        & $1.3 \times 10^7$                            & $6 \times 10^7$ \\
    Q2B     & $1.2 \times 10^7$                       & $1.2 \times 10^7$                            & $6 \times 10^7$ \\
    GNN-QE  & $ 3\times 10^6$                        & $3 \times 10^6$                            & $3 \times 10^6$  \\
    ConE    & $1.2 \times 10^7$                       & $1.2 \times 10^7$                           & $6 \times 10^7$  \\
    GQE     & $1.5 \times 10^7$                        & $1.5 \times 10^7$                           & $7.5 \times 10^7$ \\ \bottomrule
    \end{tabular}%
    }
    
}{%
\vspace{20pt}
    \caption{Number of parameters used by different complex query answering methods -- values for GNN-QE are approximated using the backbone NBFNet \citep{zhu2021neural}, while the remaining use their original studies.}
    \label{tab:parameters}
}

\end{floatrow}
\end{figure}

We use the query types \emph{2i}, \emph{3i}, \emph{2in}, \emph{3in} for training the calibration module proposed in \cref{sec:adapting}.
We selected these query types as they do not require variable assignments other than for the answer variable $A$, making the training process efficient.
%
%
As the neural link prediction model is frozen, we only train the adapter layers that have a maximum of $\mathbf{W} \in \mathbb{R}^{2 \times 2d}$ learnable weights.
Compared to previous works, we have $\approx 10^3$ times fewer \emph{trainable} parameters, as shown in \cref{tab:parameters}, while maintaining competitive results.

\subsection{Results}

\subsection{Complex Query Answering}
\cref{tab:results_ours} shows the predictive accuracy of \cqda for answering complex queries compared to the current state-of-the-art methods.
Some methods do not support queries that include negations; we leave the corresponding entries blank.
We can see that \cqda increases the MRR from $34.4$ to $35.1$ averaged across all query types and datasets. In particular, \cqda shows the most substantial increase in predictive accuracy on NELL995 by producing more accurate results than all other methods for all query types.
\cqda achieves such results using less than $30\%$ of the complex query types during training while maintaining competitive results across each dataset and query type.
For queries including negations, \cqda achieves a relative improvement of $6.8\%$ to $37.1\%$, which can be attributed to the fact that the adaptation is completed with query types \emph{2in} and \emph{3in} that include negation, which allows for learning an adaptation layer that is robust for these types of queries. 
In our experiments, we found that calculating the neural adaptation parameters $\theta$ of the adaptation function $\rho_{\theta}$ in \cref{eq:adapt} as a function of the predicate representation yields the most accurate results followed by computing $\theta$ as a function of the source entity and predicate representation, which is strictly more expressive.
In \cref{app:adapt}, we show the impact of the adaptation layers on the neural link prediction scores.
The adaptation process does not require data-intensive training and allows the model to generalise to query types not observed during training. This prompts us to investigate the minimal amount of data samples and query types required for adaptation.

\subsection{Data Efficiency}
To analyse the data efficiency of \cqda, we compare the behaviour of the pre-trained link predictors tuned with $1\%$ and $100\%$ of the training complex query examples in FB15K-237, presented in \cref{tab:data_efficient}.
For adapting on $1\%$ of the training complex queries, we used the same hyper-parameters we identified when training on the full dataset.
Even when using $1\%$ of the complex training queries ($3290$ samples) for tuning, the model still achieves competitive results, with an average MRR difference of $2.2$ compared to the model trained using the entire training set. \cqda also produces higher test MRR results than GNN-QE with an average MRR increase of $4.05$.

\begin{table*}[t]
\centering
\resizebox{\textwidth}{!}{%
\begin{tabular}{@{}=l+c+c+c+c+c+c+c+c+c+c+c+c+c+c+c@{}}
\toprule
\bf Dataset & \bf Model & \bf 1p & \bf 2p & \bf 3p & \bf 2i & \bf 3i & \bf pi & \bf ip & \bf 2u & \bf up & \bf 2in & \bf 3in & \bf inp & \bf pin & \bf pni \\

\midrule






\multirow{3}{*}{FB237, 1\%} & \cqda    & \textbf{46.7} & \textbf{11.8} & \textbf{11.4} & \textbf{33.6} & 41.2           & \textbf{24.82} & \bf 17.81          & \textbf{16.45} & \bf 8.74          & \textbf{10.8} & \textbf{13.86} & 5.93          & \textbf{5.38} & \bf 14.82          \\

& GNN-QE  & 36.82         & 8.96           & 8.13           & 33.02         & \bf 49.28 & 24.58          & 14.18          &10.73          & 8.47          & 4.89           & 12.31          & \bf 6.74 & 4.41          & 4.09 \\
& BetaE  & 36.80 & 6.89 & 5.94 & 22.84 & 34.34 & 17.12 & 8.72 & 9.23 & 5.66 & 4.44 & 6.14 & 5.18 & 2.54 & 2.94 \\
\midrule

\multirow{3}{*}{FB237 2i, 1\%}  & \cqda & \bf 46.7          & \bf 11.8          & \bf 11.2          & \bf 30.35         &  40.75          & \bf 23.36          & \textbf{18.28} & \bf 15.85          & \textbf{8.96} & \bf 9.36          & \bf 10.25          & \bf 5.17          & \bf 4.46          & \bf 4.44          \\

& GNN-QE  & 34.81         & 5.40           & 5.17           & 30.12         & \bf 48.88 & 23.06          & 12.65          & 9.85          & 5.26          & 4.26           & 12.5          & 4.43 & 0.71          & 1.98 \\
& BetaE & 37.99 & 5.62 & 4.48 & 23.73 & 35.25 & 15.63 & 7.96 & 9.73 & 4.56 & 0.15 & 0.49 & 0.62 & 0.10 & 0.14 \\

\bottomrule

\end{tabular}%
}
\caption{Comparison of test MRR results for queries on FB15K-237 using the following training sets -- FB237, 1\% (resp. FB237 2i, 1\%) means that, in addition to all 1p (atomic) queries, only 1\% of the complex queries (resp. \textit{2i} queries) was used during training. As \cqda uses a pre-trained link predictor, we also include all \textit{1p} queries when training GNN-QE for a fair comparison.}
\label{tab:data_efficient}
\end{table*}
\begin{table*}[t]
\centering

\begin{tabular}{@{}=l+c+c+c+c+c+c+c+c+c+c+c+c@{}}
\toprule
\bf Model               & \bf 2p  & \bf 2i   & \bf 3i   & \bf pi   & \bf ip   & \bf 2u & \bf up & \bf 2in & \bf 3in & \bf inp & \bf pin & \bf pni \\ \midrule
%


CQD & 13.3 & 35.0 & 48.5 & 27.1 & 20.4 &   17.6 &    9.6 & 3.4 & 8.2 & 2.8 & 1.5 & 4.6 \\

CQD$_{\text{F}}$ &  9.3 & 21.9 & 32.6 & 20.0 & 14.5 &   13.4 &    6.4 & 6.8 & 7.5 & 5.5 & 3.6 & 4.4 \\

CQD$^{\mathcal{A}}_{F}$ & 9.4 & 24.2 & 37.3 & 21.4 & 16.5 &   13.9 &    6.6 & 8.8 & 10.0 & 5.6 & 4.7 & 4.4 \\

CQD$_{\text{C}}$ & 10.9 & 33.7 & 47.3 & 25.6 & 18.9 &   16.4 &    9.4 & 7.9 & 12.2 & 6.6 & 4.2 & 5.0 \\
CQD$_{\text{R}}$ & 6.4 & 22.2 & 31.0 & 16.6 & 11.2 &   12.5 &    4.8 & 4.7 & 5.9 & 4.1 & 2.0 & 3.5 \\
\cqda & \bf 13.2 & \bf 35.0 & \bf 48.5 & \bf 27.3 & \bf 20.7 &   \bf 17.6 &  \bf 10.5 & \bf 13.2 & \bf 14.9 & \bf 7.4 & \bf 7.8 & \bf 5.5 \\
\bottomrule
\end{tabular}%
\caption{Test MRR results for FOL queries on FB15K-237 using the following CQD extensions: 
{CQD from \citet{DBLP:conf/iclr/ArakelyanDMC21,DBLP:conf/ijcai/MinerviniADC22} with the considered normalisation and negations;} CQD$_{\text{F}}$, where we fine-tune all neural link predictor parameters in CQD; CQD$^{\mathcal{A}}_{F}$, where we \emph{fine-tune all link predictor parameters} in \cqda; CQD$_{\text{R}}$, where we learn a \emph{transformation} for the entity and relation embeddings and we use it to \emph{replace} the initial entity and relation representations; and CQD$_{\text{C}}$, where we learn a transformation for the entity and relation embeddings, and we \emph{concatenate} it to the initial entity and relation representations.}
\label{tab:fintetune_results}
\end{table*}

We can also confirm that the adaptation process converges after $\leq 10\%$ of the training epochs as seen in \cref{fig:adapt_speed}. The convergence rate is not hindered when using only $1\%$ of the training queries. This shows that \cqda is a scalable method with a fast convergence rate that can be trained in a data-efficient manner.

\subsection{Out-of-Distribution Generalisation}

To study the generalisation properties of \cqda, we trained the adaptation layer on all atomic queries and only $1\%$ of samples for \emph{one} training query type \emph{2i}, one of the simplest complex query types.
We see in \cref{tab:data_efficient} that \cqda can generalise to other types of complex queries not observed during training with an average MRR difference of $2.9$ compared to training on all training query types. \cqda also produces significantly higher test MRR results than GNN-QE, with an average increase of $5.1$ MRR.
%
%
The greatest degradation in predictive accuracy occurs for the queries containing negations, with an average decrease of $2.7$.
This prompts us to conjecture that being able to answer general EPFO queries is not enough to generalise to the larger set of queries, which include atomic negation.
However, our method can generalise on all query types, using only $1\%$ of the \emph{2i} queries, with $1496$ overall samples for adaptation.

\subsection{Fine-Tuning All Model Parameters}

One of the reasons for the efficiency of \cqda is that the neural link predictor is not fine-tuned for query answering, and only the parameters in the adaptation function are learned.
We study the effect of fine-tuning the link predictor using the full training data for CQD and \cqda on FB15K-237.
We consider several variants:
\begin{inparaenum}[1)]
\item CQD$_{\text{F}}$, where we \textbf{F}ine-tune all neural link predictor parameters in CQD;
\item CQD$^{\mathcal{A}}_{F}$, where we fine-tune all link predictor parameters in \cqda,
\item CQD$_{\text{R}}$, where we learn a transformation for the entity and relation embeddings and we use it to \textbf{R}eplace the initial entity and relation representations, and
\item CQD$_{\text{C}}$, where we learn a transformation for the entity and relation embeddings, and we \textbf{C}oncatenate it to the initial entity and relation representations.
\end{inparaenum}

It can be seen from \cref{tab:fintetune_results} that \cqda yields the highest test MRR results across all query types while fine-tuning all the model parameters produces significant degradation along all query types, which we believe is due to catastrophic forgetting~\citep{goodfellow2013empirical} of the pre-trained link predictor.

\section{Conclusions}

In this work, we propose the novel method \cqda for answering complex FOL queries over KGs, which increases the averaged MRR over the previous state-of-the-art from $34.4$ to $35.1$ while using $\leq 30\%$ of query types.
Our method uses a single adaptation layer over neural link predictors, which allows for training in a data-efficient manner.
We show that the method can maintain competitive predictive accuracy even when using $1\%$ of the training data.
Furthermore, our experiments on training on a subset ($1\%$) of the training queries from a single query type (\emph{2i}) show that it can generalise to new queries that were not used during training while being data-efficient.
Our results provide further evidence for how neural link predictors exhibit a form of compositionality that generalises to the complex structures encountered in the more general problem of query answering.
\cqda is a method for improving this compositionality while preserving computational efficiency.
As a consequence, rather than designing specialised models trained end-to-end for the query answering task, we can focus our efforts on improving the representations learned by neural link predictors, which would then transfer to query answering via efficient adaptation, as well as other downstream tasks where they have already proved beneficial, such as clustering, entity classification, and information retrieval.

\subsubsection*{Acknowledgements}
Pasquale was partially funded by the European Union’s Horizon 2020 research and innovation programme under grant agreement no. 875160, ELIAI (The Edinburgh Laboratory for Integrated Artificial Intelligence) EPSRC (grant no. EP/W002876/1), an industry grant from Cisco, and a donation from Accenture LLP, and is grateful to NVIDIA for the GPU donations. 
Daniel and Michael were partially funded by Elsevier's Discovery Lab.
Michael was partially funded by the Graph-Massivizer project (Horizon Europe research and innovation program of the European Union under grant agreement 101093202).
Erik is partially funded by a DFF Sapere Aude research leader grant under grant agreement No 0171-00034B, as well as by a NEC PhD fellowship, and is supported by the Pioneer Centre for AI, DNRF grant number P1.
Isabelle is partially funded by a DFF Sapere Aude research leader grant under grant agreement No 0171-00034B, as well as by the Pioneer Centre for AI, DNRF grant number P1.
This work was supported by the Edinburgh International Data Facility (EIDF) and the Data-Driven Innovation Programme at the University of Edinburgh.
%


\section{Appendix}
\subsection{Impact of adaptation} \label{app:adapt}

\begin{figure}[!t]
    \centering
    \includegraphics[width=\textwidth]{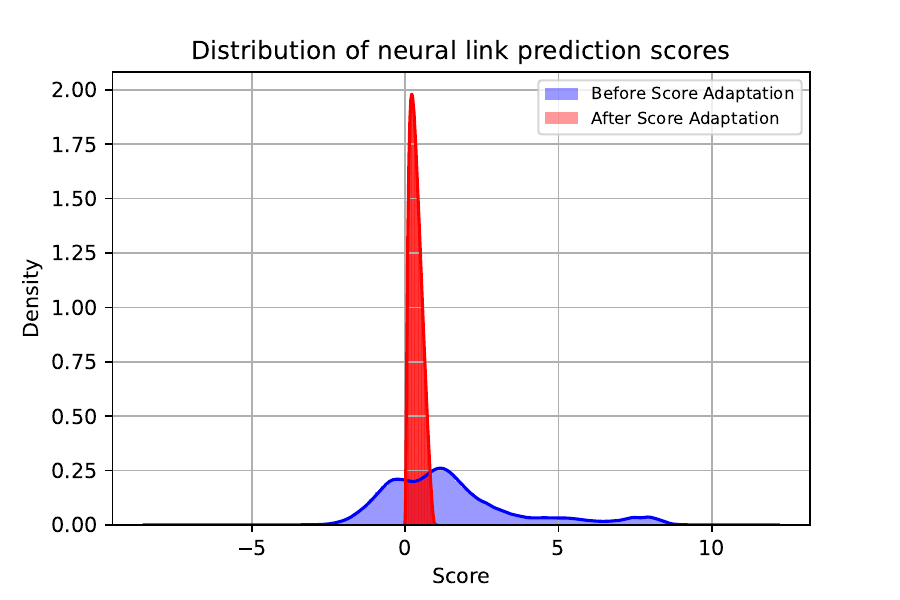}
    \caption{The distribution of the scores of the neural link predictor before applying the adaptation layer and after. }
    \label{fig:adapt_before_after}
\end{figure}

We investigate the effect of the adaptation process in \cqda by comparing the score of the neural link predictor before and after applying the adaptation layer. As we see from \cref{fig:adapt_before_after}, the scores before adaptation have a variation of $5.04$ with the boundaries at $[-8,12]$.
This makes them problematic for complex query answering as discussed in \cref{sec:adapting}. The Adapted scores have a smaller variation at $0.03$ while the maximum and minimum lie in the range $[0,1]$.

\chapter{FLARE: Faithful Logic-Aided Reasoning and Exploration}
\label{chap:flare}

\section{Introduction} \label{sec:intro_flare}
\begin{figure}[!t]
    \centering
    \includegraphics[width=\textwidth, clip]{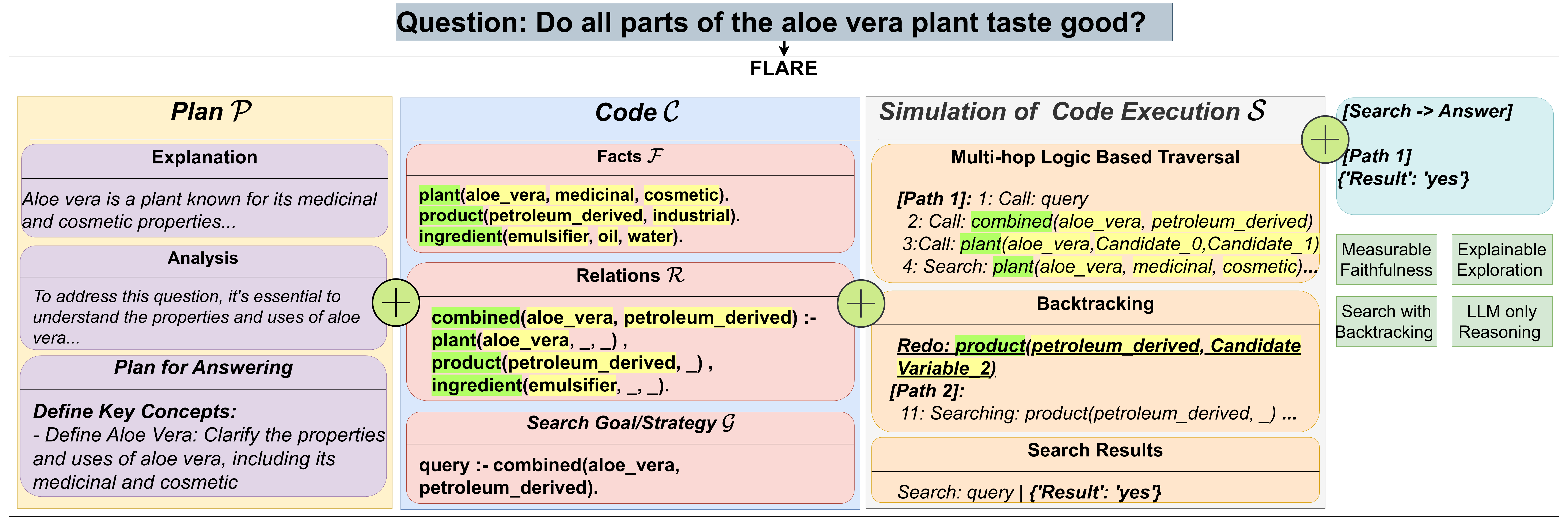}
    \caption{A depiction of the \emph{plan}, \emph{code} and simulated \emph{search} in {\flare}. Each module is generated separately and iteratively, allowing us to obtain the final answer. The green and yellow highlighted text shows the overlap between the facts and the relations between the code and the simulated search.}
    \label{fig:flare_main}
\end{figure}

Complex Reasoning in natural Question Answering (QA) tasks assumes the capability to explore the problem space of the designated query with a formalised set of facts, relations, commonsense knowledge and logical implications.
In line with this, LLMs have been enhanced with CoT \citep{wei2022chain} prompting, which supplements the QA process by generating intermediate reasoning chains given a set of in-context examples \citep{NEURIPS2020_1457c0d6}, as shown in \cref{fig:flare_main}.
This allowed for advancement in commonsense \citep{madaan2022language}, symbolic \citep{wang2022self,sprague2024cot} and mathematical \citep{jie2023design} reasoning.
Although CoT allows for a problem exploration in natural language steps, such an approach has been shown to cause performance degradation for reasoning tasks involving multi-step planning~\citep{valmeekam2022large,suzgun-etal-2023-challenging}, problem exploration~\citep{yao2022react}, and arithmetic tasks~\citep{hendrycks2021measuring, madaan2022text}.
These discrepancies arise as CoT suffers from a limited ability to decompose, search, verify and backtrack using intermediate rationale chains~\citep{yao2022react}, cascading hallucinations and errors \citep{DBLP:conf/nips/LingFLHLMS23} and that natural language might not be an optimal representation for describing the reasoning process~\citep{DBLP:conf/icml/0002LZCHSL0XI24}.
Simultaneously, LLM output has been shown to be unfaithful and inconsistent w.r.t. the intermediate CoT rationale \citep{DBLP:conf/acl/JacoviBBHHT0AG24,lanham2023measuring,DBLP:conf/nips/TurpinMPB23}.
%

%
%
%
To mitigate the problem of CoT faithfulness and allow for more robust reasoning during QA, \citet[Faithful CoT]{DBLP:conf/ijcnlp/LyuHSZRWAC23} and Logic-LM~\citep{DBLP:conf/emnlp/PanAWW23} suggested generating code which is further executed using an external symbolic solver.
Producing and executing code enables the generation of outputs guided by external solvers, leveraging search with backtracking to explore the problem space effectively.
However, strict translations of natural language queries into code, such as \emph{autoformalisation}~\citep{DBLP:conf/mkm/Szegedy20,DBLP:conf/mkm/WangKU18}, is a non-trivial task involving direct inference of implicit commonsense and domain-specific knowledge and the ability to align abstract and informal concepts directly to constrained formal definitions for further execution~\citep{DBLP:conf/nips/WuJLRSJS22}.
An example query, \emph{``Do all parts of the aloe vera plant taste good?''}, is challenging to formalize or address with a strict algorithmic solution, as it requires interpretative, deductive and context-dependent reasoning, referred to as soft or fuzzy reasoning.
Using external solvers makes such fuzzy reasoning impossible and requires consistently generating syntactically correct executable code.
While some LLMs have coding capabilities stemming from their pretraining \citep{DBLP:journals/corr/abs-2406-00515,DBLP:journals/corr/abs-2408-10914}, relative code consistency is more probable with models explicitly trained for coding \citep{DBLP:journals/corr/abs-2107-03374}.
To overcome these problems, 
we propose Faithful Logic-Aided Reasoning and Exploration (\flare), an interpretable method that allows for planning, fuzzy reasoning, and traversing the problem space with backtracking, exact task decomposition, and measuring faithfulness.
In \flare, given a natural language query, we prompt an LLM to sequentially generate a \emph{plan} that includes an analysis and the logical steps necessary for formalising and answering the question, a logic programming~\citep{wielemaker2012swi} \emph{code} that allows formalising the query into a set of facts, relations and their composition forming the space for exploring that query and the \emph{search}, which is an LLM-generated code execution simulation.
An illustration of {\flare} can be seen in \cref{fig:flare_main}.
In our framework, the generated code must not be consistently executable by an external solver, allowing for the soft-formalisation of natural language. Although we see that even generalist LLMs are able to produce executable code in $\geq 50\%$ of cases.
{\flare} allows us to measure the faithfulness of the outcome w.r.t. the simulated code execution by directly comparing the search paths produced by the external solver to that LLM generation.
This comparison also allows for pinpointing model hallucinations and inconsistencies.
We systematically study the effectiveness of our method using $4$ general-purpose LLMs of varying scales across $9$ diverse QA and $3$ logical inference benchmarks, covering Math World Problems, Multi-hop QA, Relation inference, deductive and analytical reasoning and show that our method achieves state-of-the-art results in $7$ out of $9$ QA datasets and $2$ out of $3$ logic datasets in comparison to CoT, F-CoT and Logic-LM.
We also show that the method is competitive for models tuned for coding, with an average overall increase of $16\%$ over F-Cot and $9\%$ over CoT.
%
%
Our findings show that model accuracy strongly correlates with the faithfulness of the reasoning process towards search traces from the simulated code execution.
We also provide ablations showing that the model can interpretably pinpoint hallucinations, underutilized knowledge, and the limitations of the search over the problem space. 
Our key contributions are the following: 
\begin{itemize}[leftmargin=*]
\item We introduce {\flare} a novel paradigm for logic-aided and interpretable formalisation and search over the problem space in QA and logic reasoning tasks.
\item We perform a systematic evaluation across $9$ QA and $3$ logical inference benchmarks and $4$ models of varying scales, showing the advantages of using {\flare} for QA in a few-shot setup over prior approaches.
\item The modularity of {\flare} allows defining a simple ingrained method for measuring model faithfulness, which is further shown to be strongly correlated with performance.
\item We further show that using {\flare} allows us to interpretably and rigorously detect hallucinations along with sub-optimal and inconsistent reasoning patterns.
\end{itemize}

\section{Related Work}
\label{sec:related_flare}

\subsection{Reasoning in Natural Language}
Few-shot prompting \citep{DBLP:conf/nips/BrownMRSKDNSSAA20} has been shown to be an effective approach for increasing the reasoning capabilities of LLMs in natural language generation \citep{DBLP:journals/corr/abs-2102-01672, DBLP:conf/acl/ReifIYCCW22,DBLP:conf/iclr/SanhWRBSACSRDBX22}.
LLM reasoning can be further enhanced with prompting techniques such as CoT \citep{wei2022chain}, which attempts to segment reasoning into explicitly written intermediate steps.
Concurrent work has also proposed that models \emph{``think step by step''} \citep{DBLP:conf/nips/KojimaGRMI22}, or divide the problem into subtasks before the solution \citep[Least-to-Most]{DBLP:conf/iclr/ZhouSHWS0SCBLC23}.
These approaches have been shown to suffer from arithmetic inaccuracies \citep{DBLP:conf/nips/LewkowyczADDMRS22, DBLP:conf/nips/HendrycksBKABTS21} and reasoning inconsistencies \citep{DBLP:journals/corr/abs-2209-07686}.
Further attempts have been made to add a planning stage before reasoning by dividing the process into recursive plan formulation and execution steps \citep{DBLP:conf/iclr/YaoZYDSN023,DBLP:conf/acl/WangXLHLLL23}.
The \emph{plan} generation step in {\flare} is a hybrid technique inspired by these methods but focused on generating a natural language strategy for formalising the query into code. 
\subsection{Reasoning with Search}

Several lines of work propose using techniques to expand the reasoning paths over the problem space.
Self-consistency decoding \citep{DBLP:conf/iclr/0002WSLCNCZ23} is an approach used to sample many natural language reasoning paths and take a majority vote for an answer.
Another popular approach is Tree-of-Thoughts~\citep[ToT;][]{DBLP:conf/nips/YaoYZS00N23}, which proposes to explore the query with reasoning similar to a tree traversal, where each state is created and evaluated using an LLM.
Similar techniques try to adapt symbolic search approaches akin to DFS, BFS \citep{DBLP:conf/aaai/BestaBKGPGGLNNH24}, A$^*$~\citep{DBLP:journals/corr/abs-2402-14083} or other combinations~\citep{DBLP:journals/corr/abs-2404-03683} with direct tuning~\citep{DBLP:journals/corr/abs-2402-14083}, imitation training~\citep{DBLP:conf/nips/YangSAN22} or few-shot prompting~\citep{DBLP:journals/corr/abs-2404-01230}.
It must be noted that all of these techniques have only been tested in constrained mathematical puzzle-solving and algorithmic domains like the 24 Game~\citep{DBLP:conf/nips/YangSAN22}, Countdown~\citep{wiki:Countdown}, Sorting~\citep{DBLP:conf/aaai/BestaBKGPGGLNNH24}, maze solving~\citep{DBLP:conf/nips/YangSAN22}, Sokoban~\citep{DBLP:journals/corr/abs-2402-14083}, and others.
Although the \emph{search} component of {\flare} has some similarities to these techniques, we argue that our method allows for generalistic reasoning with interpretable multi-hop search through simulated code execution.
\subsection{Reasoning with Formalisation}
Another line of research has tried formalising natural language queries into code~\citep{DBLP:conf/icml/GaoMZ00YCN23,DBLP:conf/icml/0002LZCHSL0XI24} or pseudo-code~\citep{DBLP:journals/corr/abs-2404-02575,DBLP:journals/corr/abs-2404-03683}.
This allows the translation of the query into a strict structure and offloads the reasoning and search components to deterministic solvers like Python~\cite{DBLP:journals/tmlr/ChenM0C23}, PDDL \cite{DBLP:conf/ijcnlp/LyuHSZRWAC23, DBLP:journals/corr/abs-2304-11477}, DataLog \cite{DBLP:conf/ijcnlp/LyuHSZRWAC23} and others.
While models are capable of synthesising programs~\citep{DBLP:journals/corr/abs-2108-07732, DBLP:conf/iclr/NijkampPHTWZSX23} and benefit from the use of code in numerical and algorithmic reasoning settings~\citep{DBLP:journals/tmlr/ChenM0C23,DBLP:conf/icml/GaoMZ00YCN23}, the usage of code for general QA has not been rigorously explored.
The reasons are that formalisation from natural language into a strict and executable code is challenging~\citep{DBLP:conf/nips/WuJLRSJS22}, following the exact syntactic constraints of the programming language not abundantly used during pre-training is onerous \citep{DBLP:journals/corr/abs-2404-00971} and can require models explicitly tuned for coding~\citep{DBLP:journals/corr/abs-2107-03374}.
Using an external solver for reasoning also limits the capability for soft reasoning in commonsense knowledge and implications. Although we formalise the natural language query into a logic programming Prolog program during the \emph{code} generation part of {\flare}, we do not explicitly require the code to be executable and do not use external solvers during inference. This allows for the further use of the LLM for soft-reasoning to simulate code execution in a logic-based problem space traversal similar to Prolog while circumventing the need for code tuning a generalist model.
%
\begin{table*}[t!]
\adjustbox{width=\textwidth,center}{
\begin{tabular}{@{}lccccccccc@{}}
\toprule
\multicolumn{1}{c}{} &
  \multicolumn{5}{c}{\textbf{Math Word Problems}} &
  \multicolumn{3}{c}{\textbf{Multi-hop QA}} &
  \textbf{Relation} \\ \midrule
\multicolumn{1}{c|}{\textbf{Method}} &
  GSM8K &
  SVAMP &
  MultiArith &
  ASDiv &
  \multicolumn{1}{c|}{AQuA} &
  StrategyQA &
  Date &
  \multicolumn{1}{c|}{Sport} &
  CLUTRR \\ \midrule
\multicolumn{1}{l|}{${Llama-3.1-8B}_{\flare}$} &
  \textbf{72.7} &
  \textbf{86.0} &
  \textbf{96.3} &
  \textbf{83.1} &
  \multicolumn{1}{c|}{\cellcolor[HTML]{9AFF99}\textbf{62.9}} &
  \textbf{70.2} &
  \textbf{59.3} &
  \multicolumn{1}{c|}{\underline{76.6}} &
  {\underline{36.8}} \\
\multicolumn{1}{l|}{${Llama-3.1-8B}_{F-CoT}$} &
  it{0} &
  it{0} &
  it{0} &
  it{0} &
  \multicolumn{1}{c|}{it{12.2}} &
  {\underline{53.2}} &
  it{0} &
  \multicolumn{1}{c|}{it{0}} &
  it{32} \\
\multicolumn{1}{l|}{${Llama-3.1-8B}_{CoT}$} &
  {\underline{59.2}} &
  {\underline{58.6}} &
  {\underline{60.1}} &
  {\underline{61.9}} &
  \multicolumn{1}{c|}{{\underline{35}}} &
  it{2.9} &
  \underline{20.9} &
  \multicolumn{1}{c|}{\textbf{95.8}} &
  \textbf{42.2} \\ \midrule
\multicolumn{1}{l|}{${CmDR}_{\flare}$} &
  \textbf{52.4} &
  \textbf{74.0} &
  \textbf{84.5} &
  \textbf{72.2} &
  \multicolumn{1}{c|}{\textbf{43.7}} &
  \textbf{67.0} &
  \textbf{52.3} &
  \multicolumn{1}{c|}{\textbf{78.9}} &
  {\underline{29.1}} \\
\multicolumn{1}{l|}{${CmDR}_{F-CoT}$} &
  it{0} &
  it{0} &
  it{0} &
  it{0} &
  \multicolumn{1}{c|}{it{0}} &
  {\underline{59.7}} &
  it{0} &
  \multicolumn{1}{c|}{it{0}} &
  it{8.6} \\
\multicolumn{1}{l|}{${CmDR}_{CoT}$} &
  {\underline{46.5}} &
  {\underline{57.3}} &
  {\underline{83.1}} &
  {\underline{37.2}} &
  \multicolumn{1}{c|}{{\underline{28.3}}} &
  it{21.3} &
  {\underline{47.4}} &
  \multicolumn{1}{c|}{{\underline{55.2}}} &
  \textbf{29.5} \\ \midrule
\multicolumn{1}{l|}{${CmDR+}_{\flare}$} &
  \textbf{71.4} &
  \cellcolor[HTML]{9AFF99}\textbf{83.5} &
  \textbf{90.4} &
  \textbf{81.3} &
  \multicolumn{1}{c|}{\textbf{55.9}} &
  \cellcolor[HTML]{9AFF99}\textbf{70.8} &
  {\underline{61.8}} &
  \multicolumn{1}{c|}{\textbf{77.7}} &
  \textbf{41.0} \\
\multicolumn{1}{l|}{${CmDR+}_{F-CoT}$} &
  it{0} &
  it{0} &
  it{0} &
  it{0} &
  \multicolumn{1}{c|}{it{15.4}} &
  {\underline{57.6}} &
  it{0} &
  \multicolumn{1}{c|}{it{0}} &
  it{35.3} \\
\multicolumn{1}{l|}{${CmDR+}_{CoT}$} &
  {\underline{48.7}} &
  {\underline{81.1}} &
  {\underline{86.6}} &
  {\underline{44.6}} &
  \multicolumn{1}{c|}{{\underline{44.1}}} &
  it{48.4} &
  \textbf{79.1} &
  \multicolumn{1}{c|}{{\underline{62.6}}} &
  {\underline{42.5}} \\ \midrule
\multicolumn{1}{l|}{${GPT-3.5}_{\flare}$} &
  it{68.1} &
  {\underline{82.7}} &
  \cellcolor[HTML]{9AFF99}\textbf{98.3} &
  \cellcolor[HTML]{9AFF99}\textbf{85.4} &
  \multicolumn{1}{c|}{{\underline{55.1}}} &
  \textbf{65.5} &
  \cellcolor[HTML]{9AFF99}\textbf{82.4} &
  \multicolumn{1}{c|}{{\underline{85.6}}} &
  \cellcolor[HTML]{9AFF99}\textbf{49.8} \\
\multicolumn{1}{l|}{${GPT-3.5}_{F-CoT}$} &
  {\underline{75.8}} &
  \textbf{83.0} &
  it{95.3} &
  {\underline{81.7}} &
  \multicolumn{1}{c|}{it{53.5}} &
  it{51.5} &
  {\underline{73.5}} &
  \multicolumn{1}{c|}{it{52.3}} &
  {\underline{12.1}} \\
\multicolumn{1}{l|}{${GPT-3.5}_{CoT}$} &
  \cellcolor[HTML]{9AFF99}\textbf{79.8} &
  it{82.4} &
  {\underline{98.2}} &
  it{75.8} &
  \multicolumn{1}{c|}{\textbf{59.4}} &
  {\underline{51.7}} &
  it{69.9} &
  \multicolumn{1}{c|}{\cellcolor[HTML]{9AFF99}\textbf{95.8}} &
  {\underline{4.3}} \\ \bottomrule
\end{tabular}
}
\caption{The following table shows the performance of each of the tested models given a technique for reasoning. Each \textbf{bold}, \underline{underlined}, and it{italicised} element highlights the best, second best and worst technique per specific model. The overall best method per dataset is highlighted in \colorbox[HTML]{9AFF99}{green}.}
\label{tab:results_main}
\end{table*}

\subsection{Reasoning Faithfulness}

An explanation is considered \emph{faithful} if it explicitly and accurately describes the reasoning process of the model during inference \citep{DBLP:conf/dsaa/GilpinBYBSK18, DBLP:conf/acl/JacoviG20}. In the context of prompting techniques such as CoT, we are interested in the faithfulness of the intermediate reasoning chains towards the final output. Faithful intermediate reasoning chains should not just look \emph{plausible} \citep{herman2017promise} but have exact reflections of the problem exploration and reasoning used to arrive at the final answer. Natural language reasoning chains prevalent in CoT and similar methods are shown to be unfaithful, either masking the reasoning biases \citep{DBLP:conf/nips/TurpinMPB23} of the model or outright ignoring the intermediate reasoning \citep{DBLP:journals/corr/abs-2307-13702}. In {\flare}, we introduce a method to seamlessly measure the faithfulness of the final outcome w.r.t. completed search.

\section{Methodology}
\label{sec:method}
\subsection{LLM Simulated Search}
\label{subsec:generate}

{\flare} comprises three modules for generating a \emph{plan}, \emph{code} and simulated \emph{search} for answering a natural language query $\mathcal{Q} = \{T^{\mathcal{Q}}_1 \dots T^{\mathcal{Q}}_{|\mathcal{Q}|}\}$, where each $T^{\mathcal{Q}}_i$ is a token in the query $\mathcal{Q}$.

\subsection{Generating A Plan}
For each query $\mathcal{Q}$, given an LLM ${\mathcal{M}}$, we initially use instructions $\mathcal{I}^{\mathcal{P}}$ to prompt it to generate a \emph{plan} $\mathcal{P}$, which should be comprised of task explanation, analysis and a plan for further formalising the query. An example of this can be seen in the \emph{plan} section in \cref{fig:flare_main}. We use in-context few shot examples $\mathcal{E}_{\mathcal{P}}$ of such \emph{plan} generations along with bf{greedy} decoding for obtaining the final plan.
\begin{align}
    \mathcal{P}_i \sim argmax p_{\mathcal{M}}(T^{\mathcal{P}}_i \mid T^{\mathcal{P}}_{:i-1},\mathcal{E}_{\mathcal{P}}, \mathcal{Q},\mathcal{I}^{\mathcal{P}})
\end{align}

Where $\mathcal{P}_i$ and $T^{\mathcal{P}}_i$ is the $i$-th token in the generated \emph{plan} $\mathcal{P}$ and $p_{\mathcal{M}}$ is the probability of the next token over the vocabulary obtained from model ${\mathcal{M}}$.

\subsection{Generating Code}

After generating the \emph{plan}, we use instructions $\mathcal{I}^{\mathcal{C}}$ to prompt the LLM ${\mathcal{M}}$ to generate a Prolog code $\mathcal{C}$, an example of which can be seen in \cref{fig:flare_main}. We append executable code generation samples $\mathcal{C}_{{sample}}$ to the previous in-context examples $\mathcal{E}_{\mathcal{P}}$ and obtain few-shot code generation demonstrations $\mathcal{E}_{\mathcal{C}} =[\mathcal{E}_{\mathcal{P}};\mathcal{C}_{{sample}}]$
\begin{gather}
\label{eq:code}
     \mathcal{C}_i \sim \argmax p_{\mathcal{M}}(T^{\mathcal{C}}_i \mid T^{\mathcal{C}}_{:i-1} \mathcal{E}_{\mathcal{C}}, \mathcal{Q}, \mathcal{I}^{\mathcal{P}},\mathcal{P},\mathcal{I}^{\mathcal{C}})
     \\
     \mathcal{F}_{{{code}}}, \mathcal{R}_{{{code}}} , \mathcal{G}_{{{code}}} = {EXTRACT}(\mathcal{C}_i) \nonumber
 \end{gather}

Where $\mathcal{C}_i { and }T^{\mathcal{C}}_i$ is the $i$-th token in the generated \emph{code} $\mathcal{C}$.

\begin{figure}[t!]
    \centering
    \includegraphics[clip=true,width=\textwidth]{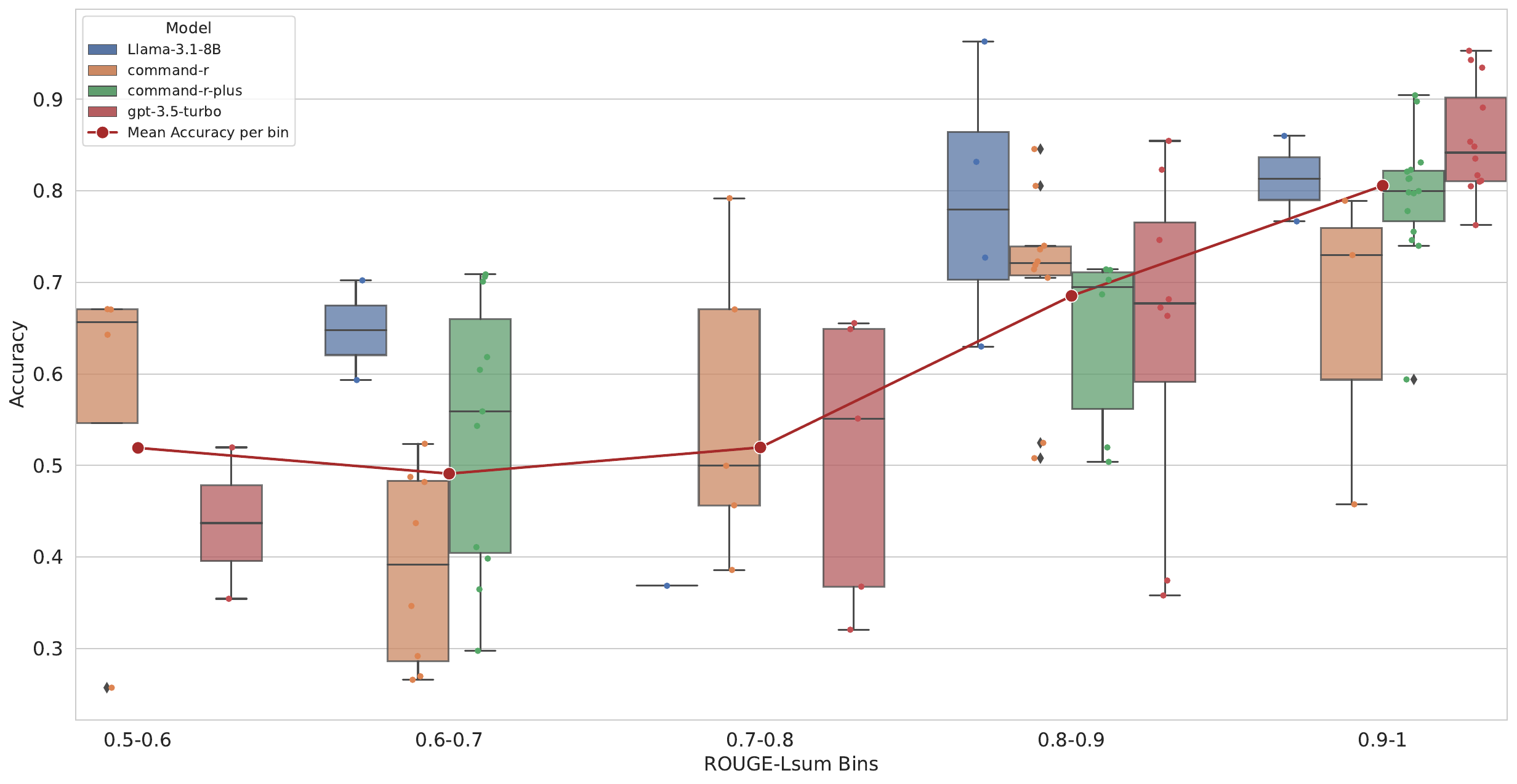}
    \caption{The trend of mean model accuracy w.r.t mean faithfulness for all the models.}
    \label{fig:faith_vs_acc}
\end{figure}

\subsection{Benefits of Prolog}  
Prolog is a symbolic logic-programming engine \citep{DBLP:conf/acm/Bowen79} designed for heuristic search over Horn Clauses \citep{DBLP:journals/jlp/ChandraH85}. As a declarative programming paradigm \citep{DBLP:conf/agp/Lloyd94}, the code is expressed as the logic of computation, expressed as a set of facts $\mathcal{F}$ and relations $\mathcal{R}$ defining the problem space, with the goal $\mathcal{G}$ being a first-order logic combination of them. Prolog employs depth-first search (DFS) \citep{DBLP:conf/acm/Bowen79} for sub-goal decomposition and traversal of the problem space, satisfying $\mathcal{G}$ through a sequence of steps known as the \emph{trace}. Each step either confirms/invalidates a sub-goal, expands the search tree, or retries failed sub-goals with new combinations. An example of such a search is shown in \cref{fig:flare_main}.  
Prolog supports exhaustive search by exploring all paths that satisfy or fail the goal. This explicit segmentation of facts, relations, and search strategies simplifies query formalization. As a declarative language, Prolog enables segmentation using a simple regexp heuristic, referred to as EXTRACT in \cref{eq:code} and \cref{eq:search}.  
Including exhaustive traces in-context allows an LLM to simulate sub-goal decomposition, backtracking, and intermediate goal invalidation, discussed further in the next paragraph.  

\subsection{Simulating Search}

After generating the logic-programming \emph{code}, we want to simulate program execution by generating a problem space traversal trace with our LLM ${\mathcal{M}}$. We use instructions $\mathcal{I}^{\mathcal{S}}$ and update our in-context samples by appending search traces $\mathcal{S}_{{sample}}$ constructed from Prolog execution of sample codes $\mathcal{C}_{{sample}}$, i.e. $\mathcal{E}_{\mathcal{S}} =[\mathcal{E}_{\mathcal{C}};\mathcal{S}_{{sample}}]$. 
\begin{gather}
\label{eq:search}
     \mathcal{S}_i \sim \argmax p_{\mathcal{M}}(T^{\mathcal{S}}_i \mid T^{\mathcal{S}}_{:i-1} \mathcal{E}_{\mathcal{C}}, \mathcal{Q}, \mathcal{I}^{\mathcal{P}},\mathcal{P},\mathcal{I}^{\mathcal{C}},\mathcal{C},\mathcal{I}^{\mathcal{S}}) \\
     {\mathcal{A}}_{{{search}}},
     \mathcal{F}_{{{search}}}, \mathcal{R}_{{{search}}} = {EXTRACT}(\mathcal{S}_i) \nonumber
 \end{gather}
Where $T^{\mathcal{S}}_i$ is the $i$-th token in the generated \emph{search} trace $\mathcal{S}$. During iterative problem space traversal, we can segment the facts $\mathcal{F}_{{{search}}}$, relations $\mathcal{R}_{{{search}}}$, completed and backtracked paths with their answers ${\mathcal{A}}_{{{search}}}$ used during the search simulation. To get the final answer we update in-context samples with their correct final answers ${\mathcal{A}}_{{sample}}$ from the executed search $\mathcal{S}_{{sample}}$, $\mathcal{E}_{{\mathcal{A}}} =[\mathcal{E}_{\mathcal{S}};{\mathcal{A}}_{{sample}}]$ and use instructions $\mathcal{I}^{\mathcal{A}}$ to obtain the final answer from the model.
 \begin{align}
      {\mathcal{A}}_{{{Final}}} \sim \argmax p_{\mathcal{M}}(T^{{\mathcal{A}}}_i \mid T^{{\mathcal{A}}}_{:i-1} \mathcal{E}_{\mathcal{C}}, \mathcal{Q}, \mathcal{I}^{\mathcal{P}},\mathcal{P},\mathcal{I}^{\mathcal{C}},\mathcal{C},\mathcal{I}^{\mathcal{S}},\mathcal{S},\mathcal{I}^{\mathcal{A}})
 \end{align}
The prompts used for generating each part in {\flare} can be seen in \cref{append:prompts} along with a complete example in \cref{tab:flare_ex_complete}.     

\subsection{Detecting Reasoning Inconsistencies}
\label{subsec:reasoning_inc}

For each query $\mathcal{Q}$ given the \emph{code} $\mathcal{C}$ and the simulated \emph{search} $\mathcal{S}$ along with the extracted facts $\mathcal{F}_{{{code}}}, \mathcal{F}_{{{search}}}$ and relations $\mathcal{R}_{{{code}}}, \mathcal{R}_{{{search}}}$ from each designated module, we aim to detect the inconsistencies during the reasoning process of the LLM. We use exact string matching between all these facts and relations in \emph{code} and simulated \emph{search}.
\begin{gather}
\forall i, \exists j \quad {such that} \quad \mathcal{F}_{{code}}^i = \mathcal{F}_{{search}}^j \quad {and} \quad \forall v, \exists q \quad \mathcal{R}_{{code}}^v = \mathcal{R}_{{search}}^q \\
\forall j, \exists i \quad {such that} \quad \mathcal{F}_{{code}}^i = \mathcal{F}_{{search}}^j \quad {and} \quad \forall q, \exists v \quad \mathcal{R}_{{code}}^v = \mathcal{R}_{{search}}^q
\end{gather}
With this framework in mind, we define two reasoning failure modes. 
%
\begin{wraptable}{r}{0.5\textwidth}
\adjustbox{width=0.9\textwidth}{
\begin{tabular}{@{}ccccl@{}}
\toprule
\multicolumn{1}{c|}{\multirow{2}{*}{Dataset}} & \multicolumn{4}{c}{ChatGPT (gpt-3.5-turbo)}        \\ \cmidrule(l){2-5} 
\multicolumn{1}{c|}{}                         & Standard & CoT   & Logic-LM       & FLARE          \\ \midrule
PrOntoQA                                      & 47.40    & 67.80 & 61.00          & \textbf{73.40} \\
LogicalDeduction                              & 40.00    & 42.33 & \textbf{65.67} & 58.60          \\
AR-LSAT                                       & 20.34    & 17.31 & 26.41          & \textbf{27.39} \\ \bottomrule
\end{tabular}
}
\caption{Comparison of Direct Prompting, CoT, Logic-LM and \flare.}
\label{tab:logiclm_res}
\end{wraptable}
In the \emph{first} failure mode, given that some fact or relation was used in the simulated \emph{search} but did not exist in the generated \emph{code}, i.e. $\exists j { such that } \mathcal{F}_{{search}}^j \notin \mathcal{F}_{{code}}$, we claim that the LLM has \emph{hallucinated}. We postulate that the model either produced incomplete knowledge during formalisation to \emph{code} or created a piece of non-existing information during the \emph{search}. We do not consider facts that emerged during a direct inference step within the simulated search during our calculation. For example, if we are dealing with a mathematical query $4\cdot (5+6) = ?$, the search would involve separately evaluating the expression $5+6=11$. In this case, $11$ will not be treated as a hallucinated fact within the search but rather as an emergent fact obtained from direct inference.   
The \emph{second} failure mode is the reciprocal case, where a fact or relation present in the \emph{code} is not used during the \emph{search}. We refer to this phenomenon as \emph{sub-optimal reasoning} as it shows that the LLM could not explore the problem space completely or injected unsuitable knowledge during formalisation into \emph{code}.

\subsection{Measuring Faithfulness}

We propose a method to measure the faithfulness of the LLM reasoning process when using {\flare}. As mentioned in \cref{subsec:generate}, for each query in a dataset $\mathcal{D} = [\mathcal{Q}_1, \dots ,\mathcal{Q}_{|\mathcal{D}|}]$, we generate a set of codes $\Phi = [\mathcal{C}_1, \dots ,\mathcal{C}_{|\Phi|}]$ and simulated problem space searches $\Psi = [\mathcal{S}_1, \dots, \mathcal{S}_{|\Psi|}]$. We use the Prolog engine to execute all of the codes $\Phi$ and obtain a set of correctly written programs $\Phi^\prime$ and exact search paths $\Psi^\prime$. As we do not require explicit programmatic correctness during inference in {\flare} for any code $\mathcal{C}_i$, some Prolog executions resulting in an error are filtered out in $\Psi^\prime$. To assess model reasoning faithfulness towards code formalisations, we compare the search paths $\Phi^\prime$ obtained from Prolog execution with their designated counterparts $\Phi_{{{gen}}}^\prime$ generated by the LLM from the same code. We use ROUGE \citep{lin-2004-rouge} to compute the matching score for each executed and simulated search path. In particular, we use ROUGE-Lsum, which uses the longest common subsequence (LCS) over each line to obtain the final score. This method fits our cause as a line in a Prolog search execution represents a single logic step within the traversal. This allows us to measure the similarity of the reasoning contents and structure in exact and simulated searches. 

\section{Experimental Setup}
\label{sec:ex_setup}
\subsection{Datasets}
To evaluate {\flare}, we use a benchmark of $9$ tasks covering Math Word Problems (MWP), multi-hop QA and relation inference, and $3$ common logical reasoning datasets. For testing numeric and mathematical reasoning, we follow CoT \citep{wei2022chain} by including GSM8K \citep{DBLP:journals/corr/abs-2110-14168}, SVAMP \citep{DBLP:conf/naacl/PatelBG21}, MultiArith \citep{DBLP:conf/emnlp/RoyR15}, ASDiv \citep{DBLP:conf/acl/MiaoLS20} and AQuA \citep{DBLP:conf/acl/LingYDB17}. Among these, GSM8K, SVAMP, MultiArith and ASDiv cover elementary and middle school arithmetic word problems with a set of integers or decimals as the answer. AQuA is a multiple-choice numerical, symbolic reasoning dataset where each answer is a mathematical expression containing notations, values and expressions not defined in the query.
We also test {\flare} using three multi-hop QA datasets. We use StrategyQA \citep{DBLP:journals/tacl/GevaKSKRB21}, which is a boolean QA task that requires sub-goal decomposition and a multi-hop reasoning strategy to answer. The example \emph{``Do all parts of the aloe vera plant taste good?''} used in \cref{fig:flare_main}, is taken from StrategyQA. The multi-hop QA testing also includes Date and Sports Understanding, subsets of BIG-Bench \citep{srivastava2023beyond}. The tasks involve inferring an exact date given some calculations in the relative time period and understanding if an artificially created sports statement is feasible.
Furthermore, we assess {\flare} on Relational Inference using CLUTRR \citep{DBLP:conf/emnlp/SinhaSDPH19}, which involves inferring the familial relation between two entities mentioned in a natural language description of the partial family graph.
We evaluate {\flare} on challenging logic inference datasets: ProntoQA \citep{DBLP:conf/iclr/Saparov023}, AR-LSAT \citep{DBLP:journals/corr/abs-2104-06598}, and LogicalDeductions from BigBench \citep{DBLP:journals/tmlr/SrivastavaRRSAF23}, focusing on harder subsets proposed by \cite{DBLP:conf/emnlp/PanAWW23}. These datasets, covering deductive, analytical, and logical reasoning, allow us to assess {\flare}'s performance. Details, including descriptions and examples, are in \cref{tab:data_stat_flare} of \cref{append:prompts}.

\subsection{Benchmarks}

We compare {\flare} with CoT \citep{wei2022chain} as a prompting method that reasons using natural language chains and with F-CoT \citep{DBLP:conf/ijcnlp/LyuHSZRWAC23} and Logic-LM \citep{DBLP:conf/emnlp/PanAWW23} that formalise the query into a code and offload the reasoning to an external symbolic solver. We use Llama3.1 (8B) \citep{DBLP:journals/corr/abs-2407-21783}, CmDR (30B) \citep{cohere2024commandr}, CmDR+ (100B) \citep{cohere2024commandr}  and GPT3.5 \citep{DBLP:conf/nips/BrownMRSKDNSSAA20} ($\geq 100B$ \citep{DBLP:journals/corr/abs-2303-10420}). As the coding model OpenAI Codex (code-DaVinci-002) \citep{DBLP:journals/corr/abs-2107-03374} used in F-CoT has been deprecated, we replace it with the new GPT3.5 as suggested by OpenAI and recalculate the results accordingly.

\section{Results}
\label{sec:results}

\subsection{Few-shot prompting}
\begin{table}[t!]
\adjustbox{width=\textwidth}{
\begin{tabular}{@{}lcccccc@{}}
\toprule
\textbf{Method} &
  \multicolumn{1}{l}{${CmDR}_{plan-only}$} &
  \multicolumn{1}{l}{${CmDR}_{\flare}$} &
  \multicolumn{1}{l}{${CmDR+}_{plan-only}$} &
  \multicolumn{1}{l}{${CmDR+}_{\flare}$} &
  \multicolumn{1}{l}{${GPT-3.5}_{plan-only}$} &
  \multicolumn{1}{l}{${GPT-3.5}_{\flare}$} \\ \midrule
GSM8K      & 24.7          & \textbf{52.4} & 40.7          & \textbf{71.4} & 36.1          & \textbf{68.1} \\
AQuA       & 35.0          & \textbf{43.7} & 55.1          & \textbf{55.9} & 54.3          & \textbf{55.1} \\
StrategyQA & 65.5          & \textbf{67.0} & \textbf{75.7} & 70.8          & 62.3          & \textbf{65.5} \\ \bottomrule
\end{tabular}
}
\caption{The table shows the accuracy of an LLM with {\flare} compared to prompting for a final answer directly after generating (plan-only) a plan $\mathcal{P}$.}
\label{tab:plan_only}
\end{table}

\begin{figure}[t!]
    \centering
    \includegraphics[clip=true,width=\textwidth]{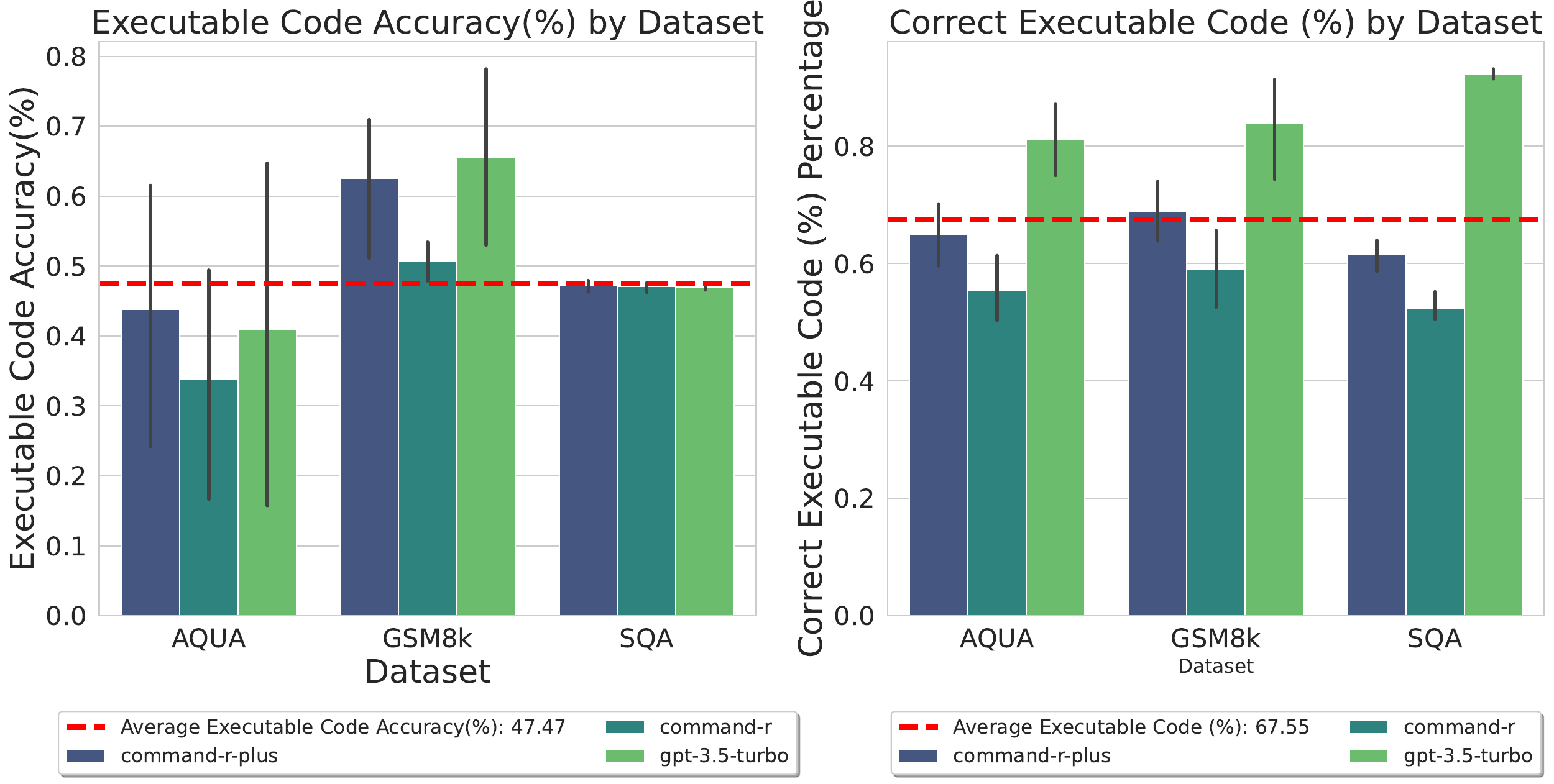}
    \caption{The figure shows the percentage of executable code per model (right) and the accuracy of the executable code when answering the queries (left).}
    \label{fig:code_perc}
\end{figure}

To evaluate {\flare}, we use a set of models of varying sizes on diverse benchmarks, as defined in \cref{sec:ex_setup}. We compare the performance of each model while using {\flare}, CoT and F-CoT prompting. The results for F-CoT and CoT on all the models are computed using the codebase of the original study \citep{DBLP:conf/ijcnlp/LyuHSZRWAC23}. We additionally compare Logic-LM and {\flare} using the logic reasoning benchmarks proposed in \cite{DBLP:conf/emnlp/PanAWW23}.

\subsection{LLMs for general reasoning} 

Our results, presented in \cref{tab:results_main}, show that using {\flare} allows the LLMs to achieve state-of-the-art results on $7$ out of $9$ datasets, with an average $28\%$ increase over CoT. We can see a clear trend that {\flare} increases the performance compared to CoT and F-CoT for all the models of varying scales. We also see that LLMs that are not explicitly tuned for coding suffer massive degeneracies when using F-CoT. We postulate that they are unable to consistently produce executable programs that satisfy a predefined scheme in F-CoT, thus resulting in an error during execution. This further highlights the value of simulating program execution using an LLM instead of using external solvers.
The results show that using {\flare} yields more benefit on datasets that require longer chains of multi-hop and symbolic reasoning, like AQuA and StrategyQA.
Our findings in \cref{tab:logiclm_res} show that \flare achieves state-of-the-art results on $2$ out of $3$ logic inference benchmarks with $10\%$ increase over CoT and $~7\%$ increase over Logic-LM.

\subsection{LLMs for code generation}

To understand the effect of {\flare} on models tuned for coding, we use GPT3.5 \citep{NEURIPS2020_1457c0d6} as it was the OpenAI suggested succession model for Codex \citep{DBLP:journals/corr/abs-2107-03374} which is used in F-CoT and possesses strong coding capabilities \citep{DBLP:journals/corr/abs-2303-10420}. The results in \cref{tab:results_main} show that using {\flare} is beneficial for models that are tuned for coding and boost the accuracy with a $16\%$ increase over F-CoT and $9\%$ over CoT. The reason is that many natural language queries with non-trivial formalisations are more suited to be tackled with more commonsense soft reasoning than direct code execution. This is evident in \cref{tab:results_main} where {\flare} and CoT are consistently better than F-CoT in StrategyQA, Sports and CLUTRR. 
The opposite case of numeric and algorithmic heavy reasoning tasks is also covered by {\flare} as it maintains strong performance similar to F-CoT on MWP problems \cref{tab:results_main}.
Consequently, {\flare} allows combining algorithmic formalisation with simulated soft-reasoning, circumventing the pitfalls of using a deterministic external solver while still producing a query formalisation and problem space traversal. 

\subsection{Is simulating search useful?}
\begin{table}[t!]
\adjustbox{width=\textwidth}{
\begin{tabular}{@{}lccccc@{}}
\toprule
\textbf{Model} &
  \multicolumn{1}{l}{Avg. Number of Paths} &
  \multicolumn{1}{l}{Avg. \#Hops per path} &
  \multicolumn{1}{l}{Avg. \#Fails per path} &
  \multicolumn{1}{l}{Avg. Total Hops} &
  \multicolumn{1}{l}{Avg. Total Fails} \\ \midrule
\multicolumn{6}{c}{\textbf{Incorrect Answers}}                                                         \\
${Llama-3.1-8B}_{\flare}$ & 1.55 & 11.12 & \cellcolor[HTML]{CBCEFB}1.52 & 15.09 & \cellcolor[HTML]{CBCEFB}2.26 \\
${CmDR}_{\flare}$             & 1.51 & 6.55  & \cellcolor[HTML]{CBCEFB}0.68 & 10.56 & \cellcolor[HTML]{CBCEFB}1.39 \\
${CmDR+}_{\flare}$        & 0.92 & 7.52  & \cellcolor[HTML]{CBCEFB}1.13 & 8.57  & \cellcolor[HTML]{CBCEFB}1.32 \\
GPT-3.5           & 0.68 & 5.22  & \cellcolor[HTML]{CBCEFB}0.71 & 5.32  & \cellcolor[HTML]{CBCEFB}0.74 \\ \midrule
\multicolumn{6}{c}{\textbf{Correct Answers}}                                                           \\ \midrule
${Llama-3.1-8B}_{\flare}$ & 1.43 & 9.12  & 0.62                         & 12.36 & 0.96                         \\
${CmDR}_{\flare}$             & 1.19 & 7.10  & 0.42                         & 11.29 & 0.66                         \\
${CmDR+}_{\flare}$        & 0.97 & 7.19  & 0.42                         & 8.22  & 0.61                         \\
${GPT-3.5}_{\flare}$           & 0.82 & 5.65  & 0.26                         & 5.69  & 0.27                         \\ \bottomrule
\end{tabular}
}
\caption{The table depicts the difference in the average explored paths, hops, and fails during the reasoning process, which leads to incorrect or correct answers. The \cellcolor[HTML]{CBCEFB}purple colour illustrates that incorrect reasoning paths have fewer explorations that led to Failed search paths.}
\label{tab:reasoning_stats}
\end{table}

To understand if simulating a search over the problem space is useful, we compare the performance of {\flare} where we only generate the \emph{plan} without the subsequent \emph{code} or \emph{search} components. We refer to this framework setup as \emph{plan-only}, which can be seen in \cref{fig:flare_main} if we were to use only the \emph{plan} for answer generation. We completed this ablation using CmDR, CmDR+, and GPT-3.5, and we used GSM8K, AQuA, and StrategyQA as our baselines. The results in \cref{tab:plan_only} confirm that all of the models suffer massive performance degradation from $61.1 \rightarrow 49.9$ when omitting the \emph{code} and the \emph{search} components of {\flare}. We hypothesise that this is caused by insufficient problem space exploration when using the \emph{plan-only} setting.   
Furthermore, we have already seen in \cref{tab:results_main} that in methods, like F-CoT, that do not use simulated problem space exploration for soft-reasoning and only rely on \emph{plan} and \emph{code}, the performance also deteriorates even resulting in a complete breakdown of reasoning over the designated datasets. This can be viewed as a constrained version of {\flare} with \emph{code-only} execution.
Consequently, our results show that simulating problem space traversal is highly beneficial as it avoids the pitfalls posed by \emph{plan-only} and \emph{code-only} modes by exploring the problem space more rigorously and soft-reasoning during that traversal instead of using external solvers.

\subsection{Faithful Reasoning Improves Performance}
\begin{table}[t!]
\adjustbox{width=\textwidth}{
\begin{tabular}{@{}lccc@{}}
\toprule
\textbf{Model} & \multicolumn{1}{r}{Unique Explorations (\%) in Search} & \multicolumn{1}{l}{Relation overlap (\%)} & \multicolumn{1}{r}{Unused Code relations (\%)} \\ \midrule
\multicolumn{4}{c}{\textbf{Correct Answers}}   \\ \midrule
${Llama-3.1-8B}_{\flare}$   & 74.14   & 43.65  & 5.73  \\
${CmDR}_{\flare}$               & 59.06   & 35.96  & 4.02  \\
${CmDR+}_{\flare}$              & 64.30   & 34.47  & 4.54  \\
${GPT-3.5}_{\flare}$             & 64.46   & 37.55  & 1.90  \\ \midrule
\multicolumn{4}{c}{\textbf{Incorrect Answers}} \\ \midrule
${Llama-3.1-8B}_{\flare}$   & 54.69   & 35.04  & 9.28  \\
${CmDR}_{\flare}$               & 54.50   & 32.76  & 6.23  \\
${CmDR+}_{\flare}$              & 44.12   & 24.98  & 8.22  \\
${GPT-3.5}_{\flare}$             & 36.02   & 24.44  & 6.94  \\ \bottomrule
\end{tabular}
}
\caption{The table shows how the percentage of unique emergent inferences in search, overlapping relations between code and search, and unused relations in code impact answer correctness.}
\label{tab:reasoning_stats_deep}
\end{table}

As described in \cref{sec:method}, using {\flare} allows us to measure the faithfulness of the LLM reasoning process by comparing the simulated problem space traversals ${\Phi}^\prime_{{gen}}$ with actual traces ${\Phi}^\prime$ produced from a symbolic Prolog solver. To do this, we initially compute the percentage of syntactically correct executable code each LLM produces. We can see from the right part of \cref{fig:code_perc} that all of the models are capable of producing correct executable Prolog code in $67\%$ of cases on average and $\geq 50\%$ of cases at the very least. This shows that the simulated searches ${\Phi}^\prime_{{gen}}$ can be considered a representative sample that will be further used to accurately measure the faithfulness of the simulated search w.r.t. the generated code. After measuring the reasoning faithfulness for each model, we want to understand what impact it has on the performance of the LLM. In \cref{fig:faith_vs_acc}, we segment the models w.r.t. their ROUGE-Lsum scores. The results show that model performance is strongly positively correlated with reasoning faithfulness. 
However, we also observe in the left part of \cref{fig:code_perc} that executing semantically precise code results in an accurate answer only in $47\%$ of cases on average. Indeed, having a simulated search trace with a ROUGE-Lsum faithfulness score of $1$, would be equivalent to simply executing the program as proposed in F-CoT. Yet we have priorly shown that F-CoT struggles with reasoning tasks that are hard to formalise and require multi-hop commonsense and soft reasoning.
These two discoveries show that optimal LLM reasoning, conditioned on a search in the problem space, should be increasingly faithful toward the facts, relations and the search strategy defined within the code while simultaneously maintaining the capability for soft-reasoning along more abstractly defined concepts. Our results show that {\flare} allows LLMs to maintain a similar reasoning capacity.

\subsection{What is important during the search?}

\begin{table}[t!]
\adjustbox{width=\textwidth}{
\begin{tabular}{@{}lcccc@{}}
\toprule
\textbf{Model} &
  \multicolumn{1}{l}{Avg. hops per Paths} &
  \multicolumn{1}{r}{Hallucination (\%)} &
  \multicolumn{1}{r}{Unutilised knowledge (\%)} \\ \midrule
Llama-3.1-8B   & 9.4 & 63.3 & 62.9  \\
CmDR           & 6.7 & 54.7 & 56.9 \\
CmDR+         & 7.2 & 54.3 & 56.3 \\
GPT-3.5       & 5.5 & 49.3 & 52.1 \\ \bottomrule
\end{tabular}
}
\caption{The table shows the changes in simulated search statistics when using {\flare} w.r.t model scale from 8B to 100B+. Hallucinations refer to facts and predicates only used in trace, while unutilised knowledge relates to the facts and relations only seen in the code. }
\label{tab:scale_effect}
\end{table}

We expand the analysis of the simulated search traces to detect the reasons which can lead to optimal reasoning within an LLM. For this purpose, we calculate several statistics, like the average number of explored paths, average and total hops and failures per path, for each model during the simulated traversal. The failure in a path is an invalidation of a solution for a sub-goal explored during the search, which is used for backtracking, as explained in \cref{sec:method}. Calculating these statistics is simple as the \emph{search} component of {\flare}, seen in \cref{fig:flare_main}, is a structured simulation of a Prolog trace, where each line contains a hop of reasoning inference.
We split these statistics for the reasoning paths that lead to correct or incorrect outcomes. Our results in \cref{tab:reasoning_stats} show that LLM performance and reasoning optimally are not directly connected to the amount of explored paths or multi-hop inferences per path. We also see that traces that lead to incorrect answers have a higher number of failures per path and in total. We explain this phenomenon with the hypothesis that LLMs with traces that were optimal for reasoning and led to correct answers could skip exploring degenerate solutions due to strong commonsense reasoning capabilities.
Further analyses focus on identifying inconsistencies and failure modes (\cref{subsec:reasoning_inc}). By comparing relations in code with those in search traces, we measure emergent hallucinations and unused relations, highlighting areas of sub-optimal reasoning. Additionally, we assess the uniqueness of emergent facts per inference hop, which indicates the extent of problem-space exploration (\cref{tab:reasoning_stats_deep}).
The results in \cref{tab:reasoning_stats_deep} show consistently over each model that, on average, traces that lead to correct answers had a higher percentage of unique emergent facts (UEF) and overlap in the relations (OR) used between the code and search, while the portion of underutilized relations (UR) was lower. This means that optimal reasoning with an LLM requires a great degree of problem-space exploration with fewer relation hallucinations during the search and more relation utilization from the defined code. This aligns with our prior discoveries, which show a strong correlation between simulated search faithfulness towards the formalised code and model performance. Our framework {\flare} has these reasoning patterns ingrained within its inference pipeline. 
\begin{figure}[t!]
    \centering
    \includegraphics[clip=true,width=\textwidth]{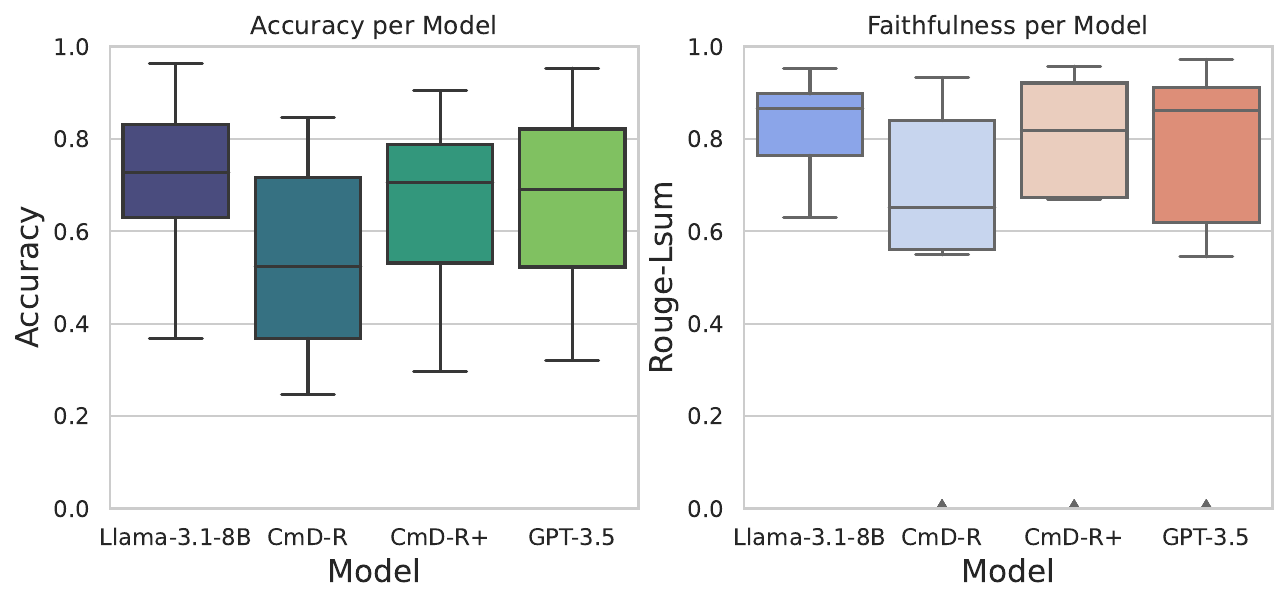}
    \caption{The effect of the model parameter scale from 8B to 100B+ on model accuracy (left) and faithfulness (right).}
    \label{fig:effect_of_scale}
\end{figure}

\subsection{The effect of scale}

We want to assess the impact of the number of parameters in the model on the overall performance and faithfulness. The results in \cref{fig:effect_of_scale} show no precise relation between model scale, performance and faithfulness. However, scaled models from the same family, i.e. CmDR (30B) and CmDR+ (100B), show improvements in reasoning faithfulness and model performance. We can also see in  \cref{tab:scale_effect} that as the model size increases, the average number of hops and the portion of hallucinations and unutilised knowledge decreases. This further confirms our prior assumptions that models with strong commonsense soft-reasoning capabilities can skip steps during the search while maintaining the knowledge and structure of the traversal strategy outlined in the code. 

\section{Conclusion}

This work introduces {\flare}, a novel approach for logic-aided interpretable formalisation and reasoning with simulated search over the problem space. We show that models of varying scales obtain state-of-the-art results compared to prompting paradigms like CoT and F-CoT. We further pinpoint that using {\flare} allows us to perform soft-reasoning with simulated search, making it flexible for diverse reasoning benchmarks. We introduce a method to measure model reasoning faithfulness w.r.t. the problem formalization ingrained within {\flare}. Our results show that model performance is positively correlated with the faithfulness of the reasoning process. The systematic studies of the method show the benefits of using simulated search compared to natural language reasoning and external symbolic solvers. We further show that using {\flare} allows us to interpretably and rigorously detect hallucinations and sub-optimal and inconsistent reasoning patterns.

\section*{Reproducibility Report}

To reproduce the results of our study, we provide the complete codebase, processing pipelines and prompts for each dataset. The only model hyper-parameter we explicitly fix is the temperature for greedy decoding. We also make the inference of all of the models using {\flare}, F-CoT and CoT across all of the datasets publicly available for further experimentation and exploration.


\section{Appendix}
\label{append:prompts}

\subsection{LLM Prompts}
We define straight-forward prompts for generating \emph{plan}, \emph{code} and \emph{search} simulation in {\flare}, which can be observed in \cref{tab:prompts}. 

\subsection{Dataset Statistics}

The datasets used in this study encompass a variety of domains, specifically targeting the performance of the models in interpreting Math Word Problems, multi-hop question answering, and relational inference. Table \ref{tab:data_stat_flare} provides a detailed breakdown of each dataset, including the number of few-shot in-context samples (shots), the number of test samples, and representative examples from each dataset. The datasets provide a comprehensive basis for evaluating the models' abilities to handle complex tasks across different domains, facilitating an in-depth analysis of model performance under few-shot conditions.

\begin{table*}[t!]
\label{tab:prompts}
    \centering
    \adjustbox{width=0.75\textwidth}{
    \begin{tabular}{@{}p{5cm}p{5cm}p{5cm}@{}}
        \toprule
        \textbf{Task} & \textbf{Prompt} & \textbf{Description} \\
        \midrule
        \textbf{Plan Generation} & 
        \begin{minipage}[t]{5cm}
        \vspace{0pt}
        Generate an explanation and analysis, and plan to generate a prompt for writing a swi-prolog code for the last task. The 3 sections should be exactly outlined. Your plan should show enough intermediate reasoning steps towards the answer. Construct the plan as much as you can and describe the logic specifically. When constructing the plan for the code prompt, actively use swi prolog search capabilities.
        \end{minipage}
        & 
        \begin{minipage}[t]{5cm}
        \vspace{0pt}
        Detailed instructions for generating an outline and plan, with an emphasis on reasoning steps and using Prolog's search capabilities.
        \end{minipage}
        \\
        \midrule
        \textbf{Code Generation} & 
        \begin{minipage}[t]{5cm}
        \vspace{0pt}
        Write a Prolog code to solve using the plan. If there are unknown or stochastic atoms or predicates, fill in the values for them as a logical assumption and add a comment in the same line Assumed atom/predicate". Do not use write and read commands within the code. The code should be very detailed and utilize swi prolog capabilities to the fullest. To run the program, at the end create a predicate named "query" that returns the correct numerical answer. The last line of the program should be the commented-out driver predicate "query". Write only the code.
        \end{minipage}
        & 
        \begin{minipage}[t]{5cm}
        \vspace{0pt}
        Instructions for generating a Prolog code based on the plan with assumptions for unknown atoms. Emphasizes code details and a final ``query'' predicate.
        \end{minipage}
        \\
        \midrule
        \textbf{Simulated Search} & 
        \begin{minipage}[t]{5cm}
        \vspace{0pt}
        Ignoring the read commands, explicitly write out the search paths that are explored by the code:
        \#\#\#\#
        Here are the paths [Starting Search Simulation]: 
        \#\#\#\# 
        [Path 1]:
        \end{minipage}
        & 
        \begin{minipage}[t]{5cm}
        \vspace{0pt}
        A task to simulate and display the search paths that the Prolog code would follow during execution.
        \end{minipage}
        \\
        \midrule
        \textbf{Final Answer} & 
        \begin{minipage}[t]{5cm}
        \vspace{0pt}
        Given the plan, the code and the explored search paths answer the question above. Answer with the correct numerical answer.
        \#\#\#\#\# Here is the answer:
        \end{minipage}
        & 
        \begin{minipage}[t]{5cm}
        \vspace{0pt}
        Final prompt asking for the correct numerical answer based on the previous steps.
        \end{minipage}
        \\
        \bottomrule
    \end{tabular}
    }
    \caption{Table of Prompts for Plan, Code, Simulated Search, and Final Answer generation for GSM8K \citep{DBLP:journals/corr/abs-2110-14168}.}
\end{table*}

\subsection{FLARE Pseudo-code}

Below, we present the pseudo-code for the execution of the \emph{plan}, \emph{code}, and \emph{search} procedures in {\flare}. The pseudo-code describes the modular pipeline in FLARE for tackling natural language queries with faithful simulated search.

\begin{itemize}
    \item bf{Plan Generation:} This stage creates a structured natural language outline of the reasoning process, breaking down the query into logical steps and analysis. The plan serves as the foundation for formalization into a logic-based representation.
    \item bf{Code Generation:} Based on the generated plan, a logic programming code (e.g., in Prolog) is synthesized. This code formalizes the query into a set of facts, relations, and goals, which collectively define the problem space for reasoning.
    \item bf{Search Simulation:} The generated code is utilized to simulate a search trace over the problem space. This includes iterative reasoning, backtracking when goals are unmet, and extracting emergent facts or relations during the process.
\end{itemize}

Each of these stages is implemented as a modular component. The generation from each of the stages feeds into the next, allowing seamless integration and incremental improvement in reasoning accuracy.

\begin{center}
\begin{tabularx}{\textwidth}{|p{2cm}|X|}
\hline
\textbf{Section} & \textbf{Content} \\ \hline

Question & 
"Can Reiki be stored in a bottle?" \\ \hline

Explanation & 
Reiki is a form of alternative therapy that involves the channeling of energy to promote healing and relaxation. It is often administered ... \\ \hline

Analysis &
To address the question of whether Reiki can be stored in a bottle, it is important to consider... \\ \hline

Plan &
\begin{itemize}
    \item Understand Reiki Energy:
        ...
    \item Explore Energy Storage:
        ...
    \item Consider Symbolism and Intention:
        ...
    \item Evaluate Practicality:
        ...
    \item Reflect on Tradition:
        ...
\end{itemize} \\ \hline

Code &
reiki\_energy(spiritual\_energy).
reiki\_energy(channeling\_through\_touch).
store\_energy\_in\_bottle(feasible) :- 
reiki\_energy(spiritual\_energy),    reiki\_energy(channeling\_through\_touch).

not\_common\_practice(store\_reiki\_in\_bottle).

query :-store\_energy\_in\_bottle(feasible),
not\_common\_practice(store\_reiki\_in\_bottle).

\\ \hline

Search Paths &
\begin{itemize}
    \item [Path 1:]
    \begin{itemize}
        \item 0: Start of execution: Beginning Search
        \item 1: Call: query
        \item 2: Call: store\_energy\_in\_bottle(feasible)
        \item 3: Call: reiki\_energy(spiritual\_energy)
        \item 4: Call: reiki\_energy(channeling\_through\_touch)
        \item 5: Call: not\_common\_practice(store\_reiki\_in\_bottle)
        \item 6: Call: feasible=no
        \item 7: Fail: feasible=no | {'Result': 'Search Failed'}
        \item 8: Fail: query | {'Result': 'Search Failed'}
    \end{itemize}
\end{itemize} \\ \hline

Answer &
"No, Reiki energy cannot be stored in a bottle based on the logical evaluation of its abstract, non-physical nature and traditional practices of Reiki." \\ \hline
\end{tabularx}
\smallskip
\captionof{table}{Complete example of \flare}
\label{tab:flare_ex_complete}
\end{center}

\begin{table}[t!]
\adjustbox{width=\textwidth}{
\begin{tabular}{@{}ccccc@{}}
\toprule
Domain &
  Dataset &
  Shots &
  Test Samples &
  Example \\ \midrule
\multirow{4}{*}{\begin{tabular}[c]{@{}c@{}}Math \\ Word \\ Problems\end{tabular}} &
  GSM8K &
  8 &
  1,319 &
  \begin{tabular}[c]{@{}c@{}}Q: A robe takes 2 bolts of blue fiber and half that much white fiber.  \\ How many bolts in total does it take?\\ A: 3\end{tabular} \\
 &
  SVAMP &
  8 &
  1,000 &
  Q: Dan had \$3 left with him after he bought a candy bar. If he had \$4 at the start, how much did the candy bar cost?A: 1 \\
 &
  MultiArith &
  8 &
  600 &
  \begin{tabular}[c]{@{}c@{}}Q: A pet store had 13 siamese cats and 5 house cats. During a sale they sold 10 cats. \\ How many cats do they have left? \\ A: 8\end{tabular} \\
 &
  ASDiv &
  8 &
  2,096 &
  \begin{tabular}[c]{@{}c@{}}Q: Adam has five more apples than Jackie. Jackie has nine apples. How many apples does Adam have?\\ \\ A: 14\end{tabular} \\
\multicolumn{1}{l}{} &
  AQuA &
  8 &
  254 &
  \begin{tabular}[c]{@{}c@{}}Q: A man walks at 5 kmph for 6 hrs and at 4 kmph for 12 hrs. His average speed is\\ Answer option: A)4 1/3 km/h, B)7 2/3 km/h, C)9 ½ km/h, D)8 km/h, E)81 km/h\\ A: A\end{tabular} \\ \midrule
\begin{tabular}[c]{@{}c@{}}Multi- \\ hop \\ QA\end{tabular} &
  StrategyQA &
  6 &
  2,290 &
  \begin{tabular}[c]{@{}c@{}}Q: Did Aristotle use a laptop? \\ A: False\end{tabular} \\
 &
  \begin{tabular}[c]{@{}c@{}}Date \\ Understanding\end{tabular} &
  10 &
  359 &
  \begin{tabular}[c]{@{}c@{}}Q: Yesterday was April 30, 2021. What is the date tomorrow in MM/DD/YYYY? \\ A: "05/02/2021"\end{tabular} \\
 &
  \begin{tabular}[c]{@{}c@{}}Sports \\ Understanding\end{tabular} &
  10 &
  977 &
  \begin{tabular}[c]{@{}c@{}}Q: Is the following sentence plausible? Lionel Messi was called for icing? \\ A: False\end{tabular} \\ \midrule
\begin{tabular}[c]{@{}c@{}}Relational \\ Inference\end{tabular} &
  CLUTRR &
  8 &
  1,042 &
  \begin{tabular}[c]{@{}c@{}}Q: [Carlos] is [Clarence]'s brother. [Carlos] and his sister, [Annie], went shopping. \\ asked her mom [Valerie] if she wanted anything, but [Valerie] said no. \\ How is [Valerie] related to [Clarence]? \\ A: "mother"\end{tabular} \\ \bottomrule
\end{tabular}
}
\caption{The statistics and examples of the datasets used in benchmarking. Shots refers to the number of few-shot in-context samples used during benchmarking.}
\label{tab:data_stat_flare}
\end{table}

\clearpage
\bibliography{references}
\bibliographystyle{acl_natbib}

\end{document}